\documentclass[10pt,journal,compsoc]{IEEEtran}
\usepackage{amsmath}
\usepackage{amssymb}
\usepackage{algorithm}
\usepackage{algorithmic}
\usepackage{multirow}
\usepackage{url}
\usepackage{amsthm}
\usepackage{mathrsfs}
\usepackage{graphicx}
\usepackage[tight,footnotesize]{subfigure}

\newtheorem{definition}{Definition}
\newtheorem{theorem}{Theorem}
\newtheorem{lemma}{Lemma}
\newtheorem{property}{Property}
\newtheorem{proposition}{Proposition}

\begin{document}

\title{Bilinear Factor Matrix Norm Minimization for Robust PCA: Algorithms and Applications}

\author{Fanhua~Shang,~\IEEEmembership{Member,~IEEE,}
        James~Cheng,
        Yuanyuan~Liu,
        Zhi-Quan~Luo,~\IEEEmembership{Fellow,~IEEE,}
        and~Zhouchen~Lin,~\IEEEmembership{Senior Member, IEEE}
\IEEEcompsocitemizethanks{\IEEEcompsocthanksitem F.\ Shang (Corresponding author), J.\ Cheng and Y.\ Liu are with the Department of Computer Science and Engineering, The Chinese University of Hong Kong,
Shatin, N.T., Hong Kong. E-mail: \{fhshang, jcheng, yyliu\}@cse.cuhk.edu.hk.\protect
\IEEEcompsocthanksitem Z.-Q.\ Luo is with Shenzhen Research Institute of Big Data, the Chinese University of Hong Kong, Shenzhen, China and Department of Electrical and Computer Engineering, University of Minnesota, Minneapolis, MN 55455, USA. E-mail: luozq@cuhk.edu.cn and luozq@umn.edu.\protect
\IEEEcompsocthanksitem Z.\ Lin is with Key Laboratory of Machine Perception (MOE), School of EECS, Peking University, Beijing 100871, P.R.\ China, and the Cooperative Medianet Innovation Center, Shanghai Jiao Tong University, Shanghai 200240, P.R.\ China. E-mail: zlin@pku.edu.cn.}
}

\markboth{IEEE TRANSACTIONS ON PATTERN ANALYSIS AND MACHINE INTELLIGENCE, VOL. 40, NO. 9, SEPTEMBER 2018}
{Shell \MakeLowercase{\textit{et al.}}: Bare Demo of IEEEtran.cls for Computer Society Journals}

\IEEEtitleabstractindextext{
\begin{abstract}
The heavy-tailed distributions of corrupted outliers and singular values of all channels in low-level vision have proven effective priors for many applications such as background modeling, photometric stereo and image alignment. And they can be well modeled by a hyper-Laplacian. However, the use of such distributions generally leads to challenging non-convex, non-smooth and non-Lipschitz problems, and makes existing algorithms very slow for large-scale applications. Together with the analytic solutions to $\ell_{p}$-norm minimization with two specific values of $p$, i.e., $p\!=\!1/2$ and $p\!=\!2/3$, we propose two novel bilinear factor matrix norm minimization models for robust principal component analysis. We first define the double nuclear norm and Frobenius/nuclear hybrid norm penalties, and then prove that they are in essence the Schatten-$1/2$ and $2/3$ quasi-norms, respectively, which lead to much more tractable and scalable Lipschitz optimization problems. Our experimental analysis shows that both our methods yield more accurate solutions than original Schatten quasi-norm minimization, even when the number of observations is very limited. Finally, we apply our penalties to various low-level vision problems, e.g., text removal, moving object detection, image alignment and inpainting, and show that our methods usually outperform the state-of-the-art methods.
\end{abstract}

}

\maketitle

\setcounter{page}{2066}

\section{Introduction}
\label{intr}

\begin{figure}[t]
\centering
\includegraphics[width=0.31\linewidth]{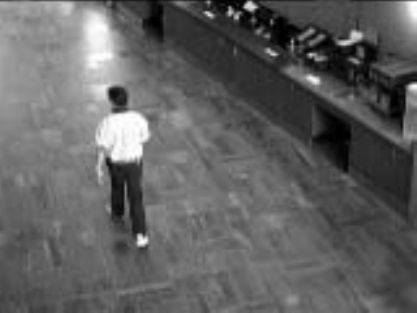}\,
\includegraphics[width=0.31\linewidth]{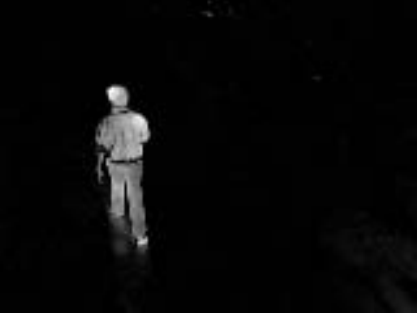}\,
\includegraphics[width=0.318\linewidth]{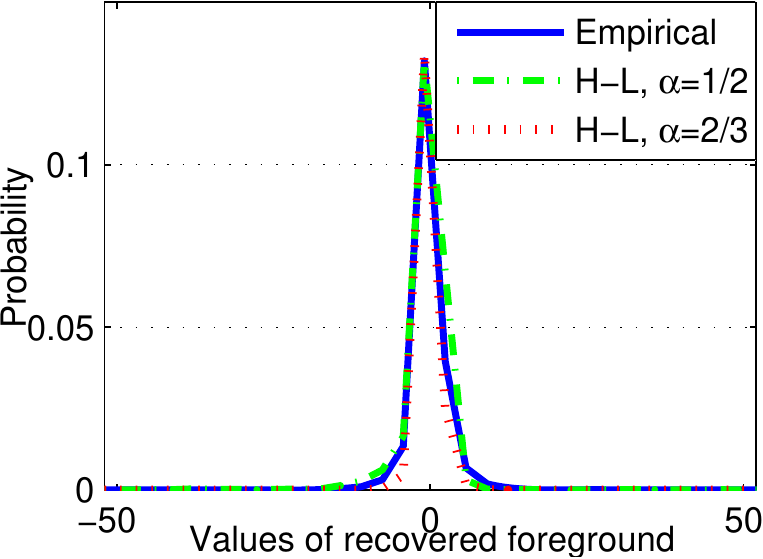}
\caption{Sparsity priors. Left: A typical image of video sequences. Middle: the sparse component recovered by~\cite{candes:rpca}. Right: the empirical distribution of the sparse component (blue solid line), along with a hyper-Laplacian fit with $\alpha\!=\!1/2$ (green dashdot line) and $\alpha\!=\!2/3$ (red dotted line).}
\label{fig1}
\end{figure}

\IEEEPARstart{T}{he} sparse and low-rank priors have been widely used in many real-world applications in computer vision and pattern recognition, such as image restoration~\cite{aharon:ksvd}, face recognition~\cite{wright:sr}, subspace clustering~\cite{torre:rsl,elhamifar:ssc,liu:lrr} and robust principal component analysis~\cite{candes:rpca} (RPCA, also called low-rank and sparse matrix decomposition in~\cite{tao:lrsd, zhou:godec} or robust matrix completion in~\cite{chen:rmc}). Sparsity plays an important role in various low-level vision tasks. For instance, it has been observed that the gradient of natural scene images can be better modeled with a heavy-tailed distribution such as hyper-Laplacian distributions ($p(x)\!\propto\!e^{-k|x|^{\alpha}}$, typically with $0.5\!\leq\!\alpha\!\leq\!0.8$, which correspond to non-convex $\ell_{p}$-norms)~\cite{krishnan:l12, mallat:wav}, as exhibited by the sparse noise/outliers in low-level vision problems~\cite{luo:nnm} shown in Fig.\ \ref{fig1}. To induce sparsity, a principled way is to use the convex $\ell_{1}$-norm~\cite{candes:rpca, eriksson:lrma, ke:rpca, kim:factEN, zheng:lrma, zhou:mod}, which is the closest convex relaxation of the sparser $\ell_{p}$-norm, with compressed sensing being a prominent example. However, it has been shown in~\cite{fan:ve} that the $\ell_{1}$-norm over-penalizes large entries of vectors and results in a biased solution. Compared with the $\ell_{1}$-norm, many non-convex surrogates of the $\ell_{0}$-norm listed in~\cite{lu:lrm} give a closer approximation, e.g., SCAD~\cite{fan:ve} and MCP~\cite{zhang:mcp}. Although the use of hyper-Laplacian distributions makes the problems non-convex, fortunately an analytic solution can be derived for two specific values of $p$, $1/2$ and $2/3$, by finding the roots of a cubic and quartic polynomial, respectively~\cite{krishnan:l12,xu:l12r,zuo:gist}. The resulting algorithm can be several orders of magnitude faster than existing algorithms~\cite{krishnan:l12}.

As an extension from vectors to matrices, the low-rank structure is the sparsity of the singular values of a matrix. Rank minimization is a crucial regularizer to induce a low-rank solution. To solve such a problem, the rank function is usually relaxed by its convex envelope~\cite{liu:lrr, cai:svt, chandrasekaran:rpca, jin:hankel, recht:nnm, shahid:grpca, zhang:op, zhang:tlrt}, the nuclear norm (i.e., the sum of the singular values, also known as the trace norm or Schatten-$1$ norm~\cite{recht:nnm,gu:norm}). By realizing the intimate relationship between the $\ell_{1}$-norm and nuclear norm, the latter also over-penalizes large singular values, that is, it may make the solution deviate from the original solution as the $\ell_{1}$-norm does~\cite{lu:lrm, nie:rmc}. Compared with the nuclear norm, the Schatten-${q}$ norm for $0\!<\!q\!<\!1$ is equivalent to the $\ell_{q}$-norm on singular values and makes a closer approximation to the rank function~\cite{lai:irls, lu:irls}. Nie \emph{et al.}~\cite{nie:rmc} presented an efficient augmented Lagrange multiplier (ALM) method to solve the joint $\ell_{p}$-norm and Schatten-$q$ norm (LpSq) minimization. Lai \emph{et al.}~\cite{lai:irls} and Lu \emph{et al.}~\cite{lu:irls} proposed iteratively reweighted least squares methods for solving Schatten quasi-norm minimization problems. However, all these algorithms have to be solved iteratively, and involve singular value decomposition (SVD) in each iteration, which occupies the largest computation cost, $\mathcal{O}(\min(m,n)mn)$~\cite{shang:snm,golub:mc}.

\begin{figure}[t]
\centering
\includegraphics[width=0.316\linewidth]{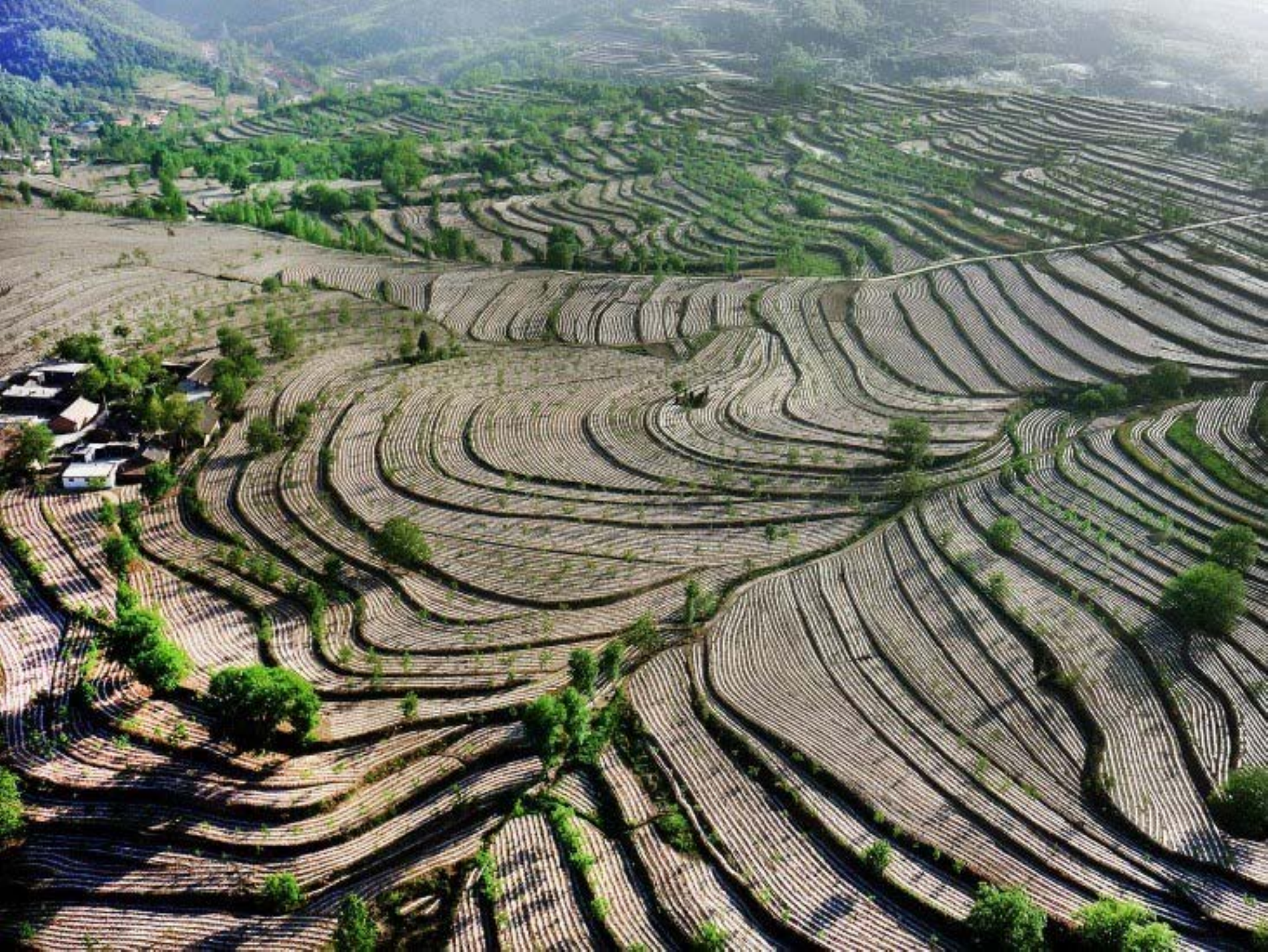}\;
\includegraphics[width=0.316\linewidth]{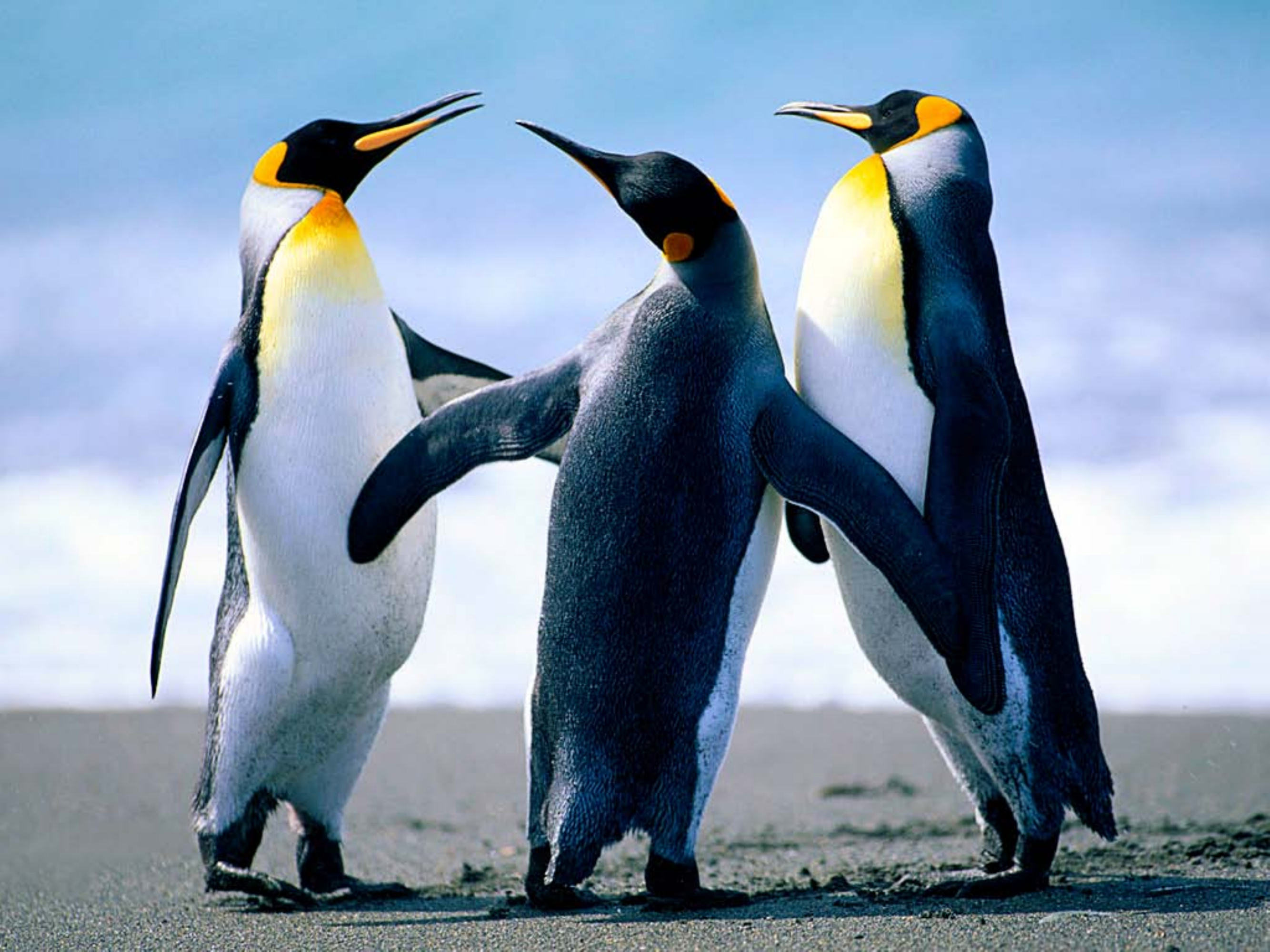}\;
\includegraphics[width=0.316\linewidth]{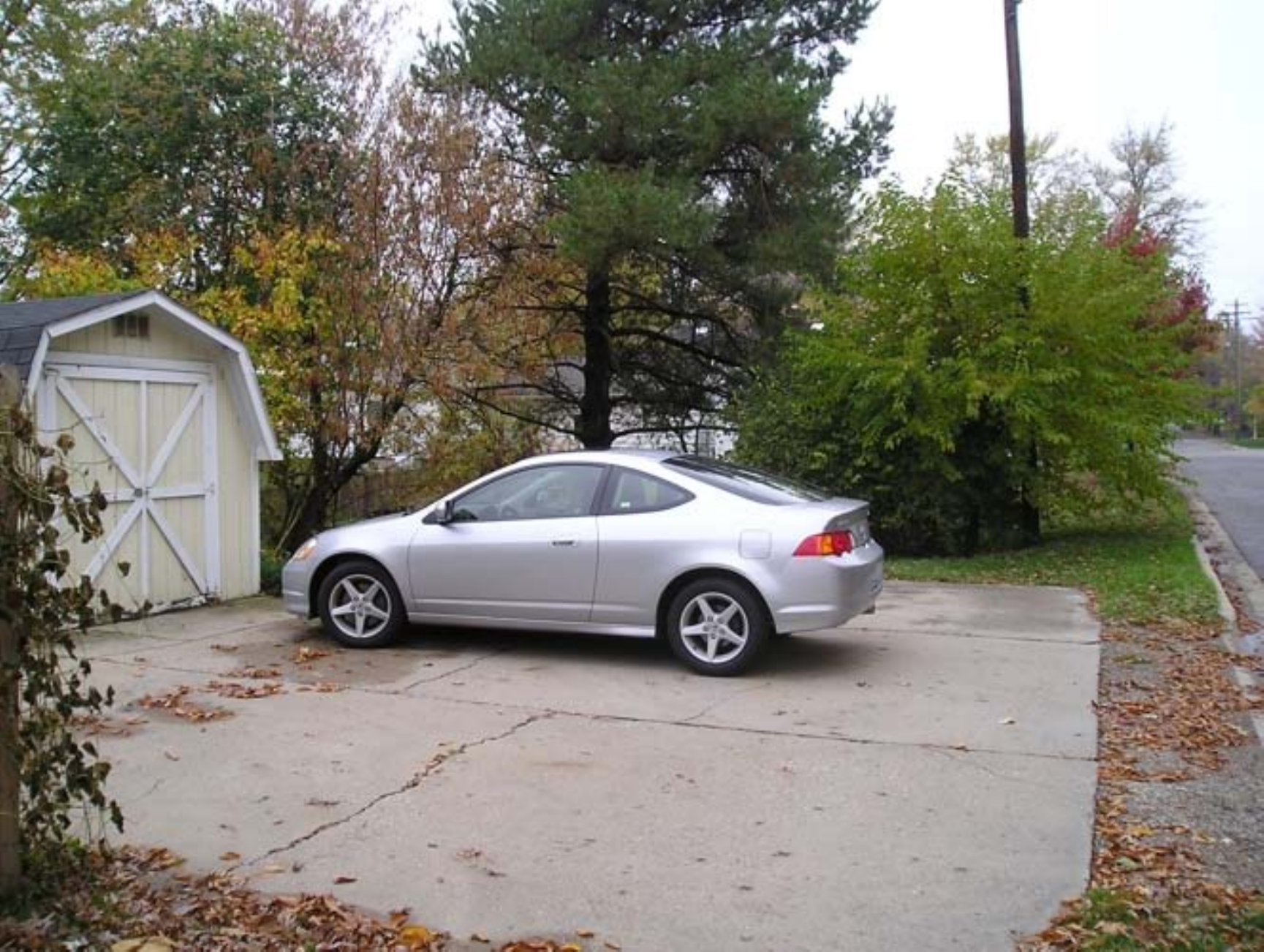}
\includegraphics[width=0.326\linewidth]{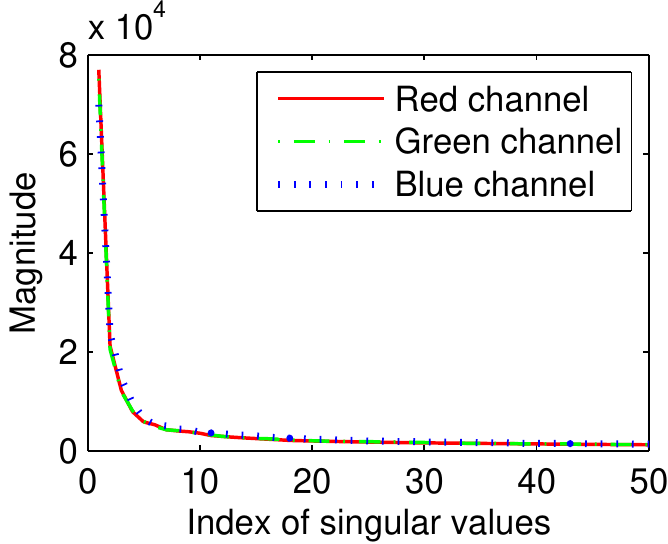}
\includegraphics[width=0.326\linewidth]{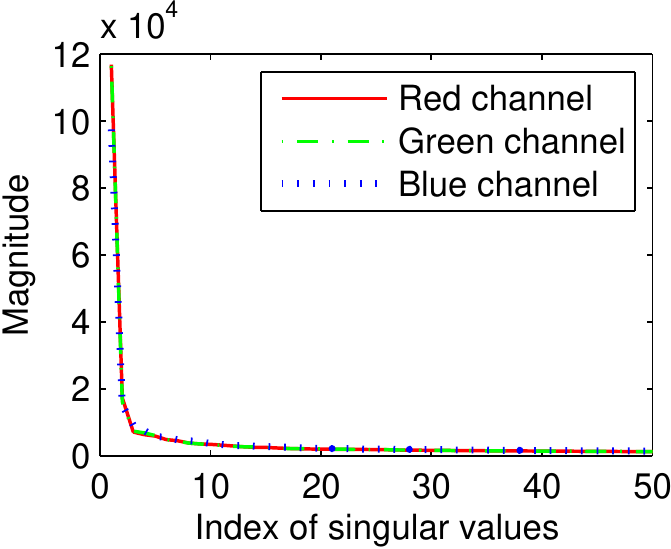}
\includegraphics[width=0.326\linewidth]{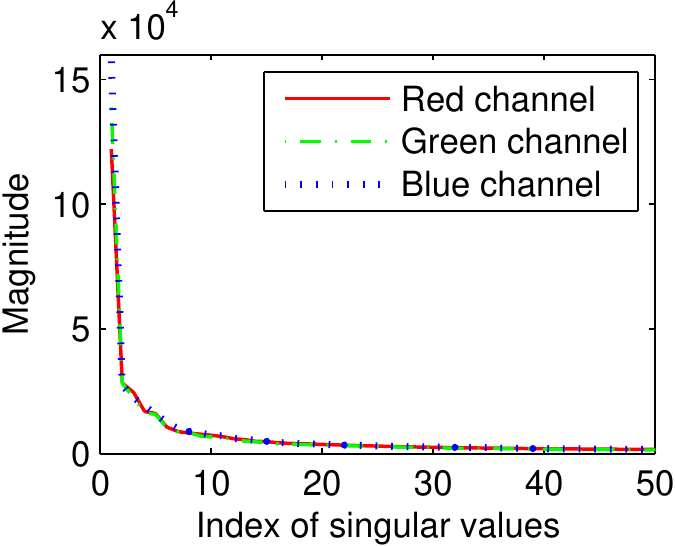}
\caption{The heavy-tailed empirical distributions (bottom) of the singular values of the three channels of these three images (top).}
\label{fig2}
\end{figure}

It has been shown in~\cite{wang:llv} that the singular values of nonlocal matrices in natural images usually exhibit a heavy-tailed distribution ($p(\sigma)\!\!\propto\!\!e^{-k|\sigma|^{\alpha}}$), as well as the similar phenomena in natural scenes~\cite{gu:wnnm, hu:tnnr}, as shown in Fig.\ \ref{fig2}. Similar to the case of heavy-tailed distributions of sparse outliers, the analytic solutions can be derived for the two specific cases of $\alpha$, $1/2$ and $2/3$. However, such algorithms have high per-iteration complexity $\mathcal{O}(\min(m,n)mn)$. Thus, we naturally want to design equivalent, tractable and scalable forms for the two cases of the Schatten-$q$ quasi-norm, $q\!=\!1/2$ and $2/3$, which can fit the heavy-tailed distribution of singular values closer than the nuclear norm, as analogous to the superiority of the $\ell_{p}$ quasi-norm (hyper-Laplacian priors) to the $\ell_{1}$-norm (Laplacian priors).

We summarize the main contributions of this work as follows. 1) By taking into account the heavy-tailed distributions of both sparse noise/outliers and singular values of matrices, we propose two novel tractable bilinear factor matrix norm minimization models for RPCA, which can fit empirical distributions very well to corrupted data. 2) Different from the definitions in our previous work~\cite{shang:snm}, we define the double nuclear norm and Frobenius/nuclear hybrid norm penalties as tractable low-rank regularizers. Then we prove that they are in essence the Schatten-${1/2}$ and ${2/3}$ quasi-norms, respectively. The solution of the resulting minimization problems only requires SVDs on two much smaller factor matrices as compared with the much larger ones required by existing algorithms. Therefore, our algorithms can reduce the per-iteration complexity from $\mathcal{O}(\min(m,n)mn)$ to $\mathcal{O}(mnd)$, where $d\!\ll\!m,n$ in general. In particular, our penalties are Lipschitz, and more tractable and scalable than original Schatten quasi-norm minimization, which is non-Lipschitz and generally NP-hard~\cite{lai:irls,bian:ipa}. 3) Moreover, we present the convergence property of the proposed algorithms for minimizing our RPCA models and provide their proofs. We also extend our algorithms to solve matrix completion problems, e.g., image inpainting. 4) We empirically study both of our bilinear factor matrix norm minimizations and show that they outperform original Schatten norm minimization, even with only a few observations. Finally, we apply the defined low-rank regularizers to address various low-level vision problems, e.g., text removal, moving object detection, and image alignment and inpainting, and obtain superior results than existing methods.

\begin{table}[t]
\setlength\tabcolsep{3pt}
\renewcommand\arraystretch{1.15}
\scriptsize
\caption{The norms of sparse and low-rank matrices. Let $\sigma\!=\!(\sigma_{1},\ldots,\sigma_{r})\!\in\!\mathbb{R}^{r}$ be the non-zero singular values of $L\!\in\!\mathbb{R}^{m\!\times\!n}$.}
\label{tab1}
\begin{tabular}[c]{c|cc|cc}
\hline
$\!p,q\!$ & \multicolumn{2}{c|}{Sparsity} & \multicolumn{2}{c}{Low-rankness} \\
\hline
$\!0\!$	               & \!$\|S\|_{\ell_{0}}$	& {$\!\ell_{0}$-norm}	& \!$\|L\|_{S_{0}}\!=\!\|\sigma\|_{\ell_{0}}$	& \!\!\!rank \\
\hline
$\!(0,1)$ & $\|S\|_{\ell_{p}}\!\!\!=\!\!(\!\sum\!|\!S_{ij}\!|^{p}\!)^{\!1\!/\!p}\!\!$	& {$\!\ell_{p}$-norm} &$\|L\|_{\!S_{q}}\!\!\!=\!\!(\!\sum\!\sigma^{q}_{\!k}\!)^{\!1\!/\!q}$	 &\!\!\!$\textup{Schatten-}q$ norm\!\\
\hline
$\!1\!$	               & \!$\|S\|_{\ell_{1}}\!\!=\!\!\sum\!|S_{ij}|$	  & {$\!\ell_{1}$-norm}	& \!$\|L\|_{*}\!\!=\!\!\sum\!\sigma_{k}$	& \!\!\!nuclear norm\\
\hline
$\!2\!$	               & \!$\|S\|_{\!F}\!=\!\sqrt{\sum\!S^{2}_{ij}}$	& \!\!\!Frobenius norm   & \!$\|L\|_{\!F}\!=\!\sqrt{\sum\!\sigma^{2}_{k}}$ & \!\!Frobenius norm\!\!\\
\hline
\end{tabular}
\end{table}

\section{Related Work}
\label{back}
In this section, we mainly discuss some recent advances in RPCA, and briefly review some existing work on RPCA and its applications in computer vision (readers may see~\cite{candes:rpca} for a review). RPCA~\cite{chandrasekaran:rpca, wright:rpca} aims to recover a low-rank matrix $L\!\in\!\mathbb{R}^{m\times n}$ ($m\!\geq\! n$) and a sparse matrix $S\!\in\! \mathbb{R}^{m\times n}$ from corrupted observations $D\!=\!L^{*}\!+\!S^{*}\!\in\! \mathbb{R}^{m\times n}$ as follows:
\begin{equation}\label{eq2.01}
\min_{L,S}\; \lambda\,\textup{rank}(L)+\|S\|_{\ell_{0}},\;\textup{s.t.}, L+S=D
\end{equation}
where $\|\!\cdot\!\|_{\ell_{0}}$ denotes the $\ell_{0}$-norm\footnote{Strictly speaking, the $\ell_{0}$-norm is not actually a norm, and is defined as the number of non-zero elements. When $p\!\geq\!1$, $\|S\|_{\ell_{p}}$ strictly defines a norm which satisfies the three norm conditions, while it defines a quasi-norm when $0\!<\!p\!<\!1$. Due to the relationship between $\|S\|_{\ell_{p}}$ and $\|L\|_{\!S_{q}}$, the latter has the same cases as the former.} and $\lambda\!>\!0$ is a regularization parameter. Unfortunately, solving \eqref{eq2.01} is NP-hard. Thus, we usually use the convex or non-convex surrogates to replace both of the terms in \eqref{eq2.01}, and formulate this problem into the following more general form:
\begin{equation}\label{eq2.02}
\min_{L,S}\; \lambda\|L\|^{q}_{\!S_{q}}+\|S\|^{p}_{\ell_{p}},\;\textup{s.t.}, \mathcal{P}_{\Omega}(L\!+\!S)=\mathcal{P}_{\Omega}(D)
\end{equation}
where in general $p,q\!\in\![0,2]$, $\|S\|_{\ell_{p}}$ and $\|L\|_{\!S_{q}}$ are depicted in Table~\ref{tab1} and can be seen as the loss term and regularized term, respectively, and $\mathcal{P}_{\Omega}$ is the orthogonal projection onto the linear subspace of matrices supported on $\Omega\!:=\!\{(i,j)|D_{ij}\,\textup{is observed}\}$: $\mathcal{P}_{\Omega}(D)_{ij}\!=\!D_{ij}$ if $(i,j)\!\in\!\Omega$ and $\mathcal{P}_{\Omega}(D)_{ij}\!=\!0$ otherwise. If $\Omega$ is a small subset of the entries of the matrix, \eqref{eq2.02} is also known as the robust matrix completion problem as in~\cite{chen:rmc}, and it is impossible to exactly recover ${S}^{*}$~\cite{wright:cpcp}. As analyzed in~\cite{shang:rbf}, we can easily see that the optimal solution $S_{\Omega^{c}}\!=\!\textbf{0}$, where $\Omega^{c}$ is the complement of $\Omega$, i.e., the index set of unobserved entries. When $p\!=\!2$ and $q\!=\!1$, \eqref{eq2.02} becomes a nuclear norm regularized least squares problem as in~\cite{toh:apg} (e.g., image inpainting in Sec.\ \ref{sec623}).

\begin{table*}
\begin{center}
\small
\caption{Comparison of various RPCA models and their properties. Note that $U\!\in\!\mathbb{R}^{m\times d}$ and $V\!\in\! \mathbb{R}^{n\times d}$ are the factor matrices of $L$, i.e., $L\!=\!UV^{T}$.}
\label{tab2}
\begin{tabular}{|l|c|c|c|c|c|}
\hline
\!\multirow{2}*{Model} & \multirow{2}*{Objective function} & \multirow{2}*{Constraints} & \!\multirow{2}*{Parameters}\! & \!\multirow{2}*{Convex?}\!\! & \!{Per-iteration}\!\!\\
& & & & &\!{Complexity}\!\!\\
\hline\hline
\!RPCA~\cite{candes:rpca,lin:almm}\!\!\!\! & $\lambda\|L\|_{\ast}+\|S\|_{\ell_{1}}$ & $L\!+\!S=D$ & $\lambda$ & Yes  & $\mathcal{O}(mn^{2})$\\
\hline
\!PSVT~\cite{oh:psvt}           & $\lambda\|L\|_{d}+\|S\|_{\ell_{1}}$ & \multirow{2}{*}{$L\!+\!S=D$}    & $\lambda,d$ & \multirow{2}{*}{No}   & \multirow{2}{*}{$\mathcal{O}(mn^{2})$}\\
\!WNNM~\cite{gu:wnnmj}          & $\lambda\|L\|_{w,*}+\|S\|_{\ell_{1}}$ &    & $\lambda$   &    & \\
\hline
\!LMaFit~\cite{shen:alad} & $\|D\!-\!L\|_{\ell_{1}}$ & $UV^{T}\!=L$ &$d$  & \multirow{5}{*}{No} & \multirow{5}{*}{$\mathcal{O}(mnd)$}\\
\!MTF~\cite{liu:ftf}\! & $\lambda\|W\|_{\ast}+\|S\|_{\ell_{1}}$ & $\!\!\!UWV^{T}\!\!\!+\!S\!=\!\!D,U^{T}\!U\!\!=\!\!V^{T}\!V\!\!=\!I_{d}\!\!\!$ & $\lambda,d$ & &\\
\!RegL1~\cite{zheng:lrma}\!\! & $\lambda\|V\|_{\ast}+\|\mathcal{P}_{\Omega}(D\!-\!UV^{T})\|_{\ell_{1}}$ & $U^{T}U\!=\!I_{d}$ & $\lambda,d$ & &\\
\!Unifying~\cite{cabral:nnbf} & $\frac{\lambda}{2}(\|U\|^{2}_{F}\!+\!\|V\|^{2}_{F})+\|\mathcal{P}_{\Omega}(D\!-\!L)\|_{\ell_{1}}$ & $UV^{T}\!=L$ & $\lambda,d$ & &\\
\!factEN~\cite{kim:factEN} & $\!\!\!\frac{\lambda_{1}}{2}\!(\|U\|^{2}_{F}\!\!+\!\!\|V\|^{2}_{F})\!+\!\!\frac{\lambda_{2}}{2}\!\|L\|^{2}_{F}\!\!+\!\!\|\mathcal{P}_{\Omega}(D\!-\!L)\|_{\ell_{1}}\!\!\!$ & $UV^{T}\!=L$ & $\lambda_{1},\lambda_{2},d$ & &\\
\hline
\!LpSq~\cite{nie:rmc}           & $\lambda\|L\|^{q}_{\!S_{q}}\!+\!\|\mathcal{P}_{\Omega}(D\!-\!L)\|^{p}_{\ell_{p}}$ $(0\!<\!p,q\!\leq\!1)$ & $L\!+\!S=D$ & $\lambda$ & No & $\mathcal{O}(mn^{2})$\\
\hline
$\!(\textup{S+L})_{1/2}$ & $\frac{\lambda}{2}(\|U\|_{*}\!+\!\|V\|_{*})+\|\mathcal{P}_{\Omega}(S)\|^{{1}/{2}}_{\ell_{{1}/{2}}}$ & \multirow{2}{*}{$L\!+\!S\!=\!D,UV^{T}\!=L$} & \multirow{2}{*}{$\lambda,d$} & \multirow{2}{*}{No} & \multirow{2}{*}{$\mathcal{O}(mnd)$}\\
$\!(\textup{S+L})_{2/3}$ & $\frac{\lambda}{3}(\|U\|^{2}_{F}\!+\!2\|V\|_{*})+\|\mathcal{P}_{\Omega}(S)\|^{2/3}_{\ell_{{2}/{3}}}$ & & & &\\
\hline
\end{tabular}
\end{center}
\end{table*}

\subsection{Convex Nuclear Norm Minimization}
In~\cite{candes:rpca,wright:rpca,lin:almm}, both of the non-convex terms in (\ref{eq2.01}) are replaced by their convex envelopes, i.e., the nuclear norm ($q\!=\!1$) and the $\ell_{1}$-norm ($p\!=\!1$), respectively.
\begin{equation}\label{eq2.03}
\min_{L,S}\; \lambda\|L\|_{*}+\|S\|_{\ell_{1}},\;\textup{s.t.}, L+S=D.
\end{equation}
Wright \emph{et al.}~\cite{wright:rpca} and Cand\`{e}s \emph{et al.}~\cite{candes:rpca} proved that, under some mild conditions, the convex relaxation formulation (\ref{eq2.03}) can exactly recover the low-rank and sparse matrices $(L^{*},S^{*})$ with high probability. The formulation (\ref{eq2.03}) has been widely used in many computer vision applications, such as object detection and background subtraction~\cite{zhou:mod}, image alignment~\cite{peng:rasl}, low-rank texture analysis~\cite{zhang:tlrt}, image and video restoration~\cite{ji:lrmc}, and subspace clustering~\cite{shahid:grpca}. This is mainly because the optimal solutions of the sub-problems involving both terms in (\ref{eq2.03}) can be obtained by two well-known proximal operators: the singular value thresholding (SVT) operator~\cite{cai:svt} and the soft-thresholding operator~\cite{donoho:st}. The $\ell_{1}$-norm penalty in~(\ref{eq2.03}) can also be replaced by the $\ell_{1,2}$-norm as in outlier pursuit~\cite{zhang:op,xu:rpca,yang:multi,ma:fea} and subspace learning~\cite{liu:lrr,shao:gtsl,ding:llrc}.

To efficiently solve the popular convex problem (\ref{eq2.03}), various first-order optimization algorithms have been proposed, especially the alternating direction method of multipliers~\cite{boyd:admm} (ADMM, or also called inexact ALM in~\cite{lin:almm}). However, they all involve computing the SVD of a large matrix of size $m\!\times\!n$ in each iteration, and thus suffer from high computational cost, which severely limits their applicability to large-scale problems~\cite{oh:frsvt}, as well as existing Schatten-$q$ quasi-norm ($0\!<\!q\!<\!1$) minimization algorithms such as LpSq~\cite{nie:rmc}. While there have been many efforts towards fast SVD computation such as partial SVD~\cite{larsen:svd}, the performance of those methods is still unsatisfactory for many real applications~\cite{oh:frsvt,cai:fsvt}.

\subsection{Non-Convex Formulations}
To address this issue, Shen \emph{et al.}~\cite{shen:alad} efficiently solved the RPCA problem by factorizing the low-rank component into two smaller factor matrices, i.e., $L\!=\!UV^{T}$ as in~\cite{jiang:lrmf}, where $U\!\in\!\mathbb{R}^{m\times d}$, $V\!\in\!\mathbb{R}^{n\times d}$, and usually $d\!\ll\!\min(m,n)$, as well as the matrix tri-factorization (MTF)~\cite{liu:ftf} and factorized data~\cite{xiao:lrr} cases. In~\cite{zheng:lrma,shang:rbf,shu:rosl,shang:rpca,liu:as}, the column-orthonormal constraint is imposed on the first factor matrix $U$. According to the following matrix property, the original convex problem (\ref{eq2.03}) can be reformulated as a smaller matrix nuclear norm minimization problem.

\begin{property}
For any matrix $L\!\in\!\mathbb{R}^{m\times n}$ and $L\!=\!UV^{T}$. If $U$ has orthonormal columns, i.e., $U^TU\!=\!I_{d}$, then $\|L\|_{*}\!=\!\|V\|_{*}$.
\end{property}

Cabral \emph{et al.}~\cite{cabral:nnbf} and Kim \emph{et al.}~\cite{kim:factEN} replaced the nuclear norm regularizer in (\ref{eq2.03}) with the equivalent non-convex formulation stated in Lemma~\ref{lem1}, and proposed scalable bilinear spectral regularized models, similar to collaborative filtering applications in~\cite{recht:nnm, srebro:mmmf}. In~\cite{kim:factEN}, an elastic-net regularized matrix factorization model was proposed for subspace learning and low-level vision problems.

Besides the popular nuclear norm, some variants of the nuclear norm were presented to yield better performance. Gu \emph{et al.}~\cite{gu:wnnm} proposed a weighted nuclear norm (i.e., $\|L\|_{w,*}\!=\!\sum_{i}\!w_{i}\sigma_{i}$, where $w\!=\![w_{1},\ldots,w_{n}]^{T}$), and assigned different weights $w_{i}$ to different singular values such that the shrinkage operator becomes more effective. Hu \emph{et al.}~\cite{hu:tnnr} first used the truncated nuclear norm (i.e., $\|L\|_{d}\!=\!\sum^{n}_{i=d+\!1}\!\sigma_{i}$) to address image recovery problems. Subsequently, Oh \emph{et al.}~\cite{oh:psvt} proposed an efficient partial singular value thresholding (PSVT) algorithm to solve many RPCA problems of low-level vision. The formulations mentioned above are summarized in Table~\ref{tab2}.

\section{Bilinear Factor Matrix Norm Minimization}
\label{dfnp}
In this section, we first define the double nuclear norm and Frobenius/nuclear hybrid norm penalties, and then prove the equivalence relationships between them and the Schatten quasi-norms. Incorporating with hyper-Laplacian priors of both sparse noise/outliers and singular values, we propose two novel bilinear factor matrix norm regularized models for RPCA. Although the two models are still non-convex and even non-smooth, they are more tractable and scalable optimization problems, and their each factor matrix term is convex. On the contrary, the original Schatten quasi-norm minimization problem is very difficult to solve because it is generally non-convex, non-smooth, and non-Lipschitz, as well as the $\ell_{q}$ quasi-norm~\cite{bian:ipa}.

As in some collaborative filtering applications~\cite{recht:nnm, srebro:mmmf}, the nuclear norm has the following alternative non-convex formulations.

\begin{lemma}\label{lem1}
For any matrix $X\!\in\! \mathbb{R}^{m\times n}$ of rank at most $r\leq d$, the following equalities hold:
\begin{equation}\label{eq3.01}
\begin{split}
\|X\|_{*}&=\min_{U\in\mathbb{R}^{m\!\times\! d},V\in\mathbb{R}^{n\!\times\! d}:X=UV^{T}}\frac{1}{2}(\|U\|^{2}_{F}\!+\!\|V\|^{2}_{F})\\
&=\min_{U,V:X=UV^{T}}\|U\|_{F}\|V\|_{F}.
\end{split}
\end{equation}
\end{lemma}

The bilinear spectral penalty in the first equality of~\eqref{eq3.01} has been widely used in low-rank matrix completion and recovery problems, such as RPCA~\cite{kim:factEN,cabral:nnbf}, online RPCA~\cite{feng:rpca}, matrix completion~\cite{sun:gmc}, and image inpainting~\cite{jin:hankel}.

\subsection{Double Nuclear Norm Penalty}
Inspired by the equivalence relation between the nuclear norm and the bilinear spectral penalty, our double nuclear norm (D-N) penalty is defined as follows.

\begin{definition}\label{def1}
For any matrix $X\!\in\!\mathbb{R}^{m\times n}$ of rank at most $r\leq d$, we decompose it into two factor matrices $U\!\in\!\mathbb{R}^{m\times d}$ and $V\!\in\!\mathbb{R}^{n\times d}$ such that $X\!=\!UV^{T}$. Then the double nuclear norm penalty of $X$ is defined as
\begin{equation}\label{eq3.02}
\|X\|_{\textup{D-N}}=\!\min_{U,V:X=UV^{T}}\frac{1}{4}(\|U\|_{*}\!+\!\|V\|_{*})^{2}.
\end{equation}
\end{definition}

Different from the definition in~\cite{shang:snm,shang:tsnm}, i.e., $\min_{U,V:X=UV^{T}}\|U\|_{*}\|V\|_{*}$, which cannot be used directly to solve practical problems, Definition~\ref{def1} can be directly used in practical low-rank matrix completion and recovery problems, e.g., RPCA and image recovery. Analogous to the well-known Schatten-${q}$ quasi-norm~\cite{nie:rmc,lai:irls,lu:irls}, the double nuclear norm penalty is also a quasi-norm, and their relationship is stated in the following theorem.

\begin{theorem}\label{the1}
The double nuclear norm penalty $\|\!\cdot\!\|_{\textup{D-N}}$ is a quasi-norm, and also the Schatten-${1/2}$ quasi-norm, i.e.,
\begin{equation}\label{eq3.03}
\|X\|_{\textup{D-N}}=\|X\|_{S_{1/2}}.
 \end{equation}
\end{theorem}

To prove Theorem~\ref{the1}, we first give the following lemma, which is mainly used to extend the well-known trace inequality of John von Neumann~\cite{mirsky:ti,neumann:ti}.

\begin{lemma}\label{lem2}
Let $X\!\in\!\mathbb{R}^{n\times n}$ be a symmetric positive semi-definite (PSD) matrix and its full SVD be $U_{X}\Sigma_{X}U^{T}_{X}$ with $\Sigma_{X}\!=\!\textup{diag}(\lambda_{1},\ldots,\lambda_{n})$. Suppose $Y$ is a diagonal matrix (i.e., $Y\!=\!\textup{diag}(\tau_{1},\ldots,\tau_{n})$), and if $\lambda_{1}\!\geq\!\ldots\!\geq\!\lambda_{n}\!\geq\!0$ and $0\!\leq\!\tau_{1}\!\leq\!\ldots\!\leq\!\tau_{n}$, then
\begin{equation*}
\textup{Tr}(X^{T}Y)\geq \sum^{n}_{i=1}\lambda_{i}\tau_{i}.
\end{equation*}
\end{lemma}
Lemma~\ref{lem2} can be seen as a special case of the well-known von Neumann's trace inequality, and its proof is provided in the Supplementary Materials.

\begin{lemma}\label{lem3}
For any matrix $X\!=\!{U}{V}^{T}\!\in\mathbb{R}^{m\times n}$, ${U}\!\in\! \mathbb{R}^{m\times d}$ and ${V}\!\in\! \mathbb{R}^{n\times d}$, the following inequality holds:
\begin{equation*}
\frac{1}{2}\!\left(\|{U}\|_{*}+\|{V}\|_{*}\right)\geq\|X\|^{1/2}_{S_{1/2}}.
\end{equation*}
\end{lemma}

\begin{IEEEproof}
Let $U\!=\!L_{U}\Sigma_{U}R^{T}_{U}$ and $V\!=\!L_{V}\Sigma_{V}R^{T}_{V}$ be the thin SVDs of $U$ and $V$, where $L_{U}\!\in\!\mathbb{R}^{m\times d}$, $L_{V}\!\in\!\mathbb{R}^{n\times d}$, and $R_{U},\Sigma_{U},R_{V},\Sigma_{V}\!\in\! \mathbb{R}^{d\times d}$. Let $X\!=\!L_{X}\Sigma_{X}R^{T}_{X}$, where the columns of $L_{X}\!\in\! \mathbb{R}^{m\times d}$ and $R_{X}\!\in\! \mathbb{R}^{n\times d}$ are the left and right singular vectors associated with the top $d$ singular values of $X$ with rank at most $r$ $(r\!\leq\! d)$, and $\Sigma_{X}\!=\!\textup{diag}([\sigma_{1}(X),\!\cdots\!,\sigma_{r}(X),0,\!\cdots\!,0])\!\in\!\mathbb{R}^{d\times d}$. Suppose $W_{1}\!=\!L_{X}\Sigma_{U}L^{T}_{X}$, $W_{2}\!=\!L_{X}\Sigma^{\frac{1}{2}}_{U}\Sigma^{\frac{1}{2}}_{X}L^{T}_{X}$ and $W_{3}\!=\!L_{X}\Sigma_{X}L^{T}_{X}$, we first construct the following PSD matrices $\mathcal{M}_{1}\!\in\!\mathbb{R}^{2m\times 2m}$ and $\mathcal{S}_{1}\!\in\!\mathbb{R}^{2m\times 2m}$:
\begin{equation*}
\mathcal{M}_{1}\!=\!\!\left[\begin{aligned}\!-L_{X}\Sigma^{\frac{1}{2}}_{U}\\\!\!\!L_{X}\Sigma^{\frac{1}{2}}_{X}\!\end{aligned}\right ]\!\!\left[\begin{aligned}-\Sigma^{\frac{1}{2}}_{U}L^{T}_{X}\;\Sigma^{\frac{1}{2}}_{X}L^{T}_{X}\end{aligned}\right ]\!=\!\!\left[
\begin{aligned}
&\;\;W_{1}\quad \!\!-\!W_{2} \\
&\!\!-\!W^{T}_{2}\quad W_{3}
\end{aligned} \right ]\!\succeq\! 0,
\end{equation*}
\begin{equation*}
\!\!\mathcal{S}_{1}\!=\!\!\left[\begin{aligned}\!I_{n}\qquad\\\!L_{U}\!\Sigma^{-\!\frac{1}{2}}_{U}\!L^{T}_{U}\!\end{aligned}\right ]\!\!\left[\begin{aligned}\!I_{n}\;L_{U}\!\Sigma^{-\!\frac{1}{2}}_{U}\!L^{T}_{U}\!\end{aligned}\right ]\!\!=\!\!\!\left [
\begin{aligned}
&\quad\; I_{n} \;\;\;\quad\; L_{U}\!\Sigma^{-\!\frac{1}{2}}_{U}\!L^{T}_{U} \\
&\!L_{U}\!\Sigma^{-\!\frac{1}{2}}_{U}\!L^{T}_{U}\; L_{U}\!\Sigma^{-\!1}_{U}\!L^{T}_{U}\!
\end{aligned} \right ]\!\!\succeq \!\!0.
\end{equation*}

Because the trace of the product of two PSD matrices is always non-negative (see the proof of Lemma 6 in~\cite{mazumder:sr}), we have
\begin{equation*}
\begin{split}
\textrm{Tr} \left(\left [
\begin{aligned}
&\quad\; I_{n} \;\;\;\quad\; L_{U}\!\Sigma^{-\frac{1}{2}}_{U}\!L^{T}_{U} \\
&\!L_{U}\!\Sigma^{-\frac{1}{2}}_{U}\!L^{T}_{U}\; L_{U}\!\Sigma^{-1}_{U}\!L^{T}_{U}
\end{aligned}
\right ]\!\left [
\begin{aligned}
&\;\;W_{1}\quad \!\!-\!W_{2} \\
&\!\!-\!W^{T}_{2}\quad W_{3}
\end{aligned} \right ]
\right )\geq 0.
\end{split}
\end{equation*}
By further simplifying the above expression, we obtain
\begin{equation}\label{equ37}
\textrm{Tr}(W_{1})\!-\!2\textrm{Tr}(L_{U}\!\Sigma^{-\frac{1}{2}}_{U}\!L^{T}_{U}\!W^{T}_{2})\!+\!\textrm{Tr}(L_{U}\!\Sigma^{-1}_{U}\!L^{T}_{U}\!W_{3})\!\geq\!0.
\end{equation}

Recalling $X\!=\!UV^{T}$ and $L_{X}\!\Sigma_{X}\!R^{T}_{X}\!=\!L_{U}\!\Sigma_{U}\!R^{T}_{U}\!R_{V}\!\Sigma_{V}\!L^{T}_{V}$, thus $L^{T}_{U}\!L_{X}\!\Sigma_{X}\!=\!\Sigma_{U}\!R^{T}_{U}\!R_{V}\!\Sigma_{V}\!L^{T}_{V}\!R_{X}$. Due to the orthonormality of the columns of $L_{U}$, $R_{U}$, $L_{V}$, $R_{V}$, $L_{X}$ and $R_{X}$, and using the well-known von Neumann's trace inequality~\cite{mirsky:ti,neumann:ti}, we obtain
\begin{equation}\label{equ38}
\begin{split}
&\,\textrm{Tr}(L_{U}\Sigma^{-1}_{U}L^{T}_{U}W_{3})=\textrm{Tr}(L_{U}\Sigma^{-1}_{U}L^{T}_{U}L_{X}\Sigma_{X}L^{T}_{X})\\
=&\,\textrm{Tr}(L_{U}\Sigma^{-1}_{U}\Sigma_{U}R^{T}_{U}R_{V}\Sigma_{V}L^{T}_{V}R_{X}L^{T}_{X})\\
=&\,\textrm{Tr}(L_{U}R^{T}_{U}R_{V}\Sigma_{V}L^{T}_{V}R_{X}L^{T}_{X})\\
\leq&\,\textrm{Tr}(\Sigma_{V})=\|V\|_{*}.
\end{split}
\end{equation}

Using Lemma~\ref{lem2}, we have
\begin{equation}\label{equ39}
\begin{split}
&\,\textrm{Tr}(\!L_{U}\!\Sigma^{-\frac{1}{2}}_{U}\!L^{T}_{U}\!W^{T}_{2})\!=\!\textrm{Tr}(\!L_{U}\!\Sigma^{-\frac{1}{2}}_{U}\!L^{T}_{U}\!L_{X}\!\Sigma^{\frac{1}{2}}_{U}\!\Sigma^{\frac{1}{2}}_{X}\!L^{T}_{X})\\
=&\,\textrm{Tr}(\Sigma^{-\frac{1}{2}}_{U}L^{T}_{U}L_{X}\Sigma^{\frac{1}{2}}_{U}\Sigma^{\frac{1}{2}}_{X}L^{T}_{X}L_{U})\\
=&\,\textrm{Tr}(\Sigma^{-\frac{1}{2}}_{U}O_{1}\Sigma^{\frac{1}{2}}_{U}\Sigma^{\frac{1}{2}}_{X}O^{T}_{1})\\
\geq&\,\textrm{Tr}(\Sigma^{-\frac{1}{2}}_{U}\Sigma^{\frac{1}{2}}_{U}\Sigma^{\frac{1}{2}}_{X})=\textrm{Tr}(\Sigma^{\frac{1}{2}}_{X})=\|X\|^{1/2}_{S_{1/2}}
\end{split}
\end{equation}
where $O_{1}\!=\!L^{T}_{U}L_{X}\in \mathbb{R}^{d\times d}$, and it is easy to verify that $O_{1}\Sigma^{\frac{1}{2}}_{U}\Sigma^{\frac{1}{2}}_{X}O^{T}_{1}$ is a symmetric PSD matrix. Using \eqref{equ37}, \eqref{equ38}, \eqref{equ39} and $\textrm{Tr}(W_{1})\!=\!\|U\|_{*}$, we have $\|U\|_{*}\!+\!\|V\|_{*}\geq 2\|X\|^{1/2}_{S_{1/2}}$.
\end{IEEEproof}

The detailed proof of Theorem \ref{the1} is provided in the Supplementary Materials. According to Theorem \ref{the1}, it is easy to verify that the double nuclear norm penalty possesses the following property~\cite{shang:snm}.

\begin{property}\label{prop1}
Given a matrix $X\!\in\!\mathbb{R}^{m\times n}$ with $\textrm{rank}(X)\!\leq\! d$, the following equalities hold:
\begin{equation*}
\begin{split}
\|X\|_{\textup{D-N}}&=\min_{U\in\mathbb{R}^{m\!\times\! d},V\in\mathbb{R}^{n\!\times\! d}:X=UV^{T}}\left(\frac{\|U\|_{*}\!+\!\|V\|_{*}}{2}\right)^{2}\\
&=\min_{U,V:X=UV^{T}}\!\|U\|_{*}\|V\|_{*}.
\end{split}
\end{equation*}
\end{property}

\subsection{Frobenius/Nuclear Norm Penalty}
Inspired by the definitions of the bilinear spectral and double nuclear norm penalties mentioned above, we define a Frobenius/nuclear hybrid norm (F-N) penalty as follows.

\begin{definition}\label{def2}
For any matrix $X\!\in\! \mathbb{R}^{m\times n}$ of rank at most $r\leq d$, we decompose it into two factor matrices $U\!\in\!\mathbb{R}^{m\times d}$ and $V\!\in\! \mathbb{R}^{n\times d}$ such that $X\!=\!UV^{T}$. Then the Frobenius/nuclear hybrid norm penalty of $X$ is defined as
\begin{equation}\label{eq3.04}
\|X\|_{\textup{F-N}}=\!\min_{U,V:X=UV^{T}}\,\left[(\|U\|^{2}_{F}\!+\!2\|V\|_{*})/3\right]^{3/2}.
\end{equation}
\end{definition}

Different from the definition in~\cite{shang:snm}, i.e., $\min_{U,V:X=UV^{T}}\!\|U\|_{F}\|V\|_{*}$, Definition~\ref{def2} can also be directly used in practical problems. Analogous to the double nuclear norm penalty, the Frobenius/nuclear hybrid norm penalty is also a quasi-norm, as stated in the following theorem.
\begin{theorem}\label{the2}
The Frobenius/nuclear hybrid norm penalty, $\|\!\cdot\!\|_{\textup{F-N}}$, is a quasi-norm, and is also the Schatten-${2/3}$ quasi-norm, i.e.,
\begin{equation}\label{eq3.05}
\|X\|_{\textup{F-N}}=\|X\|_{S_{2/3}}.
\end{equation}
\end{theorem}

To prove Theorem~\ref{the2}, we first give the following lemma.

\begin{lemma}\label{lem4}
For any matrix $X\!=\!{U}{V}^{T}\!\in\!\mathbb{R}^{m\times n}$, ${U}\!\in\!\mathbb{R}^{m\times d}$ and ${V}\!\in\!\mathbb{R}^{n\times d}$, the following inequality holds:
\begin{equation*}
\frac{1}{3}\!\left(\|{U}\|^{2}_{F}+2\|{V}\|_{*}\right)\geq\|X\|^{2/3}_{S_{2/3}}.
\end{equation*}
\end{lemma}

\begin{IEEEproof}
To prove this lemma, we use the same notations as in the proof of Lemma~\ref{lem3}, e.g., $X\!=\!L_{X}\Sigma_{X}R^{T}_{X}$, $U\!=\!L_{U}\Sigma_{U}R^{T}_{U}$ and $V\!=\!L_{V}\Sigma_{V}R^{T}_{V}$ denote the thin SVDs of $U$ and $V$, respectively. Suppose $\overline{W}_{\!1}\!=\!R_{X}\Sigma_{V}R^{T}_{X}$, $\overline{W}_{\!2}\!=\!R_{X}\Sigma^{\frac{1}{2}}_{V}\Sigma^{\frac{2}{3}}_{X}R^{T}_{X}$ and $\overline{W}_{\!3}\!=\!R_{X}\Sigma^{\frac{4}{3}}_{X}R^{T}_{X}$, we first construct the following PSD matrices $\mathcal{M}_{2}\!\in\!\mathbb{R}^{2m\times 2m}$ and $\mathcal{S}_{2}\!\in\!\mathbb{R}^{2m\times 2m}$:
\begin{equation*}
\mathcal{M}_{2}\!=\!\!\left[\begin{aligned}\!-R_{X}\Sigma^{\frac{1}{2}}_{V}\\\!\!\!R_{X}\Sigma^{\frac{2}{3}}_{X}\!\end{aligned}\right ]\!\!\left[\begin{aligned}-\Sigma^{\frac{1}{2}}_{V}R^{T}_{X}\;\Sigma^{\frac{2}{3}}_{X}R^{T}_{X}\end{aligned}\right ]\!=\!\!\left[
\begin{aligned}
&\;\;\overline{W}_{\!1}\quad \!\!-\!\overline{W}_{\!2} \\
&\!\!-\!\overline{W}^{T}_{\!2}\quad \overline{W}_{\!3}
\end{aligned} \right ]\!\succeq\! 0,
\end{equation*}
\begin{equation*}
\!\!\mathcal{S}_{2}\!=\!\!\left[\begin{aligned}\!I_{n}\qquad\\\!L_{V}\!\Sigma^{-\!\frac{1}{2}}_{V}\!L^{T}_{V}\!\end{aligned}\right ]\!\!\left[\begin{aligned}\!I_{n}\;L_{V}\!\Sigma^{-\!\frac{1}{2}}_{V}\!L^{T}_{V}\!\end{aligned}\right ]\!\!=\!\!\left [
\begin{aligned}
&\quad\; I_{n} \;\;\;\quad\; L_{V}\!\Sigma^{-\!\frac{1}{2}}_{V}\!L^{T}_{V} \\
&\!L_{V}\!\Sigma^{-\!\frac{1}{2}}_{V}\!L^{T}_{V}\; L_{V}\!\Sigma^{-\!1}_{V}\!L^{T}_{V}\!
\end{aligned} \right ]\!\!\succeq \!\!0.
\end{equation*}

Similar to Lemma~\ref{lem3}, we have the following inequality:
\begin{equation*}
\begin{split}
\textrm{Tr} \left(\left [
\begin{aligned}
&\quad\; I_{n} \;\;\;\quad\; L_{V}\!\Sigma^{-\frac{1}{2}}_{V}\!L^{T}_{V} \\
&\!L_{V}\!\Sigma^{-\frac{1}{2}}_{V}\!L^{T}_{V}\; L_{V}\!\Sigma^{-1}_{V}\!L^{T}_{V}
\end{aligned}
\right ]\!\left [
\begin{aligned}
&\;\;\overline{W}_{\!1}\quad \!\!-\!\overline{W}_{\!2} \\
&\!\!-\!\overline{W}^{T}_{\!2}\quad \overline{W}_{\!3}
\end{aligned} \right ]
\right )\geq 0.
\end{split}
\end{equation*}
By further simplifying the above expression, we also obtain
\begin{equation}\label{equ57}
\textrm{Tr}(\overline{W}_{\!1})-2\textrm{Tr}(L_{V}\!\Sigma^{-\frac{1}{2}}_{V}\!L^{T}_{V}\overline{W}^{T}_{\!2})+\textrm{Tr}(L_{V}\!\Sigma^{-1}_{V}\!L^{T}_{V}\overline{W}_{\!3})\geq0.
\end{equation}

Since $L_{X}\Sigma_{X}R^{T}_{X}\!=\!L_{U}\Sigma_{U}R^{T}_{U}R_{V}\Sigma_{V}L^{T}_{V}$, $L^{T}_{V}R_{X}\Sigma_{X}\!=\!(L^{T}_{X}L_{U}\Sigma_{U}R^{T}_{U}R_{V}\Sigma_{V})^{T}\!=\!\Sigma_{V}R^{T}_{V}R_{U}\Sigma_{U}L^{T}_{U}L_{X}$. Due to the orthonormality of the columns of $L_{U}$, $R_{U}$, $L_{V}$, $R_{V}$, $L_{X}$ and $R_{X}$, and using the von Neumann's trace inequality~\cite{mirsky:ti,neumann:ti}, we have
\begin{equation}\label{equ58}
\begin{split}
&\,\textrm{Tr}(L_{V}\Sigma^{-1}_{V}L^{T}_{V}\overline{W}_{\!3})=\textrm{Tr}(L_{V}\Sigma^{-1}_{V}L^{T}_{V}R_{X}\Sigma_{X}\Sigma^{\frac{1}{3}}_{X}R^{T}_{X})\\
=&\,\textrm{Tr}(L_{V}\Sigma^{-1}_{V}\Sigma_{V}R^{T}_{V}R_{U}\Sigma_{U}L^{T}_{U}L_{X}\Sigma^{\frac{1}{3}}_{X}R^{T}_{X})\\
=&\,\textrm{Tr}(R^{T}_{X}L_{V}R^{T}_{V}R_{U}\Sigma_{U}L^{T}_{U}L_{X}\Sigma^{\frac{1}{3}}_{X})\\
\leq&\,\textrm{Tr}(\Sigma_{U}\Sigma^{\frac{1}{3}}_{X})\leq\frac{1}{2}\left(\|U\|^{2}_{F}+\|X\|^{2/3}_{S_{2/3}}\right).
\end{split}
\end{equation}

By Lemma~\ref{lem2}, we also have
\begin{equation}\label{equ59}
\begin{split}
&\,\textrm{Tr}(L_{V}\Sigma^{-\frac{1}{2}}_{V}L^{T}_{V}\overline{W}^{T}_{\!2})\!=\!\textrm{Tr}(L_{V}\Sigma^{-\frac{1}{2}}_{V}L^{T}_{V}R_{X}\Sigma^{\frac{1}{2}}_{V}\Sigma^{\frac{2}{3}}_{X}R^{T}_{X})\\
=&\,\textrm{Tr}(\Sigma^{-\frac{1}{2}}_{V}L^{T}_{V}R_{X}\Sigma^{\frac{1}{2}}_{V}\Sigma^{\frac{2}{3}}_{X}R^{T}_{X}L_{V})\\
=&\,\textrm{Tr}(\Sigma^{-\frac{1}{2}}_{V}O_{2}\Sigma^{\frac{1}{2}}_{V}\Sigma^{\frac{2}{3}}_{X}O^{T}_{2})\\
\geq&\,\textrm{Tr}(\Sigma^{-\frac{1}{2}}_{V}\Sigma^{\frac{1}{2}}_{V}\Sigma^{\frac{2}{3}}_{X})=\textrm{Tr}(\Sigma^{\frac{2}{3}}_{X})=\|X\|^{2/3}_{S_{2/3}}
\end{split}
\end{equation}
where $O_{2}\!=\!L^{T}_{V}R_{X}\!\in\!\mathbb{R}^{d\times d}$, and it is easy to verify that $O_{2}\Sigma^{\frac{1}{2}}_{V}\Sigma^{\frac{2}{3}}_{X}O^{T}_{2}$ is a symmetric PSD matrix. Using \eqref{equ57}, \eqref{equ58}, \eqref{equ59}, and $\textrm{Tr}(\overline{W}_{\!1})\!=\!\|V\|_{*}$, then we have $\|U\|^{2}_{F}\!+\!2\|V\|_{*}\geq 3\|X\|^{2/3}_{S_{2/3}}.$
\end{IEEEproof}

According to Theorem~\ref{the2} (see the Supplementary Materials for its detailed proof), it is easy to verify that the Frobenius/nuclear hybrid norm penalty possesses the following property~\cite{shang:snm}.
\begin{property}\label{prop2}
For any matrix $X\!\in\! \mathbb{R}^{m\times n}$ with $\textrm{rank}(X)=r\leq d$, the following equalities hold:
\begin{equation*}
\begin{split}
\|X\|_{\textup{F-N}}=&\min_{U\in\mathbb{R}^{m\!\times\! d},V\in\mathbb{R}^{n\!\times\! d}:X=UV^{T}}\left(\frac{\|U\|^{2}_{F}+2\|V\|_{*}}{3}\right)^{3/2}\\
=&\min_{U,V:X=UV^{T}}\|U\|_{F}\|V\|_{*}.
\end{split}
\end{equation*}
\end{property}

Similar to the relationship between the Frobenius norm and nuclear norm, i.e., $\|X\|_{F}\!\leq\!\|X\|_{*}\!\leq\! \textup{rank}(X)\|X\|_{F}$~\cite{recht:nnm}, the bounds hold for between both the double nuclear norm and Frobenius/nuclear hybrid norm penalties and the nuclear norm, as stated in the following property.

\begin{property}\label{prop3}
For any matrix $X\in \mathbb{R}^{m\times n}$, the following inequalities hold:
\begin{displaymath}
\|X\|_{*}\leq\|X\|_{\textup{F-N}}\leq\|X\|_{\textup{D-N}}\leq \textup{rank}(X)\|X\|_{*}.
\end{displaymath}
\end{property}

The proof of Property~\ref{prop3} is similar to that in~\cite{shang:snm}. Moreover, both the double nuclear norm and Frobenius/nuclear hybrid norm penalties naturally satisfy many properties of quasi-norms, e.g., the unitary-invariant property. Obviously, we can find that Property~\ref{prop3} in turn implies that any low F-N or D-N penalty is also a low nuclear norm approximation.

\subsection{Problem Formulations}
Without loss of generality, we can assume that the unknown entries of $D$ are set to zero (i.e., $\mathcal{P}_{\Omega^{c}}(D)\!=\!\textbf{0}$), and $S_{\Omega^{c}}$ may be any values\footnote{Considering the optimal solution $S_{\Omega^{c}}\!=\!\textbf{0}$, $S_{\Omega^{c}}$ must be set to 0 for the expected output $S$.} (i.e., $S_{\Omega^{c}}\!\in\!(-\infty,+\infty)$) such that $\mathcal{P}_{\Omega^{c}}(L\!+\!S)\!=\!\mathcal{P}_{\Omega^{c}}(D)$. Thus, the constraint with the projection operator $\mathcal{P}_{\Omega}$ in~\eqref{eq2.02} is considered instead of just $L\!+\!S\!=\!D$ as in~\cite{shang:rbf}. Together with hyper-Laplacian priors of sparse components, we use $\|L\|^{1/2}_{\textup{D-N}}$ and $\|L\|^{2/3}_{\textup{F-N}}$ defined above to replace $\|L\|^{q}_{\!S_{q}}$ in (\ref{eq2.02}), and present the following double nuclear norm and Frobenius/nuclear hybrid norm penalized RPCA models:
\begin{equation}\label{eq3.06}
\begin{split}
&\min_{U,V,L,S} \frac{\lambda}{2}\!\left(\|U\|_{*}+\|V\|_{*}\right)+\|\mathcal{P}_{\Omega}(S)\|^{1/2}_{\ell_{{1}/{2}}},\\
&\;\;\;\textup{s.t.},\,UV^{T}\!=L, L+S=\!D
\end{split}
\vspace{-2mm}
\end{equation}
\begin{equation}\label{eq3.07}
\begin{split}
&\min_{U,V,L,S} \frac{\lambda}{3}\!\left(\|U\|^{2}_{F}+2\|V\|_{*}\right)+\|\mathcal{P}_{\Omega}(S)\|^{2/3}_{\ell_{{2}/{3}}},\\
&\;\;\;\textup{s.t.},\,UV^{T}\!=L, L+S=\!D.
\end{split}
\end{equation}

From the two proposed models~\eqref{eq3.06} and \eqref{eq3.07}, one can easily see that the norm of each bilinear factor matrix is convex, and they are much more tractable and scalable optimization problems than the original Schatten quasi-norm minimization problem as in (\ref{eq2.02}).

\section{Optimization Algorithms}
\label{algo}
To efficiently solve both our challenging problems~\eqref{eq3.06} and \eqref{eq3.07}, we need to introduce the auxiliary variables $\widehat{U}$ and $\widehat{V}$, or only $\widehat{V}$ to split the interdependent terms such that they can be solved independently. Thus, we can reformulate Problems~\eqref{eq3.06} and \eqref{eq3.07} into the following equivalent forms:
\begin{equation}\label{eq4:02}
\begin{split}
&\min_{U,V,L,S,\widehat{U},\widehat{V}} \frac{\lambda}{2}\!\left(\|\widehat{U}\|_{*}\!+\!\|\widehat{V}\|_{*}\right)+\|\mathcal{P}_{\Omega}(S)\|^{1/2}_{\ell_{{1}/{2}}},\\
&\quad\;\;\textup{s.t.},\,\widehat{U}=U,\, \widehat{V}=V,\, UV^{T}\!=L,\, L+S=\!D
\end{split}
\end{equation}
\vspace{-2mm}
\begin{equation}\label{eq4:03}
\begin{split}
&\min_{U,V,L,S,\widehat{V}} \frac{\lambda}{3}\!\left(\|U\|^{2}_{F}\!+\!2\|\widehat{V}\|_{*}\right)+\|\mathcal{P}_{\Omega}(S)\|^{2/3}_{\ell_{{2}/{3}}},\\
&\quad\;\textup{s.t.},\,\widehat{V}=V,\, UV^{T}\!=L,\, L+S=\!D.
\end{split}
\end{equation}

\subsection{Solving \eqref{eq4:02} via ADMM}
Inspired by recent progress on alternating direction methods~\cite{lin:almm,boyd:admm}, we mainly propose an efficient algorithm based on the alternating direction method of multipliers~\cite{boyd:admm} (ADMM, also known as the inexact ALM~\cite{lin:almm}) to solve the more complex problem \eqref{eq4:02}, whose augmented Lagrangian function is given by
\vspace{-1mm}
\begin{equation*}
\begin{split}
&\;\;\,\mathcal{L}_{\mu}(U,V,L,S,\widehat{U},\widehat{V},\{Y_{i}\})\!=\!\frac{\lambda}{2}(\|\widehat{U}\|_{*}\!+\!\|\widehat{V}\|_{*})\!+\!\|\mathcal{P}_{\Omega}(S)\|^{\frac{1}{2}}_{\ell_{\!\frac{1}{2}}}\\
&+\!\langle Y_{1},\widehat{U}\!-\!U\rangle \!+\!\langle Y_{2},\widehat{V}\!-\!V\rangle \!+\!\langle Y_{3},UV^{T}\!\!-\!L\rangle \!+\langle Y_{4}, L\!+\!S\!-\!D\rangle\\
&\;+\!\frac{\mu}{2}\!\left(\|\widehat{U}\!-\!U\|^{2}_{F}\!+\!\|\widehat{V}\!-\!V\|^{2}_{F}\!+\!\|UV^{T}\!\!-\!L\|^{2}_{F}\!+\!\|L\!+\!S\!-\!D\|^{2}_{F}\right)
\end{split}
\end{equation*}
where $\mu\!>\!0$ is the penalty parameter, $\langle\cdot,\cdot\rangle$ represents the inner product operator, and $Y_{1}\!\in\! \mathbb{R}^{m\times d}$, $Y_{2}\!\in\! \mathbb{R}^{n\times d}$ and $Y_{3},Y_{4}\!\in\!\mathbb{R}^{m\times n}$ are Lagrange multipliers.

\subsubsection{Updating $U_{k+\!1}$ and $V_{k+\!1}$}
To update ${U}_{k+\!1}$ and ${V}_{k+\!1}$, we consider the following optimization problems:
\begin{equation}\label{eq4:04}
\min_{U}\|\widehat{U}_{k}\!-\!{U}\!+\!\mu^{-\!1}_{k}Y^{k}_{1}\|^{2}_{F}+\|UV^{T}_{k}\!-L_{k}\!+\!\mu^{-\!1}_{k}Y^{k}_{3}\|^{2}_{F},\;
\end{equation}
\vspace{-2mm}
\begin{equation}\label{eq4:05}
\min_{V}\|\widehat{V}_{k}\!-\!{V}\!+\!\mu^{-\!1}_{k}Y^{k}_{2}\|^{2}_{F}+\|U_{k\!+\!1}\!V^{T}\!-L_{k}\!+\!\mu^{-\!1}_{k}Y^{k}_{3}\|^{2}_{F}.
\end{equation}
Both \eqref{eq4:04} and \eqref{eq4:05} are least squares problems, and their optimal solutions are given by
\begin{equation}\label{eq4:06}
U_{k+\!1}=(\widehat{U}_{k}+\mu^{-\!1}_{k}Y^{k}_{1}+M_{k}V_{k})(I_{d}\!+\!V^{T}_{k}V_{k})^{-\!1},\,\;
\end{equation}
\vspace{-2mm}
\begin{equation}\label{eq4:07}
V_{k+\!1}=(\widehat{V}_{k}\!+\!\mu^{-\!1}_{k}Y^{k}_{2}\!+\!M^{T}_{k}U_{k+\!1})(I_{d}\!+\!U^{T}_{k+\!1}U_{k+\!1})^{-\!1}
\end{equation}
where $M_{k}\!=\!L_{k}-\mu^{-\!1}_{k}Y^{k}_{3}$, and $I_{d}$ denotes an identity matrix of size $d\times d$.

\begin{algorithm}[t]
\caption{ADMM for solving $(\textup{S+L})_{1/2}$ problem \eqref{eq4:02}}
\label{alg1}
\renewcommand{\algorithmicrequire}{\textbf{Input:}}
\renewcommand{\algorithmicensure}{\textbf{Initialize:}}
\renewcommand{\algorithmicoutput}{\textbf{Output:}}
\begin{algorithmic}[1]
\REQUIRE $\mathcal{P}_{\Omega}(D)\!\in\!\mathbb{R}^{m\times n}$, the given rank $d$, and $\lambda$.
\ENSURE $\mu_{0}$, $\rho\!>\!1$, $k\!=\!0$, and $\epsilon$.\\
\WHILE {not converged}
\WHILE {not converged}
\STATE {Update $U_{k+\!1}$ and $V_{k+\!1}$ by \eqref{eq4:06} and \eqref{eq4:07}.}
\STATE {Compute $\widehat{U}_{k+\!1}$ and $\widehat{V}_{k+\!1}$ via the SVT operator~\cite{cai:svt}.}
\STATE {Update $L_{k+\!1}$ and $S_{k+\!1}$ by \eqref{eq4:11} and \eqref{eq4:14}.}
\ENDWHILE \;\,$\verb|//|$ Inner loop
\STATE {Update the multipliers by\\
$\quad Y^{k+\!1}_{1}\!\!=\!Y^{k}_{1}\!+\!\mu_{k}\!(\widehat{U}_{\!k\!+\!1}\!-\!U_{\!k\!+\!1}\!),$ $Y^{k+\!1}_{2}\!\!=\!Y^{k}_{2}\!+\!\mu_{k}\!(\widehat{V}_{\!k\!+\!1}\!-\!V_{\!k\!+\!1}\!),$\\
$\quad Y^{k+\!1}_{3}\!=\!Y^{k}_{3}\!+\!\mu_{k}(U_{\!k+\!1}\!V^{T}_{\!k+\!1}\!-\!L_{k+\!1}),$\\
$\quad Y^{k+\!1}_{4}\!=\!Y^{k}_{4}\!+\!\mu_{k}(L_{k+\!1}+S_{k+\!1}\!-D)$.}
\STATE {Update $\mu_{k+\!1}$ by $\mu_{k+\!1}=\rho\mu_{k}$.}
\STATE $k\leftarrow k+1$.
\ENDWHILE \;\,$\verb|//|$ Outer loop
\OUTPUT $U_{k+\!1}$ and $V_{k+\!1}$.
\end{algorithmic}
\end{algorithm}

\subsubsection{Updating $\widehat{U}_{k+\!1}$ and $\widehat{V}_{k+\!1}$}
To solve $\widehat{U}_{k+\!1}$ and $\widehat{V}_{k+\!1}$, we fix the other variables and solve the following optimization problems
\begin{equation}\label{eq4:08}
\min_{\widehat{U}}\frac{\lambda}{2}\|\widehat{U}\|_{*}+\frac{\mu_{k}}{2}\|\widehat{U}-U_{k+\!1}+Y^{k}_{1}/\mu_{k}\|^{2}_{F},
\end{equation}
\vspace{-2mm}
\begin{equation}\label{eq4:09}
\min_{\widehat{V}}\frac{\lambda}{2}\|\widehat{V}\|_{*}+\frac{\mu_{k}}{2}\|\widehat{V}-V_{k+\!1}+Y^{k}_{2}/\mu_{k}\|^{2}_{F}.
\end{equation}

Both \eqref{eq4:08} and \eqref{eq4:09} are nuclear norm regularized least squares problems, and their closed-form solutions can be given by the so-called SVT operator~\cite{cai:svt}, respectively.

\subsubsection{Updating $L_{k+\!1}$}
To update $L_{k+\!1}$, we can obtain the following optimization problem:
\begin{equation}\label{eq4:10}
\min_{L} \|U_{k+\!1}\!V^{T}_{k+\!1}\!-\!L\!+\!\mu^{\!-\!1}_{k}Y^{k}_{3}\|^{2}_{F}\!+\!\|L\!+\!S_{k}\!-\!D\!+\!\mu^{\!-\!1}_{k}Y^{k}_{4}\|^{2}_{F}.
\end{equation}
Since \eqref{eq4:10} is a least squares problem, and thus its closed-form solution is given by
\begin{equation}\label{eq4:11}
L_{k+\!1}=\frac{1}{2}\!\left(U_{k+\!1}\!V^{T}_{k+\!1}+\mu^{\!-\!1}_{k}Y^{k}_{3}-S_{k}+D-\mu^{\!-\!1}_{k}Y^{k}_{4}\right).
\end{equation}

\subsubsection{Updating $S_{k+\!1}$}
By keeping all other variables fixed, $S_{k+\!1}$ can be updated by solving the following problem
\begin{equation}\label{eq4:12}
\min_{S}\|\mathcal{P}_{\Omega}(S)\|^{1/2}_{\ell_{{1}/{2}}}+\frac{\mu_{k}}{2}\|S\!+\!L_{k+1}\!-\!D\!+\!\mu^{\!-\!1}_{k}Y^{k}_{4}\|^{2}_{F}.
\end{equation}

Generally, the $\ell_{p}$-norm ($0\!<\!p\!<\!1$) leads to a non-convex, non-smooth, and non-Lipschitz optimization problem~\cite{bian:ipa}. Fortunately, we can efficiently solve \eqref{eq4:12} by introducing the following half-thresholding operator~\cite{xu:l12r}.

\begin{proposition}\label{the3}
For any matrix $A\!\in\!\mathbb{R}^{m\times n}$, and $X^{*}\!\in\!\mathbb{R}^{m\times n}$ is an $\ell_{1/2}$ quasi-norm solution of the following minimization
\begin{equation}\label{eq4:13}
\min_{X}\,\|X-A\|^{2}_{F}+\gamma\|X\|^{1/2}_{\ell_{{1}/{2}}},
\end{equation}
then the solution $X^{*}$ can be given by $X^{*}\!=\!\mathcal{H}_{\gamma}(A)$, where the half-thresholding operator $\mathcal{H}_{\gamma}(\cdot)$ is defined as
\begin{equation*}
\mathcal{H}_{\gamma}(a_{ij})=\left\{ \begin{array}{ll}
\!\!\frac{2}{3}a_{ij}[1\!+\!\cos(\frac{2\pi-2\phi_{\gamma}(a_{ij})}{3})], \!\!&|a_{ij}|\!>\! \frac{\sqrt[3]{54\gamma^{2}}}{4},\\
\!\!0, \!\!&\textup{otherwise},
\end{array}\right.
\end{equation*}
where $\phi_{\gamma}(a_{ij})\!=\!\arccos(\frac{\gamma}{8}({|a_{ij}|}/{3})^{-{3}/{2}})$.
\end{proposition}

Before giving the proof of Proposition~\ref{the3}, we first give the following lemma~\cite{xu:l12r}.

\begin{lemma}\label{lem5}
Let $y\!\in\!\mathbb{R}^{l\times 1}$ be a given vector, and $\tau\!>\!0$. Suppose that $x^{*}\!\in\! \mathbb{R}^{l\times 1}$ is a solution of the following problem,
\begin{equation*}
\min_{x} \|Bx-y\|^{2}_{2}+\tau\|x\|^{1/2}_{\ell_{{1}/{2}}}.
\end{equation*}
Then for any real parameter $\mu\!\in\!(0,\infty)$, $x^{*}$ can be expressed as $x^{*}\!=\!\mathcal{H}_{\tau\mu}(\varphi_{\mu}(x^{*}))$, where
$\varphi_{\mu}(x^{*})=x^{*}+\mu B^{T}(y\!-\!Bx^{*})$.
\end{lemma}

\begin{IEEEproof}
The formulation \eqref{eq4:13} can be reformulated as the following equivalent form:
\begin{equation*}
\min_{\textup{vec}(X)}\,\|\textup{vec}(X)-\textup{vec}(A)\|^{2}_{2}+\gamma\|\textup{vec}(X)\|^{1/2}_{\ell_{{1}/{2}}}.
\end{equation*}
Let $B\!=\!I_{mn}$, $\mu\!=\!1$, and using Lemma~\ref{lem5}, the closed-form solution of~\eqref{eq4:13} is given by $\textup{vec}(X^{*})\!=\!\mathcal{H}_{\gamma}(\textup{vec}(A))$.
\end{IEEEproof}

Using Proposition~\ref{the3}, the closed-form solution of~\eqref{eq4:12} is
\begin{equation}\label{eq4:14}
\begin{split}
S_{k+\!1}=&\,\mathcal{P}_{\Omega}\!\left(\mathcal{H}_{2/\mu_{k}}\!(D-L_{k+\!1}-\mu^{-\!1}_{k}Y^{k}_{4})\right)\\
&+\mathcal{P}^{\perp}_{\Omega}\!\left(D-L_{k+\!1}-\mu^{-\!1}_{k}Y^{k}_{4}\right)
\end{split}
\end{equation}
where $\mathcal{P}^{\perp}_{\Omega}$ is the complementary operator of $\mathcal{P}_{\Omega}$. Alternatively, Zuo \emph{et al.}~\cite{zuo:gist} proposed a generalized shrinkage-thresholding operator to iteratively solve $\ell_{p}$-norm minimization with arbitrary $p$ values, i.e., $0\!\leq\! p\!<\!1$, and achieve a higher efficiency.

Based on the description above, we develop an efficient ADMM algorithm to solve the double nuclear norm penalized problem~\eqref{eq4:02}, as outlined in \textbf{Algorithm}~\ref{alg1}. To further accelerate the convergence of the algorithm, the penalty parameter $\mu$, as well as $\rho$, are adaptively updated by the strategy as in~\cite{lin:almm}. The varying $\mu$, together with shrinkage-thresholding operators such as the SVT operator, sometimes play the role of adaptive selection on the rank of matrices or the number of non-zeros elements. We found that updating $\{U_{k},V_{k}\}$, $\{\widehat{U}_{k},\widehat{V}_{k}\}$, $L_{k}$ and $S_{k}$ just once in the inner loop is sufficient to generate a satisfying accurate solution of~\eqref{eq4:02}, so also called inexact ALM, which is used for computational efficiency. In addition, we initialize all the elements of the Lagrange multipliers $Y_{1}$, $Y_{2}$ and $Y_{3}$ to 0, while all elements in $Y_{4}$ are initialized by the same way as in~\cite{lin:almm}.

\subsection{Solving \eqref{eq4:03} via ADMM}
Similar to Algorithm~\ref{alg1}, we also propose an efficient ADMM algorithm to solve~\eqref{eq4:03} (i.e., Algorithm 2), and provide the details in the Supplementary Materials. Since the update schemes of $U_{k}$, $V_{k}$, $\widehat{V}_{k}$ and $L_{k}$ are very similar to that of Algorithm~\ref{alg1}, we discuss their major differences below.

By keeping all other variables fixed, $S_{k+\!1}$ can be updated by solving the following problem
\begin{equation}\label{eq4:20}
\min_{S}\|\mathcal{P}_{\Omega}(S)\|^{2/3}_{\ell_{{2}/{3}}}+\frac{\mu_{k}}{2}\|S\!+\!L_{k+\!1}\!-\!D\!+\!\mu^{-\!1}_{k}Y^{k}_{3}\|^{2}_{F}.
\end{equation}
Inspired by~\cite{krishnan:l12,zuo:gist,cao:st}, we introduce the following two-thirds-thresholding operator to efficiently solve~\eqref{eq4:20}.

\begin{proposition}\label{the4}
For any matrix $C\!\in\!\mathbb{R}^{m\times n}$, and $X^{*}\!\in\!\mathbb{R}^{m\times n}$ is an $\ell_{2/3}$ quasi-norm solution of the following minimization
\begin{equation}\label{eq4:21}
\min_{X}\,\|X-C\|^{2}_{F}+\gamma\|X\|^{2/3}_{\ell_{2/3}},
\end{equation}
then the solution $X^{*}$ can be given by $X^{*}\!=\!\mathcal{T}_{\gamma}(C)$, where the two-thirds-thresholding operator $\mathcal{T}_{\gamma}(\cdot)$ is defined as
\begin{equation*}
\mathcal{T}_{\gamma}(c_{i\!j})\!=\!\left\{ \begin{array}{ll}
\!\!\!\frac{\textup{sgn}(c_{i\!j}\!)\!\left(\!\psi_{\gamma}\!(c_{i\!j}\!)\!+\!\sqrt{\frac{2|c_{\!i\!j}\!|}{\psi_{\!\gamma}\!(c_{\!i\!j}\!)}-\psi^{2}_{\gamma}\!(c_{i\!j}\!)}\right)^{\!\!3}}{8}, \!\!\!\!&|c_{i\!j}|\!>\!\!\frac{2\sqrt[4]{3\gamma^{3}}}{3},\\
\!\!\!0, \!\!\!\!&\textup{otherwise},
\end{array}\right.
\end{equation*}
where $\psi_{\gamma}(c_{ij})\!=\!\!\frac{2}{\sqrt{3}}\sqrt{\sqrt{\gamma}\cosh(\textup{arccosh}(\frac{27c^{2}_{ij}}{16}\gamma^{-3/2})/3)}$, and $\textup{sgn}(\cdot)$ is the sign function.
\end{proposition}

Before giving the proof of Proposition~\ref{the4}, we first give the following lemma~\cite{zuo:gist,cao:st}.

\begin{lemma}\label{lem6}
Let $y\!\in\! \mathbb{R}$ be a given real number, and $\tau\!>\!0$. Suppose that $x^{*}\!\in\! \mathbb{R}$ is a solution of the following problem,
\begin{equation}\label{eq4:22}
\min_{x}\,x^{2}-2xy+\tau|x|^{2/3}.
\end{equation}
Then $x^{*}$ has the following closed-form thresholding formula:
\begin{displaymath}
x^{*}=\left\{ \begin{array}{ll}
\!\!\!\frac{\textup{sgn}(y)\!\left(\psi_{\tau}\!(y)+\sqrt{\frac{2|y|}{\psi_{\tau}\!(y)}-\psi^{2}_{\tau}\!(y)}\right)^{\!3}}{8}, \!\!&\!\!|y|\!>\!\frac{2\sqrt[4]{3\tau^{3}}}{3},\\
\!\!\!0, \!\!\!\!&\textup{otherwise}.
\end{array}\right.
\end{displaymath}
\end{lemma}

\begin{IEEEproof}
It is clear that the operator in Lemma~\ref{lem6} can be extended to vectors and matrices by applying it element-wise. Using Lemma~\ref{lem6}, the closed-form thresholding formula of~\eqref{eq4:21} is given by $X^{*}=\mathcal{T}_{\gamma}(C)$.
\end{IEEEproof}

By Proposition~\ref{the4}, the closed-form solution of~\eqref{eq4:20} is
\begin{equation}\label{eq4:23}
\begin{split}
S_{k+\!1}=&\,\mathcal{P}_{\Omega}\!\left(\mathcal{T}_{2/\mu_{k}}\!(D-L_{k+\!1}-\mu^{-\!1}_{k}Y^{k}_{3})\right)\\
&+\mathcal{P}^{\perp}_{\Omega}\!\left(D-L_{k+\!1}-\mu^{-\!1}_{k}Y^{k}_{3}\right).
\end{split}
\end{equation}

\section{Algorithm Analysis}
\label{anal}
We mainly analyze the convergence property of Algorithm~\ref{alg1}. Naturally, the convergence of Algorithm 2 can also be guaranteed in a similar way. Moreover, we also analyze their per-iteration complexity.

\subsection{Convergence Analysis}
Before analyzing the convergence of Algorithm~\ref{alg1}, we first introduce the definition of the critical points (or stationary points) of a non-convex function given in~\cite{bolte:palm}.

\begin{definition}
Given a function $f\!:\!\mathbb{R}^{n}\!\rightarrow\!(-\infty, +\infty]$, we denote the domain of $f$ by $\textup{dom}f$, i.e., $\textup{dom}f \!:=\!\{x\!\in\! \mathbb{R}^{n}\!:\! f(x)\!\leq\!+\infty\}$. A function $f$ is said to be proper if $\textup{dom}f\!\neq\!\emptyset$; lower semicontinuous at the point $x_{0}$ if $\lim_{x\rightarrow x_{0}}\inf f(x)\geq f(x_{0})$. If $f$ is lower semicontinuous at every point of its domain, then it is called a lower semicontinuous function.
\end{definition}

\begin{definition}
Let a non-convex function $f\!:\!\mathbb{R}^{n}\!\rightarrow\!(-\infty, +\infty]$ be a proper and lower semi-continuous function. $x$ is a critical point of $f$ if $0\!\in\!\partial f(x)$, where $\partial f(x)$ is the limiting sub-differential of $f$ at $x$, i.e., $\partial f(x)\!=\!\{u\!\in\!\mathbb{R}^{n}\!:\exists x^{k}\!\rightarrow\! x,\; f(x^{k})\!\rightarrow\! f(x)$ and $u^{k}\!\in\!\widehat{\partial}f(x^{k})\!\rightarrow\! u\;\textup{as}\;k\!\rightarrow\!\infty\}$, and $\widehat{\partial}f(x^{k})$ is the Fr\`{e}chet sub-differential of $f$ at $x$ (see \cite{bolte:palm} for more details).
\end{definition}

As stated in~\cite{oh:psvt,hong:admm}, the general convergence property of the ADMM for non-convex problems has not been answered yet, especially for multi-block cases~\cite{hong:admm,chen:admm}. For such challenging problems \eqref{eq3.06} and \eqref{eq3.07}, although it is difficult to guarantee the convergence to a local minimum, our empirical convergence tests showed that our algorithms have strong convergence behavior (see the Supplementary Materials for details). Besides the empirical behavior, we also provide the convergence property for Algorithm~\ref{alg1} in the following theorem.

\begin{theorem}\label{the51}
Let $\{(U_{k},V_{k},\widehat{U}_{k},\widehat{V}_{k},L_{k},S_{k},\{Y^{k}_{i}\})\}$ be the sequence generated by Algorithm~\ref{alg1}. Suppose that the sequence $\{Y^{k}_{3}\}$ is bounded, and $\mu_{k}$ is non-decreasing and $\sum^{\infty}_{k=0}({\mu_{k+\!1}}/{\mu^{4/3}_{k}})\!<\!\infty$, then
\begin{enumerate}[(I)]
\item {$\{(U_{k},V_{k})\}$, $\{(\widehat{U}_{k},\widehat{V}_{k})\}$, $\{L_{k}\}$ and $\{S_{k}\}$ are all Cauchy sequences;}
\item Any accumulation point of the sequence $\{(U_{k}, V_{k}, \widehat{U}_{k},\widehat{V}_{k}, L_{k}, S_{k})\}$ satisfies the Karush-Kuhn-Tucker (KKT) conditions for Problem~\eqref{eq4:02}.
\end{enumerate}
\end{theorem}

The proof of Theorem~\ref{the51} can be found in the Supplementary Materials. Theorem~\ref{the51} shows that under mild conditions, any accumulation point (or limit point) of the sequence generated by Algorithm~\ref{alg1} is a critical point of the Lagrangian function $\mathcal{L}_{\mu}$, i.e., $(U_{\star},V_{\star},\widehat{U}_{\star},\widehat{V}_{\star},L_{\star},S_{\star},\{Y^{\star}_{i}\})$, which satisfies the first-order optimality conditions (i.e., the KKT conditions) of \eqref{eq4:02}: $0\!\in\! \frac{\lambda}{2}\partial \|\widehat{U}_{\star}\|_{*}\!+\!Y^{\star}_{1}$, $0\!\in\! \frac{\lambda}{2}\partial\|\widehat{V}_{\star}\|_{*}\!+\!Y^{\star}_{2}$, $0\!\in\! \partial\Phi(S_{\star})\!+\!\mathcal{P}_{\Omega}(Y^{\star}_{4})$, $\mathcal{P}_{\Omega^{c}}(Y^{\star}_{4})\!=\!\textbf{0}$, $L_{\star}\!=\!U_{\star}V^{T}_{\star}$, $\widehat{U}_{\star}\!=\!U_{\star}$, $\widehat{V}_{\star}\!=\!V_{\star}$, and $L_{\star}\!+\!S_{\star}\!=\!D$, where $\Phi(S)\!:=\!\|\mathcal{P}_{\Omega}(S)\|^{1/2}_{\ell_{1/2}}$. Similarly, the convergence of Algorithm 2 can also be guaranteed. Theorem~\ref{the51} is established for the proposed ADMM algorithm, which has only a single iteration in the inner loop. When the inner-loop iterations of Algorithm~\ref{alg1} iterate until convergence, it may lead to a simpler proof. We leave further theoretical analysis of convergence as future work.

Theorem~\ref{the51} also shows that our ADMM algorithms have much weaker convergence conditions than the ones in~\cite{kim:factEN,oh:psvt}, e.g., the sequence of only one Lagrange multiplier is required to be bounded for our algorithms, while the ADMM algorithms in~\cite{kim:factEN,oh:psvt} require the sequences of all Lagrange multipliers to be bounded.

\subsection{Convergence Behavior}
According to the KKT conditions of \eqref{eq4:02} mentioned above, we take the following conditions as the stopping criterion for our algorithms (see details in Supplementary Materials),
\begin{equation*}
\max\{{\epsilon_{1}}/{\|D\|_{F}},\,\epsilon_{2}\}<\epsilon
\end{equation*}
where $\epsilon_{1}\!=\!\max\{\|U_{k}\!V^{T}_{k}\!-\!L_{k}\|_{F},\|L_{k}\!+\!S_{k}\!-\!D\|_{F},\|Y^{k}_{1}(\widehat{V}_{k})^{\dagger}\!-\!(\widehat{U}^{T}_{k})^{\dagger}(Y^{k}_{2})^{T}\|_{F}\}$, $\epsilon_{2}\!=\!\max\{\|\widehat{U}_{k}\!-\!U_{k}\|_{F}/\|U_{k}\|_{F},\|\widehat{V}_{k}\!-\!V_{k}\|_{F}/\|V_{k}\|_{F}\}$, $(\widehat{V}_{k})^{\dagger}$ is the pseudo-inverse of $\widehat{V}_{k}$, and $\varepsilon$ is the stopping tolerance. In this paper, we set the stopping tolerance to $\epsilon\!=\!10^{-5}$ for synthetic data and $\epsilon\!=\!10^{-4}$ for real-world problems. As shown in the Supplementary Materials, the stopping tolerance and relative squared error (RSE) of our methods decrease fast, and they converge within only a small number of iterations (usually within 50 iterations).

\begin{figure}[t]
\centering
\includegraphics[width=0.489\linewidth]{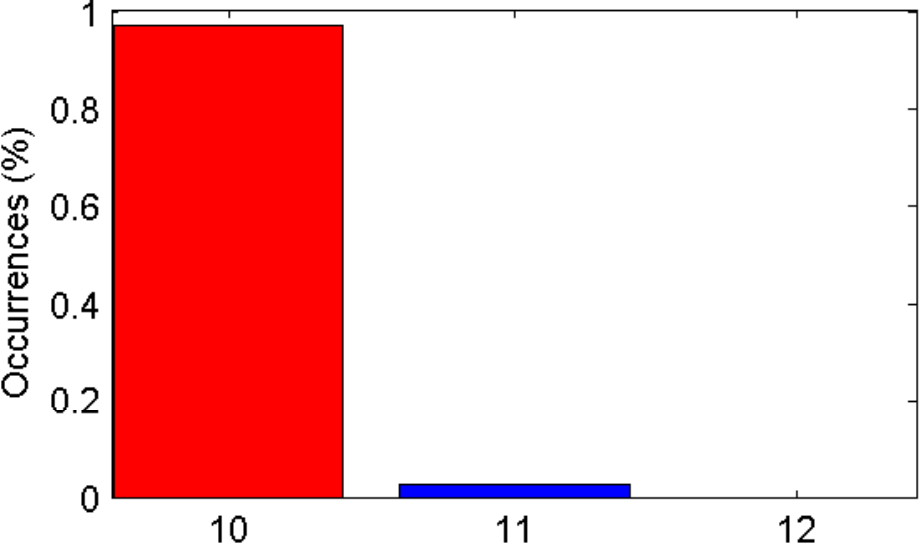}\;
\includegraphics[width=0.489\linewidth]{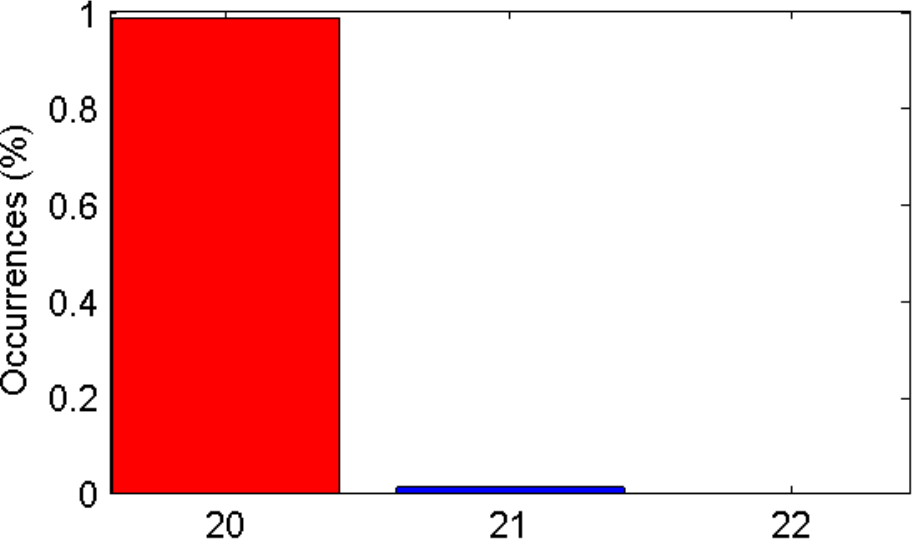}
\caption{Histograms with the percentage of estimated ranks for Gaussian noise and outlier corrupted matrices of size $500\!\times\!500$ (left) and $1,000\!\times\!1,000$ (right), whose true ranks are 10 and 20, respectively.}
\label{fig3}
\end{figure}

\begin{figure*}[t]
\centering
\includegraphics[width=0.1471\linewidth]{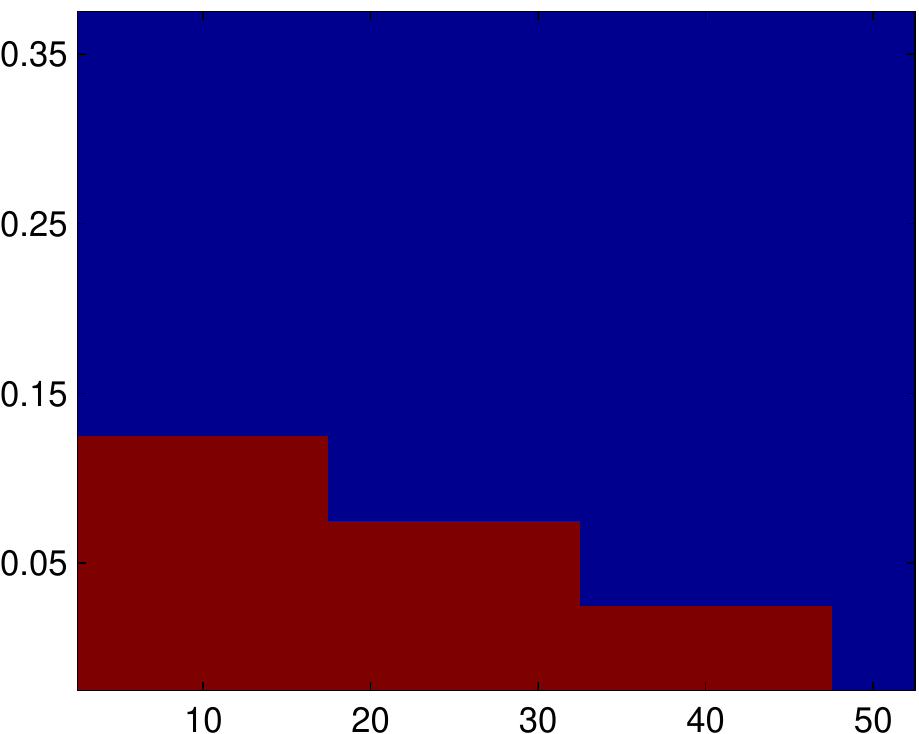}\;\!\!
\includegraphics[width=0.134\linewidth]{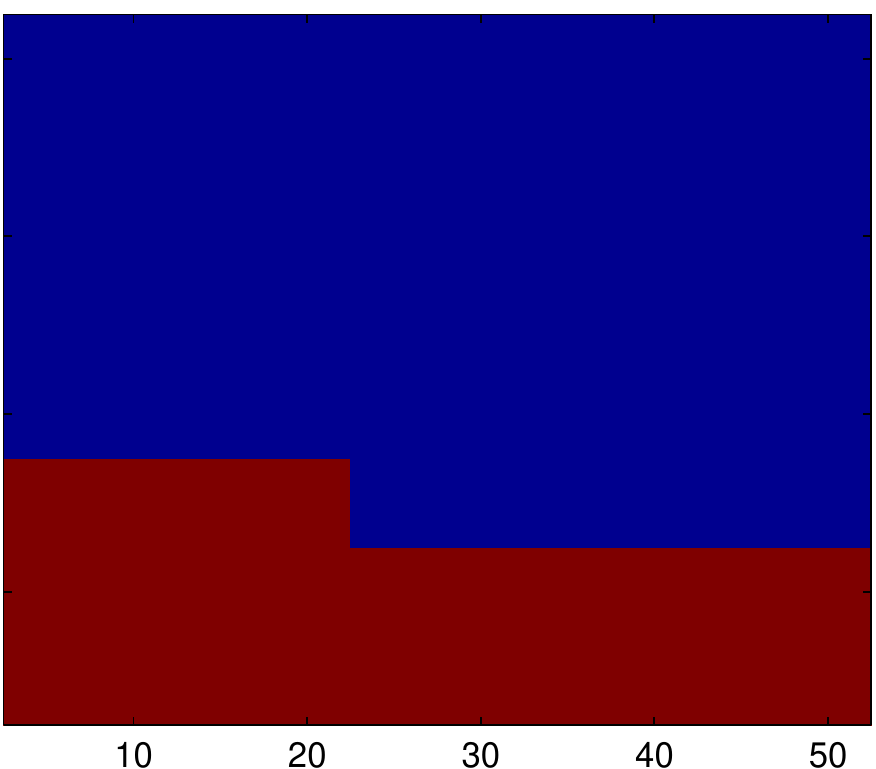}\;\!\!
\includegraphics[width=0.134\linewidth]{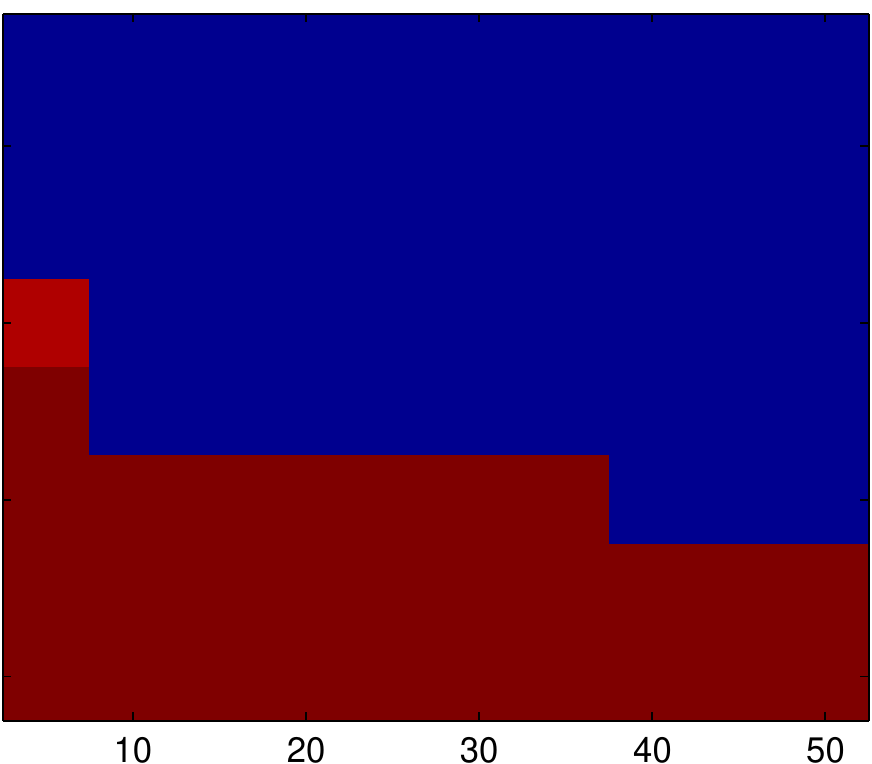}\;\!\!
\includegraphics[width=0.1345\linewidth]{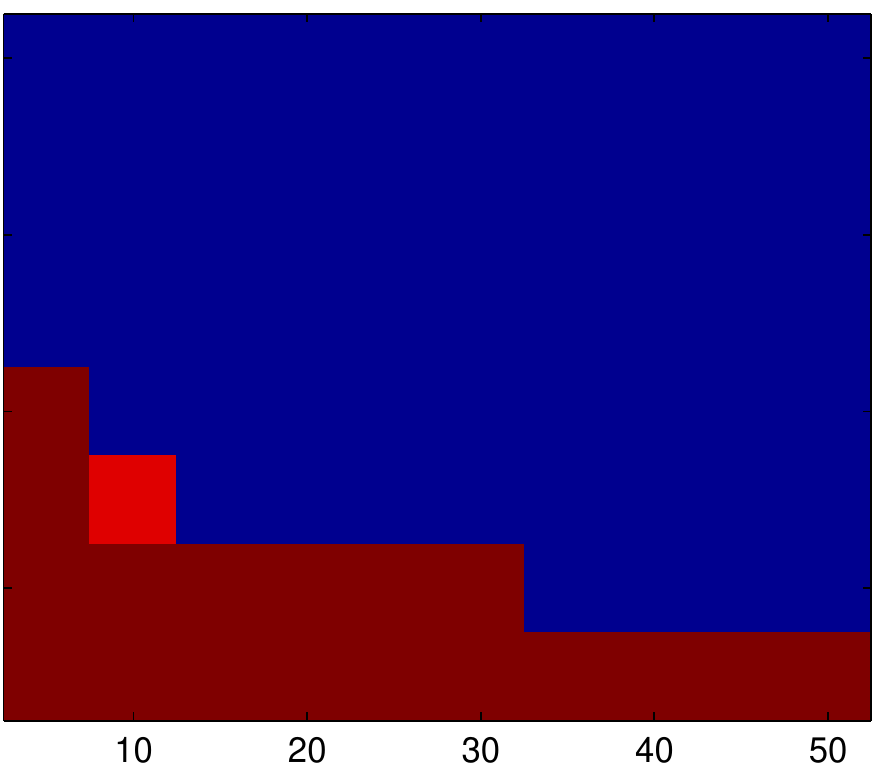}\;\!\!
\includegraphics[width=0.1346\linewidth]{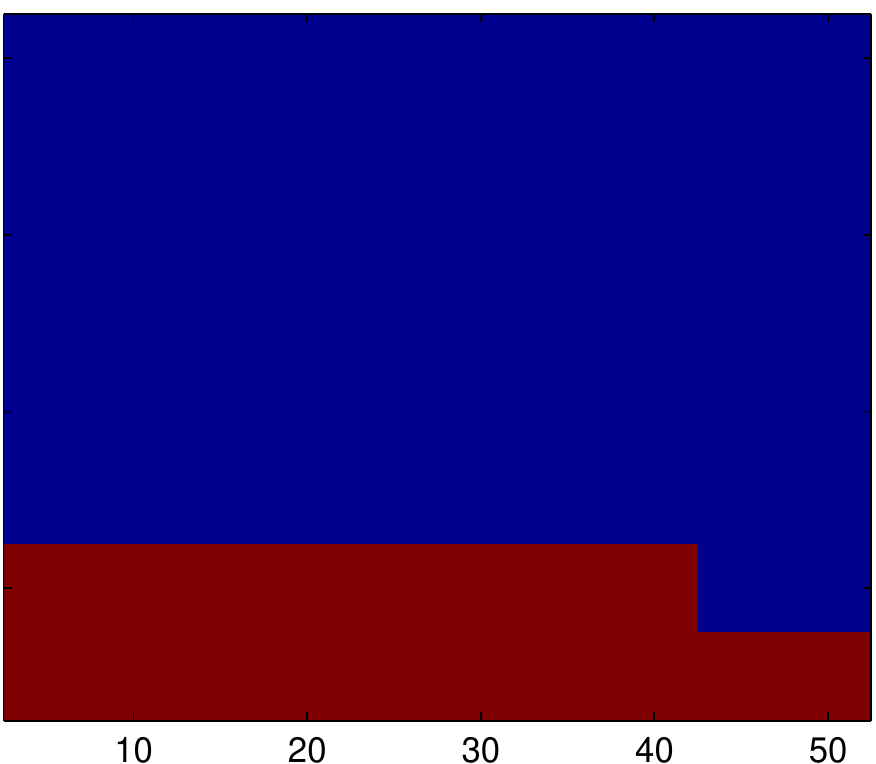}\;\!\!
\includegraphics[width=0.13405\linewidth]{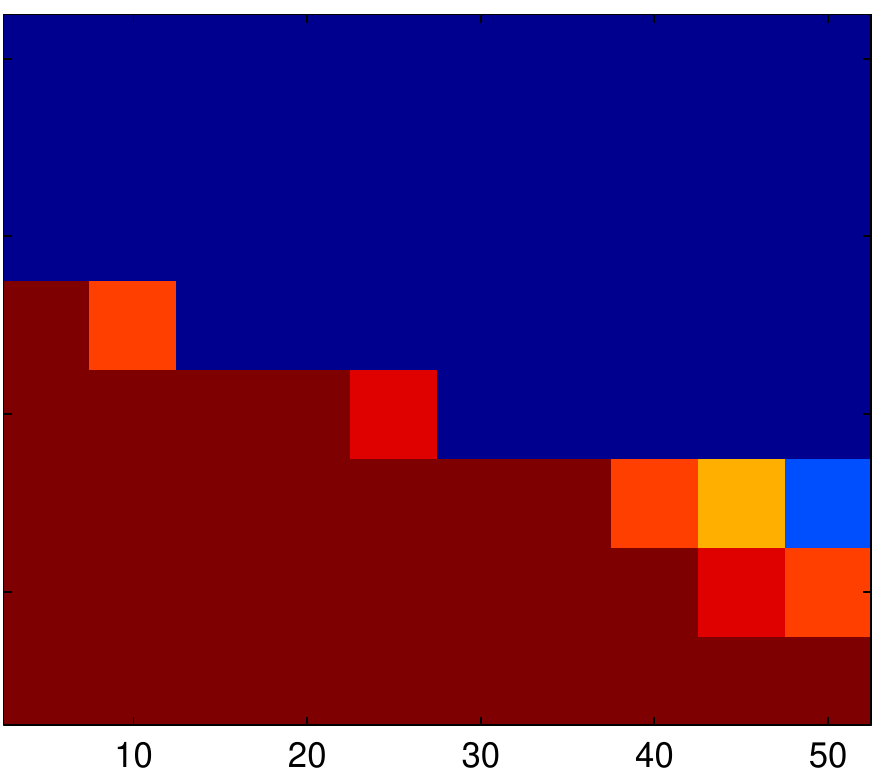}\;\!\!
\includegraphics[width=0.1506\linewidth]{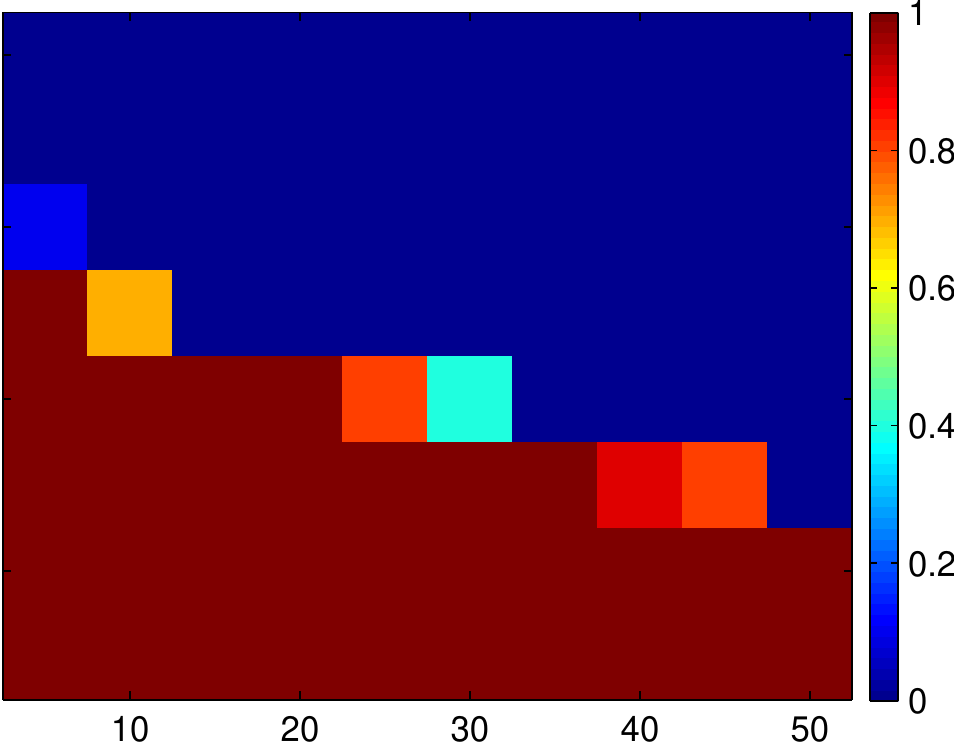}\\~\\
\vspace{-6.6mm}

\subfigure[RPCA~\cite{lin:almm}]{\includegraphics[width=0.1472\linewidth]{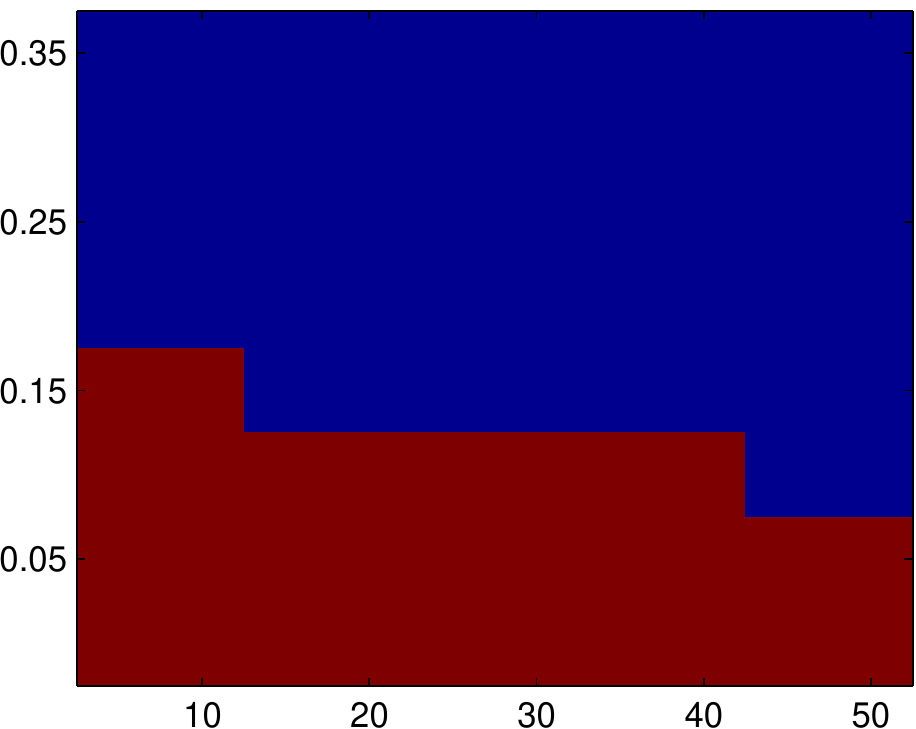}}\;\!\!
\subfigure[PSVT~\cite{oh:psvt}]{\includegraphics[width=0.134\linewidth]{PSVT1}}\;\!\!
\subfigure[WNNM~\cite{gu:wnnmj}]{\includegraphics[width=0.134\linewidth]{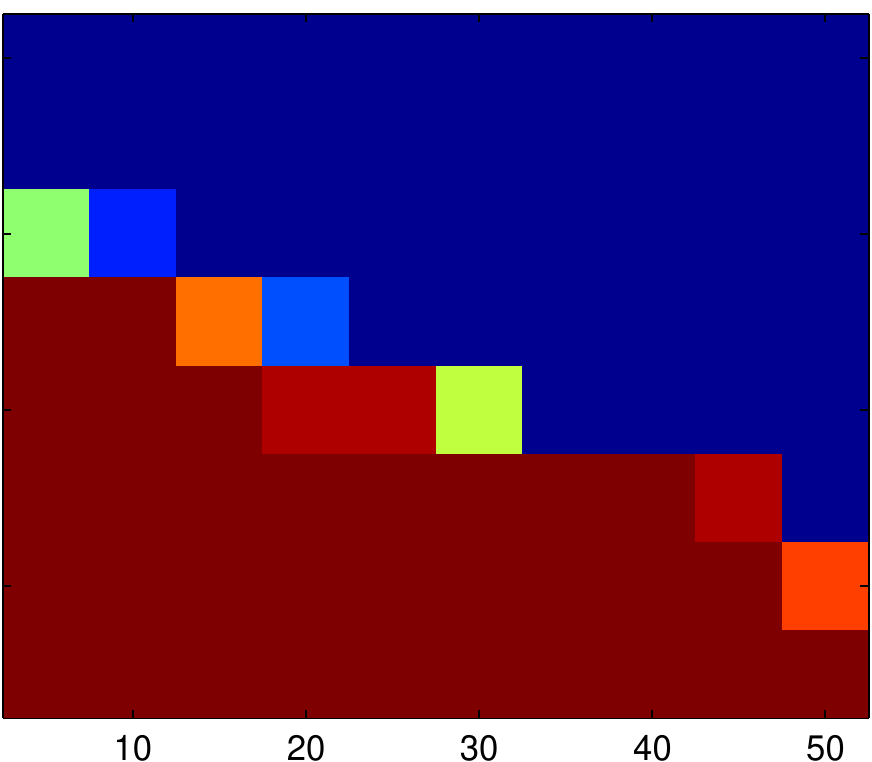}}\;\!\!
\subfigure[Unifying~\cite{cabral:nnbf}]{\includegraphics[width=0.134\linewidth]{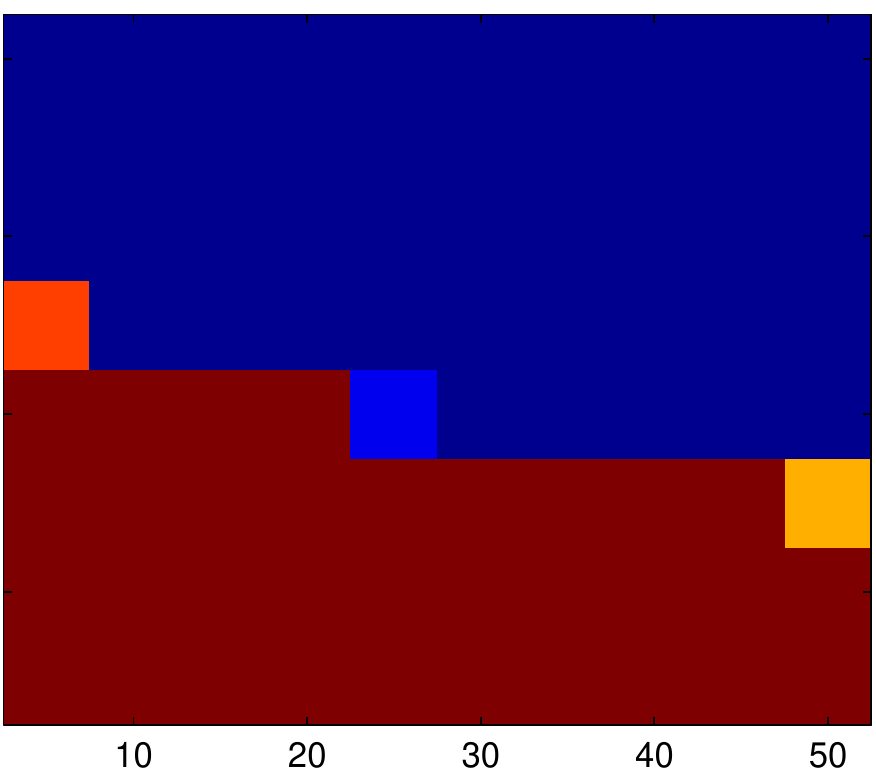}}\;\!\!
\subfigure[LpSq~\cite{nie:rmc}]{\includegraphics[width=0.1345\linewidth]{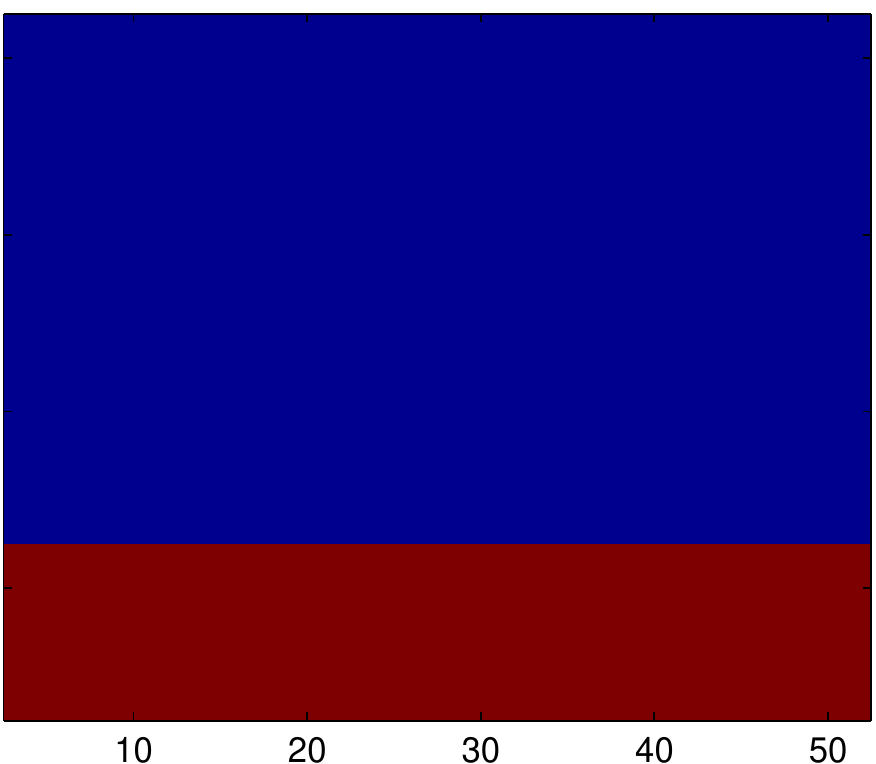}}\;\!\!
\subfigure[$(\textup{S+L})_{1/2}$]{\includegraphics[width=0.1341\linewidth]{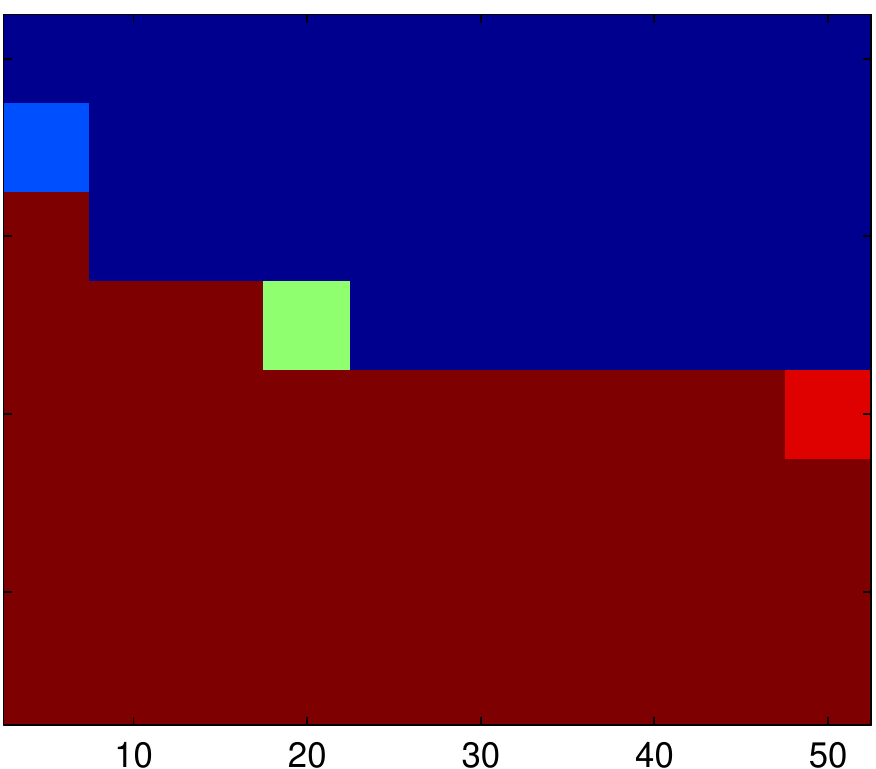}}\;\!\!
\subfigure[$(\textup{S+L})_{2/3}$]{\includegraphics[width=0.1514\linewidth]{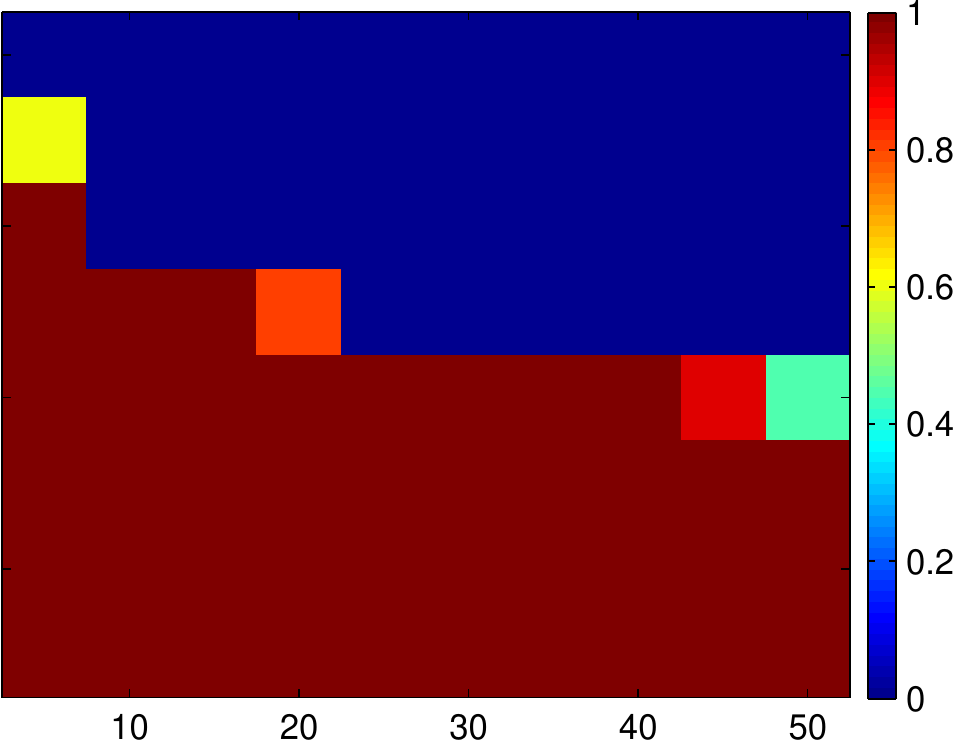}}\;\!\!
\caption{Phase transition plots for different algorithms on corrupted matrices of size $200\!\times\!200$ (top) and $500\!\times\!500$ (bottom). X-axis denotes the matrix rank, and  Y-axis indicates the corruption ratio. The color magnitude indicates the success ratio $[0,1]$, and a larger red area means a better performance of the algorithm (best viewed in colors).}
\label{fig4}
\end{figure*}

\subsection{Computational Complexity}
The per-iteration cost of existing Schatten quasi-norm minimization methods such as LpSq~\cite{nie:rmc} is dominated by the computation of the thin SVD of an $m\!\times\! n$ matrix with $m\!\geq\! n$, and is $\mathcal{O}(mn^2)$. In contrast, the time complexity of computing SVD for~\eqref{eq4:08} and~\eqref{eq4:09} is $\mathcal{O}(md^{2}\!+\!nd^{2})$. The dominant cost of each iteration in Algorithm~\ref{alg1} corresponds to the matrix multiplications in the update of $U$, $V$ and $L$, which take $\mathcal{O}(mnd\!+\!d^{3})$. Given that $d\!\ll\! m, n$, the overall complexity of Algorithm~\ref{alg1}, as well as Algorithm 2, is thus $\mathcal{O}(mnd)$, which is the same as the complexity of LMaFit~\cite{shen:alad}, RegL1~\cite{zheng:lrma}, ROSL~\cite{shu:rosl}, Unifying~\cite{cabral:nnbf}, and factEN~\cite{kim:factEN}.

\section{Experimental Results}
\label{expe}
In this section, we evaluate both the effectiveness and efficiency of our methods (i.e., $(\textup{S+L})_{1/2}$ and $(\textup{S+L})_{2/3}$) for solving extensive synthetic and real-world problems. We also compare our methods with several state-of-the-art methods, such as LMaFit\footnote{\url{http://lmafit.blogs.rice.edu/}}\!~\cite{shen:alad}, RegL1\footnote{\url{https://sites.google.com/site/yinqiangzheng/}}\!~\cite{zheng:lrma}, Unifying~\cite{cabral:nnbf}, factEN\footnote{\url{http://cpslab.snu.ac.kr/people/eunwoo-kim}}\!~\cite{kim:factEN}, RPCA\footnote{\url{http://www.cis.pku.edu.cn/faculty/vision/zlin/zlin.htm}}\!~\cite{lin:almm}, PSVT\footnote{\url{http://thohkaistackr.wix.com/page}}\!~\cite{oh:psvt}, WNNM\footnote{\url{http://www4.comp.polyu.edu.hk/~cslzhang/}}\!~\cite{gu:wnnmj}, and LpSq\footnote{\url{https://sites.google.com/site/feipingnie/}}\!~\cite{nie:rmc}.

\subsection{Rank Estimation}
As suggested in~\cite{candes:rpca,zhang:op}, the regularization parameter $\lambda$ of our two methods is generally set to $\sqrt{\max(m,n)}$. Analogous to other matrix factorization methods~\cite{kim:factEN,zheng:lrma,shen:alad,cabral:nnbf}, two proposed methods also have another important rank parameter, $d$. To estimate it, we design a simple rank estimation procedure. Since the observed data may be corrupted by noise/outlier and/or missing data, our rank estimation procedure combines two key techniques. First, we efficiently compute the $k$ largest singular values of the input matrix (usually $k\!=\!100$), and then use the basic spectral gap technique for determining the number of clusters~\cite{luxburg:sc}. Moreover, the rank estimator for incomplete matrices is exploited to look for an index for which the ratio between two consecutive singular values is minimized, as suggested in~\cite{keshavan:lrmc}.
We conduct some experiments on corrupted matrices to test the performance of our rank estimation procedure, as shown in Fig.\ \ref{fig3}. Note that the input matrices are corrupted by both sparse outliers and Gaussian noise as shown below, where the fraction of sparse outliers varies from $0$ to $25\%$, and the noise factor of Gaussian noise is changed from $0$ to $0.5$. It can be seen that our rank estimation procedure performs well in terms of robustness to noise and outliers.

\begin{table*}[t]
\renewcommand{\arraystretch}{1.12}
\setlength{\tabcolsep}{1.53pt}
\scriptsize
\caption{Comparison of average RSE, F-M and time (seconds) on corrupted matrices. We highlight the best results in \textbf{bold} and the second best in \emph{italic} for each of three performance metrics. Note that WNNM and LpSq could not yield experimental results on the largest problem within 24 hours.}
\label{tab3}
\begin{tabular}[c]{c|ccc|ccccccccc|ccccccccc}
\hline \hline
\multirow{2}{*}{\ } & \multicolumn{3}{c|}{RPCA~\cite{lin:almm}} & \multicolumn{3}{c}{PSVT~\cite{oh:psvt}} & \multicolumn{3}{c}{WNNM~\cite{gu:wnnmj}} &  \multicolumn{3}{c|}{Unifying~\cite{cabral:nnbf}} & \multicolumn{3}{c}{LpSq~\cite{nie:rmc}} &  \multicolumn{3}{c}{$(\textup{S+L})_{1/2}$} & \multicolumn{3}{c}{$(\textup{S+L})_{2/3}$}\\
\ $\!\!\!\!m\!=\!n\!\!\!$   & RSE & F-M & Time  & RSE & F-M & Time  & RSE & F-M  & Time & RSE & F-M & Time	    & RSE & F-M & Time     & RSE & F-M & Time     & RSE & F-M & Time\\
\hline
\ \!\!500   &0.1265	&0.8145	 &6.87 &0.1157  &0.8298 &2.46    &0.0581 &0.8419  &30.59   &0.0570 &0.8427 &1.96 &0.1173  &0.8213 &245.81  &\emph{0.0469} &\emph{0.8469} &\emph{1.70} &\textbf{0.0453} &\textbf{0.8474} &\textbf{1.35}\\
\ \!\!1,000	&0.1138	&0.8240	 &35.29 &0.1107  &0.8325 &16.91   &0.0448 &0.8461  &203.52  &0.0443 &0.8462 &6.93 &0.1107  &0.8305 &985.69 &\emph{0.0335} &\emph{0.8495} &\emph{6.65}  &\textbf{0.0318} &\textbf{0.8498} &\textbf{5.89}\\
\ \!\!5,000  &0.1029	&0.8324	 &1,772.55 &0.0980  &0.8349 &1,425.25   &0.0315 &0.8483  &24,370.85  &0.0313 &0.8488 &171.28 &---  &--- &---  &\emph{0.0152} &\emph{0.8520}  &\emph{134.15} &\textbf{0.0145} &\textbf{0.8521} &\textbf{128.11}\\
\ \!\!10,000 &0.1002	&0.8340	 &20,321.36 &0.0969  &0.8355 &1,5437.60   &--- &---  &---    &0.0302 &0.8489 &657.41 &---  &--- &---  &\emph{0.0109} &\emph{0.8524} &\emph{528.80}  &\textbf{0.0104} &\textbf{0.8525} &\textbf{487.46}\\
\hline\hline
\end{tabular}
\end{table*}

\subsection{Synthetic Data}
We generated the low-rank matrix $L^{*}\!\in\!\mathbb{R}^{m\times n}$ of rank $r$ as the product $PQ^{T}$, where $P\!\in\!\mathbb{R}^{m\times r}$ and $Q\!\in\!\mathbb{R}^{n\times r}$ are independent matrices whose elements are independent and identically distributed (i.i.d.) random variables sampled from standard Gaussian distributions. The sparse matrix $S^{*}\!\in\!\mathbb{R}^{m\times n}$ is generated by the following procedure: its support is chosen uniformly at random and the non-zero entries are i.i.d.\ random variables sampled uniformly in the interval $[-5,5]$. The input matrix is $D\!=\!L^{*}\!+\!S^{*}\!+\!N$, where the Gaussian noise is $N\!=\!nf\!\times\!\mathbf{randn}$ and $nf\!\geq\!0$ is the noise factor. For quantitative evaluation, we measured the performance of low-rank component recovery by the RSE, and evaluated the accuracy of outlier detection by the F-measure (abbreviated to F-M) as in~\cite{zhou:mod}. The higher F-M or lower RSE, the better is the quality of the recovered results.

\begin{figure}[t]
\centering
\includegraphics[width=0.491\linewidth]{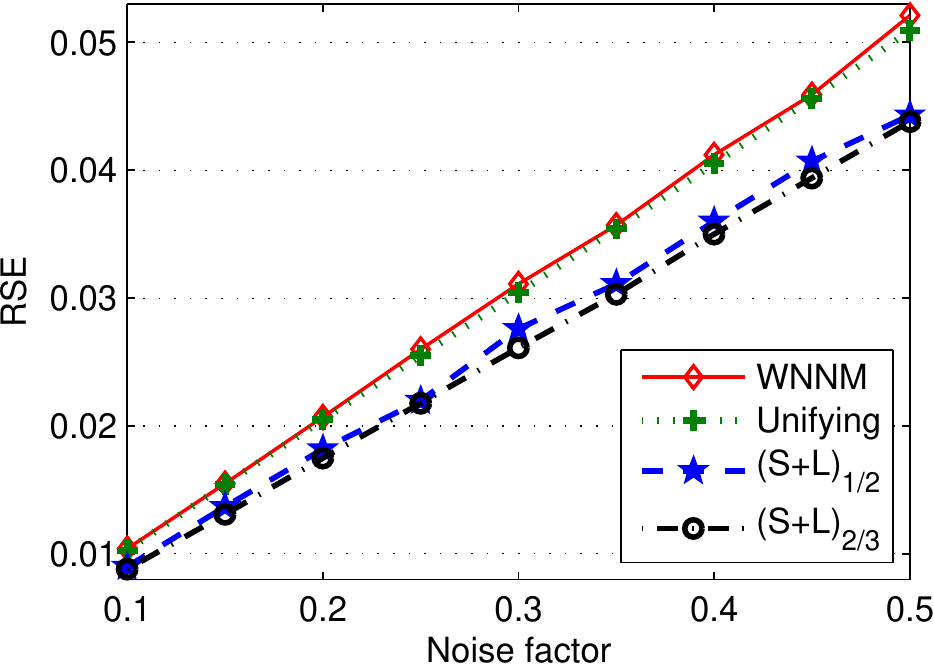}\:
\includegraphics[width=0.491\linewidth]{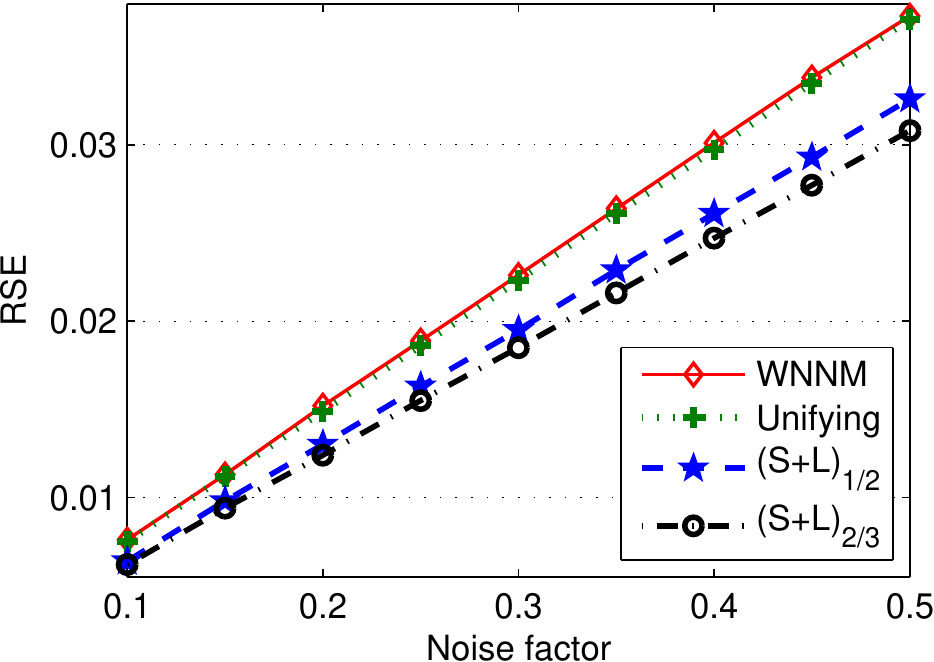}
\caption{Comparison of RSE results on corrupted matrices of size $500\!\times\!500$ (left) and $1,000\!\times\!1,000$ (right) with different noise factors.}
\label{fig5}
\end{figure}

\subsubsection{Model Comparison}
We first compared our methods with RPCA (nuclear norm \& $\ell_{1}$-norm), PSVT (truncated nuclear norm \& $\ell_{1}$-norm), WNNM (weighted nuclear norm \& $\ell_{1}$-norm), LpSq (Schatten $q$-norm \& $\ell_{p}$-norm), and Unifying (bilinear spectral penalty \& $\ell_{1}$-norm), where $p$ and $q$ in LpSq are chosen from the range of $\{0.1,0.2,\ldots,1\}$. A phase transition plot uses magnitude to depict how likely a certain kind of low-rank matrices can be recovered by those methods for a range of different matrix ranks and corruption ratios. If the recovered matrix $L$ has a RSE smaller than $10^{-2}$, we consider the estimation of both $L$ and $S$ is regarded as successful. Fig.\ \ref{fig4} shows the phase transition results of RPCA, PSVT, WNNM, LpSq, Unifying and both our methods on outlier corrupted matrices of size $200\times200$ and $500\times500$, where the corruption ratios varied from 0 to 0.35 with increment 0.05, and the true rank $r$ from 5 to 50 with increment 5. Note that the rank parameter of PSVT, Unifying, and both our methods is set to $d\!=\!\lfloor1.25r\rfloor$ as suggested in~\cite{shang:snm,shen:alad}. The results show that both our methods perform significantly better than the other methods, which justifies the effectiveness of the proposed RPCA models~\eqref{eq3.06} and \eqref{eq3.07}.

To verify the robustness of our methods, the observed matrices are corrupted by both Gaussian noise and outliers, where the noise factor and outlier ratio are set to $nf\!=\!0.5$ and $20\%$. The average results (including RSE, F-M, and running time) of 10 independent runs on corrupted matrices with different sizes are reported in Table~\ref{tab3}. Note that the rank parameter $d$ of PSVT, Unifying, and both our methods is computed by our rank estimation procedure. It is clear that both our methods significantly outperform all the other methods in terms of both RSE and F-M in all settings. Those non-convex methods including PSVT, WNNM, LpSq, Unifying, and both our methods consistently perform better than the convex method, RPCA, in terms of both RSE and F-M. Impressively, both our methods are much faster than the other methods, and at least 10 times faster than RPCA, PSVT, WNNM, and LpSq in the case when the size of matrices exceeds $5,000\!\times\!5,000$. This actually shows that our methods are more scalable, and have even greater advantage over existing methods for handling large matrices.

We also report the RSE results of both our methods on corrupted matrices of size $500\!\times\!500$ and $1,000\!\times\!1,000$ with outlier ratio 15\%, as shown in Fig.\ \ref{fig5}, where the true ranks of those matrices are $10$ and $20$, and the noise factor ranges from $0.1$ to $0.5$. In addition, we provide the best results of two baseline methods, WNNM and Unifying. Note that the parameter $d$ of Unifying and both our methods is computed via our rank estimation procedure. From the results, we can observe that both our methods are much more robust against Gaussian noise than WNNM and Unifying, and have much greater advantage over them in cases when the noise level is relatively large, e.g., $0.5$.

\begin{figure}[t]
\centering
\subfigure[F-measure vs.\ outlier ratio]{\includegraphics[width=0.491\linewidth]{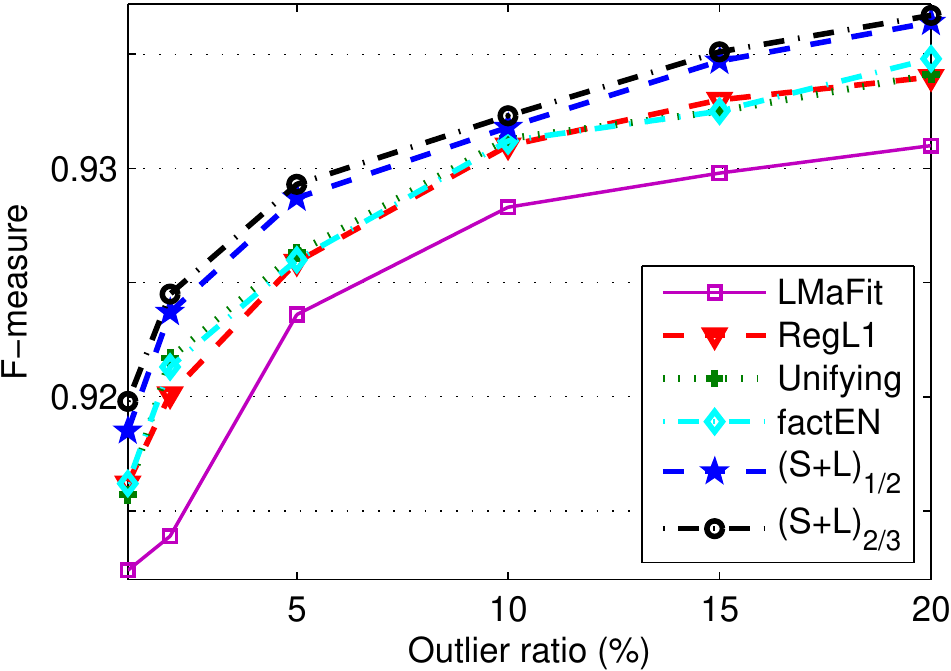}\label{fig6_first_case}}\,
\subfigure[RSE vs.\ outlier ratio]{\includegraphics[width=0.491\linewidth]{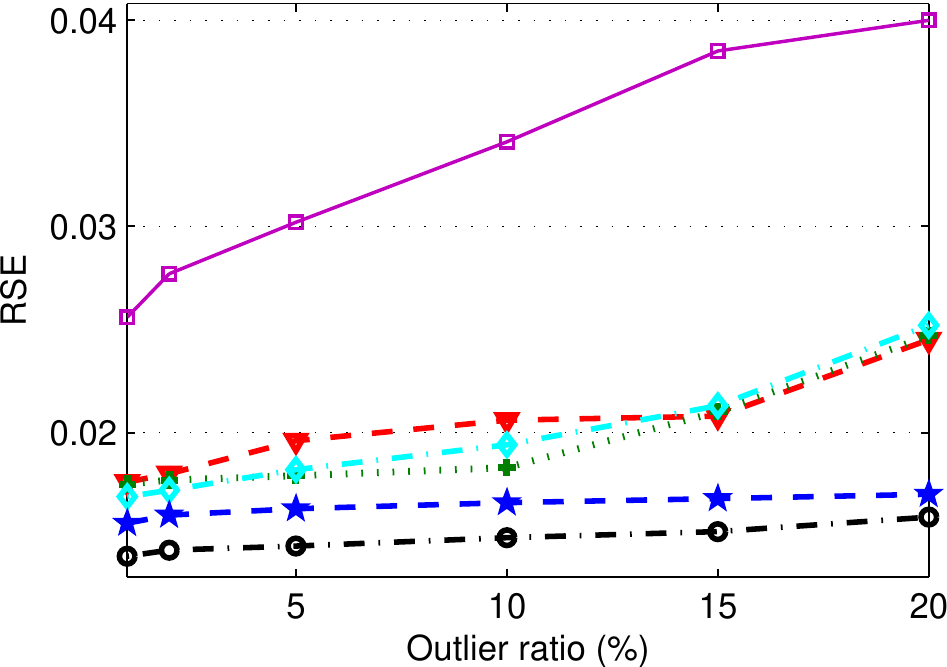}\label{fig6_second_case}}
\subfigure[RSE vs.\ high missing ratio]{\includegraphics[width=0.491\linewidth]{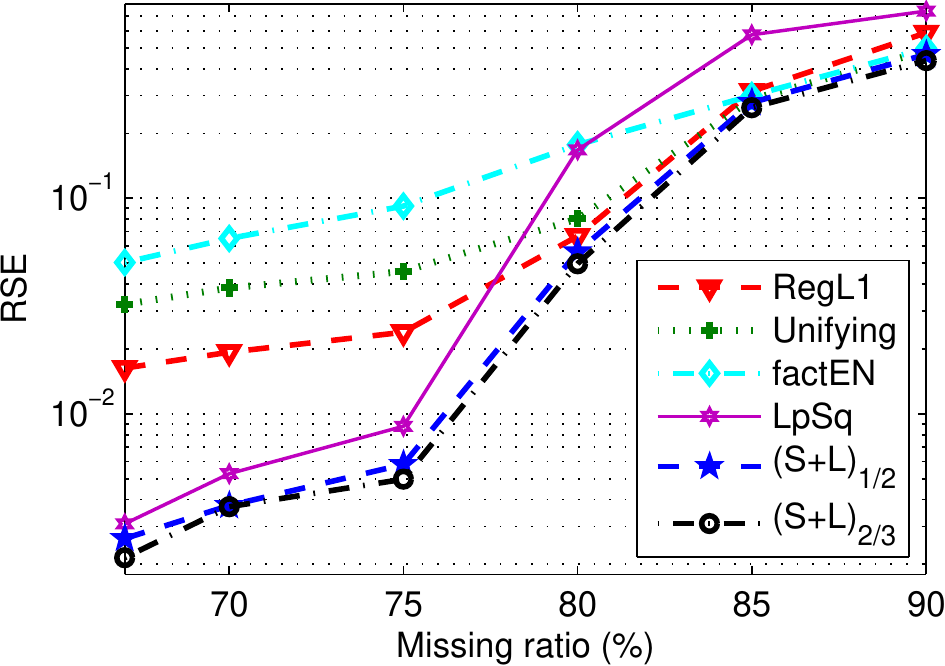}\label{fig6_third_case}}\,
\subfigure[RSE vs.\ low missing ratio]{\includegraphics[width=0.491\linewidth]{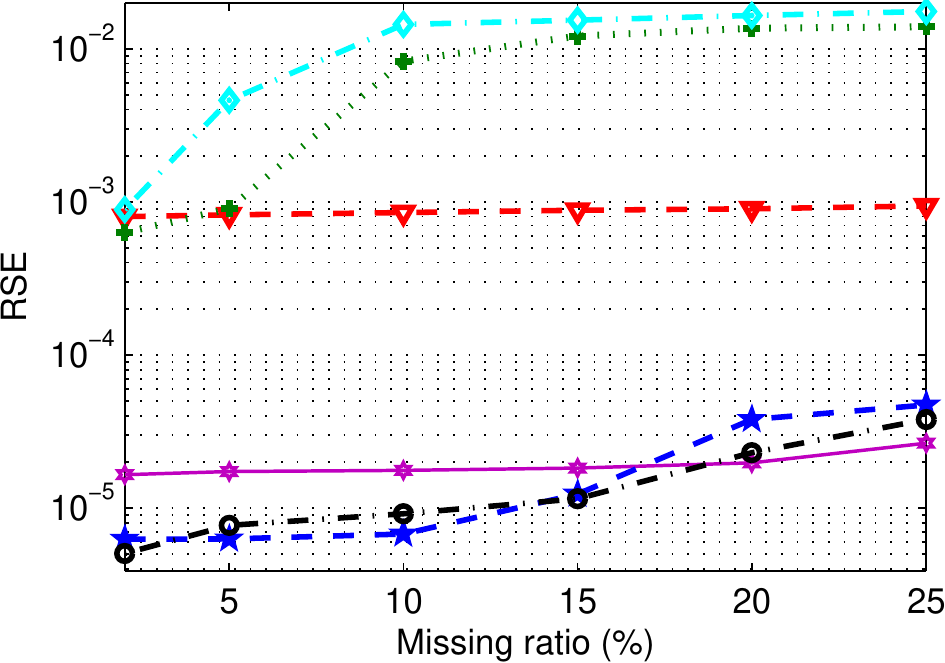}\label{fig6_fourth_case}}
\caption{Comparison of different methods on corrupted matrices under varying outlier and missing ratios in terms of RSE and F-measure.}
\label{fig6}
\end{figure}

\begin{figure}[t]
\centering
\includegraphics[width=0.491\linewidth]{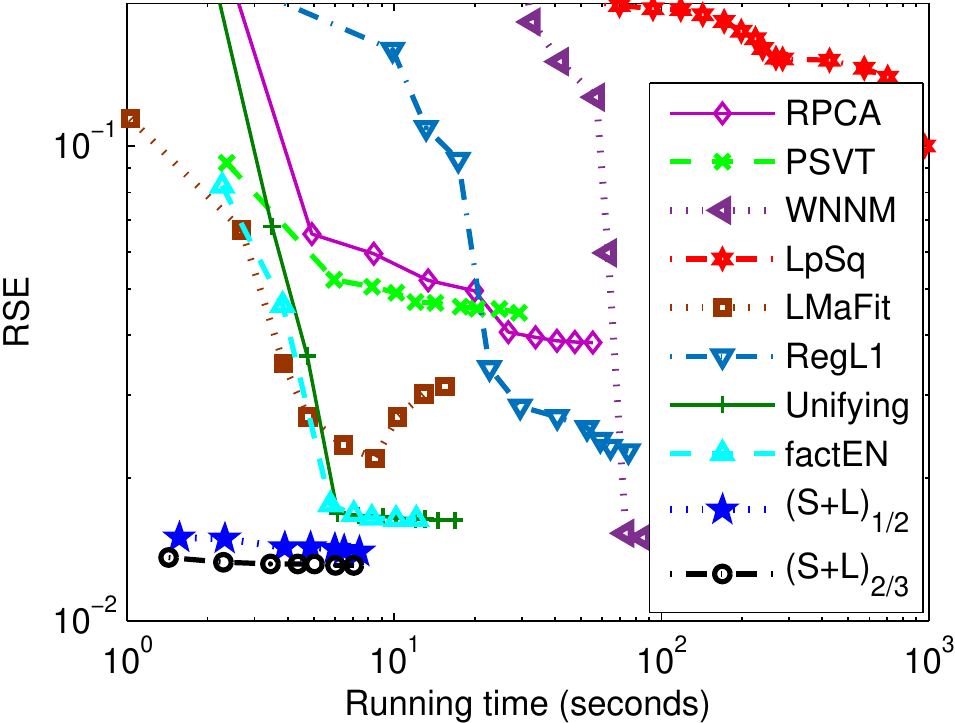}
\includegraphics[width=0.491\linewidth]{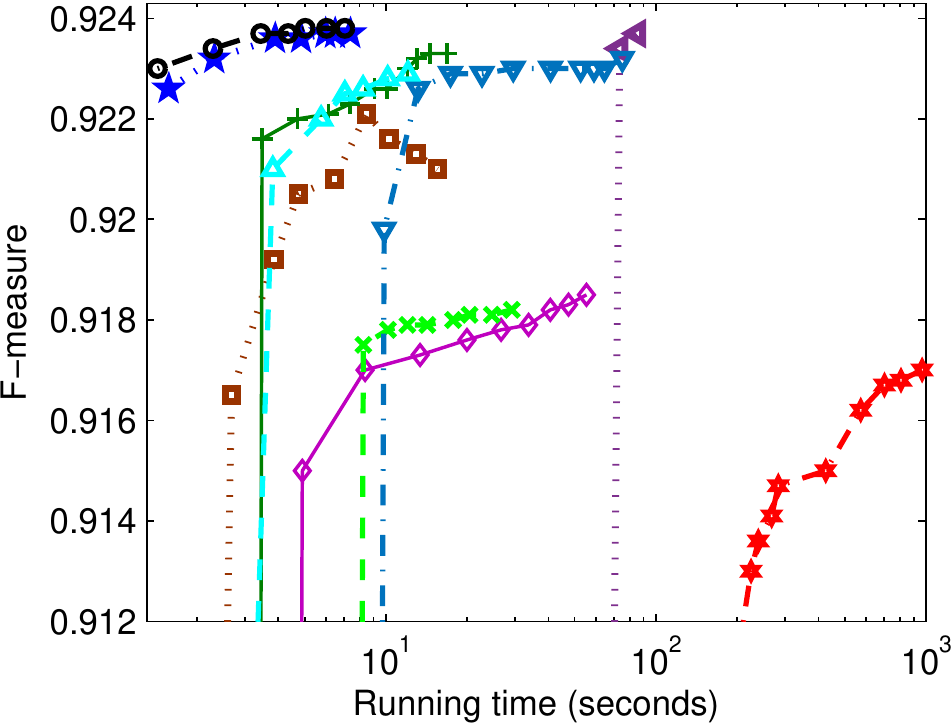}
\caption{Comparison of RSE (left) and F-measure (right) of different methods on corrupted matrices of size $1000\!\times\!1000$ vs.\ running time.}
\label{fig7}
\end{figure}

\begin{figure*}[t]
\centering
\includegraphics[width=0.1212\linewidth]{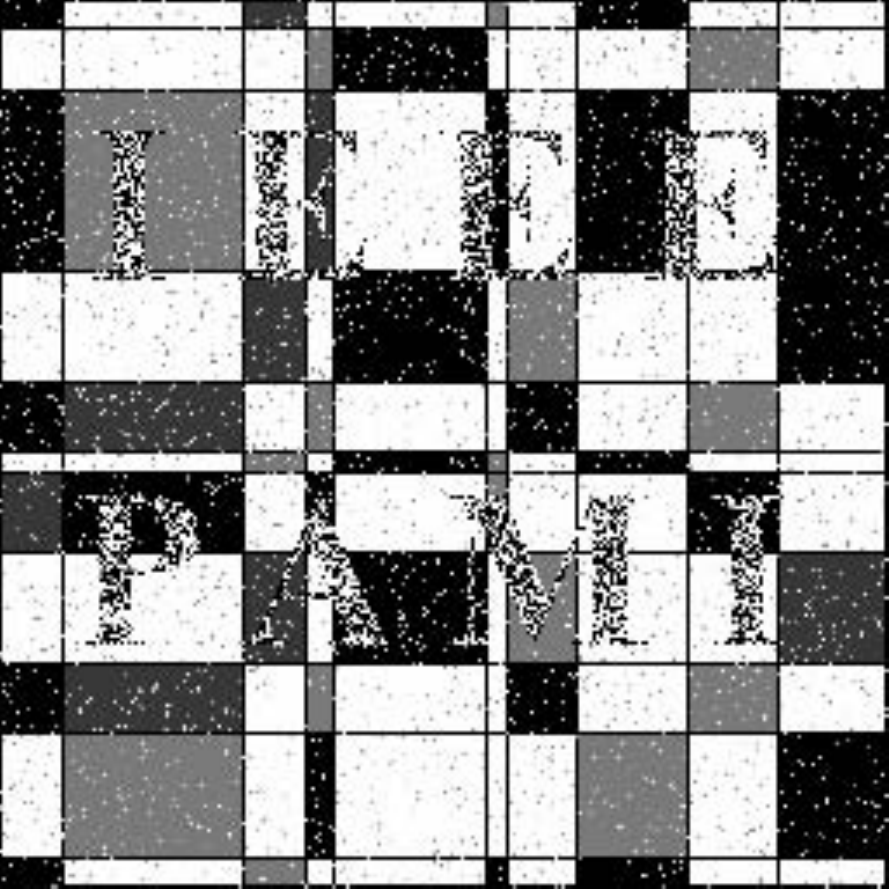}\;\!\!
\includegraphics[width=0.1212\linewidth]{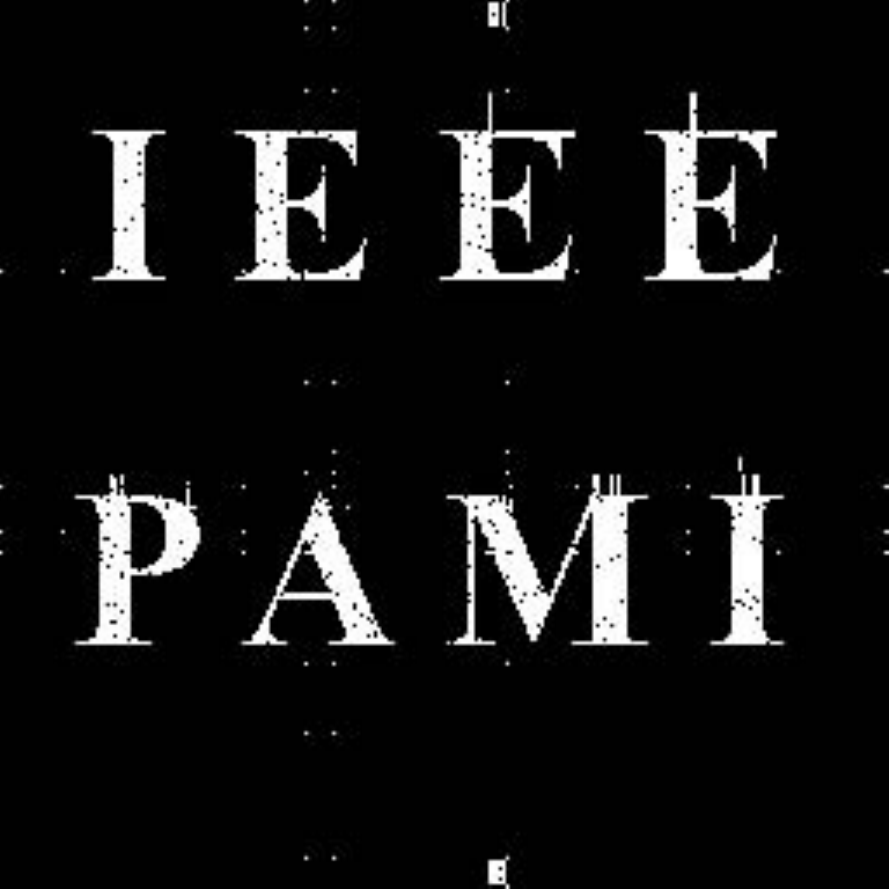}\;\!\!
\includegraphics[width=0.1212\linewidth]{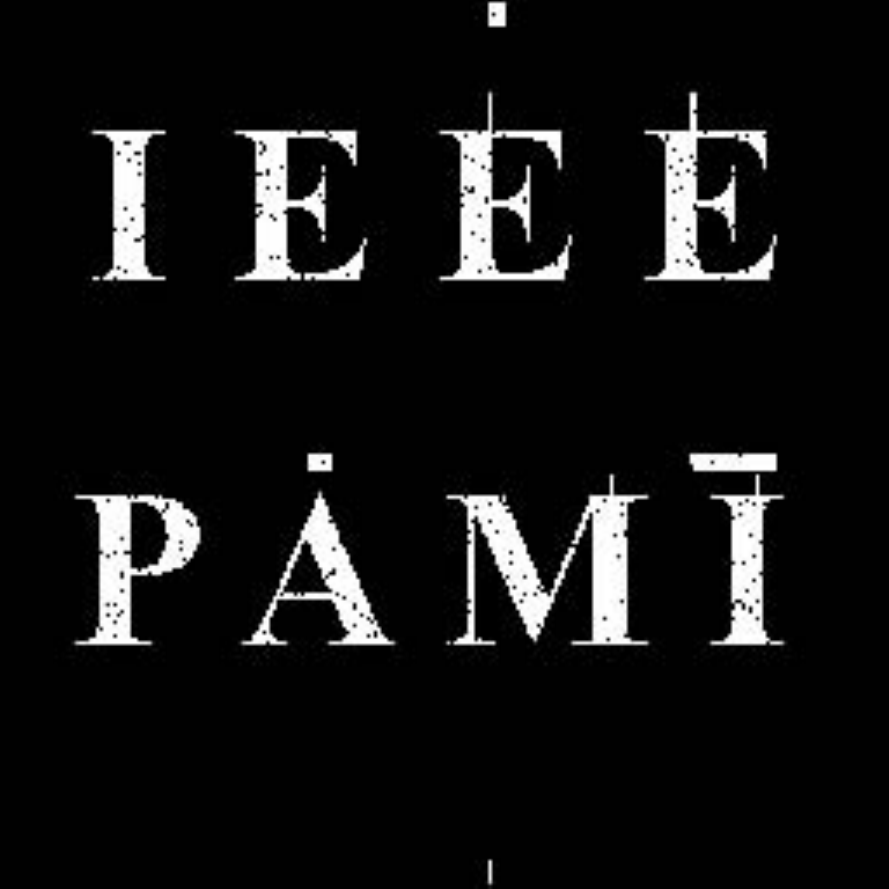}\;\!\!
\includegraphics[width=0.1212\linewidth]{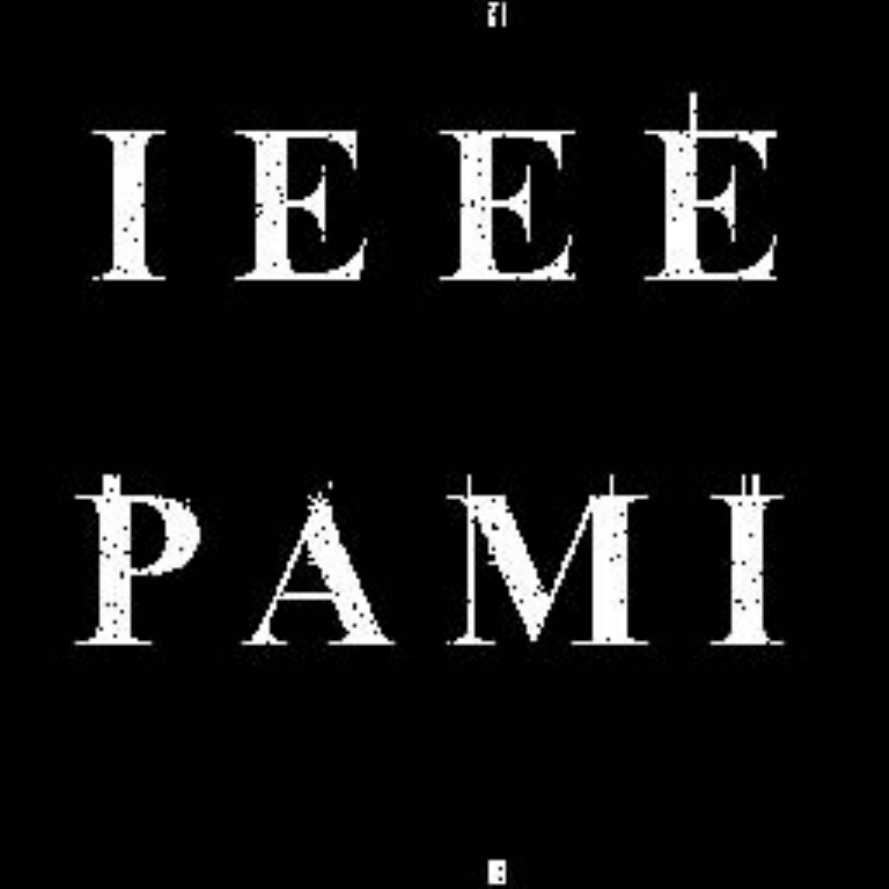}\;\!\!
\includegraphics[width=0.1212\linewidth]{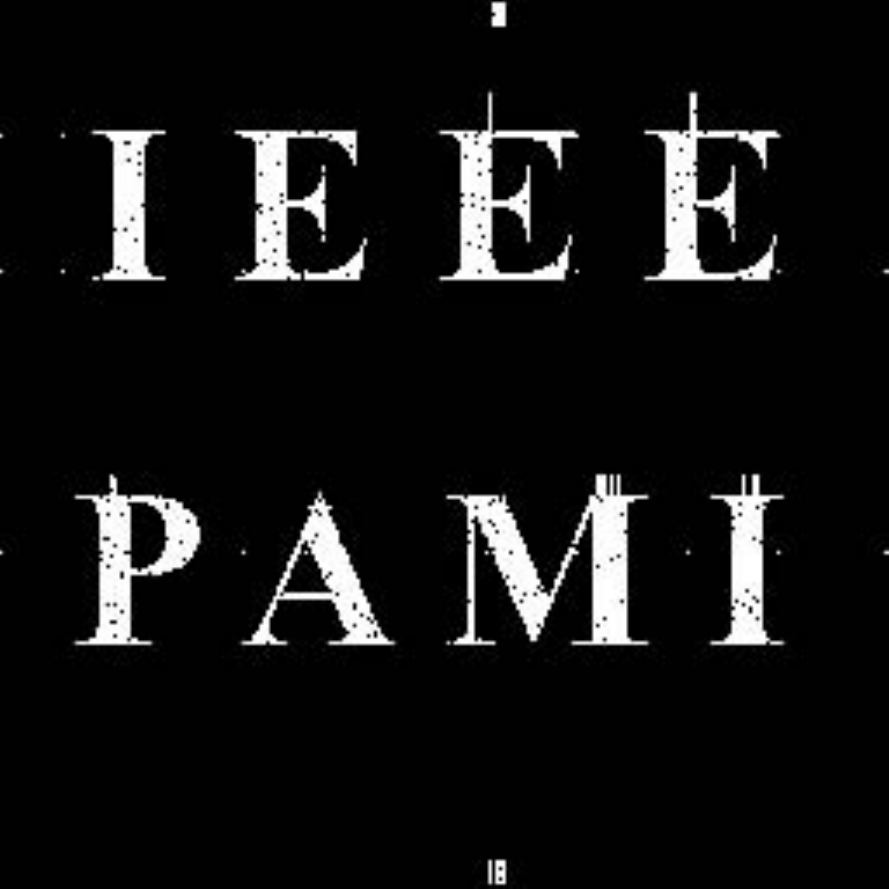}\;\!\!
\includegraphics[width=0.1212\linewidth]{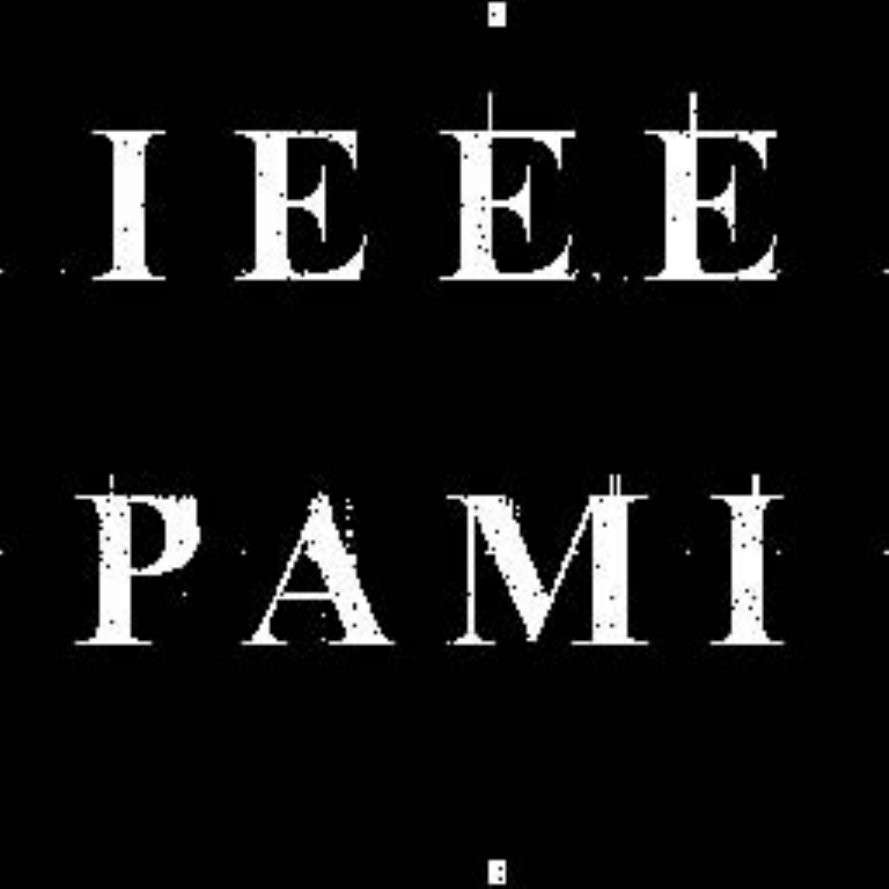}\;\!\!
\includegraphics[width=0.1212\linewidth]{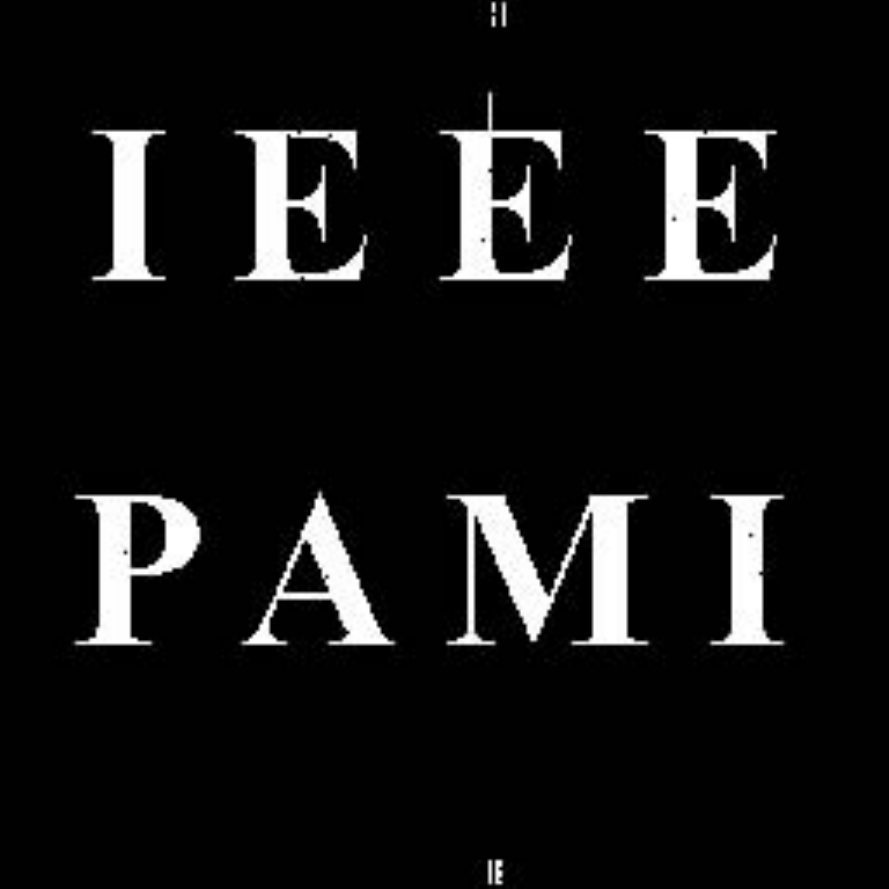}\;\!\!
\includegraphics[width=0.1212\linewidth]{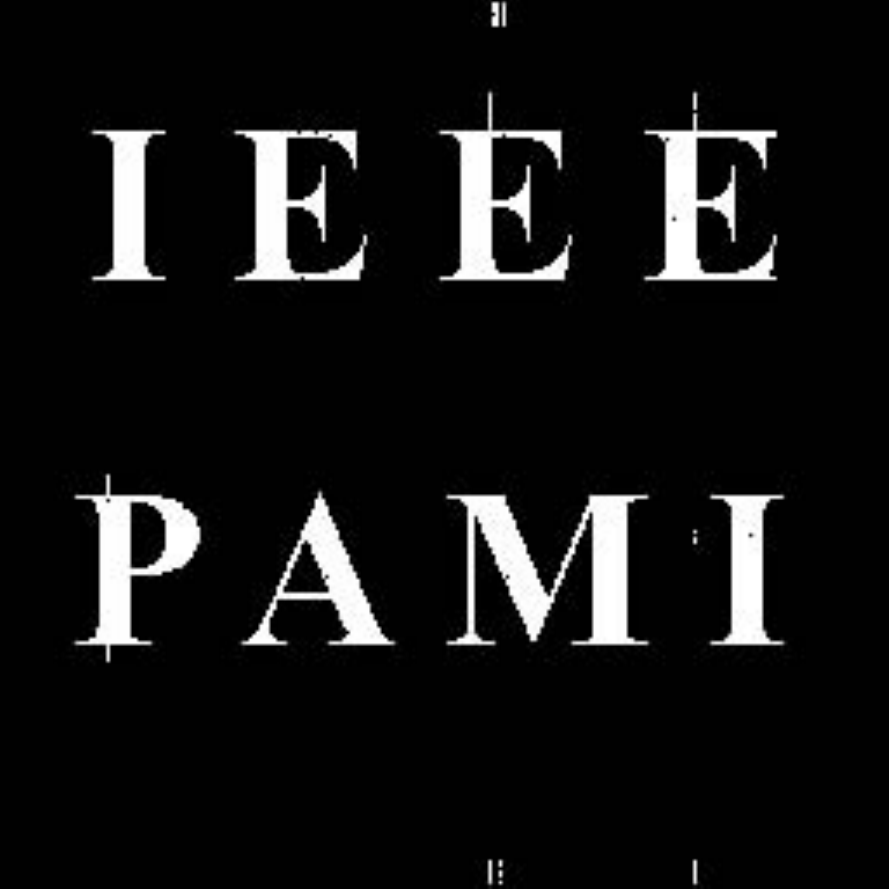}

\vspace{-0.5mm}

\subfigure[Images]{\includegraphics[width=0.1212\linewidth]{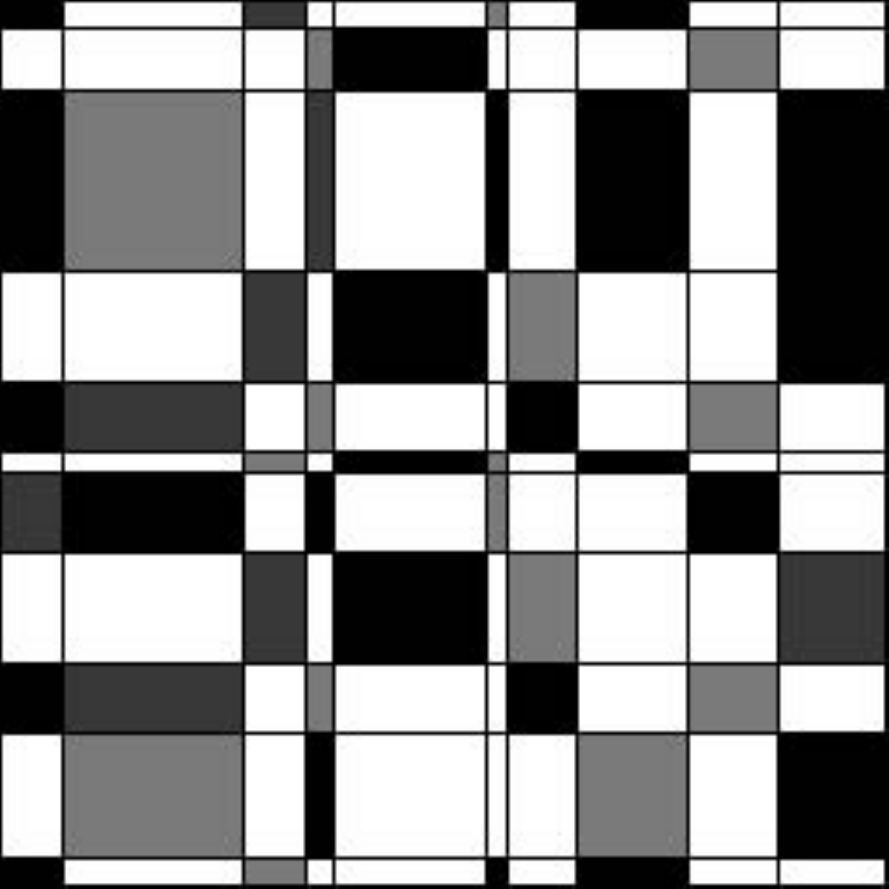}\label{fig81_first_case}}\;\!\!
\subfigure[RPCA~\cite{lin:almm}]{\includegraphics[width=0.1212\linewidth]{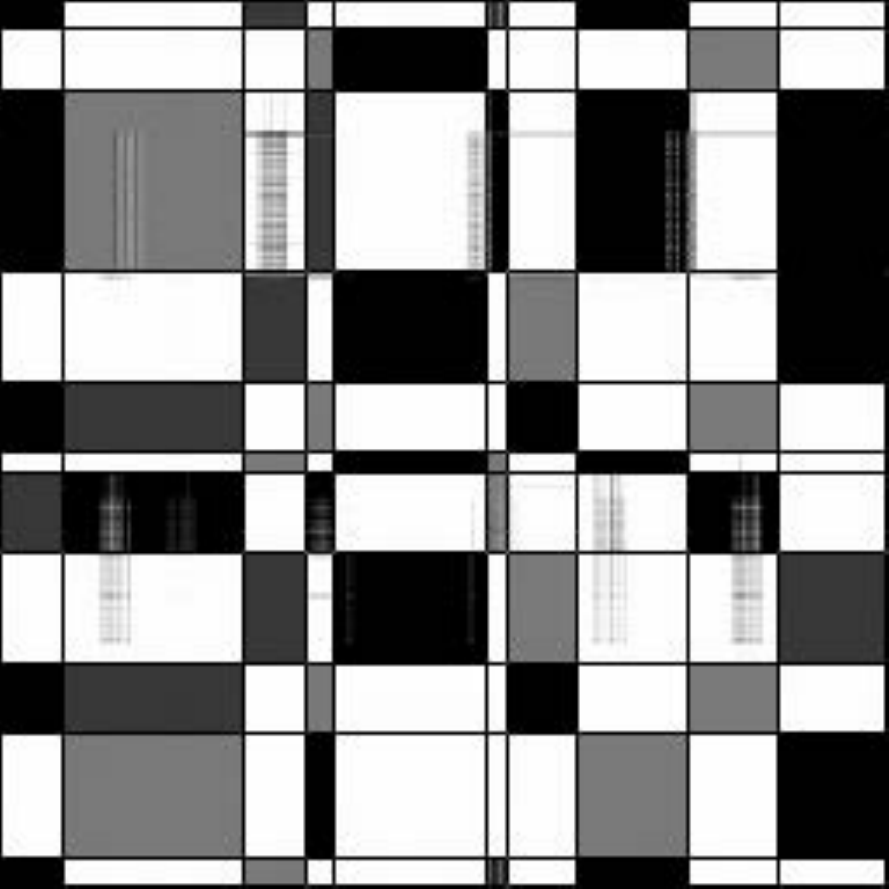}}\;\!\!
\subfigure[PSVT~\cite{oh:psvt}]{\includegraphics[width=0.1212\linewidth]{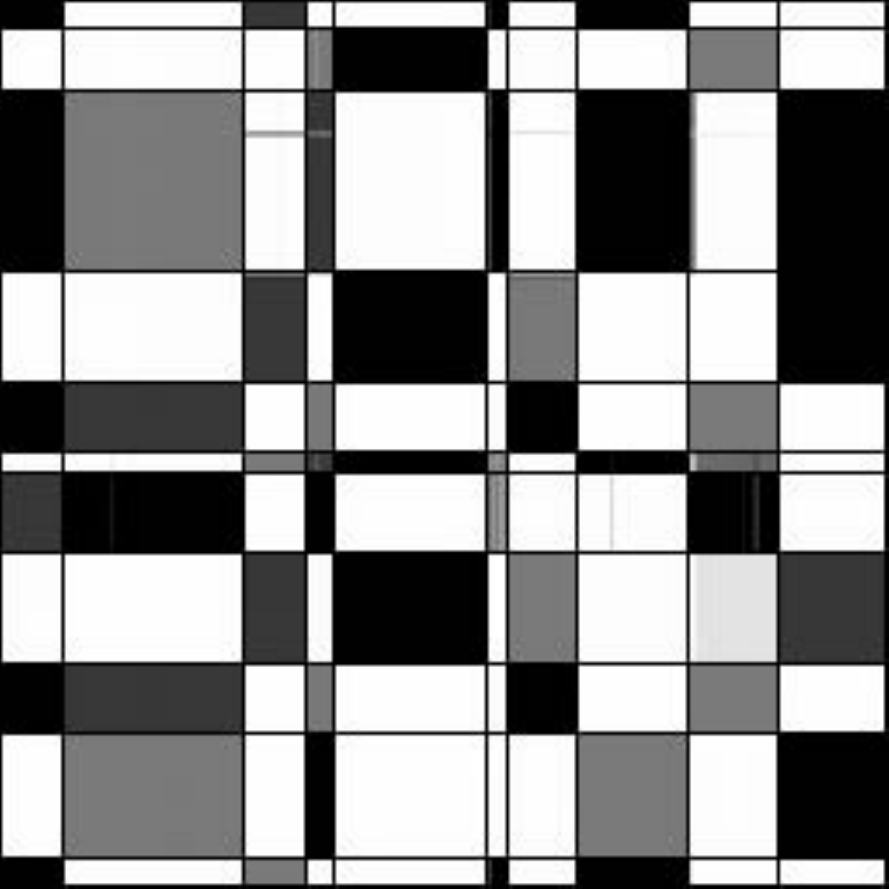}}\;\!\!
\subfigure[WNNM~\cite{gu:wnnmj}]{\includegraphics[width=0.1212\linewidth]{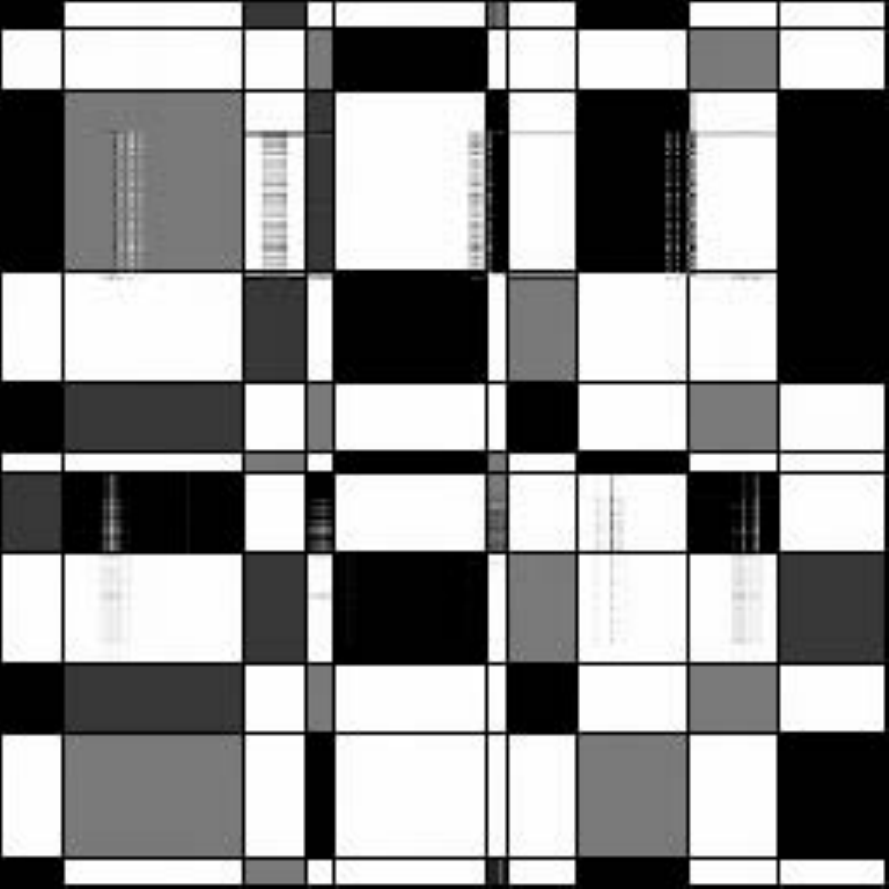}}\;\!\!
\subfigure[Unifying~\cite{cabral:nnbf}]{\includegraphics[width=0.1212\linewidth]{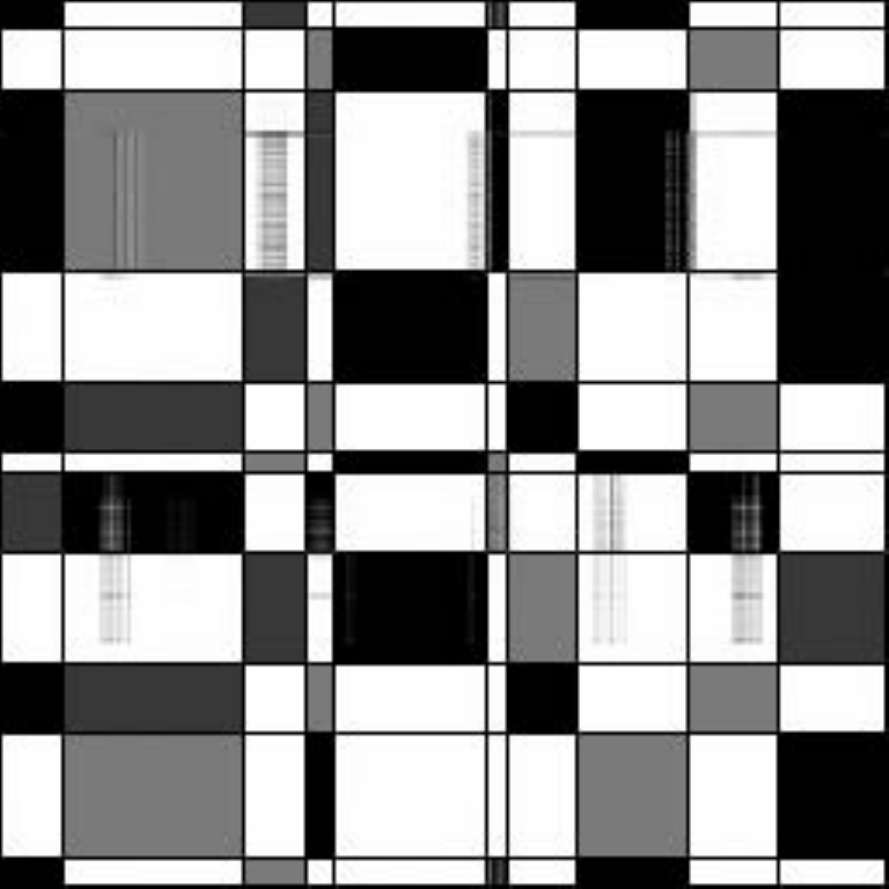}}\;\!\!
\subfigure[LpSq~\cite{nie:rmc}]{\includegraphics[width=0.1212\linewidth]{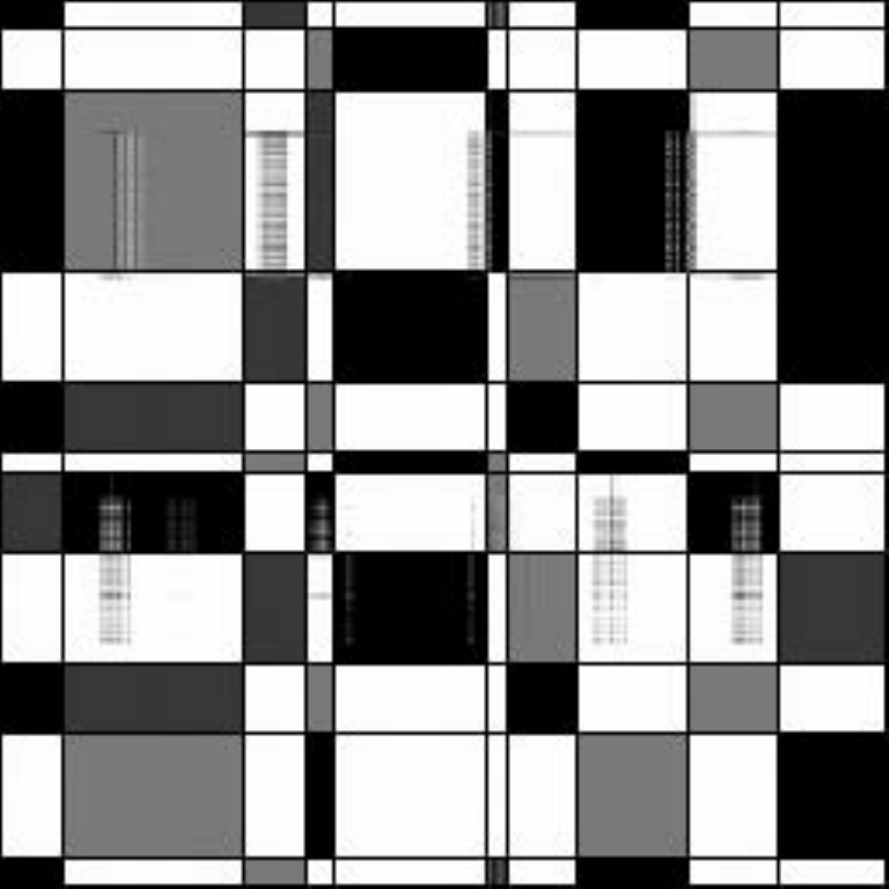}}\;\!\!
\subfigure[$(\textup{S+L})_{1/2}$]{\includegraphics[width=0.1212\linewidth]{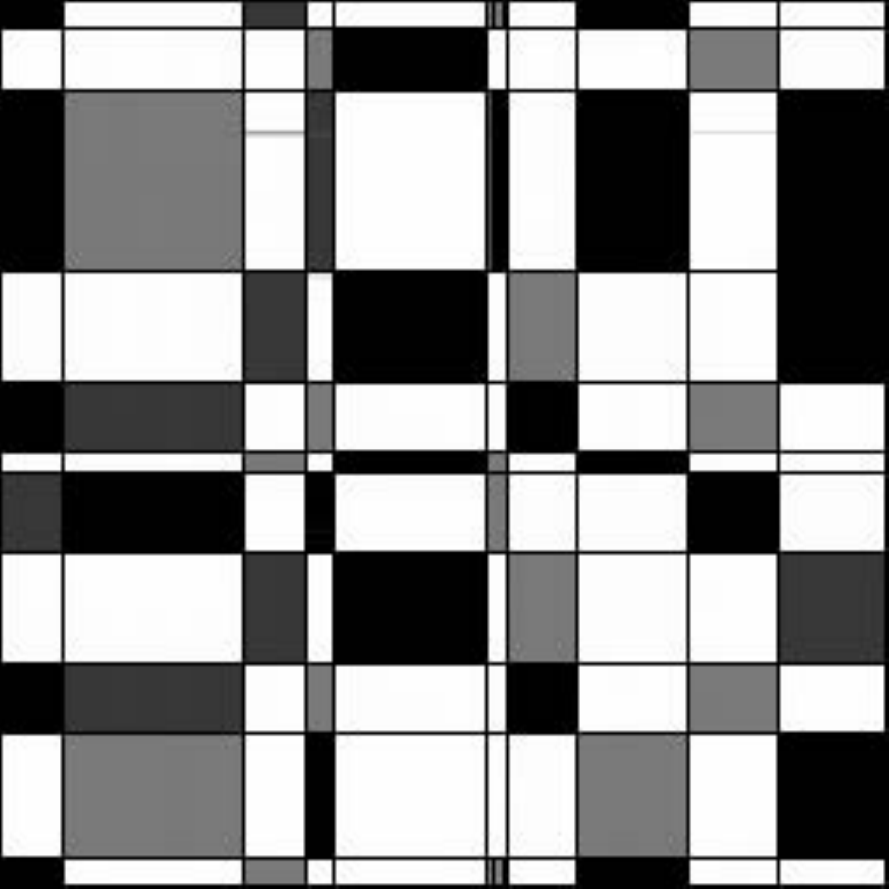}}\;\!\!
\subfigure[$(\textup{S+L})_{2/3}$]{\includegraphics[width=0.1212\linewidth]{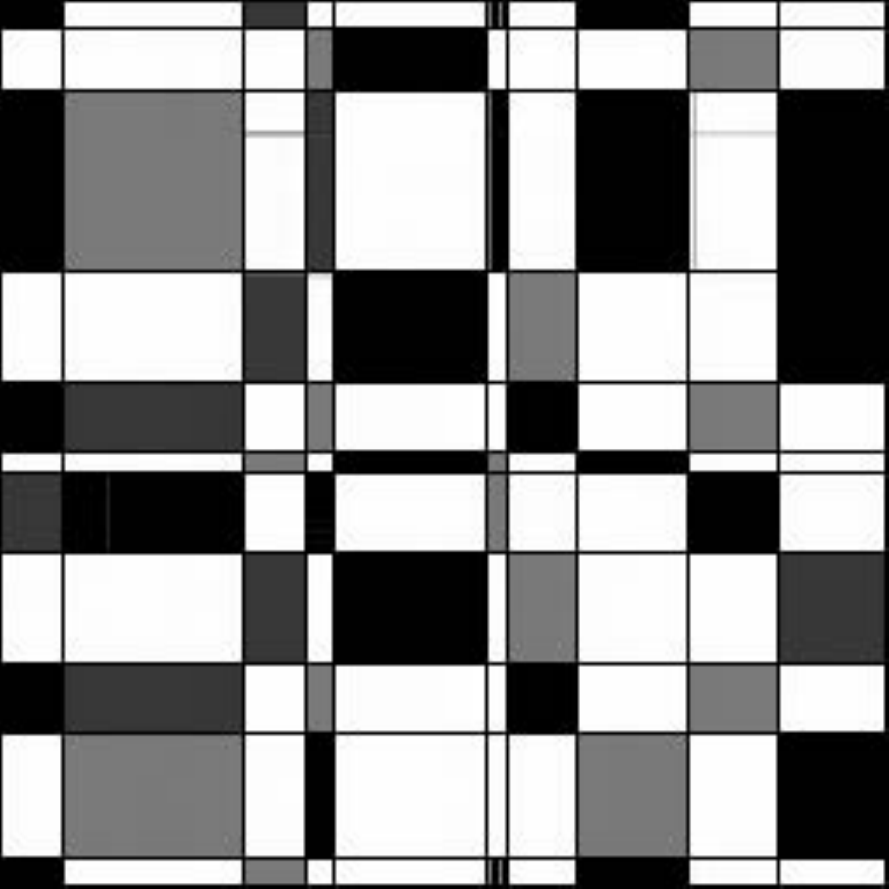}}
\caption{Text removal results of different methods: detected text masks (top) and recovered background images (bottom). (a) Input image (top) and original image (bottom); (b) RPCA (F-M: 0.9543, RSE: 0.1051); (c) PSVT (F-M: 0.9560, RSE: 0.1007); (d) WNNM (F-M: 0.9536, RSE: 0.0943); (e) Unifying (F-M: 0.9584, RSE: 0.0976); (f) LpSq (F-M: 0.9665, RSE: 0.1097); (g) $(\textup{S+L})_{1/2}$ (F-M: \textbf{0.9905}, RSE: \textbf{0.0396}); and (h) $(\textup{S+L})_{2/3}$ (F-M: \emph{0.9872}, RSE: \emph{0.0463}).}
\label{fig81}
\end{figure*}

\subsubsection{Comparisons with Other Methods}
Figs.\ \ref{fig6_first_case} and \ref{fig6_second_case} show the average F-measure and RSE results of different matrix factorization based methods on $1,000\!\times\!1,000$ matrices with different outliers ratios, where the noise factor is set to $0.2$. For fair comparison, the rank parameter of all these methods is set to $d\!=\!\lfloor1.25r\rfloor$ as in~\cite{shang:snm,shen:alad}. In all cases, RegL1~\cite{zheng:lrma}, Unifying~\cite{cabral:nnbf}, factEN~\cite{kim:factEN}, and both our methods have significantly better performance than LMaFit~\cite{shen:alad}, where the latter has no regularizers. This empirically verifies the importance of low-rank regularizers including our defined bilinear factor matrix norm penalties. Moreover, we report the average RSE results of these matrix factorization based methods with outlier ratio $5\%$ and various missing ratios in Figs.\ \ref{fig6_third_case} and \ref{fig6_fourth_case}, in which we also present the results of LpSq. One can see that only with a very limited number of observations (e.g., $80\%$ missing ratio), both our methods yield much more accurate solutions than the other methods including LpSq, while more observations are available, both LpSq and our methods significantly outperform the other methods in terms of RSE.

Finally, we report the performance of all those methods mentioned above on corrupted matrices of size $1000\!\times\!1000$ as running time goes by, as shown in Fig.\ \ref{fig7}. It is clear that both our methods obtain significantly more accurate solutions than the other methods with much shorter running time. Different from all other methods, the performance of LMaFit becomes even worse over time, which may be caused by the intrinsic model without a regularizer. This also empirically verifies the importance of all low-rank regularizers, including our defined bilinear factor matrix norm penalties, for recovering low-rank matrices.

\begin{figure*}[t]
\centering
\includegraphics[width=0.157\linewidth]{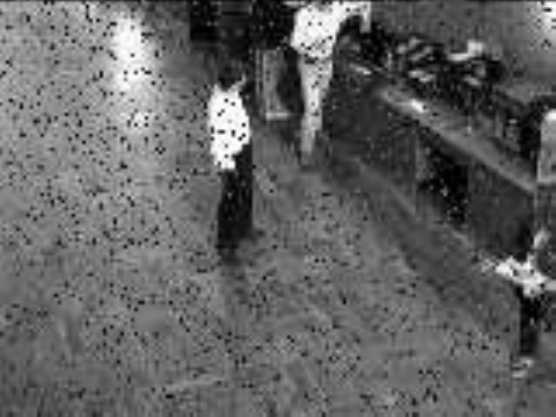}\,
\includegraphics[width=0.157\linewidth]{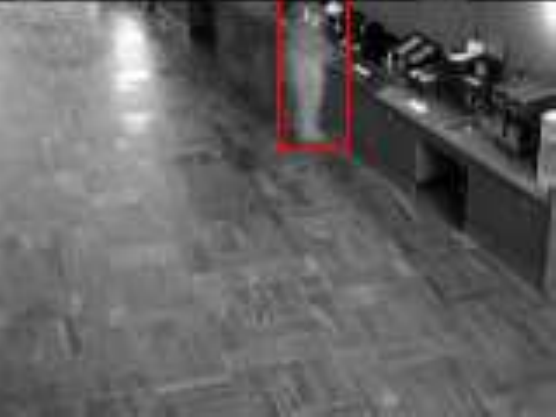}\,
\includegraphics[width=0.157\linewidth]{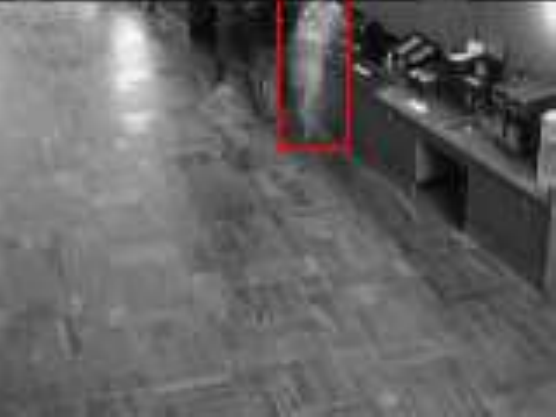}\,
\includegraphics[width=0.157\linewidth]{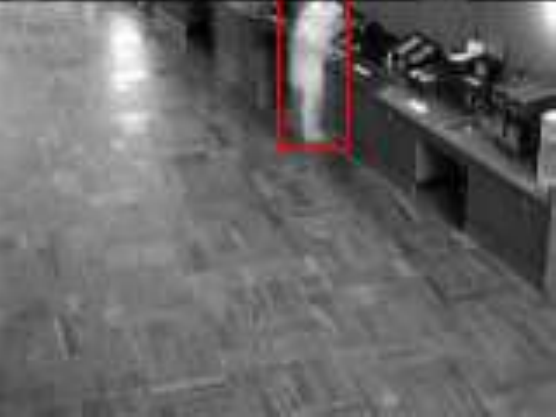}\,
\includegraphics[width=0.157\linewidth]{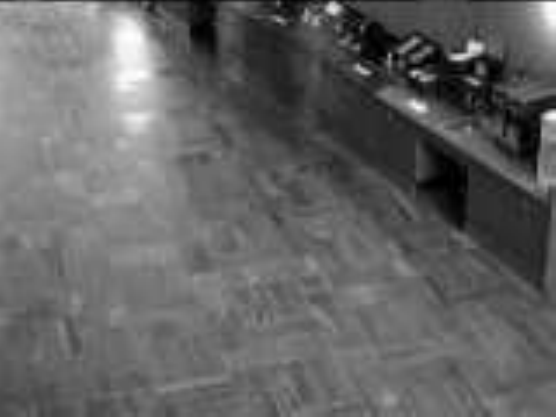}\,
\includegraphics[width=0.157\linewidth]{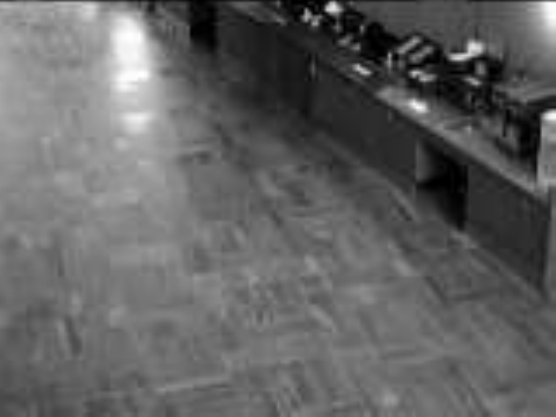}
\subfigure[\!Input, \!segmentation]{\includegraphics[width=0.157\linewidth]{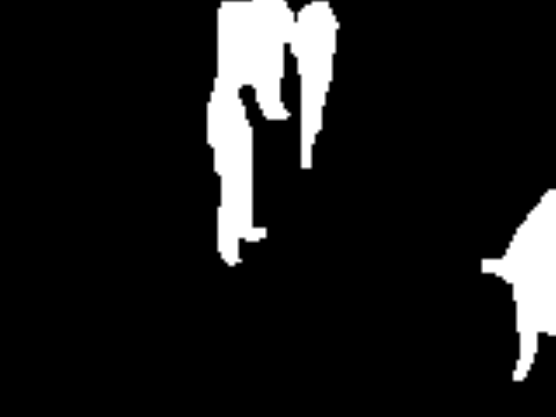}\label{fig8_first_case}}\,
\subfigure[RegL1~\cite{zheng:lrma}]{\includegraphics[width=0.157\linewidth]{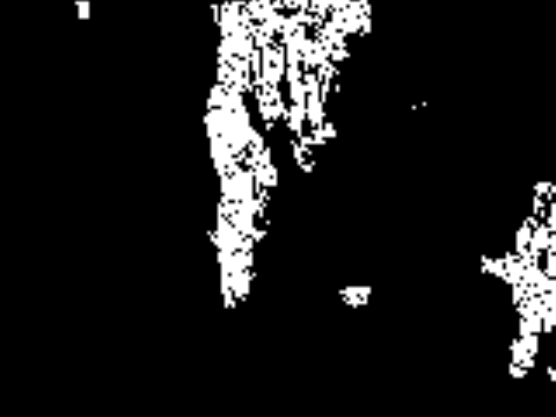}\label{fig8_second_case}}\,
\subfigure[Unifying~\cite{cabral:nnbf}]{\includegraphics[width=0.157\linewidth]{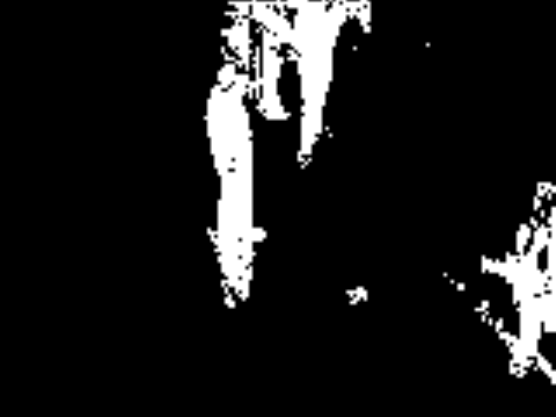}}\,
\subfigure[factEN~\cite{kim:factEN}]{\includegraphics[width=0.157\linewidth]{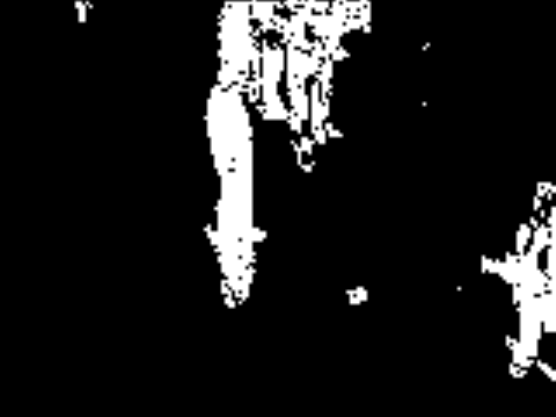}}\,
\subfigure[$(\textup{S+L})_{1/2}$]{\includegraphics[width=0.157\linewidth]{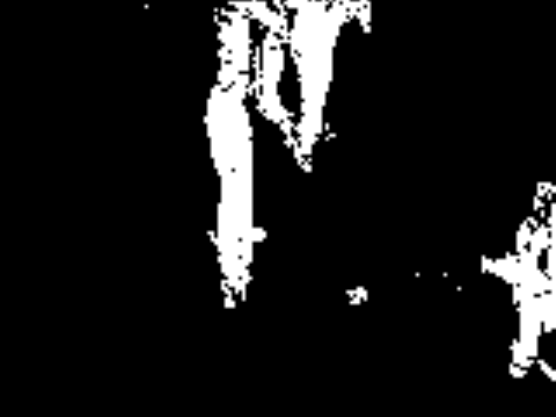}}\,
\subfigure[$(\textup{S+L})_{2/3}$]{\includegraphics[width=0.157\linewidth]{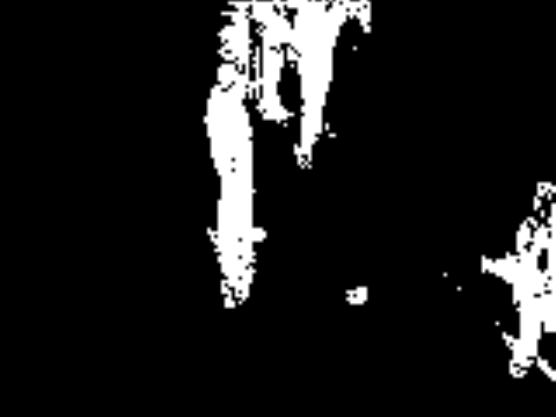}\label{fig8_last_case}}
\caption{Background and foreground separation results of different algorithms on the Bootstrap data set. The one frame with missing data of each sequence (top) and its manual segmentation (bottom) are shown in (a). The results of different algorithms are presented from (b) to (f), respectively. The top panel is the recovered background, and the bottom panel is the segmentation.}
\label{fig8}
\end{figure*}

\begin{figure}[t]
\includegraphics[width=0.495\linewidth]{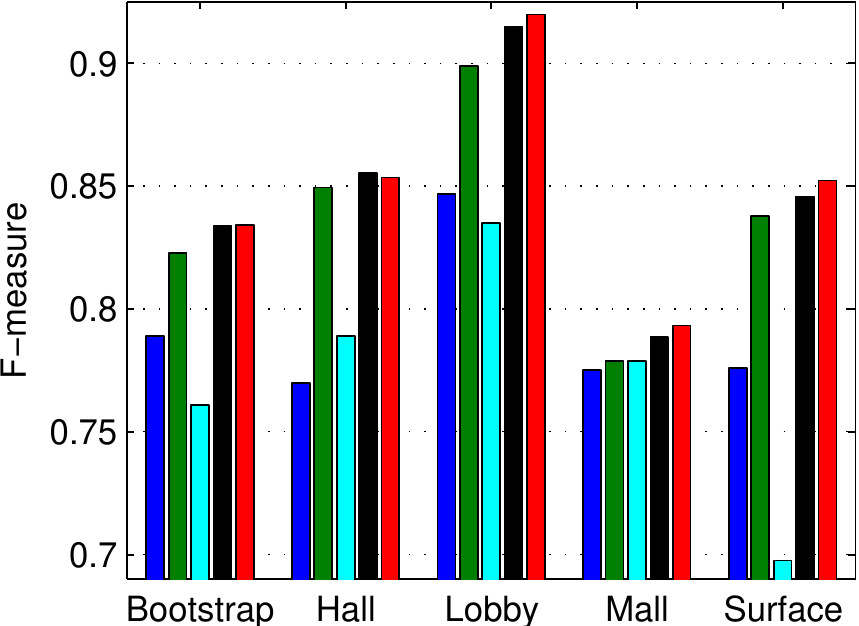}
\includegraphics[width=0.495\linewidth]{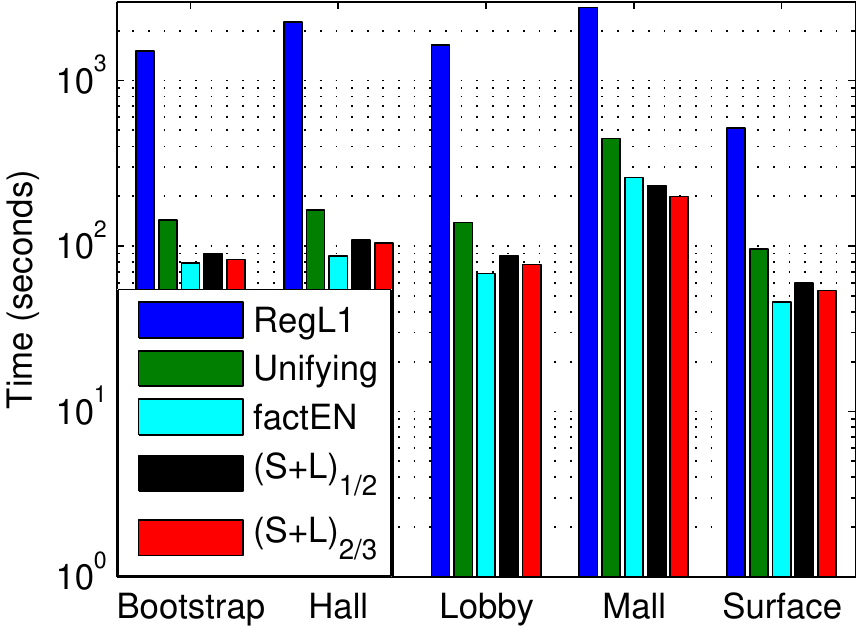}
\caption{Quantitative comparison for different methods in terms of F-measure (left) and running time (right, in seconds and in logarithmic scale) on five subsequences of surveillance videos.}
\label{fig9}
\end{figure}

\subsection{Real-world Applications}
In this subsection, we apply our methods to solve various low-level vision problems, e.g., text removal, moving object detection, and image alignment and inpainting.

\subsubsection{Text Removal}
We first apply our methods to detect some text and remove them from the image used in~\cite{shang:tsnm}. The ground-truth image is of size $256\!\times\!256$ with rank equal to 10, as shown in Fig.\ \ref{fig81_first_case}. The input data are generated by setting 5\% of the randomly selected pixels as missing entries. For fairness, we set the rank parameter of PSVT~\cite{oh:psvt}, Unifying~\cite{cabral:nnbf} and our methods to $20$, and the stopping tolerance $\epsilon\!=\!10^{-4}$ for all these algorithms. The text detection and removal results are shown in Fig.\ \ref{fig81}, where the text detection accuracy (F-M) and the RSE of recovered low-rank component are also reported. The results show that both our methods significantly outperform the other methods not only visually but also quantitatively. The running time of our methods and the Schatten quasi-norm minimization method, LpSq~\cite{nie:rmc}, is 1.36sec, 1.21sec and 77.65sec, respectively, which show that both our methods are more than 50 times faster than LpSq.

\subsubsection{Moving Object Detection}
We test our methods on real surveillance videos for moving object detection and background subtraction as a RPCA plus matrix completion problem. Background modeling is a crucial task for motion segmentation in surveillance videos. A video sequence satisfies the low-rank and sparse structures, because the background of all the frames is controlled by few factors and hence exhibits low-rank property, and the foreground is detected by identifying spatially localized sparse residuals~\cite{candes:rpca,zhou:mod,wright:rpca}. We test our methods on real surveillance videos for object detection and background subtraction on five surveillance videos: Bootstrap, Hall, Lobby, Mall and WaterSurface databases\footnote{\url{http://perception.i2r.a-star.edu.sg/}}. The input data are generated by setting 10\% of the randomly selected pixels of each frame as missing entries, as shown in Fig.\ \ref{fig8_first_case}.

Figs.\ \ref{fig8_second_case}-\ref{fig8_last_case} show the foreground and background separation results on the Bootstrap data set. We can see that the background can be effectively extracted by RegL1, Unifying, factEN and both our methods, where their rank parameter is computed via our rank estimation procedure. It is clear that the decomposition results of both our methods are significantly better than that of RegL1 and factEN visually in terms of both background components and foreground segmentation. In addition, we also provide the running time and F-measure of different algorithms on all the five data sets, as shown in Fig.\ \ref{fig9}, from which we can observe that both our methods consistently outperform the other methods in terms of F-measure. Moreover, Unifying and our methods are much faster than RegL1. Although factEN is slightly faster than Unifying and our methods, it usually has poorer quality of the results.

\begin{figure*}[t]
\centering
\includegraphics[width=0.117\linewidth]{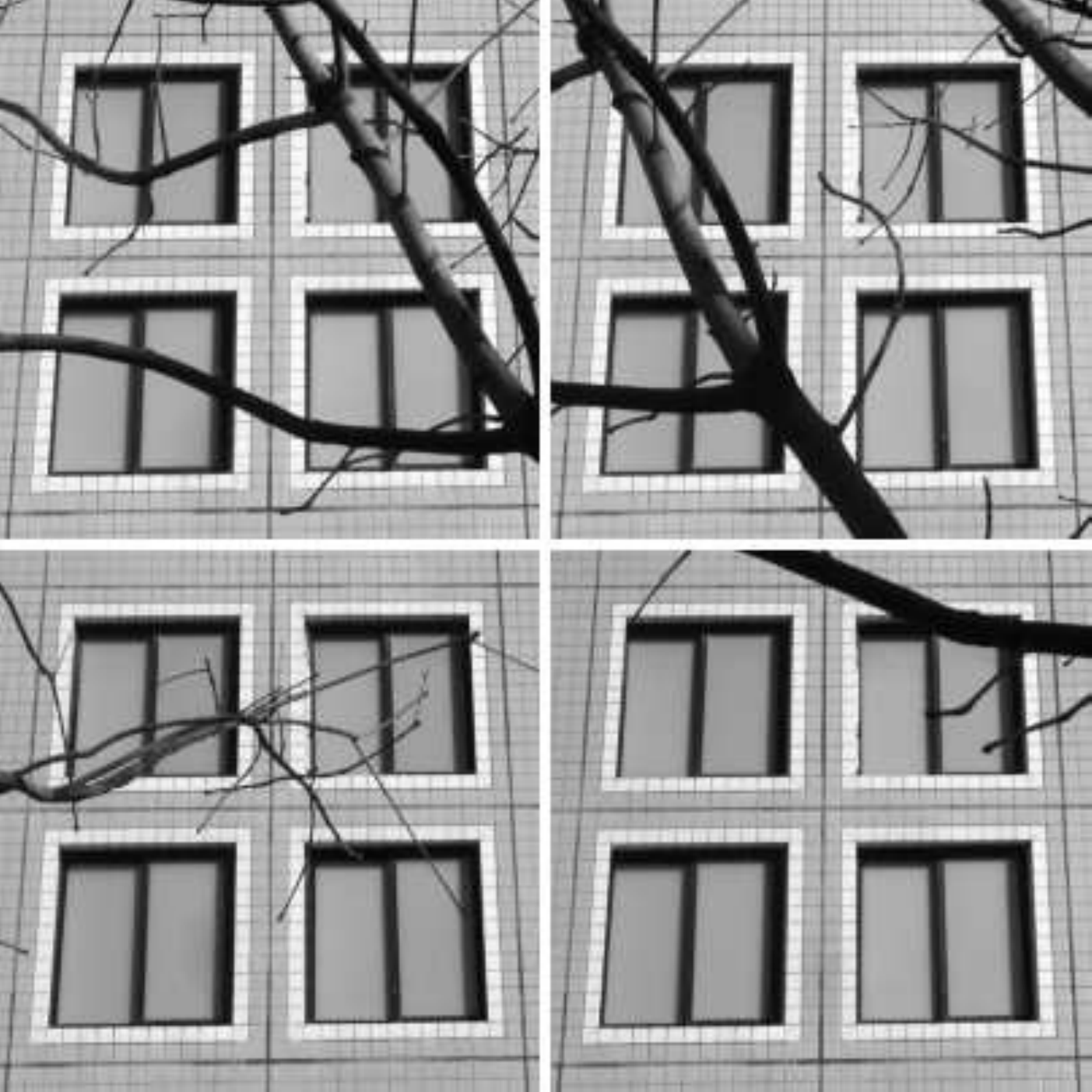}
\includegraphics[width=0.117\linewidth]{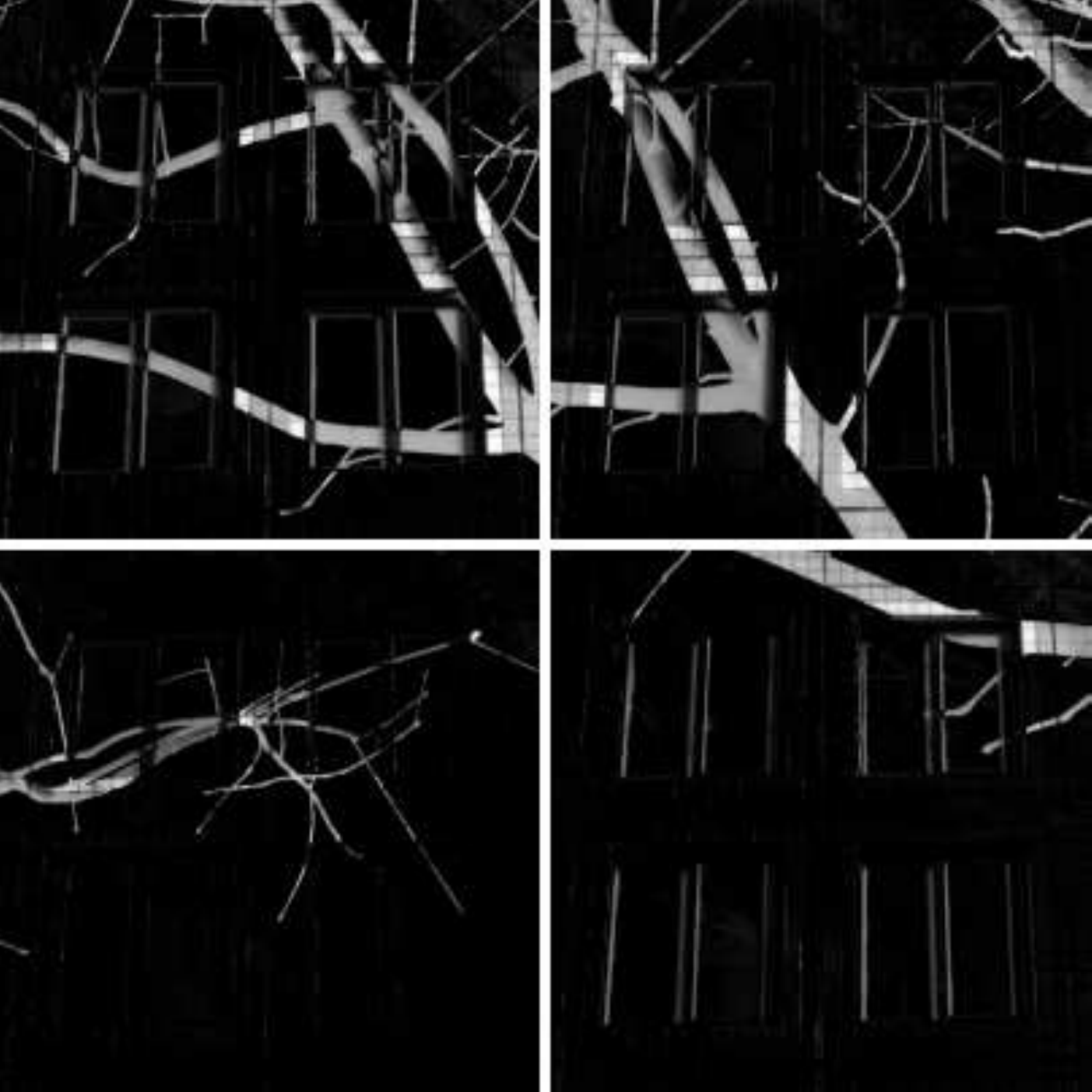}
\includegraphics[width=0.117\linewidth]{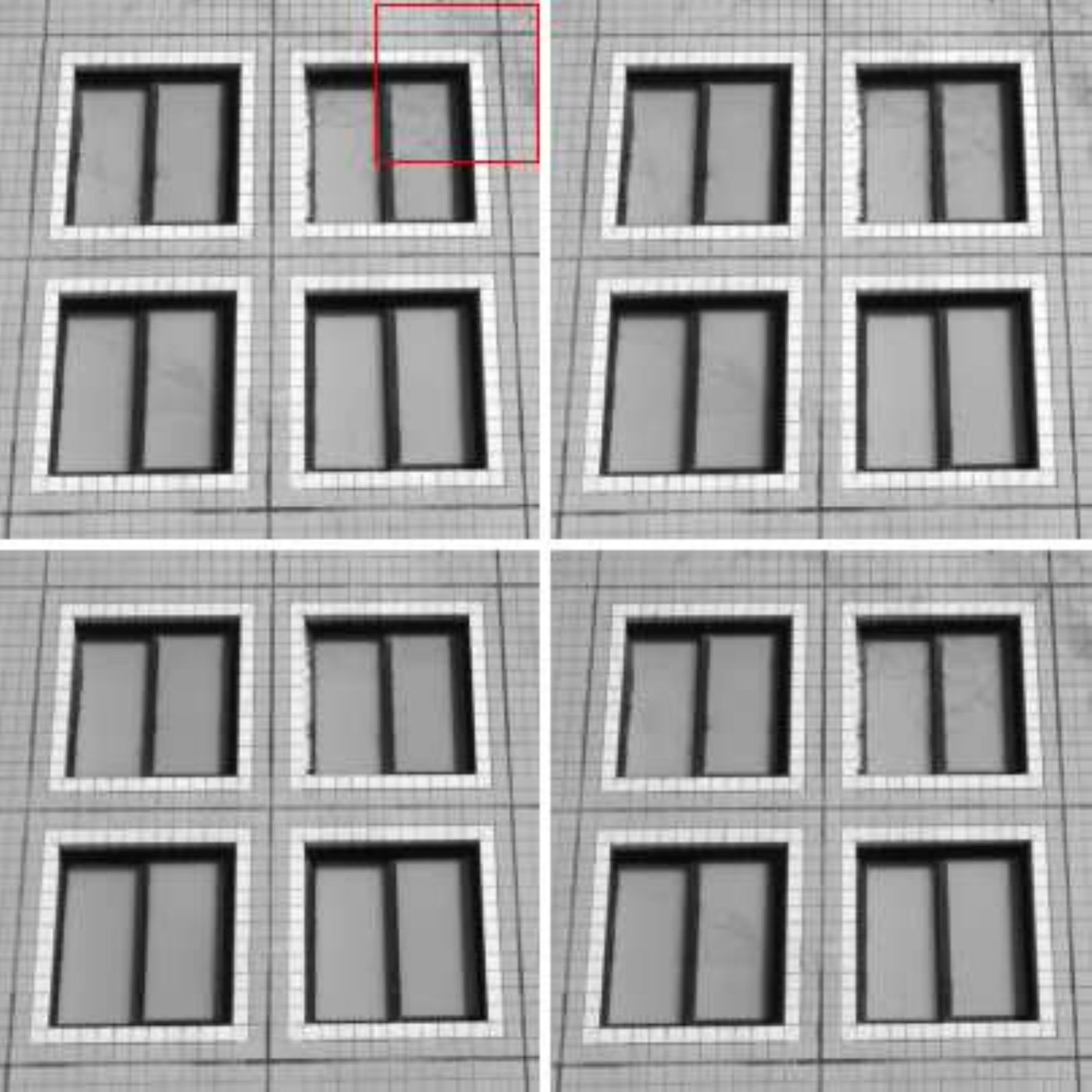}
\includegraphics[width=0.1209\linewidth]{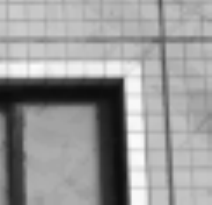}\;\;\;
\includegraphics[width=0.117\linewidth]{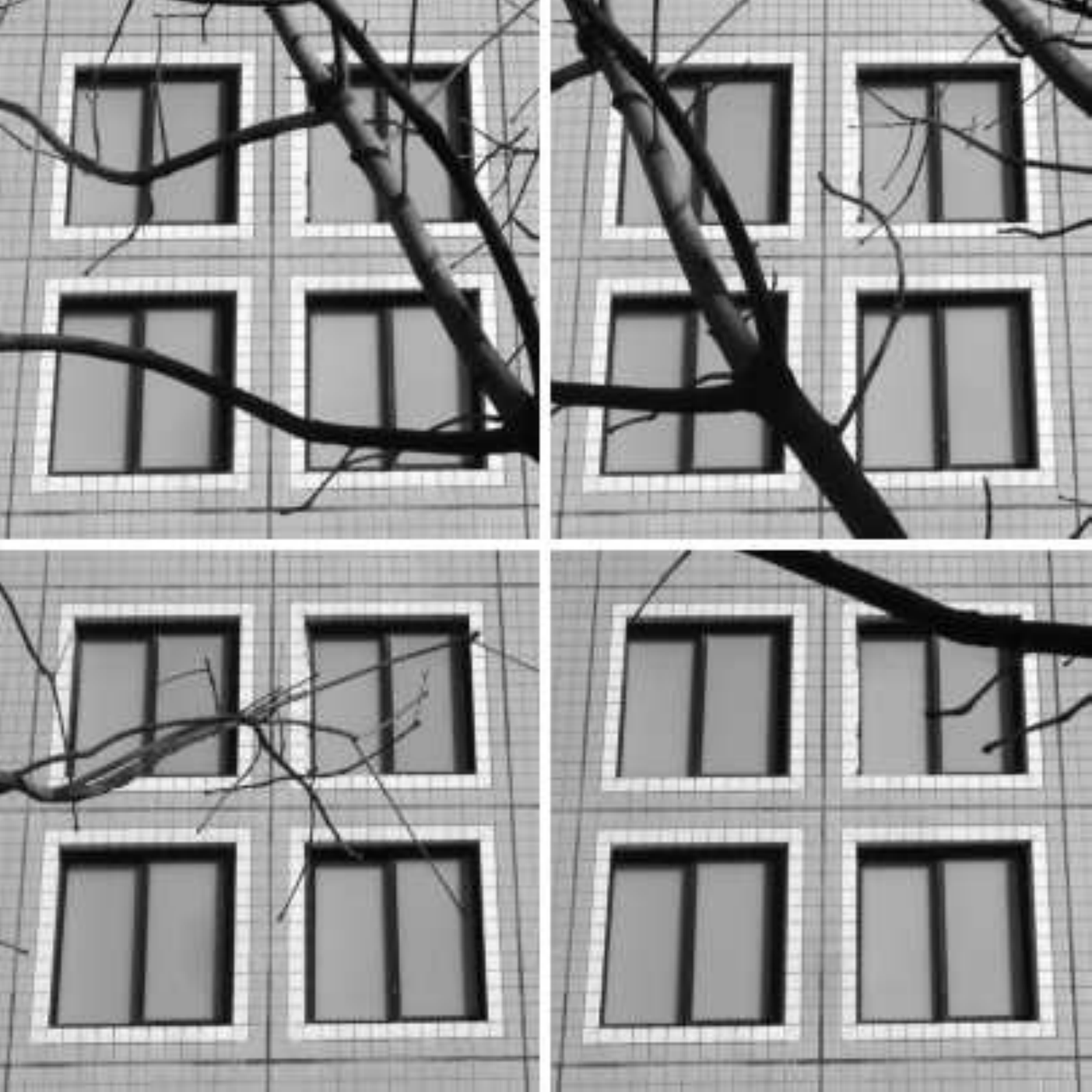}
\includegraphics[width=0.117\linewidth]{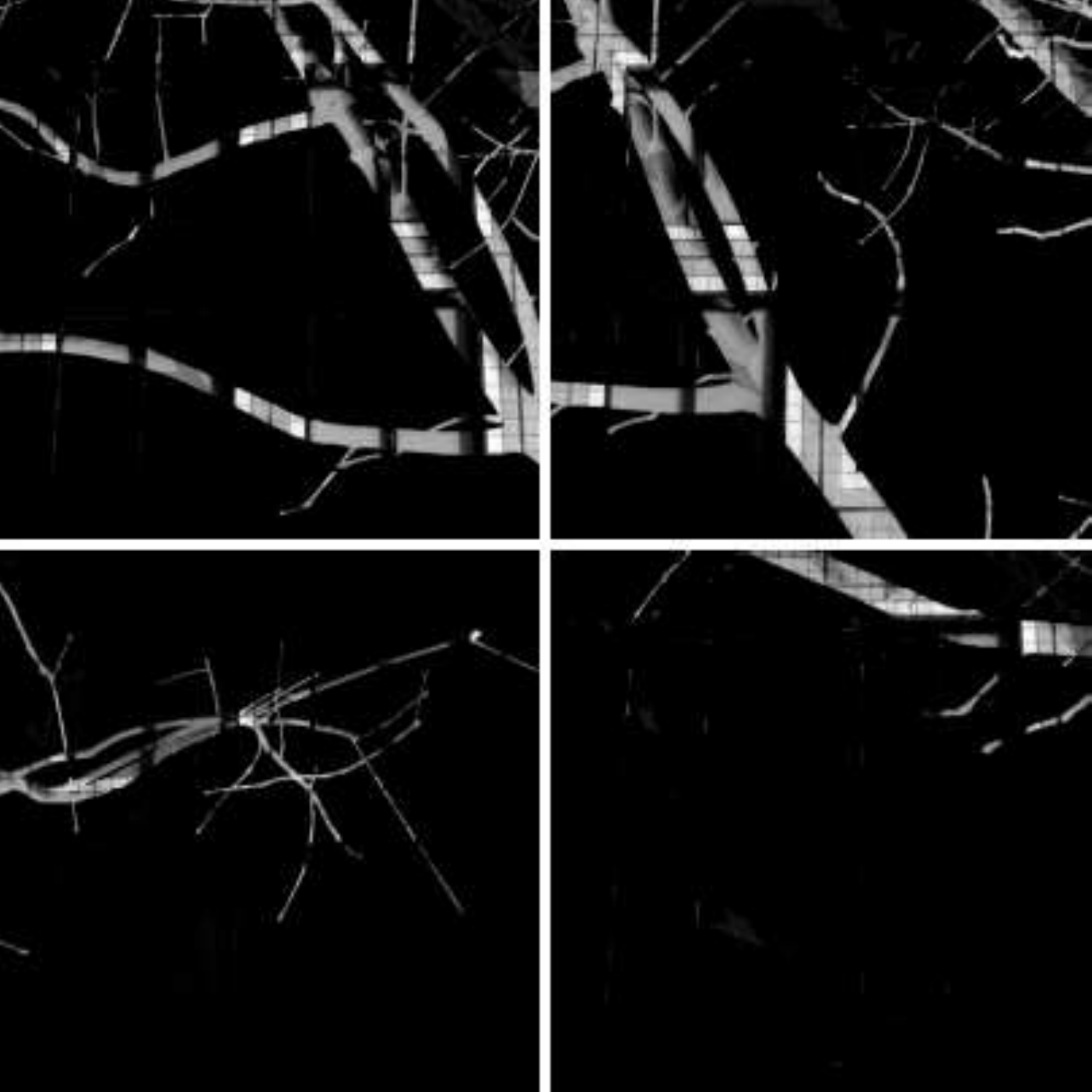}
\includegraphics[width=0.117\linewidth]{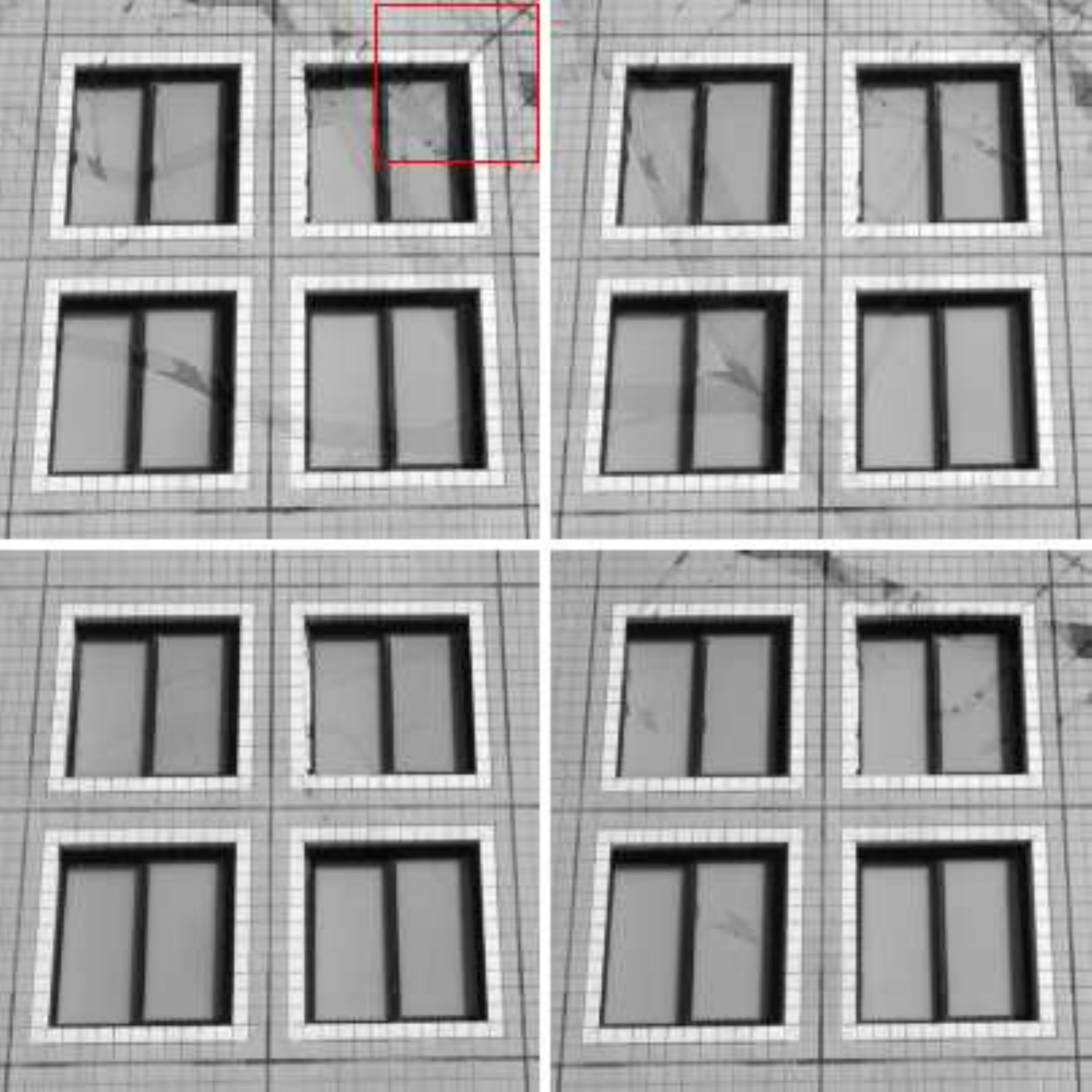}
\includegraphics[width=0.1209\linewidth]{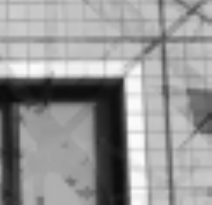}

\vspace{-0.5mm}

\subfigure[]{\includegraphics[width=0.117\linewidth]{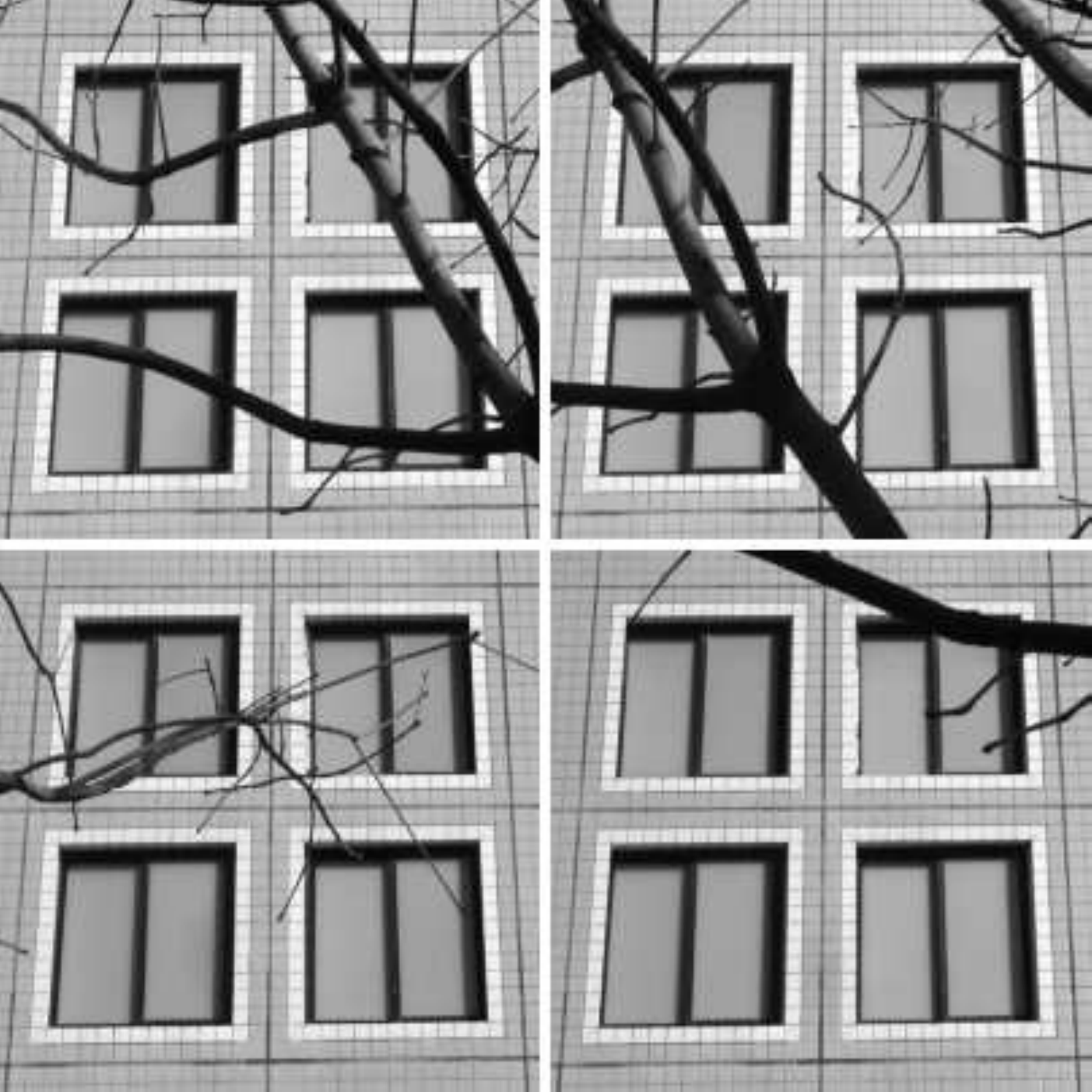}}
\subfigure[]{\includegraphics[width=0.117\linewidth]{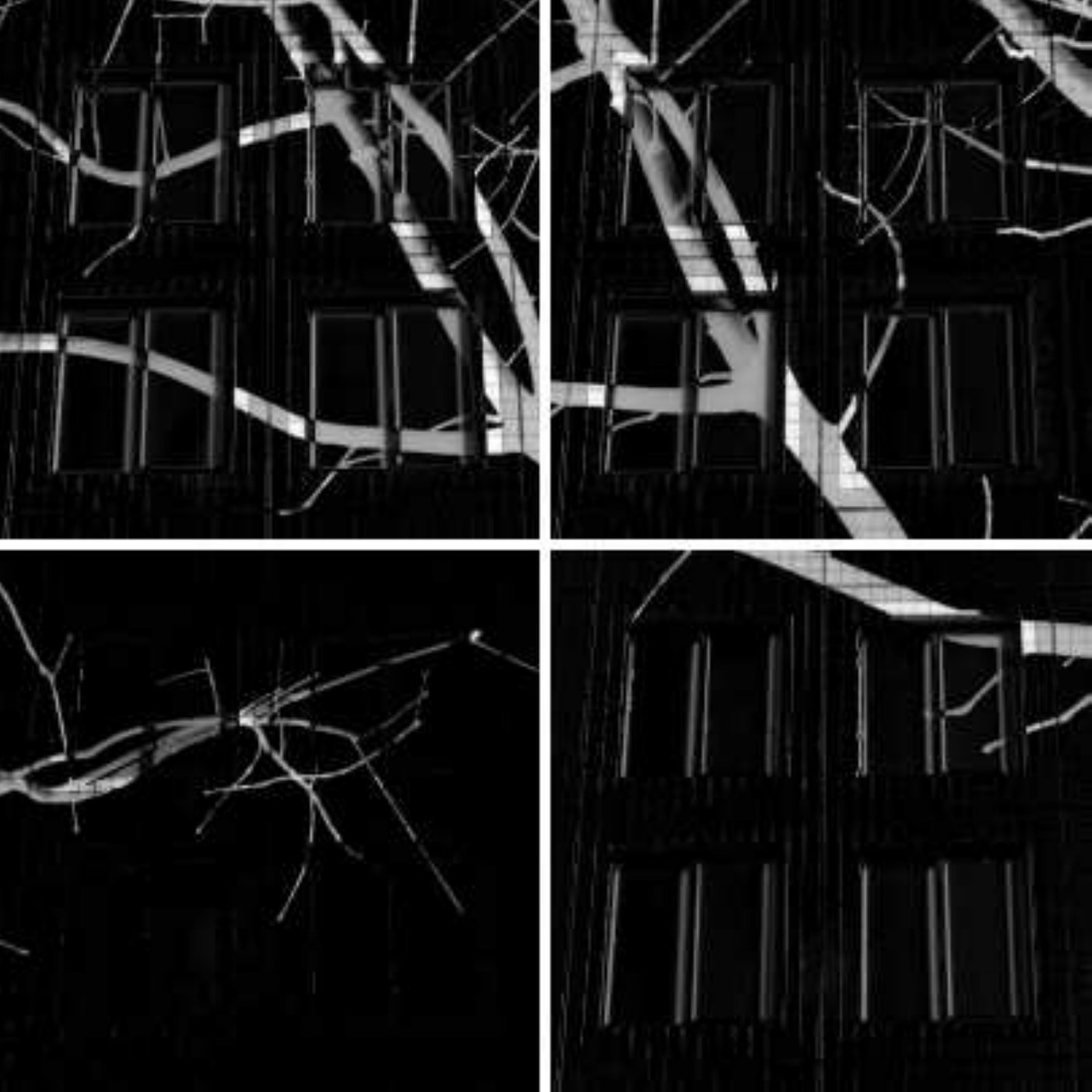}}
\subfigure[]{\includegraphics[width=0.117\linewidth]{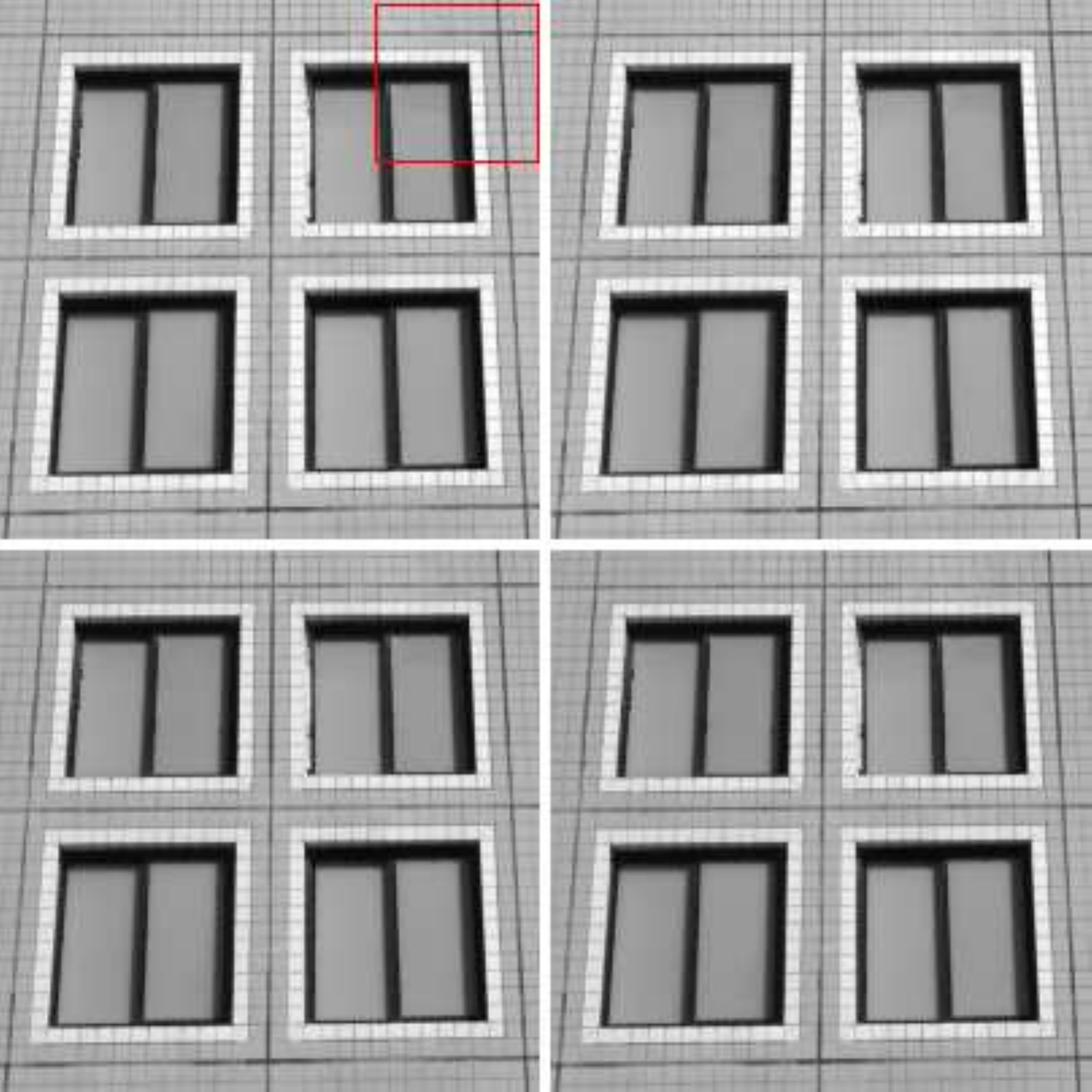}\label{fig101}}
\subfigure[]{\includegraphics[width=0.1209\linewidth]{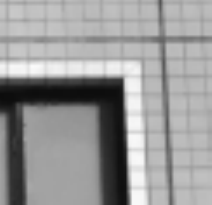}\label{fig102}}\;\;\;
\subfigure[]{\includegraphics[width=0.117\linewidth]{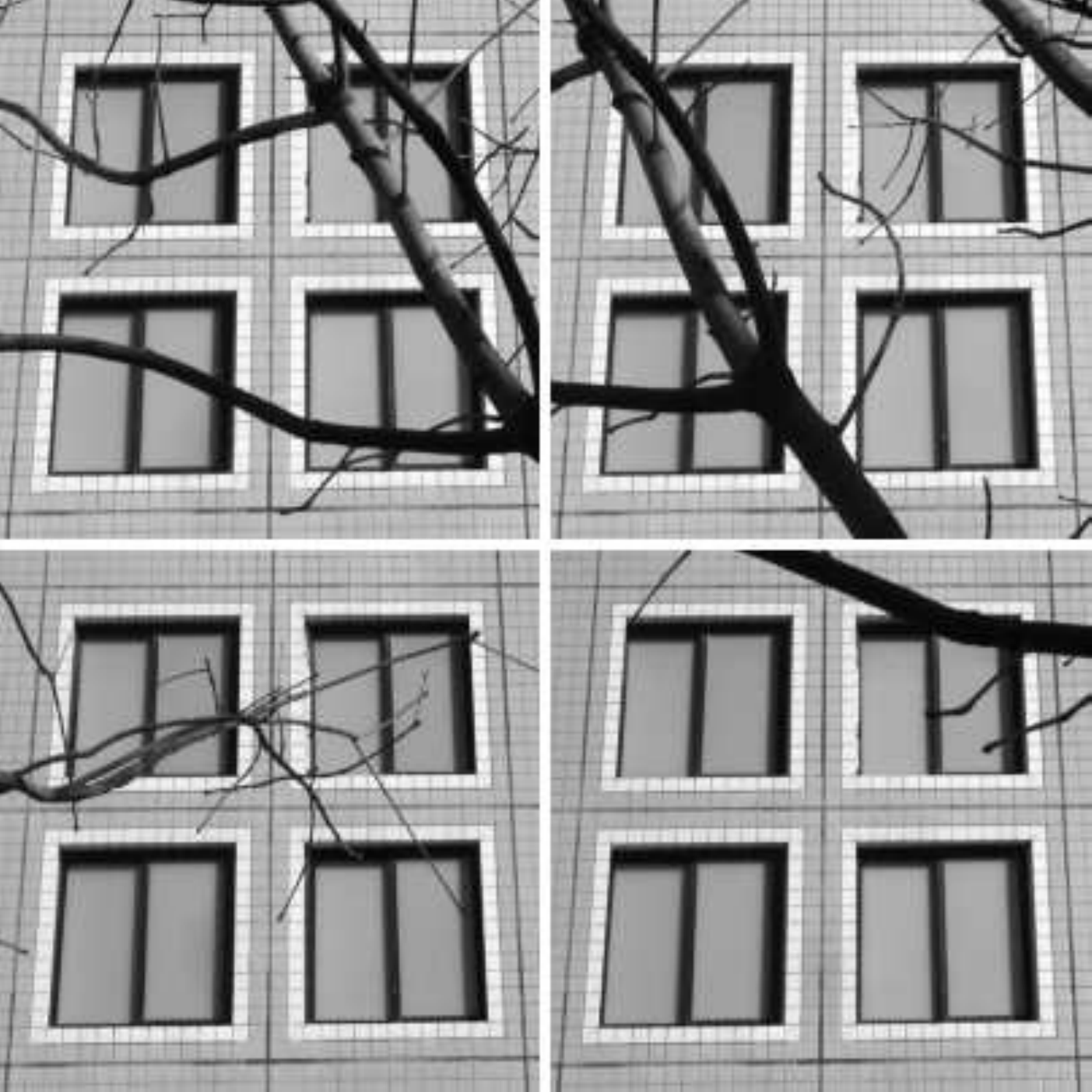}}
\subfigure[]{\includegraphics[width=0.117\linewidth]{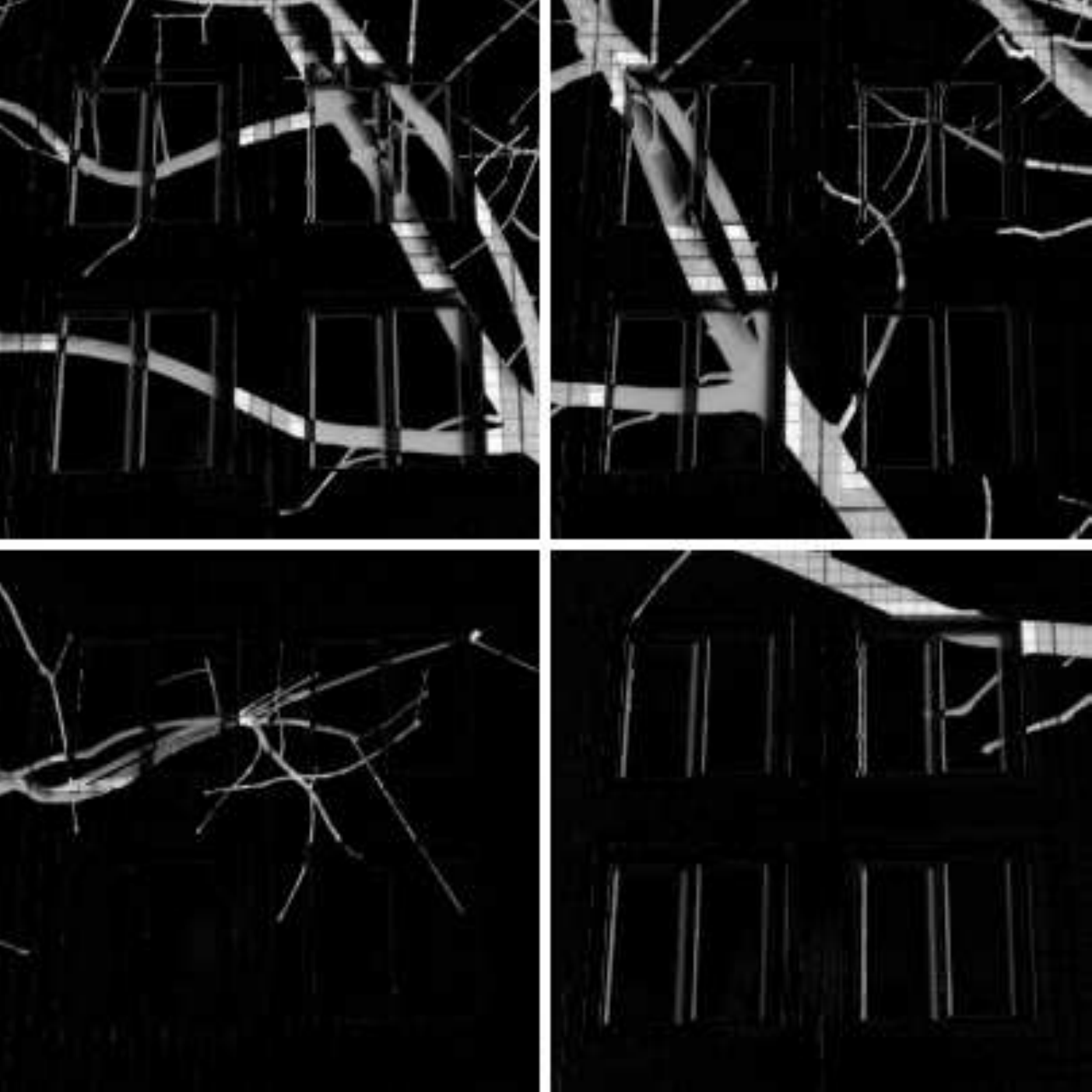}}
\subfigure[]{\includegraphics[width=0.117\linewidth]{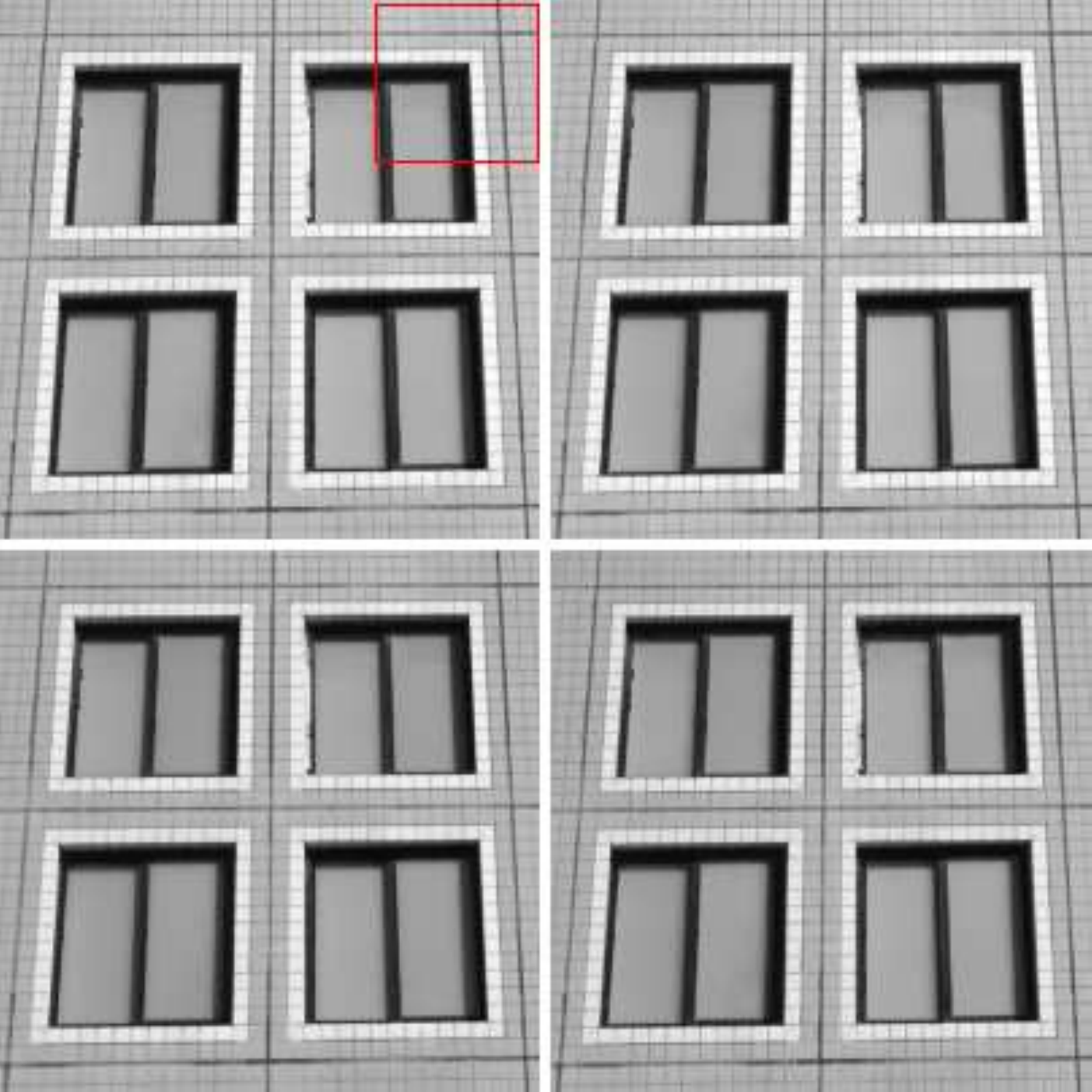}\label{fig103}}
\subfigure[]{\includegraphics[width=0.1209\linewidth]{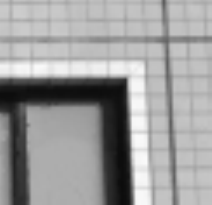}\label{fig104}}
\caption{Alignment results on the Windows data set used in~\cite{peng:rasl}. The first row: the results generated by RASL~\cite{peng:rasl} (left) and PSVT~\cite{oh:psvt} (right); The second row: the results generated by our $(\textup{S+L})_{1/2}$ (left) and $(\textup{S+L})_{2/3}$ (right) methods, respectively. (a) and (e): Aligned images; (b) and (f): Sparse occlusions; (c) and (g): Low-rank components; (d) and (h): Close-up (see red box in the left).}
\label{fig10}
\end{figure*}

\subsubsection{Image Alignment}
We also study the performance of our methods in the application of robust image alignment: Given $n$ images $\{I_{1},\ldots,I_{n}\}$ of an object of interest, the image alignment task aims to align them using a fixed geometric transformation model, such as affine~\cite{peng:rasl}. For this problem, we search for a transformation $\tau\!=\!\{\tau_{1},\ldots,\tau_{n}\}$ and write $D\!\circ\!\tau\!=\![\textup{vec}(I_{1}\!\circ\!\tau_{1})|\cdots|\textup{vec}(I_{n}\!\circ\!\tau_{n})]\!\in\! \mathbb{R}^{m\times n}$. In order to robustly align the set of linearly correlated images despite sparse outliers, we consider the following double nuclear norm regularized model
\begin{equation}\label{equ61}
\min_{U,V,S,\tau} \frac{\lambda}{2}(\|U\|_{*}\!+\!\|V\|_{*}\!)\!+\!\|S\|^{{1}/{2}}_{\ell_{1/2}},\;\textup{s.t.}, D\!\circ\!\tau\!=\!UV^{T}\!+\!S.
\end{equation}
Alternatively, our Frobenius/nuclear norm penalty can also be used to address the image alignment problem above.

We first test both our methods on the Windows data set (which contains 16 images of a building, taken from various viewpoints by a perspective camera, and with occlusions due to tree branches) used in~\cite{peng:rasl} and report the aligned results of RASL\footnote{\url{http://perception.csl.illinois.edu/matrix-rank/}}~\cite{peng:rasl}, PSVT~\cite{oh:psvt} and our methods in Fig.\ \ref{fig10}, from which it is clear that, compared with RASL and PSVT, both our methods not only robustly align the images, correctly detect and remove the occlusion, but also achieve much better performance in terms of low-rank components, as shown in Figs.\ \ref{fig102} and \ref{fig104}, which give the close-up views of the red boxes in Figs.\ \ref{fig101} and \ref{fig103}, respectively.

\subsubsection{Image Inpainting}
\label{sec623}
Finally, we applied the defined D-N and F-N penalties to image inpainting. As shown by Hu \emph{et al.}~\cite{hu:tnnr}\footnote{\url{https://sites.google.com/site/zjuyaohu/}}, the images of natural scenes can be viewed as approximately low rank matrices. Naturally, we consider the following D-N penalty regularized least squares problem:
\begin{equation}\label{equ62}
\min_{U,V,L} \!\frac{\lambda}{2}(\|U\|_{*}\!+\!\|V\|_{*}\!)\!+\!\frac{1}{2}\|\mathcal{P}_{\Omega}(L\!-\!D)\|^{2}_{F},\;\textup{s.t.}, L\!=\!U\!V^{T}.
\end{equation}
The F-N penalty regularized model and the corresponding ADMM algorithms for solving both models are provided in the Supplementary Materials. We compared our methods with one nuclear norm solver~\cite{toh:apg}, one weighted nuclear norm method~\cite{gu:wnnm,gu:wnnmj}, two truncated nuclear norm methods~\cite{hu:tnnr,oh:psvt}, and one Schatten-$q$ quasi-norm method~\cite{lu:lrm}\footnote{As suggested in~\cite{lu:lrm}, we chose the $\ell_{q}$-norm penalty, where $q$ was chosen from the range of $\{0.1,0.2,\ldots,1\}$.}. Since both of the ADMM algorithms in~\cite{hu:tnnr,oh:psvt} have very similar performance as shown in~\cite{oh:psvt}, we only report the results of~\cite{hu:tnnr}. For fair comparison, we set the same values to the parameters $d$, $\rho$ and $\mu_{0}$ for both our methods and~\cite{hu:tnnr,oh:psvt}, e.g., $d\!=\!9$ as in~\cite{hu:tnnr}.

\begin{figure}[t]
\centering
\includegraphics[width=0.105\linewidth]{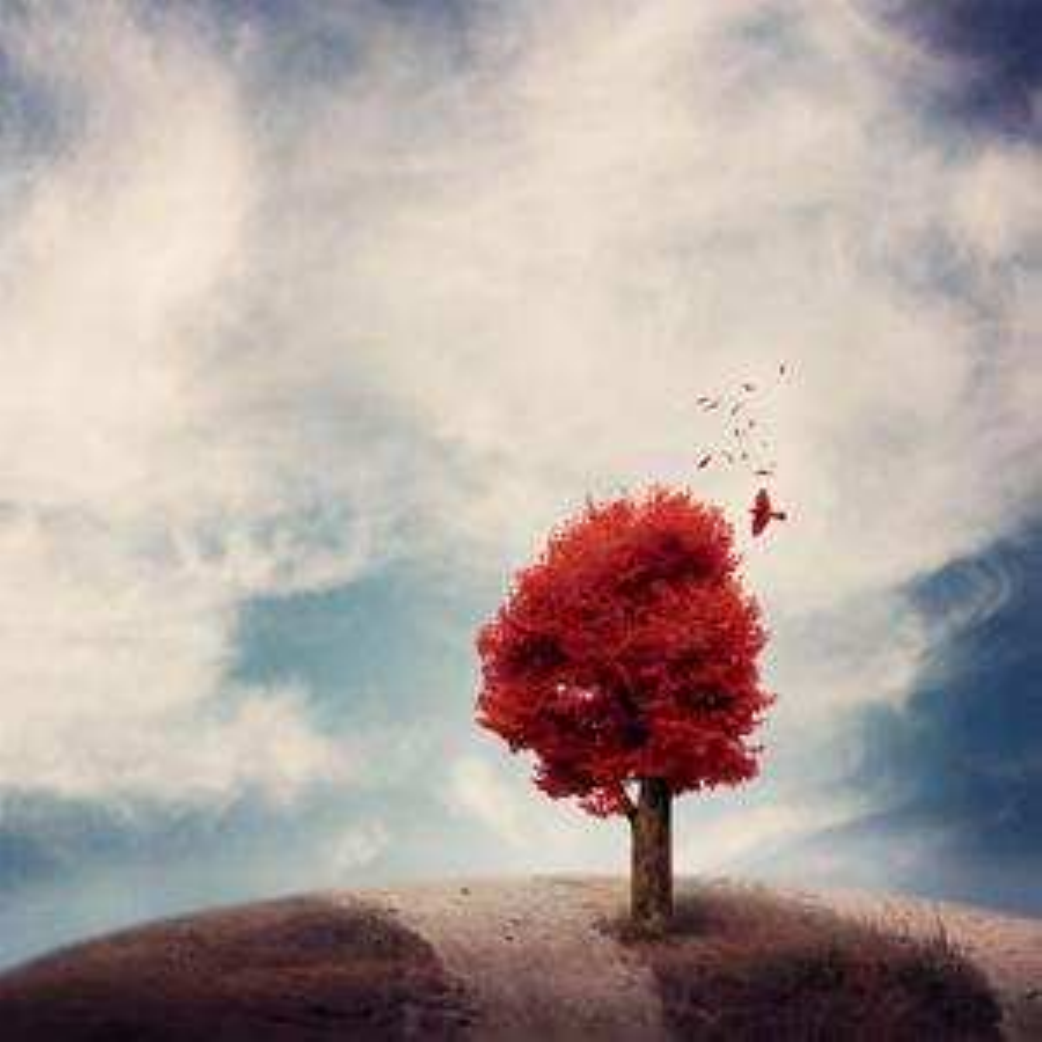}\!
\includegraphics[width=0.105\linewidth]{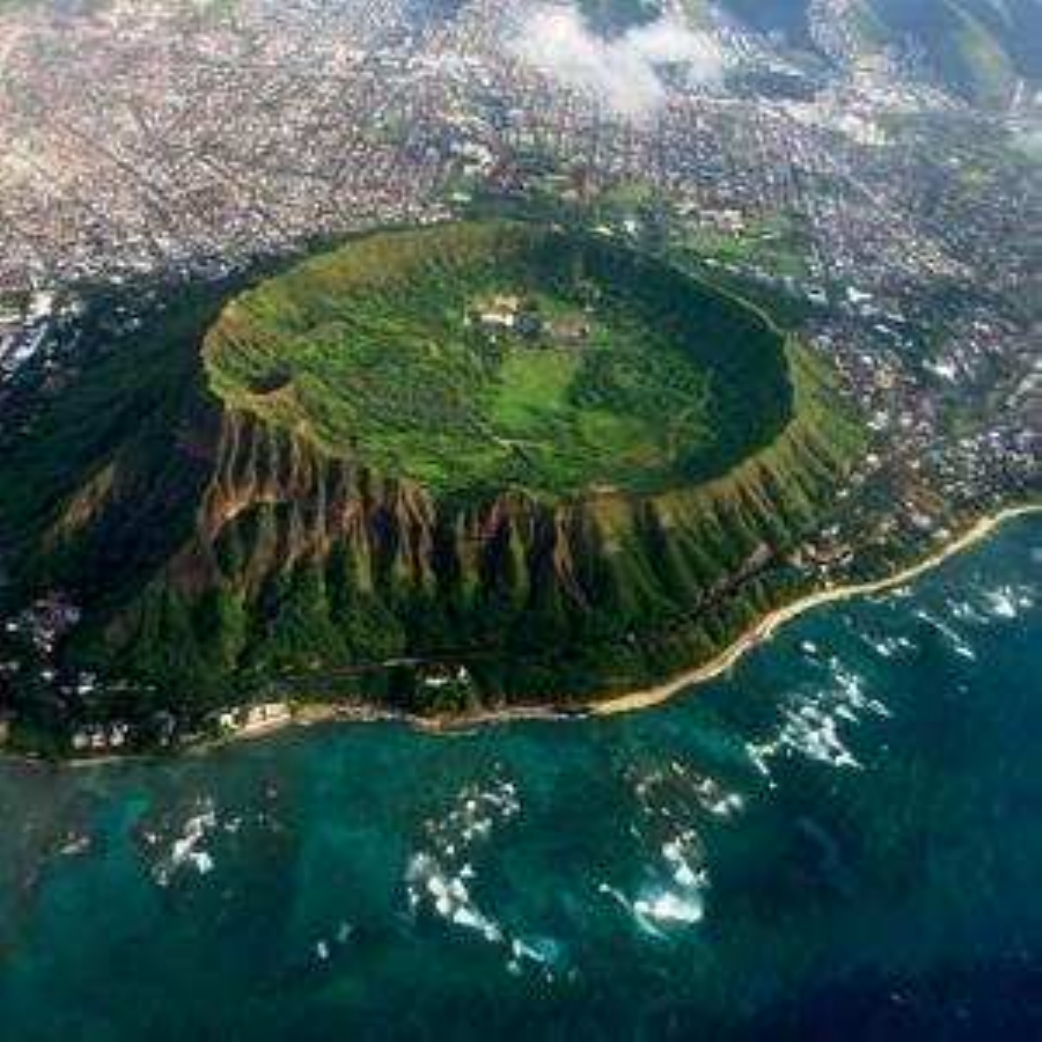}\!
\includegraphics[width=0.105\linewidth]{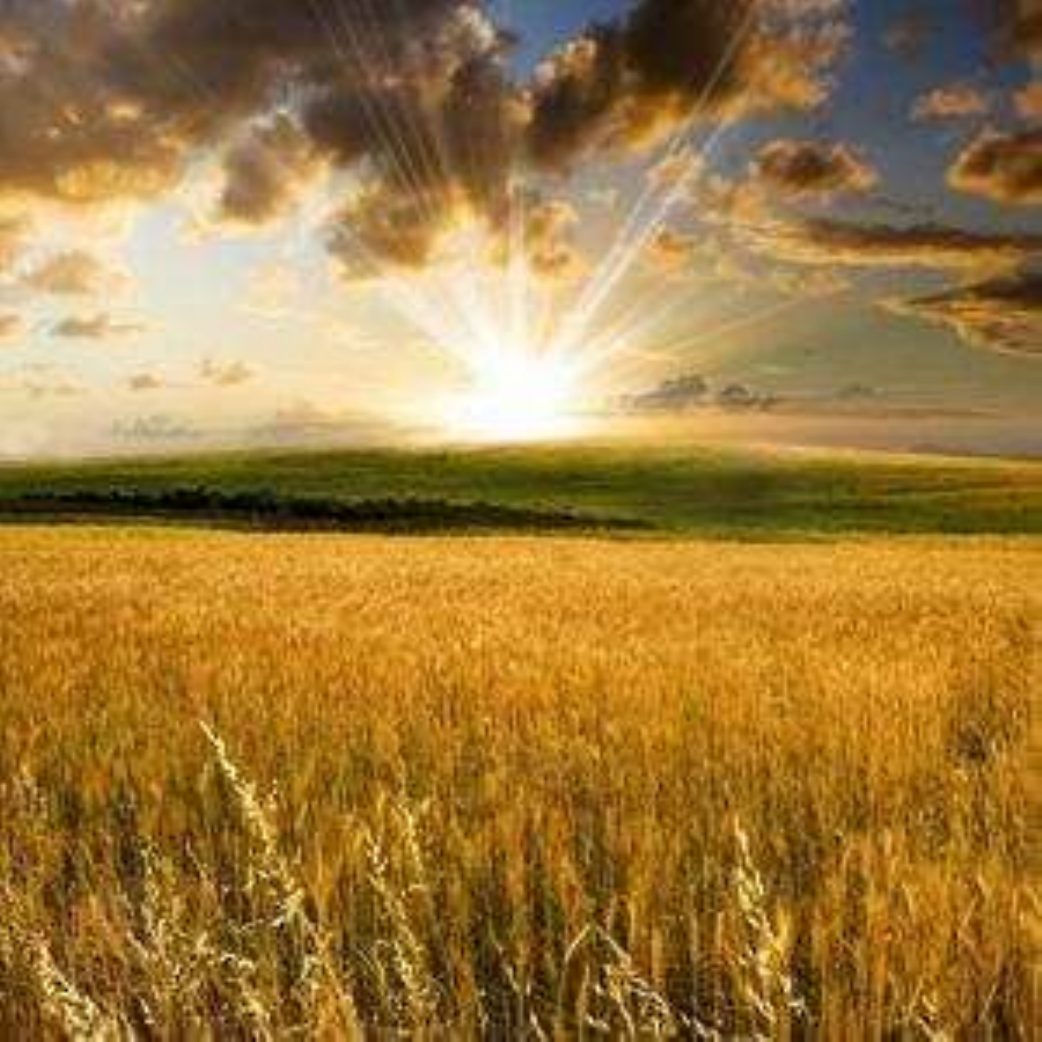}\!
\includegraphics[width=0.105\linewidth]{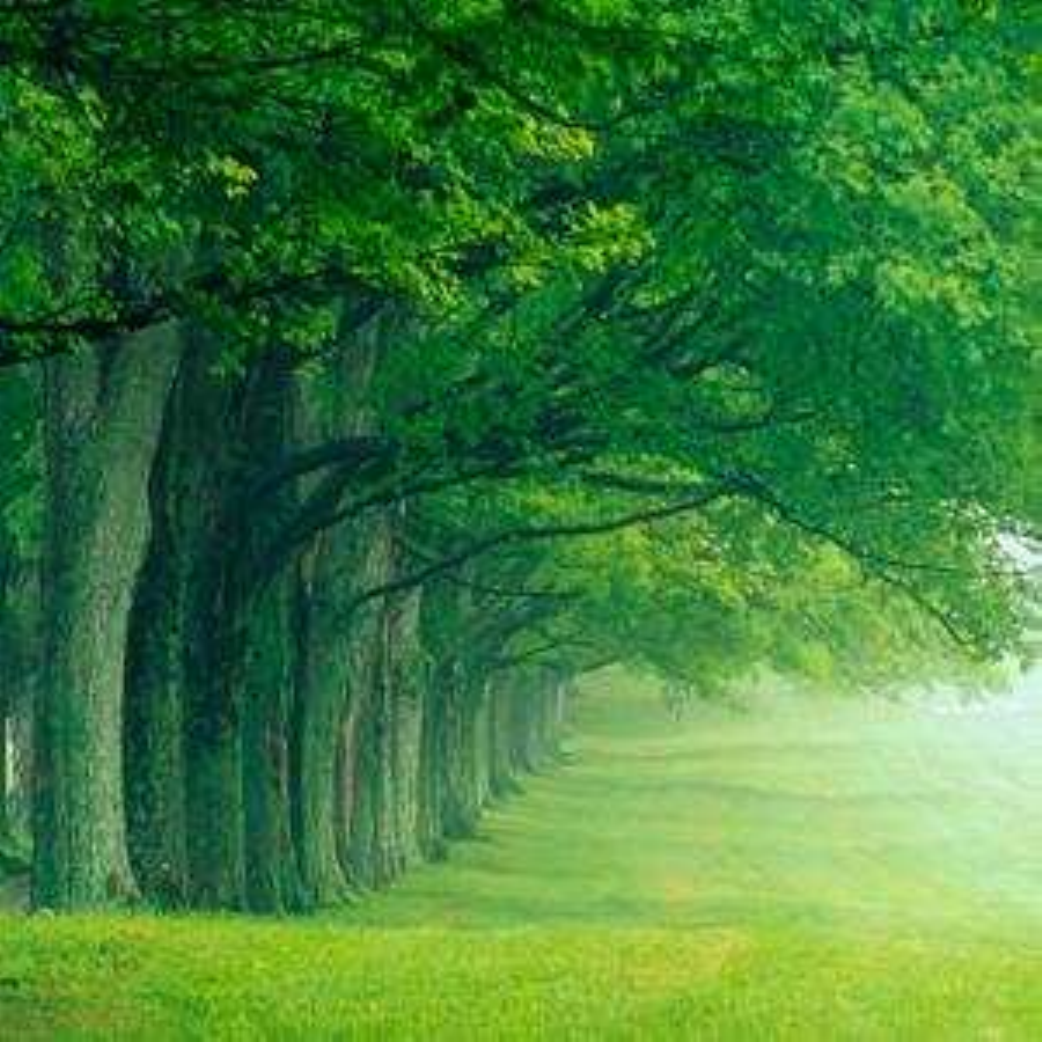}\!
\includegraphics[width=0.105\linewidth]{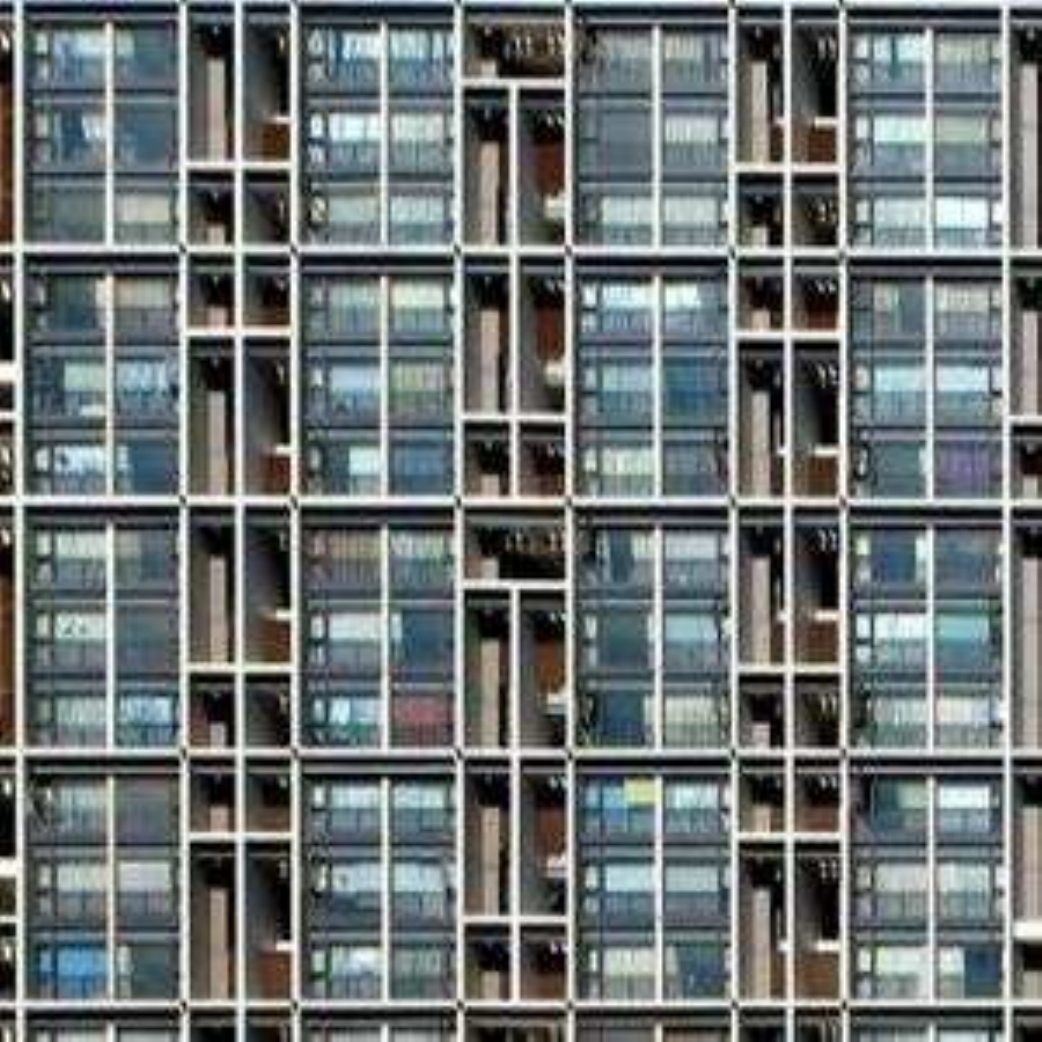}\!
\includegraphics[width=0.1543\linewidth]{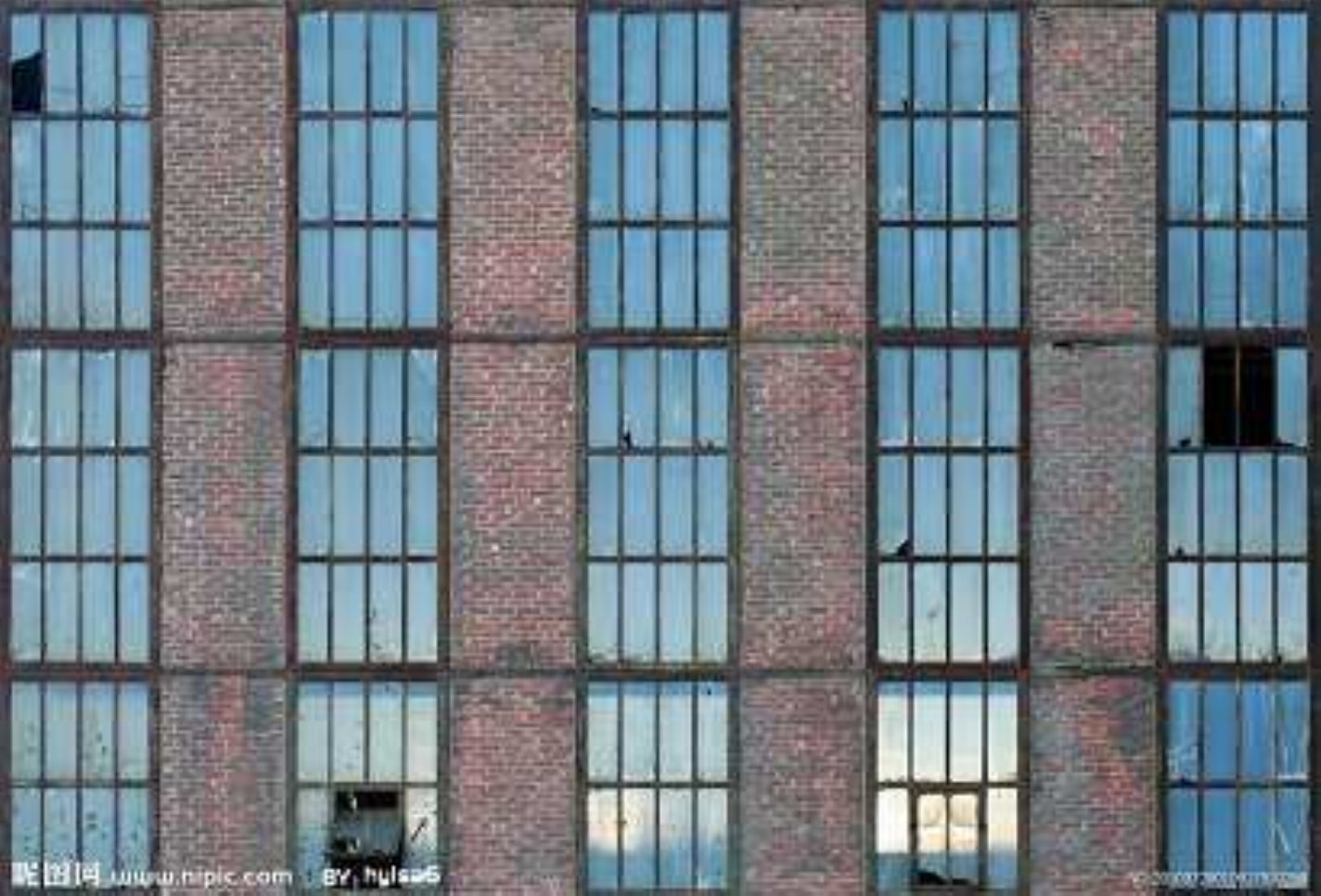}\!
\includegraphics[width=0.1746\linewidth]{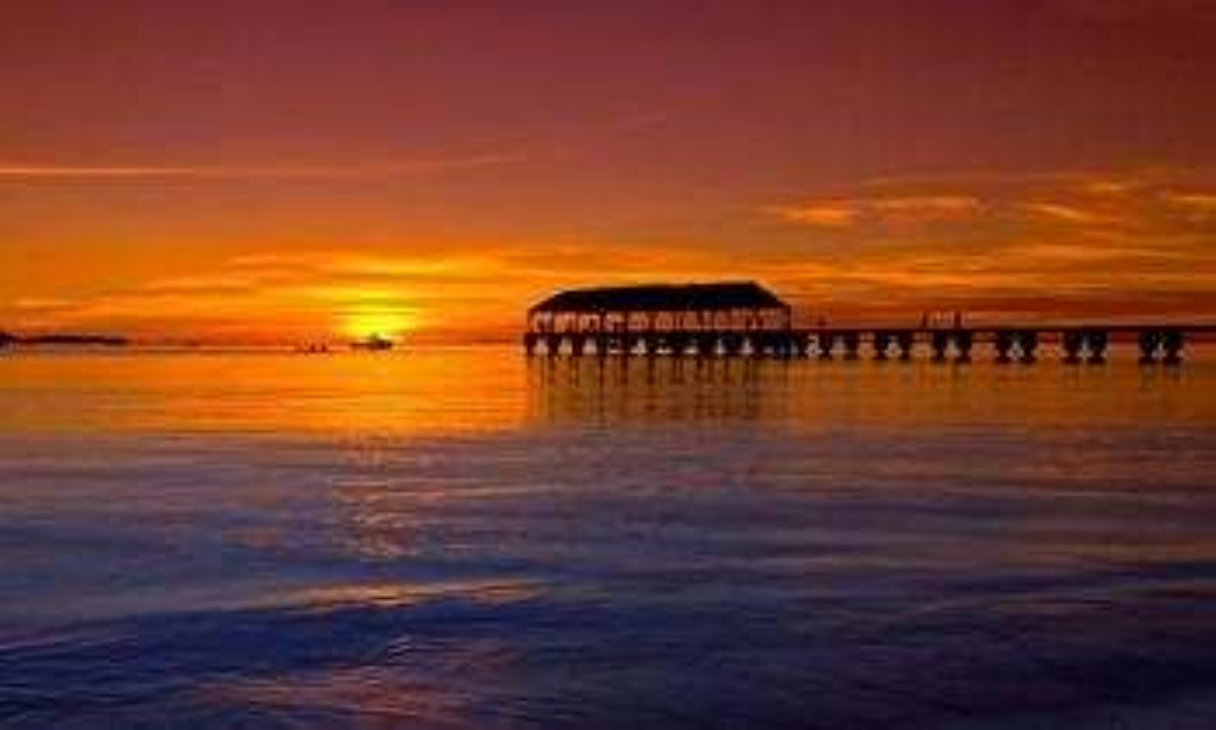}\!
\includegraphics[width=0.105\linewidth]{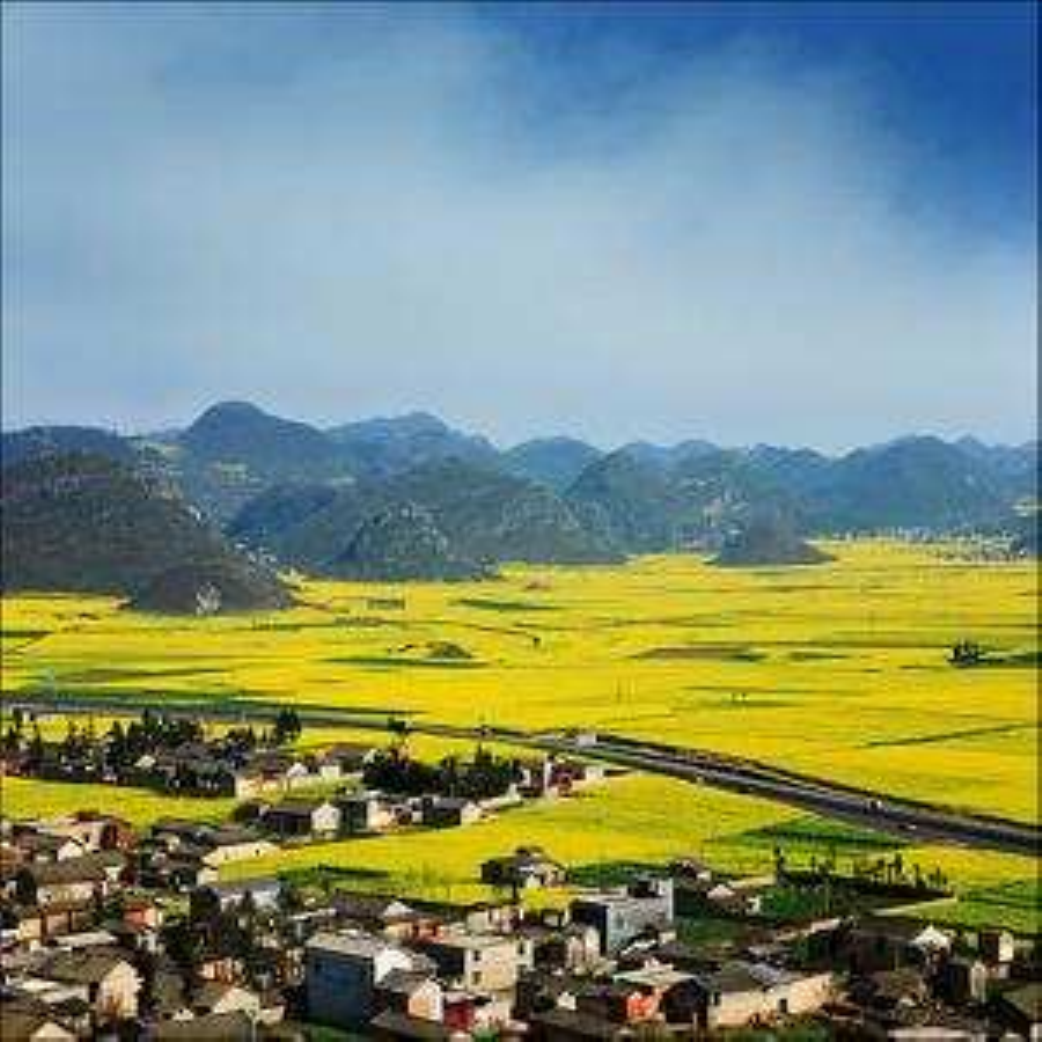}

\vspace{1mm}
\includegraphics[width=0.98\linewidth]{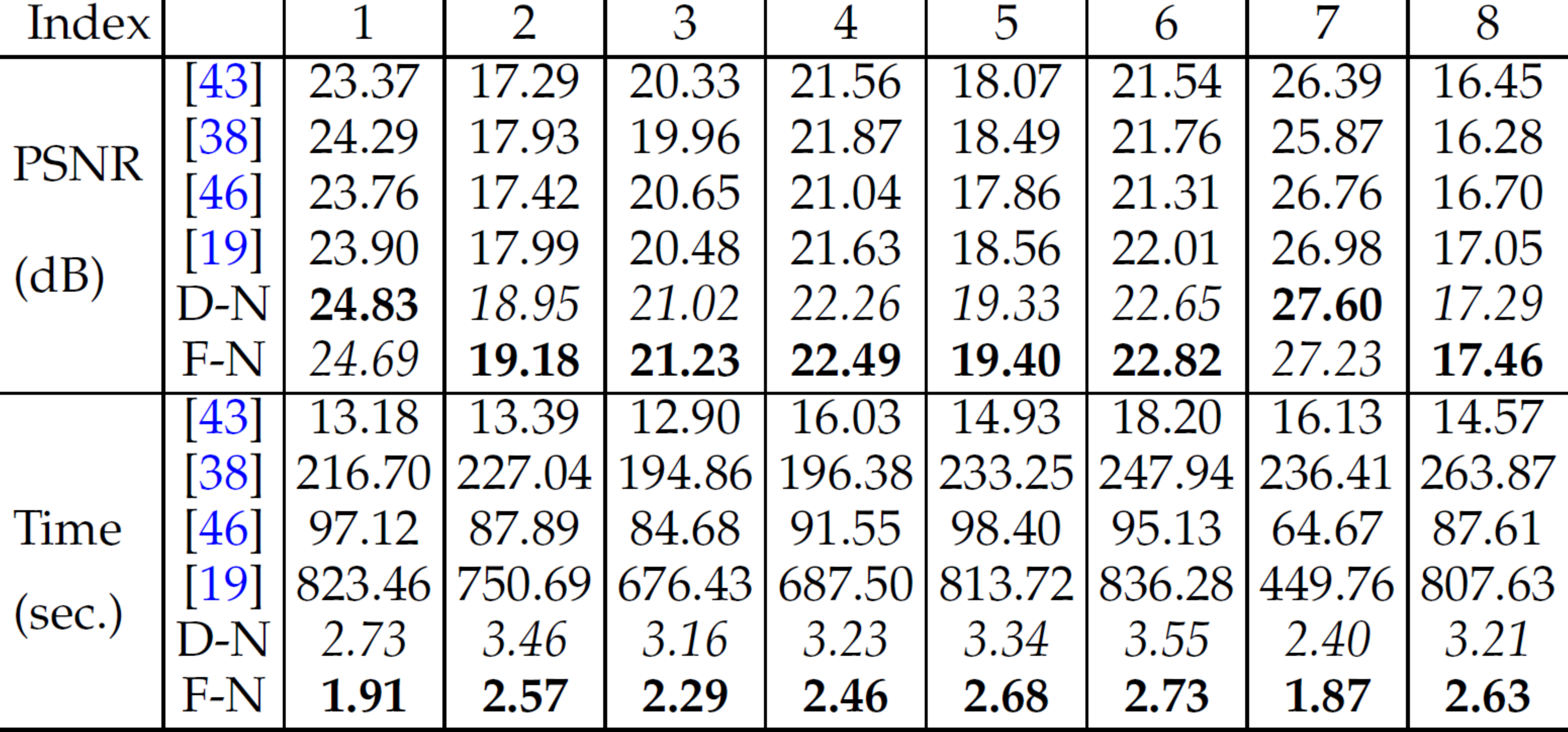}
\caption{The natural images used in image inpainting~\cite{hu:tnnr} (top), and quantitative comparison of inpainting results (bottom).}
\label{fig11}
\end{figure}

Fig.\ \ref{fig11} shows the 8 test images and some quantitative results (including average PSNR and running time) of all those methods with 85\% random missing pixels. We also show the inpainting results of different methods for random mask of 80\% missing pixels in Fig.\ \ref{fig12} (see the Supplementary Materials for more results with different missing ratios and rank parameters). The results show that both our methods consistently produce much better PSNR results than the other methods in all the settings. As analyzed in~\cite{shang:snm, shang:tsnm}, our D-N and F-N penalties not only lead to two scalable optimization problems, but also require significantly fewer observations than traditional nuclear norm solvers, e.g., \cite{toh:apg}. Moreover, both our methods are much faster than the other methods, in particular, more than 25 times faster than the methods~\cite{lu:lrm,hu:tnnr,gu:wnnmj}.

\begin{figure*}[t]
\centering
\subfigure[Ground truth]{\includegraphics[width=0.243\linewidth]{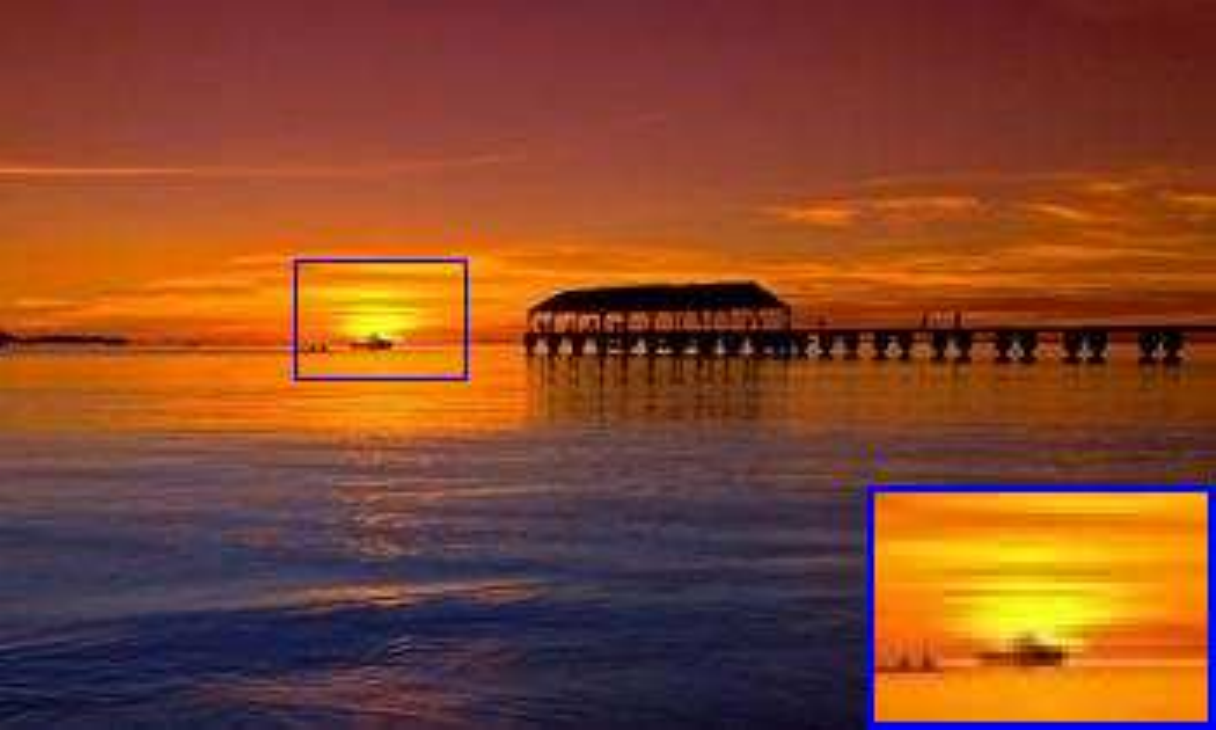}}\,
\subfigure[Input image]{\includegraphics[width=0.243\linewidth]{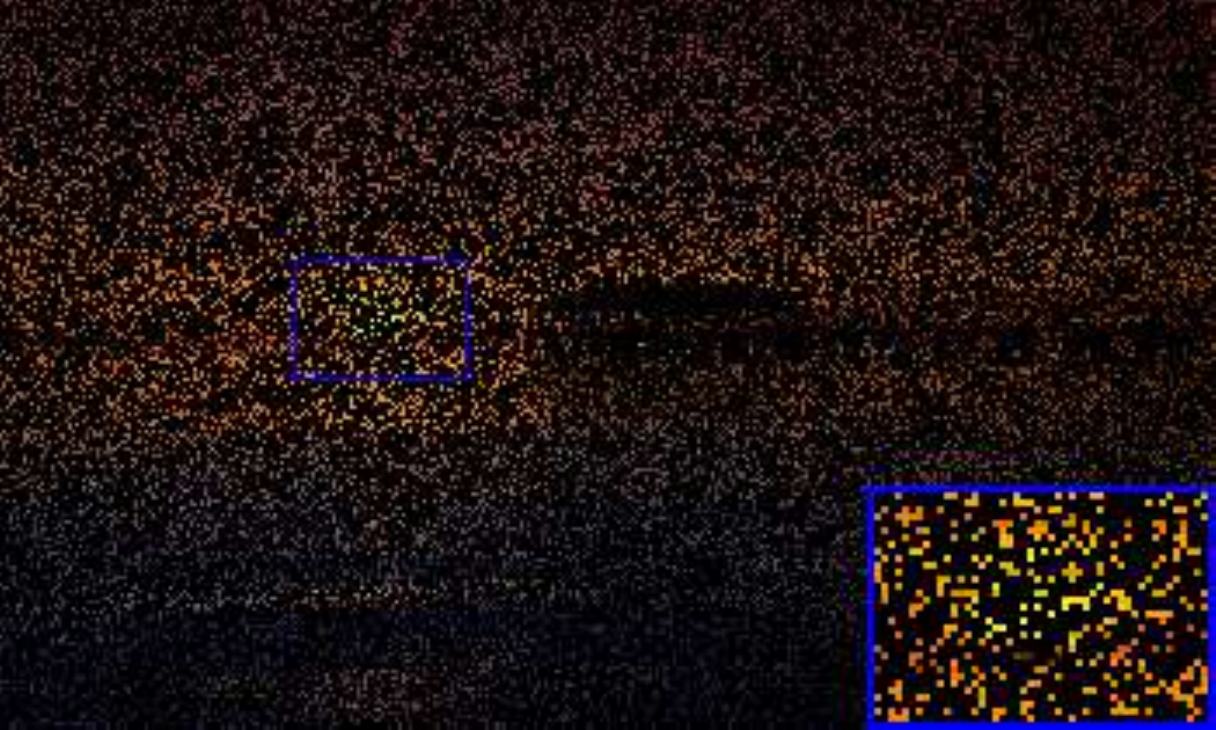}}\,
\subfigure[APGL~\cite{toh:apg} (PSNR: 26.63 dB)]{\includegraphics[width=0.243\linewidth]{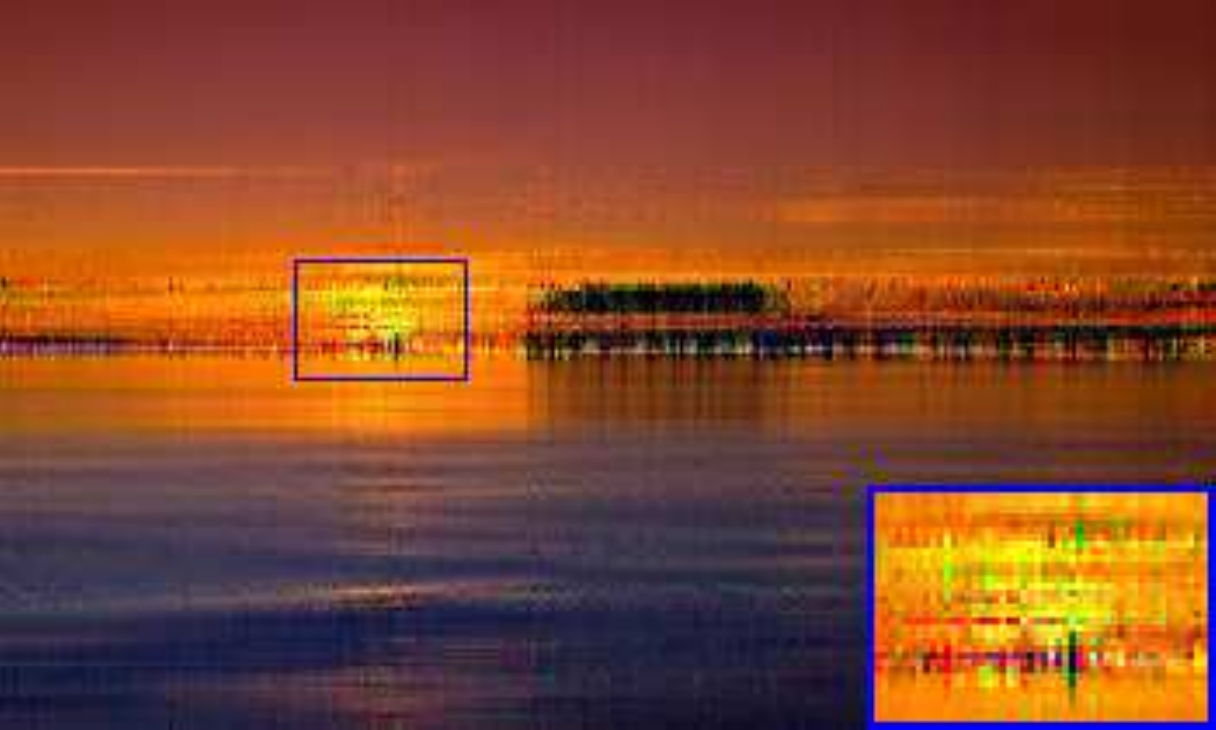}}\,
\subfigure[TNNR~\cite{hu:tnnr} (PSNR: 27.47 dB)]{\includegraphics[width=0.243\linewidth]{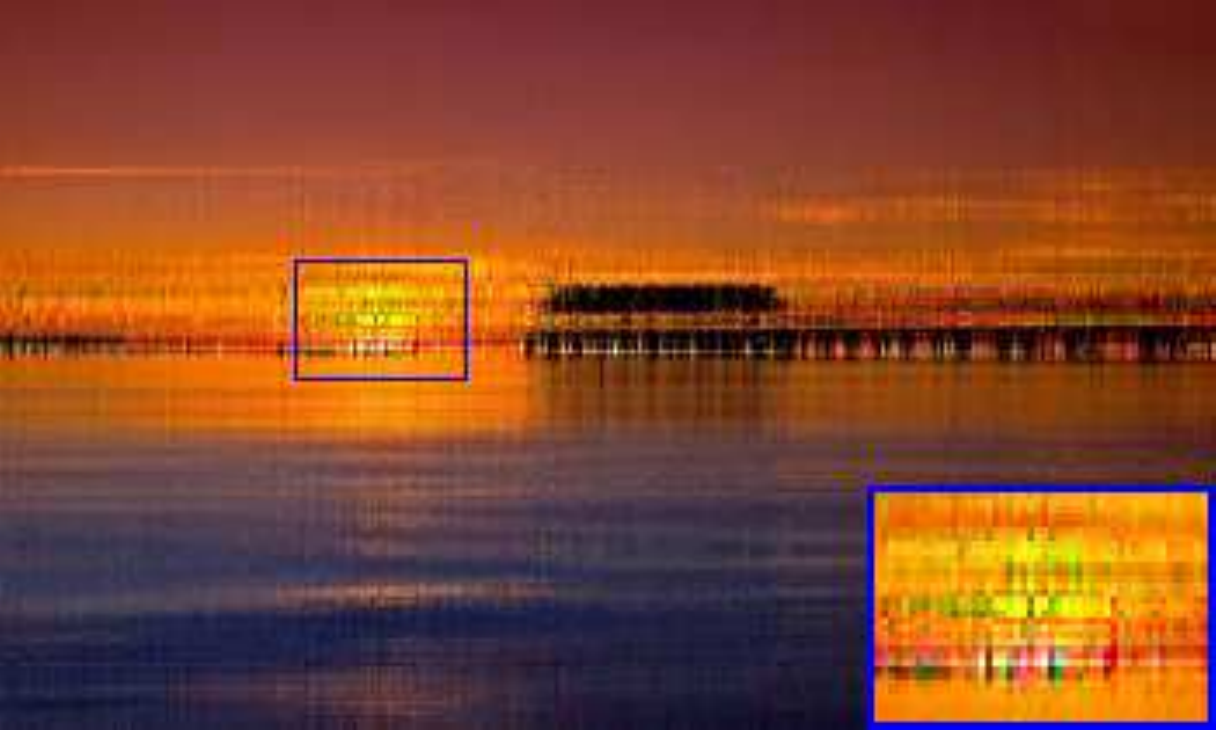}}
\subfigure[WNNM~\cite{gu:wnnmj} (PSNR: 27.95 dB)]{\includegraphics[width=0.243\linewidth]{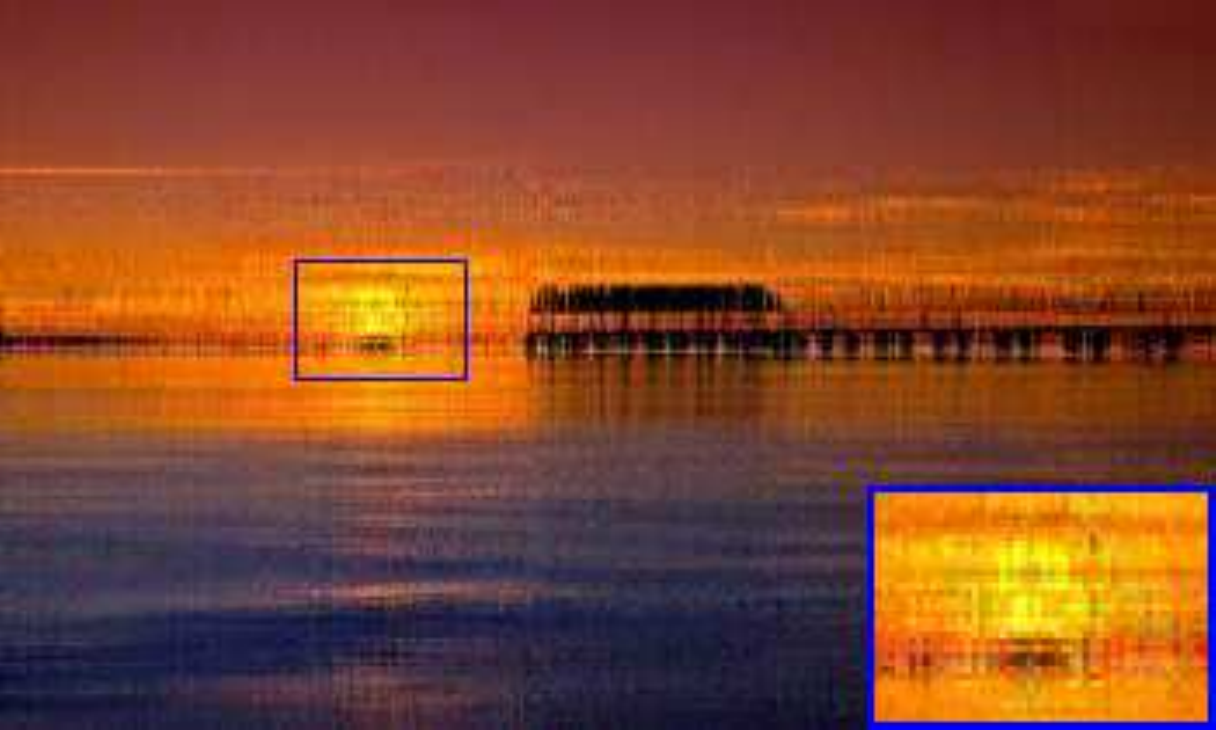}}\,
\subfigure[IRNN~\cite{lu:lrm} (PSNR: 26.86 dB)]{\includegraphics[width=0.243\linewidth]{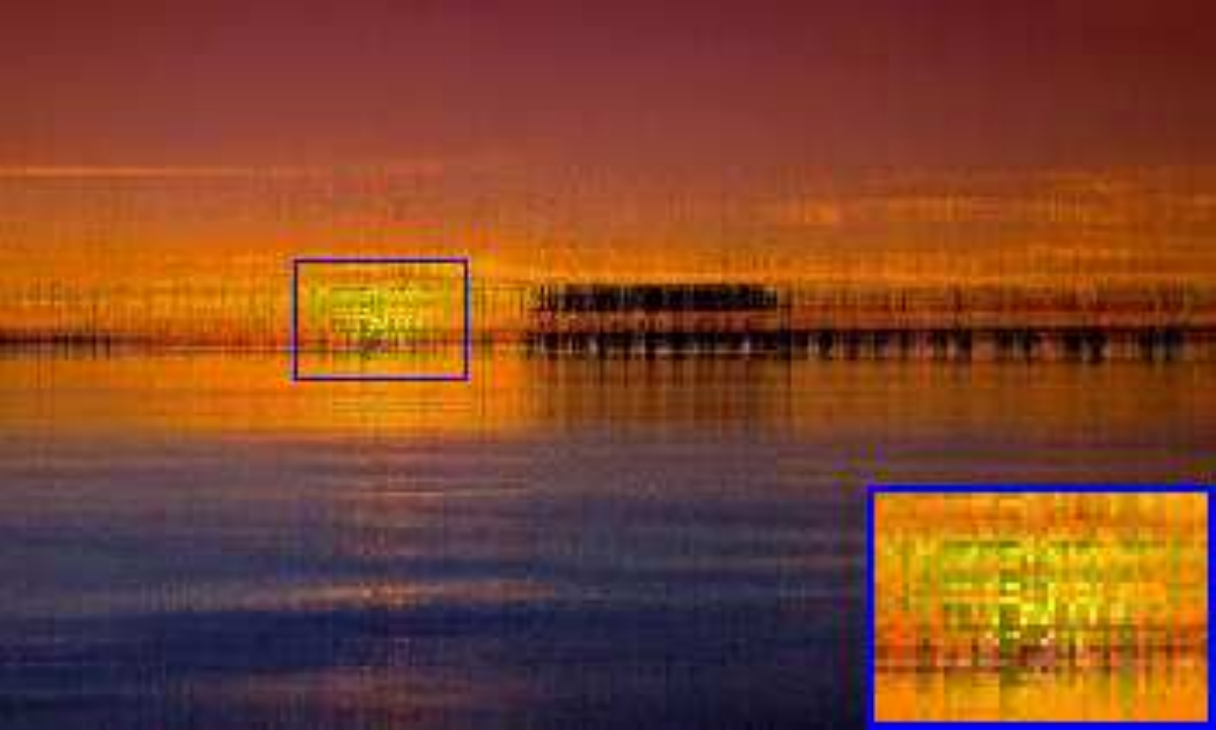}}\,
\subfigure[D-N (PSNR: 28.79 dB)]{\includegraphics[width=0.243\linewidth]{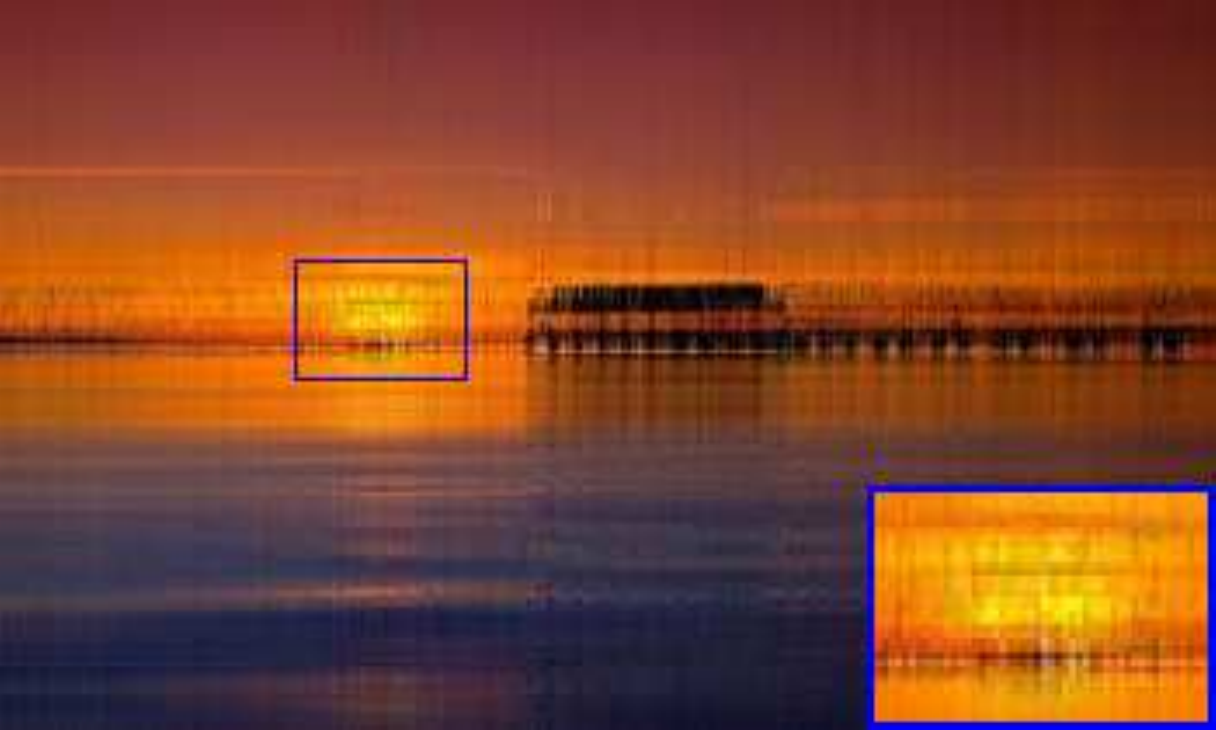}}\,
\subfigure[F-N (PSNR: 28.64 dB)]{\includegraphics[width=0.243\linewidth]{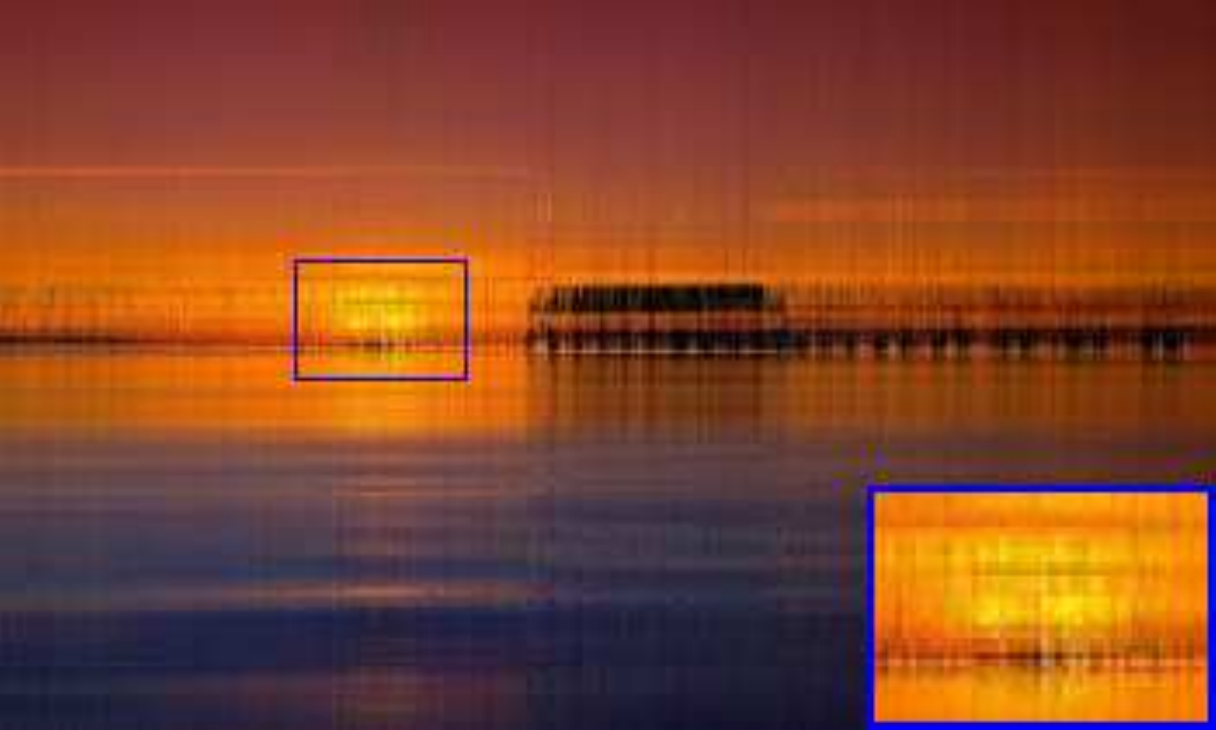}}
\caption{Image inpainting results of different methods on Image 7 with 80\% random missing pixels (best viewed in colors).}
\label{fig12}
\end{figure*}

\section{Conclusions and Discussions}
\label{concl}
In this paper we defined the double nuclear norm and Frobenius/nuclear hybrid norm penalties, which are in essence the Schatten-$1/2$ and $2/3$ quasi-norms, respectively. To take advantage of the hyper-Laplacian priors of sparse noise/outliers and singular values of low-rank components, we proposed two novel tractable bilinear factor matrix norm penalized methods for low-level vision problems. Our experimental results show that both our methods can yield more accurate solutions than original Schatten quasi-norm minimization when the number of observations is very limited, while the solutions obtained by the three methods are almost identical when a sufficient number of observations is observed. The effectiveness and generality of both our methods are demonstrated through extensive experiments on both synthetic data and real-world applications, whose results also show that both our methods perform more robust to outliers and missing ratios than existing methods.

An interesting direction of future work is the theoretical analysis of the properties of both of our bilinear factor matrix norm penalties compared to the nuclear norm and the Schatten quasi-norm. For example, how many observations are sufficient for both our models to reliably recover low-rank matrices, although in our experiments we found that our methods perform much better than existing Schatten quasi-norm methods with a limited number of observations. In addition, we are interested in exploring ways to regularize our models with auxiliary information, such as graph Laplacian~\cite{shahid:grpca,yang:lg,yang:ssr} and hyper-Laplacian matrix~\cite{yin:lrr}, or the elastic-net~\cite{kim:factEN}. We can apply our bilinear factor matrix norm penalties to various structured sparse and low-rank problems~\cite{liu:lrr,zhou:mod,zhang:op,lu:irls}, e.g., corrupted columns~\cite{chen:rmc} and Hankel matrix for image restoration~\cite{jin:hankel}.

\section*{Acknowledgments}
Fanhua Shang, James Cheng and Yuanyuan Liu were supported in part by Grants (CUHK 14206715 \& 14222816) from the Hong Kong RGC, ITF 6904079, Grant 3132821 funded by the Research Committee of CUHK. Zhi-Quan Luo was supported in part by the National Science Foundation (NSF) under Grant CCF-1526434, the National Natural Science Foundation of China under Grants 61571384 and 61731018, and the Leading Talents of Guang Dong Province program, Grant No. 00201510. Zhouchen Lin was supported by National Basic Research Program of China (973 Program) (grant no. 2015CB352502), National Natural Science Foundation (NSF) of China (grant nos. 61625301, 61731018, and 61231002), and Qualcomm. Fanhua Shang and Yuanyuan Liu are the corresponding authors.

\bibliographystyle{IEEEtran}
\bibliography{IEEEabrv,egbib}

\begin{thebibliography}{10}
\providecommand{\url}[1]{#1}
\csname url@samestyle\endcsname
\providecommand{\newblock}{\relax}
\providecommand{\bibinfo}[2]{#2}
\providecommand{\BIBentrySTDinterwordspacing}{\spaceskip=0pt\relax}
\providecommand{\BIBentryALTinterwordstretchfactor}{4}
\providecommand{\BIBentryALTinterwordspacing}{\spaceskip=\fontdimen2\font plus
\BIBentryALTinterwordstretchfactor\fontdimen3\font minus
  \fontdimen4\font\relax}
\providecommand{\BIBforeignlanguage}[2]{{%
\expandafter\ifx\csname l@#1\endcsname\relax
\typeout{** WARNING: IEEEtran.bst: No hyphenation pattern has been}%
\typeout{** loaded for the language `#1'. Using the pattern for}%
\typeout{** the default language instead.}%
\else
\language=\csname l@#1\endcsname
\fi
#2}}
\providecommand{\BIBdecl}{\relax}
\BIBdecl

\bibitem{aharon:ksvd}
M.~Aharon, M.~Elad, and A.~Bruckstein, ``K-{SVD}: An algorithm for designing
  overcomplete dictionaries for sparse representation,'' \emph{IEEE Trans.
  Signal Proces.}, vol.~54, no.~11, pp. 4311--4322, Nov. 2006.

\bibitem{wright:sr}
J.~Wright, A.~Yang, A.~Ganesh, S.~Sastry, and Y.~Ma, ``Robust face recognition
  via sparse representation,'' \emph{IEEE Trans. Pattern Anal. Mach. Intell.},
  vol.~31, no.~2, pp. 210--227, Feb. 2009.

\bibitem{torre:rsl}
F.~{De la Torre} and M.~J. Black, ``A framework for robust subspace learning,''
  \emph{Int. J. Comput. Vis.}, vol.~54, no.~1, pp. 117--142, 2003.

\bibitem{elhamifar:ssc}
E.~Elhamifar and R.~Vidal, ``Sparse subspace clustering: Algorithm, theory, and
  applications,'' \emph{IEEE Trans. Pattern Anal. Mach. Intell.}, vol.~35,
  no.~11, pp. 2765--2781, Nov. 2013.

\bibitem{liu:lrr}
G.~Liu, Z.~Lin, S.~Yan, J.~Sun, Y.~Yu, and Y.~Ma, ``Robust recovery of subspace
  structures by low-rank representation,'' \emph{IEEE Trans. Pattern Anal.
  Mach. Intell.}, vol.~35, no.~1, pp. 171--184, Jan. 2013.

\bibitem{candes:rpca}
E.~J. Cand\`{e}s, X.~Li, Y.~Ma, and J.~Wright, ``Robust principal component
  analysis?'' \emph{J. ACM}, vol.~58, no.~3, pp. 1--37, 2011.

\bibitem{tao:lrsd}
M.~Tao and X.~Yuan, ``Recovering low-rank and sparse components of matrices
  from incomplete and noisy observations,'' \emph{SIAM J. Optim.}, vol.~21,
  no.~1, pp. 57--81, 2011.

\bibitem{zhou:godec}
T.~Zhou and D.~Tao, ``Go{D}ec: Randomized low-rank \& sparse matrix
  decomposition in noisy case,'' in \emph{Proc. 28th Int. Conf. Mach. Learn.},
  2011, pp. 33--40.

\bibitem{chen:rmc}
Y.~Chen, H.~Xu, C.~Caramanis, and S.~Sanghavi, ``Robust matrix completion with
  corrupted columns,'' in \emph{Proc. 28th Int. Conf. Mach. Learn.}, 2011, pp.
  873--880.

\bibitem{krishnan:l12}
D.~Krishnan and R.~Fergus, ``Fast image deconvolution using hyper-{L}aplacian
  priors,'' in \emph{Proc. Adv. Neural Inf. Process. Syst.}, 2009, pp.
  1033--1041.

\bibitem{mallat:wav}
S.~Mallat, ``A theory for multiresolution signal decomposition: The wavelet
  representation,'' \emph{IEEE Trans. Pattern Anal. Mach. Intell.}, vol.~11,
  no.~7, pp. 674--693, Jul. 1989.

\bibitem{luo:nnm}
L.~Luo, J.~Yang, J.~Qian, and Y.~Tai, ``Nuclear-${L}_{1}$ norm joint regression
  for face reconstruction and recognition with mixed noise,'' \emph{Pattern
  Recogn.}, vol.~48, no.~12, pp. 3811--3824, 2015.

\bibitem{eriksson:lrma}
A.~Eriksson and A.~van~den Hengel, ``Efficient computation of robust low-rank
  matrix approximations in the presence of missing data using the ${L}_{1}$
  norm,'' in \emph{Proc. IEEE Conf. Comput. Vis. Pattern Recognit.}, 2010, pp.
  771--778.

\bibitem{ke:rpca}
Q.~Ke and T.~Kanade, ``Robust ${L}_{1}$ norm factorization in the presence of
  outliers and missing data by alternative convex programming,'' in \emph{Proc.
  IEEE Conf. Comput. Vis. Pattern Recognit.}, 2005, pp. 739--746.

\bibitem{kim:factEN}
E.~Kim, M.~Lee, and S.~Oh, ``Elastic-net regularization of singular values for
  robust subspace learning,'' in \emph{Proc. IEEE Conf. Comput. Vis. Pattern
  Recognit.}, 2015, pp. 915--923.

\bibitem{zheng:lrma}
Y.~Zheng, G.~Liu, S.~Sugimoto, S.~Yan, and M.~Okutomi, ``Practical low-rank
  matrix approximation under robust ${L}_{1}$-norm,'' in \emph{Proc. IEEE Conf.
  Comput. Vis. Pattern Recognit.}, 2012, pp. 1410--1417.

\bibitem{zhou:mod}
X.~Zhou, C.~Yang, and W.~Yu, ``Moving object detection by detecting contiguous
  outliers in the low-rank representation,'' \emph{IEEE Trans. Pattern Anal.
  Mach. Intell.}, vol.~35, no.~3, pp. 597--610, Mar. 2013.

\bibitem{fan:ve}
J.~Fan and R.~Li, ``Variable selection via nonconcave penalized likelihood and
  its {O}racle properties,'' \emph{J. Am. Statist. Assoc.}, vol.~96, no. 456,
  pp. 1348--1360, 2001.

\bibitem{lu:lrm}
C.~Lu, J.~Tang, S.~Yan, and Z.~Lin, ``Generalized nonconvex nonsmooth low-rank
  minimization,'' in \emph{Proc. IEEE Conf. Comput. Vis. Pattern Recognit.},
  2014, pp. 4130--4137.

\bibitem{zhang:mcp}
C.~Zhang, ``Nearly unbiased variable selection under minimax concave penalty,''
  \emph{Ann. Statist.}, vol.~38, no.~2, pp. 894--942, 2010.

\bibitem{xu:l12r}
Z.~Xu, X.~Chang, F.~Xu, and H.~Zhang, ``${L}_{1/2}$ regularization: A
  thresholding representation theory and a fast solver,'' \emph{IEEE Trans.
  Neural Netw. Learn. Syst.}, vol.~23, no.~7, pp. 1013--1027, Jul. 2012.

\bibitem{zuo:gist}
W.~Zuo, D.~Meng, L.~Zhang, X.~Feng, and D.~Zhang, ``A generalized iterated
  shrinkage algorithm for non-convex sparse coding,'' in \emph{Proc. IEEE Int.
  Conf. Comput. Vis.}, 2013, pp. 217--224.

\bibitem{cai:svt}
J.~Cai, E.~J. Cand\`{e}s, and Z.~Shen, ``A singular value thresholding
  algorithm for matrix completion,'' \emph{SIAM J. Optim.}, vol.~20, no.~4, pp.
  1956--1982, 2010.

\bibitem{chandrasekaran:rpca}
V.~Chandrasekaran, S.~Sanghavi, P.~Parrilo, and A.~Willsky, ``Rank-sparsity
  incoherence for matrix decomposition,'' \emph{SIAM J. Optim.}, vol.~21,
  no.~2, pp. 572--596, 2011.

\bibitem{jin:hankel}
K.~Jin and J.~Ye, ``Annihilating filter-based low-rank {H}ankel matrix approach
  for image inpainting,'' \emph{IEEE Trans. Image Process.}, vol.~24, no.~11,
  pp. 3498--3511, Nov. 2015.

\bibitem{recht:nnm}
B.~Recht, M.~Fazel, and P.~Parrilo, ``Guaranteed minimum-rank solutions of
  linear matrix equations via nuclear norm minimization,'' \emph{SIAM Rev.},
  vol.~52, no.~3, pp. 471--501, 2010.

\bibitem{shahid:grpca}
N.~Shahid, V.~Kalofolias, X.~Bresson, M.~Bronstein, and P.~Vandergheynst,
  ``Robust principal component analysis on graphs,'' in \emph{Proc. IEEE Int.
  Conf. Comput. Vis.}, 2015, pp. 2812--2820.

\bibitem{zhang:op}
H.~Zhang, Z.~Lin, C.~Zhang, and E.~Chang, ``Exact recoverability of robust
  {PCA} via outlier pursuit with tight recovery bounds,'' in \emph{Proc. 29th
  AAAI Conf. Artif. Intell.}, 2015, pp. 3143--3149.

\bibitem{zhang:tlrt}
Z.~Zhang, A.~Ganesh, X.~Liang, and Y.~Ma, ``{TILT}: Transform invariant
  low-rank textures,'' \emph{Int. J. Comput. Vis.}, vol.~99, no.~1, pp. 1--24,
  2012.

\bibitem{gu:norm}
Z.~Gu, M.~Shao, L.~Li, and Y.~Fu, ``Discriminative metric: Schatten norm vs.
  vector norm,'' in \emph{Proc. 21st Int. Conf. Pattern Recog.}, 2012, pp.
  1213--1216.

\bibitem{nie:rmc}
F.~Nie, H.~Wang, X.~Cai, H.~Huang, and C.~Ding, ``Robust matrix completion via
  joint {S}chatten $p$-norm and ${L}_{p}$-norm minimization,'' in \emph{Proc.
  IEEE Int. Conf. Data Mining}, 2012, pp. 566--574.

\bibitem{lai:irls}
M.~Lai, Y.~Xu, and W.~Yin, ``Improved iteratively rewighted least squares for
  unconstrained smoothed $\ell_{p}$ minimization,'' \emph{SIAM J. Numer.
  Anal.}, vol.~51, no.~2, pp. 927--957, 2013.

\bibitem{lu:irls}
C.~Lu, Z.~Lin, and S.~Yan, ``Smoothed low rank and sparse matrix recovery by
  iteratively reweighted least squares minimization,'' \emph{IEEE Trans. Image
  Process.}, vol.~24, no.~2, pp. 646--654, Feb. 2015.

\bibitem{shang:snm}
F.~Shang, Y.~Liu, and J.~Cheng, ``Scalable algorithms for tractable {S}chatten
  quasi-norm minimization,'' in \emph{Proc. 30th AAAI Conf. Artif. Intell.},
  2016, pp. 2016--2022.

\bibitem{golub:mc}
G.~H. Golub and C.~F.~V. Loan, \emph{Matrix Computations}, {F}ourth~ed.\hskip
  1em plus 0.5em minus 0.4em\relax Johns Hopkins University Press, 2013.

\bibitem{wang:llv}
S.~Wang, L.~Zhang, and Y.~Liang, ``Nonlocal spectral prior model for low-level
  vision,'' in \emph{Proc. Asian Conf. Comput. Vis.}, 2013, pp. 231--244.

\bibitem{gu:wnnm}
S.~Gu, L.~Zhang, W.~Zuo, and X.~Feng, ``Weighted nuclear norm minimization with
  application to image denoising,'' in \emph{Proc. IEEE Conf. Comput. Vis.
  Pattern Recognit.}, 2014, pp. 2862--2869.

\bibitem{hu:tnnr}
Y.~Hu, D.~Zhang, J.~Ye, X.~Li, and X.~He, ``Fast and accurate matrix completion
  via truncated nuclear norm regularization,'' \emph{IEEE Trans. Pattern Anal.
  Mach. Intell.}, vol.~35, no.~9, pp. 2117--2130, Sep. 2013.

\bibitem{bian:ipa}
W.~Bian, X.~Chen, and Y.~Ye, ``Complexity analysis of interior point algorithms
  for non-{L}ipschitz and nonconvex minimization,'' \emph{Math. Program.}, vol.
  149, no.~1, pp. 301--327, 2015.

\bibitem{wright:rpca}
J.~Wright, A.~Ganesh, S.~Rao, Y.~Peng, and Y.~Ma, ``Robust principal component
  analysis: exact recovery of corrupted low-rank matrices by convex
  optimization,'' in \emph{Proc. Adv. Neural Inf. Process. Syst.}, 2009, pp.
  2080--2088.

\bibitem{wright:cpcp}
J.~Wright, A.~Ganesh, K.~Min, and Y.~Ma, ``Compressive principal component
  pursuit,'' \emph{Inform. Infer.}, vol.~2, no.~1, pp. 32--68, 2013.

\bibitem{shang:rbf}
F.~Shang, Y.~Liu, H.~Tong, J.~Cheng, and H.~Cheng, ``Robust bilinear
  factorization with missing and grossly corrupted observations,''
  \emph{Inform. Sciences}, vol. 307, pp. 53--72, 2015.

\bibitem{toh:apg}
K.-C. Toh and S.~Yun, ``An accelerated proximal gradient algorithm for nuclear
  norm regularized least squares problems,'' \emph{Pac. J. Optim.}, vol.~6, pp.
  615--640, 2010.

\bibitem{lin:almm}
Z.~Lin, M.~Chen, and L.~Wu, ``The augmented {L}agrange multiplier method for
  exact recovery of corrupted low-rank matrices,'' Univ. Illinois,
  Urbana-Champaign, Tech. Rep., Mar. 2009.

\bibitem{oh:psvt}
T.~Oh, Y.~Tai, J.~Bazin, H.~Kim, and I.~Kweon, ``Partial sum minimization of
  singular values in robust {PCA}: Algorithm and applications,'' \emph{IEEE
  Trans. Pattern Anal. Mach. Intell.}, vol.~38, no.~4, pp. 744--758, Apr. 2016.

\bibitem{gu:wnnmj}
S.~Gu, Q.~Xie, D.~Meng, W.~Zuo, X.~Feng, and L.~Zhang, ``Weighted nuclear norm
  minimization and its applications to low level vision,'' \emph{Int. J.
  Comput. Vis.}, vol. 121, no.~2, pp. 183--208, 2017.

\bibitem{shen:alad}
Y.~Shen, Z.~Wen, and Y.~Zhang, ``Augmented {L}agrangian alternating direction
  method for matrix separation based on low-rank factorization,'' \emph{Optim.
  Method Softw.}, vol.~29, no.~2, pp. 239--263, 2014.

\bibitem{liu:ftf}
Y.~Liu, L.~Jiao, and F.~Shang, ``A fast tri-factorization method for low-rank
  matrix recovery and completion,'' \emph{Pattern Recogn.}, vol.~46, no.~1, pp.
  163--173, 2013.

\bibitem{cabral:nnbf}
R.~Cabral, F.~{De la Torre}, J.~Costeira, and A.~Bernardino, ``Unifying nuclear
  norm and bilinear factorization approaches for low-rank matrix
  decomposition,'' in \emph{Proc. IEEE Int. Conf. Comput. Vis.}, 2013, pp.
  2488--2495.

\bibitem{peng:rasl}
Y.~Peng, A.~Ganesh, J.~Wright, W.~Xu, and Y.~Ma, ``{RASL}: Robust alignment by
  sparse and low-rank decomposition for linearly correlated images,''
  \emph{IEEE Trans. Pattern Anal. Mach. Intell.}, vol.~34, no.~11, pp.
  2233--2246, Nov. 2012.

\bibitem{ji:lrmc}
H.~Ji, C.~Liu, Z.~Shen, and Y.~Xu, ``Robust video denoising using low rank
  matrix completion,'' in \emph{Proc. IEEE Conf. Comput. Vis. Pattern
  Recognit.}, 2010, pp. 1791--1798.

\bibitem{donoho:st}
D.~L. Donoho, ``De-noising by soft-thresholding,'' \emph{IEEE Trans. Inform.
  Theory}, vol.~41, no.~3, pp. 613--627, May 1995.

\bibitem{xu:rpca}
H.~Xu, C.~Caramanis, and S.~Sanghavi, ``Robust {PCA} via outlier pursuit,'' in
  \emph{Proc. Adv. Neural Inf. Process. Syst.}, 2010, pp. 2496--2504.

\bibitem{yang:multi}
Y.~Yang, Z.~Ma, A.~G. Hauptmann, and N.~Sebe, ``Feature selection for
  multimedia analysis by sharing information among multiple tasks,'' \emph{IEEE
  Trans. Multimedia}, vol.~15, no.~3, pp. 661--669, Apr. 2013.

\bibitem{ma:fea}
Z.~Ma, Y.~Yang, N.~Sebe, and A.~G. Hauptmann, ``Knowledge adaptation with
  partially shared features for event detection using few exemplars,''
  \emph{IEEE Trans. Pattern Anal. Mach. Intell.}, vol.~36, no.~9, pp.
  1789--1802, Sep. 2014.

\bibitem{shao:gtsl}
M.~Shao, D.~Kit, and Y.~Fu, ``Generalized transfer subspace learning through
  low-rank constraint,'' \emph{Int. J. Comput. Vis.}, vol. 109, no.~1, pp.
  74--93, 2014.

\bibitem{ding:llrc}
Z.~Ding, M.~Shao, and S.~Yan, ``Missing modality transfer learning via latent
  low-rank constraint,'' \emph{IEEE Trans. Image Process.}, vol.~24, no.~11,
  pp. 4322--4334, Nov. 2015.

\bibitem{boyd:admm}
S.~Boyd, N.~Parikh, E.~Chu, B.~Peleato, and J.~Eckstein, ``Distributed
  optimization and statistical learning via the alternating direction method of
  multipliers,'' \emph{Found. Trends Mach. Learn.}, vol.~3, no.~1, pp. 1--122,
  2011.

\bibitem{oh:frsvt}
T.~Oh, Y.~Matsushita, Y.~Tai, and I.~Kweon, ``Fast randomized singular value
  thresholding for nuclear norm minimization,'' in \emph{Proc. IEEE Conf.
  Comput. Vis. Pattern Recognit.}, 2015, pp. 4484--4493.

\bibitem{larsen:svd}
R.~Larsen, ``{PROPACK}-software for large and sparse {SVD} calculations,''
  2005.

\bibitem{cai:fsvt}
J.~Cai and S.~Osher, ``Fast singular value thresholding without singular value
  decomposition,'' \emph{Methods Anal. appl.}, vol.~20, no.~4, pp. 335--352,
  2013.

\bibitem{jiang:lrmf}
F.~J. amd M.~Oskarsson and K.~{\AA}strom, ``On the minimal problems of low-rank
  matrix factorization,'' in \emph{Proc. IEEE Conf. Comput. Vis. Pattern
  Recognit.}, 2015, pp. 2549--2557.

\bibitem{xiao:lrr}
S.~Xiao, W.~Li, D.~Xu, and D.~Tao, ``Fa{LRR}: A fast low rank representation
  solver,'' in \emph{Proc. IEEE Int. Conf. Comput. Vis.}, 2015, pp. 4612--4620.

\bibitem{shu:rosl}
X.~Shu, F.~Porikli, and N.~Ahuja, ``Robust orthonormal subspace learning:
  Efficient recovery of corrupted low-rank matrices,'' in \emph{Proc. IEEE
  Conf. Comput. Vis. Pattern Recognit.}, 2014, pp. 3874--3881.

\bibitem{shang:rpca}
F.~Shang, Y.~Liu, J.~Cheng, and H.~Cheng, ``Robust principal component analysis
  with missing data,'' in \emph{Proc. 23rd ACM Int. Conf. Inf. Knowl. Manag.},
  2014, pp. 1149--1158.

\bibitem{liu:as}
G.~Liu and S.~Yan, ``Active subspace: {T}oward scalable low-rank learning,''
  \emph{Neur. Comp.}, vol.~24, no.~12, pp. 3371--3394, 2012.

\bibitem{srebro:mmmf}
N.~Srebro, J.~Rennie, and T.~Jaakkola, ``Maximum-margin matrix factorization,''
  in \emph{Proc. Adv. Neural Inf. Process. Syst.}, 2004, pp. 1329--1336.

\bibitem{feng:rpca}
J.~Feng, H.~Xu, and S.~Yan, ``Online robust {PCA} via stochastic
  optimization,'' in \emph{Proc. Adv. Neural Inf. Process. Syst.}, 2013, pp.
  404--412.

\bibitem{sun:gmc}
R.~Sun and Z.-Q. Luo, ``Guaranteed matrix completion via non-convex
  factorization,'' \emph{IEEE Trans. Inform. Theory}, vol.~62, no.~11, pp.
  6535--6579, Nov. 2016.

\bibitem{shang:tsnm}
F.~Shang, Y.~Liu, and J.~Cheng, ``Tractable and scalable {S}chatten quasi-norm
  approximations for rank minimization,'' in \emph{Proc. 19th Int. Conf. Artif.
  Intell. Statist.}, 2016, pp. 620--629.

\bibitem{mirsky:ti}
L.~Mirsky, ``A trace inequality of {J}ohn von {N}eumann,'' \emph{Monatsh.
  Math.}, vol.~79, pp. 303--306, 1975.

\bibitem{neumann:ti}
J.~von Neumann, ``Some matrix-inequalities and metrization of matric-space,''
  \emph{Tomsk Univ. Rev.}, vol.~1, pp. 286--300, 1937.

\bibitem{mazumder:sr}
R.~Mazumder, T.~Hastie, and R.~Tibshirani, ``Spectral regularization algorithms
  for learning large incomplete matrices,'' \emph{J. Mach. Learn. Res.},
  vol.~11, pp. 2287--2322, 2010.

\bibitem{cao:st}
W.~Cao, J.~Sun, and Z.~Xu, ``Fast image deconvolution using closed-form
  thresholding formulas of $l_{q}$ ($q\!=\!\frac{1}{2},\frac{2}{3}$)
  regularization,'' \emph{J. Vis. Commun. Image R.}, vol.~24, no.~1, pp.
  31--41, 2013.

\bibitem{bolte:palm}
J.~Bolte, S.~Sabach, and M.~Teboulle, ``Proximal alternating linearized
  minimization for nonconvex and nonsmooth problems,'' \emph{Math. Program.},
  vol. 146, no.~1, pp. 459--494, 2014.

\bibitem{hong:admm}
M.~Hong, Z.-Q. Luo, and M.~Razaviyayn, ``Convergence analysis of alternating
  direction method of multipliers for a family of nonconvex problems,''
  \emph{SIAM J. Optim.}, vol.~26, no.~1, pp. 337--364, 2016.

\bibitem{chen:admm}
C.~Chen, B.~He, Y.~Ye, and X.~Yuan, ``The direct extension of {ADMM} for
  multi-block convex minimization problems is not necessarily convergent,''
  \emph{Math. Program.}, vol. 155, no. 1-2, pp. 57--79, 2016.

\bibitem{luxburg:sc}
U.~von Luxburg, ``A tutorial on spectral clustering,'' \emph{Stat. Comput.},
  vol.~17, no.~4, pp. 395--416, 2007.

\bibitem{keshavan:lrmc}
R.~Keshavan, A.~Montanari, and S.~Oh, ``Low-rank matrix completion with noisy
  observations: a quantitative comparison,'' in \emph{47th Conf. Commun.
  Control Comput.}, 2009, pp. 1216--1222.

\bibitem{yang:lg}
Y.~Yang, D.~Xu, F.~Nie, S.~Yan, and Y.~Zhuang, ``Image clustering using local
  discriminant models and global integration,'' \emph{IEEE Trans. Image
  Process.}, vol.~19, no.~10, pp. 2761--2773, Oct. 2010.

\bibitem{yang:ssr}
Y.~Yang, F.~Nie, D.~Xu, J.~Luo, Y.~Zhuang, and Y.~Pan, ``A multimedia retrieval
  framework based on semi-supervised ranking and relevance feedback,''
  \emph{IEEE Trans. Pattern Anal. Mach. Intell.}, vol.~34, no.~4, pp. 723--742,
  Apr. 2012.

\bibitem{yin:lrr}
M.~Yin, J.~Gao, and Z.~Lin, ``Laplacian regularized low-rank representation and
  its applications,'' \emph{IEEE Trans. Pattern Anal. Mach. Intell.}, vol.~38,
  no.~3, pp. 504--517, Mar. 2016.

\bibitem{daubechies:it}
I.~Daubechies, M.~Defrise, and C.~{De Mol}, ``An iterative thresholding
  algorithm for linear inverse problems with a sparsity constraint,''
  \emph{Commun. Pur. Appl. Math.}, vol.~57, no.~11, pp. 1413--1457, 2004.

\bibitem{hale:fp}
E.~T. Hale, W.~Yin, and Y.~Zhang, ``Fixed-point continuation for
  $\ell_{1}$-minimization: Methodology and convergence,'' \emph{SIAM J.
  Optim.}, vol.~19, no.~3, pp. 1107--1130, 2008.

\bibitem{wang:admm}
F.~Wang, W.~Cao, and Z.~Xu, ``Convergence of multi-block {B}regman {ADMM} for
  nonconvex composite problems,'' \emph{arXiv:1505.03063}, 2015.

\end{thebibliography}

\begin{IEEEbiography}[{\includegraphics[width=1in,height=1.25in,clip,keepaspectratio]{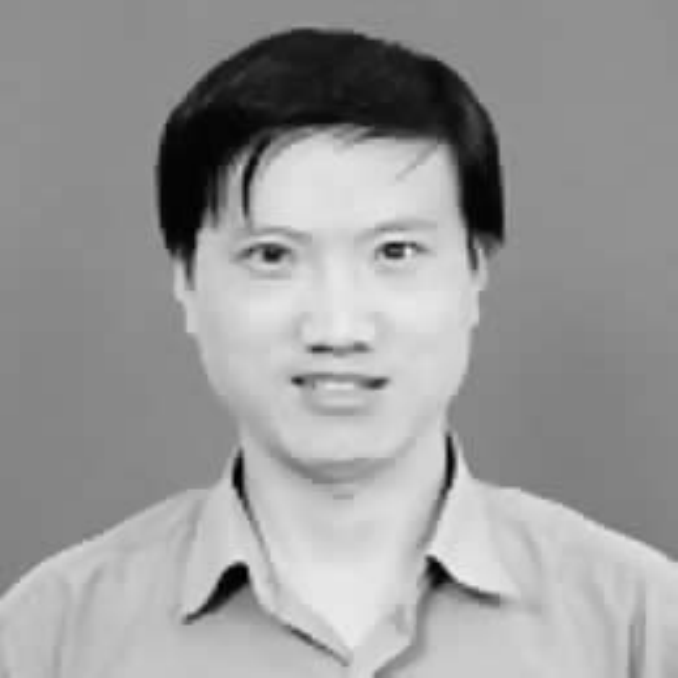}}]{Fanhua Shang} (M'14) received the Ph.D.\ degree in Circuits and Systems from Xidian University, Xi'an, China, in 2012.

He is currently a Research Associate with the Department of Computer Science and Engineering, The Chinese University of Hong Kong. From 2013 to 2015, he was a Post-Doctoral Research Fellow with the Department of Computer Science and Engineering, The Chinese University of Hong Kong. From 2012 to 2013, he was a Post-Doctoral Research Associate with the Department of Electrical and Computer Engineering, Duke University, Durham, NC, USA. His current research interests include machine learning, data mining, pattern recognition, and computer vision.
\end{IEEEbiography}

\vspace{-5mm}

\begin{IEEEbiography}[{\includegraphics[width=1in,height=1.25in,clip,keepaspectratio]{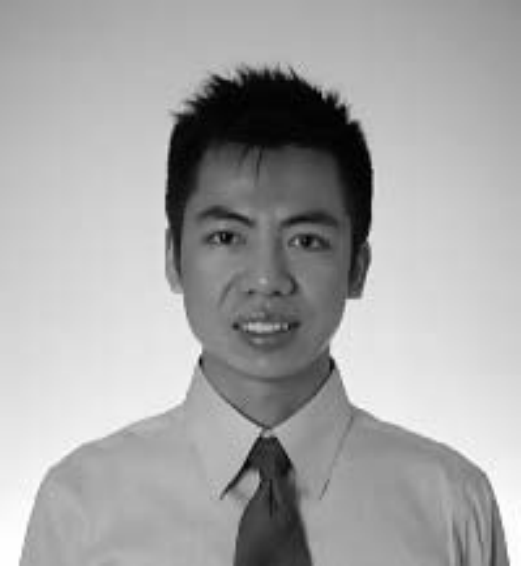}}]{James Cheng} is an Assistant Professor with the Department of Computer Science and Engineering, The Chinese University of Hong Kong, Hong Kong. His current research interests include distributed computing systems, large-scale network analysis, temporal networks, and big data.
\end{IEEEbiography}

\vspace{-5mm}
\begin{IEEEbiography}[{\includegraphics[width=1in,height=1.25in,clip,keepaspectratio]{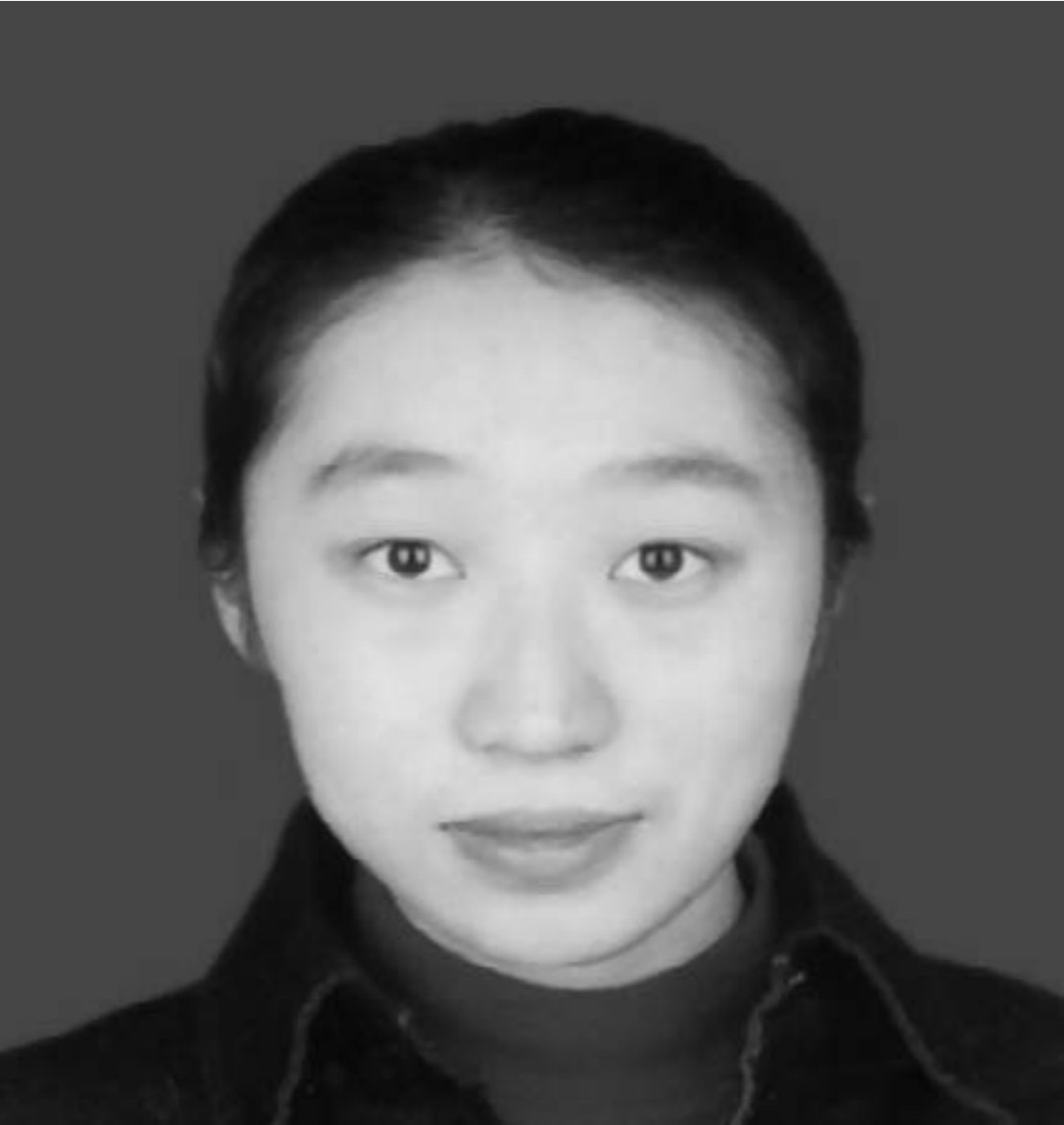}}]{Yuanyuan Liu} received the Ph.D.\ degree in Pattern Recognition and Intelligent System from Xidian University, Xi'an, China, in 2013.

She is currently a Post-Doctoral Research Fellow with the Department of Computer Science and Engineering, The Chinese University of Hong Kong. Prior to that, she was a Post-Doctoral Research Fellow with the Department of Systems Engineering and Engineering Management, CUHK. Her current research interests include image processing, pattern recognition, and machine learning.
\end{IEEEbiography}

\vspace{-5mm}

\begin{IEEEbiography}[{\includegraphics[width=1in,height=1.25in,clip,keepaspectratio]{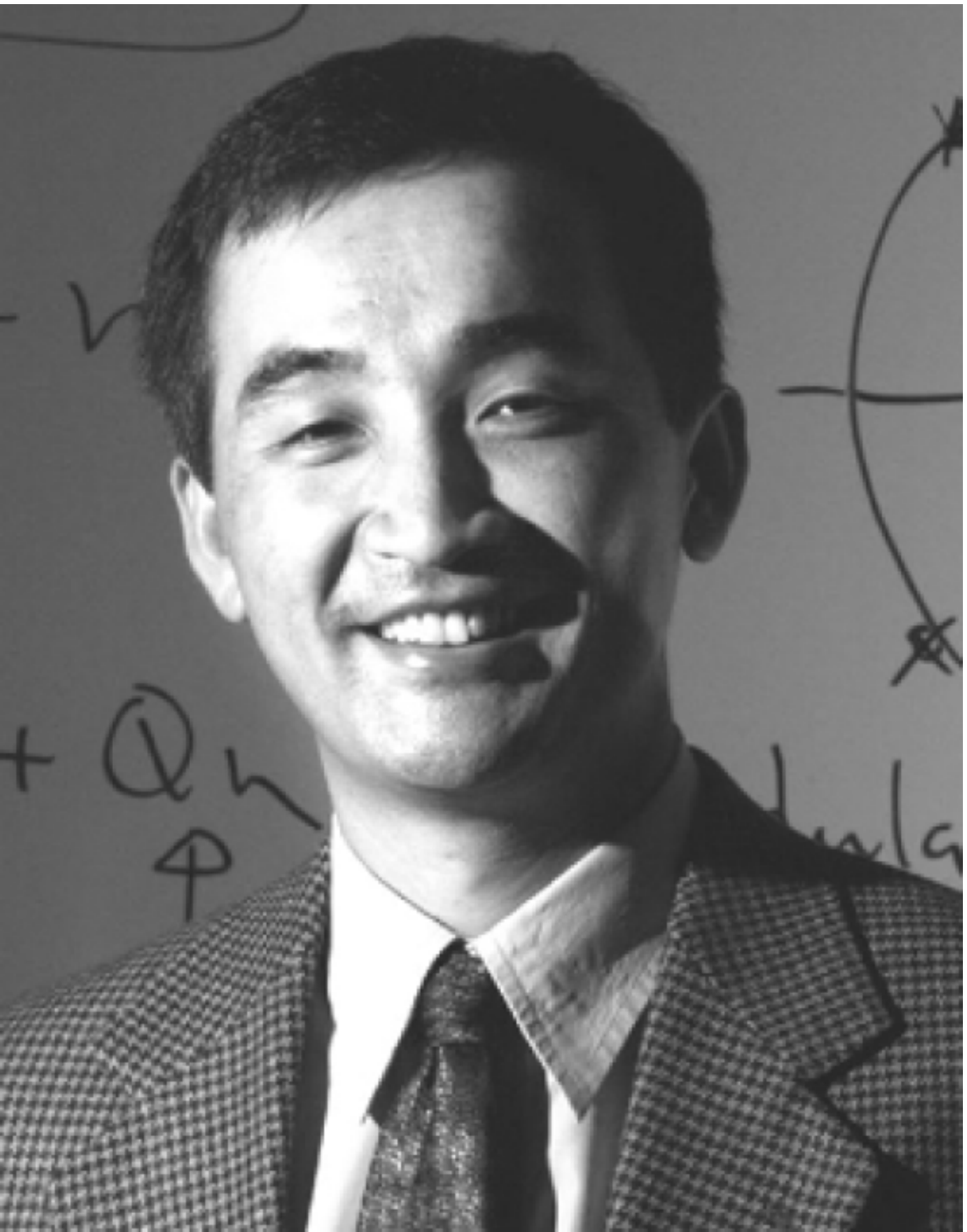}}]{Zhi-Quan (Tom) Luo} (F'07) received the B.Sc.\ degree in Applied Mathematics in 1984 from Peking University, Beijing, China. Subsequently, he was selected by a joint committee of the American Mathematical Society and the Society of Industrial and Applied Mathematics to pursue Ph.D.\ study in the USA. He received the Ph.D.\ degree in operations research in 1989 from the Operations Research Center and the Department of Electrical Engineering and Computer Science, MIT, Cambridge, MA, USA.

From 1989 to 2003, Dr.\ Luo held a faculty position with the Department of Electrical and Computer Engineering, McMaster University, Hamilton, Canada, where he eventually became the department head and held a Canada Research Chair in Information Processing. Since April of 2003, he has been with the Department of Electrical and Computer Engineering at the University of Minnesota (Twin Cities) as a full professor. His research interests lies in the union of optimization algorithms, signal processing and digital communication. Currently, he is with the Chinese University of Hong Kong, Shenzhen, where he is a Professor and serves as the Vice President Academic.

Dr.\ Luo is a fellow of SIAM. He is a recipient of the 2004, 2009 and 2011 IEEE Signal Processing Society Best Paper Awards, the 2011 EURASIP Best
Paper Award and the 2011 ICC Best Paper Award. He was awarded the 2010 Farkas Prize from the INFRMS Optimization Society. Dr.\ Luo chaired the IEEE Signal Processing Society Technical Committee on the Signal Processing for Communications (SPCOM) from 2011-2012. He has held editorial positions for several international journals including Journal of Optimization Theory and Applications, SIAM Journal on Optimization, Mathematics of Computation and IEEE TRANSACTIONS ON SIGNAL PROCESSING. In 2014 he is elected to the Royal Society of Canada.
\end{IEEEbiography}

\vspace{-5mm}
\begin{IEEEbiography}[{\includegraphics[width=1in,height=1.25in,clip,keepaspectratio]{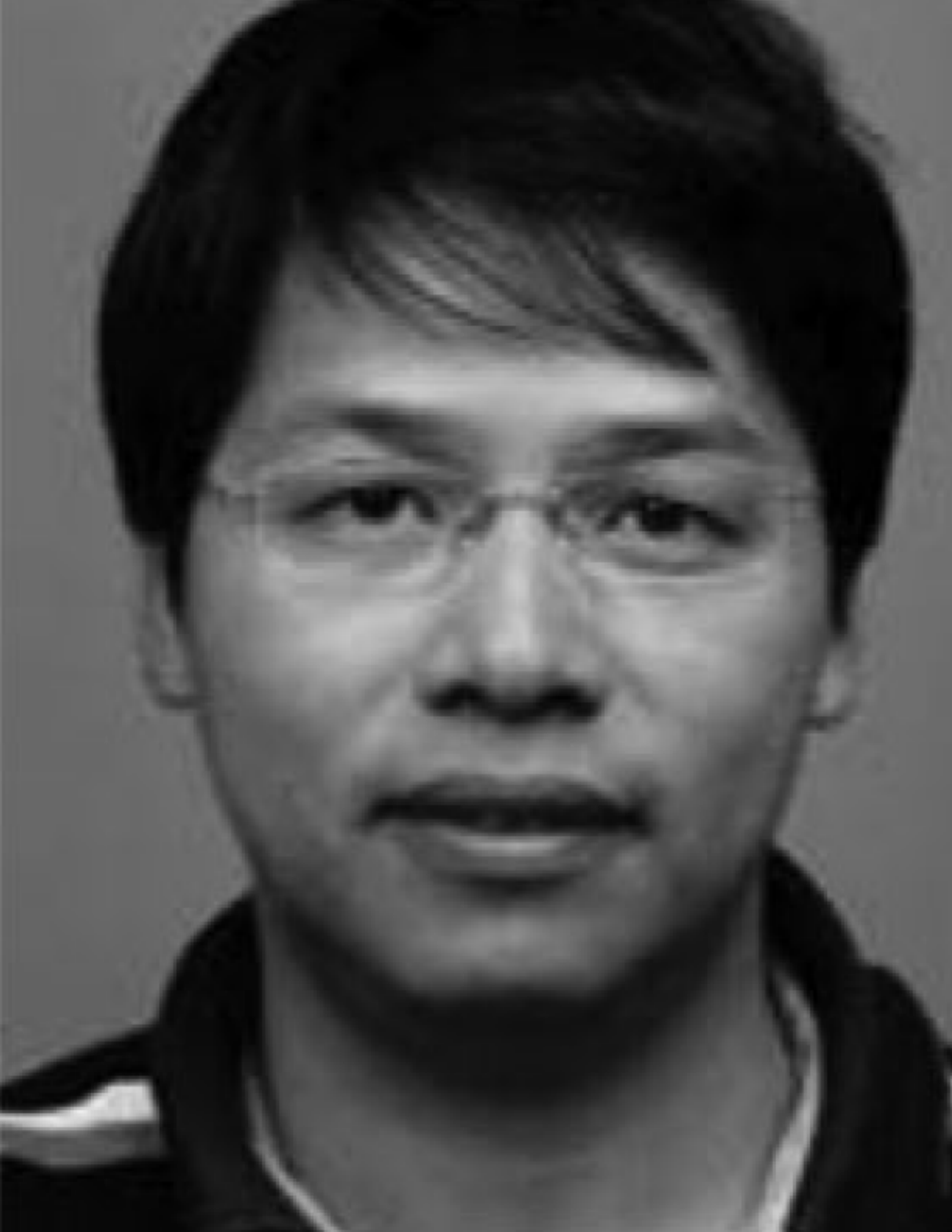}}]{Zhouchen Lin} (M'00-SM'08) received the PhD degree in applied mathematics from Peking University in 2000.

He is currently a professor with the Key Laboratory of Machine Perception, School of Electronics Engineering and Computer Science, Peking University. His research interests include computer vision, image processing, machine learning, pattern recognition, and numerical optimization. He is an area chair of CVPR 2014/2016, ICCV 2015, and NIPS 2015, and a senior program committee member of AAAI 2016/2017/2018 and IJCAI 2016. He is an associate editor of the IEEE Transactions on Pattern Analysis and Machine Intelligence and the International Journal of Computer Vision. He is an IAPR Fellow.
\end{IEEEbiography}

\onecolumn

\title{Supplementary Materials:\\ Bilinear Factor Matrix Norm Minimization for Robust PCA: Algorithms and Applications}

\author{Fanhua~Shang,~\IEEEmembership{Member,~IEEE,}
        James~Cheng,
        Yuanyuan~Liu,
        Zhi-Quan~Luo,~\IEEEmembership{Fellow,~IEEE,}
        and~Zhouchen~Lin,~\IEEEmembership{Senior Member, IEEE}
\IEEEcompsocitemizethanks{\IEEEcompsocthanksitem F.\ Shang, J.\ Cheng and Y.\ Liu are with the Department of Computer Science and Engineering, The Chinese University of Hong Kong,
Shatin, N.T., Hong Kong. E-mail: \{fhshang, jcheng, yyliu\}@cse.cuhk.edu.hk.\protect
\IEEEcompsocthanksitem Z.-Q. Luo is with Shenzhen Research Institute of Big Data, the Chinese University of Hong Kong, Shenzhen, China and Department of Electrical and Computer Engineering, University of Minnesota, Minneapolis, MN 55455, USA. E-mail: luozq@cuhk.edu.cn and luozq@umn.edu.\protect
\IEEEcompsocthanksitem Z.\ Lin is with Key Laboratory of Machine Perception (MOE), School of EECS, Peking University, Beijing 100871, P.R.\ China, and the Cooperative Medianet Innovation Center, Shanghai Jiao Tong University, Shanghai 200240, P.R.\ China. E-mail: zlin@pku.edu.cn.}
}

\markboth{IEEE TRANSACTIONS ON PATTERN ANALYSIS AND MACHINE INTELLIGENCE}
{Shell \MakeLowercase{\textit{et al.}}: Bare Demo of IEEEtran.cls for Computer Society Journals}

\maketitle
\linespread{1.69}

\vspace{16mm}
\centerline{\textbf{\large{Supplementary Materials:}}}
\centerline{\textbf{\large{Bilinear Factor Matrix Norm Minimization for Robust PCA: Algorithms and Applications}}}
\vspace{6mm}

In this supplementary material, we give the detailed proofs of some theorems, lemmas and properties. We also provide the stopping criterion of our algorithms and the details of Algorithm 2. In addition, we present two new ADMM algorithms for image recovery application and their pseudo-codes, and some additional experimental results for both synthetic and real-world datasets.

\setcounter{page}{16}
\setcounter{equation}{35}
\setcounter{lemma}{6}
\setcounter{definition}{4}
\setcounter{algorithm}{1}
\setcounter{theorem}{3}

\section*{Notations}
$\mathbb{R}^{l}$ denotes the $l$-dimensional Euclidean space, and the set of all $m\!\times\! n$ matrices\footnote{Without loss of generality, we assume $m\!\geq\! n$ in this paper.} with real entries is denoted by $\mathbb{R}^{m\times n}$. $\textrm{Tr}(X^{T}Y)=\sum_{ij}X_{ij}Y_{ij}$, where $\textrm{Tr}(\cdot)$ denotes the trace of a matrix. We assume the singular values of $X\!\in\!\mathbb{R}^{m\times n}$ are ordered as $\sigma_{1}(X)\geq \sigma_{2}(X)\geq\cdots\geq\sigma_{r}(X)>\sigma_{r+1}(X)=\cdots=\sigma_{n}(X)=0$, where $r=\textrm{rank}(X)$. Then the SVD of $X$ is denoted by $X=U\Sigma V^{T}$, where $\Sigma=\textrm{diag}(\sigma_{1},\ldots,\sigma_{n})$. $I_{n}$ denotes an identity matrix of size $n\!\times\! n$.

\begin{definition}
For any vector $x\in\mathbb{R}^{l}$, its $\ell_{p}$-norm for $0\!<\!p\!<\!\infty$ is defined as
\begin{displaymath}
\|x\|_{\ell_{p}}=\left(\sum^{l}_{i=1}|x_{i}|^{p}\right)^{1/p}
\end{displaymath}
where $x_{i}$ is the $i$-th element of $x$. When $p\!=\!1$, the $\ell_{1}$-norm of $x$ is $\|x\|_{\ell_{1}}\!=\!\sum_{i}|x_{i}|$ (which is convex), while the $\ell_{p}$-norm of $x$ is a quasi-norm when $0\!<\!p\!<\!1$, which is non-convex and violates the triangle inequality. In addition, the $\ell_{2}$-norm of $x$ is $\|x\|_{\ell_{2}}\!=\!\sqrt{\sum_{i}x^{2}_{i}}$.
\end{definition}

The above definition can be naturally extended from vectors to matrices by the following form
\begin{displaymath}
\|S\|_{\ell_{p}}=\left(\sum_{i,j}|S_{ij}|^{p}\right)^{1/p}.
\end{displaymath}

\begin{definition}
The Schatten-$p$ norm ($0\!<\!p\!<\!\infty$) of a matrix $X\!\in\!\mathbb{R}^{m\times n}$ is defined as follows:
\begin{displaymath}
\|X\|_{S_{p}}=\left(\sum^{n}_{i=1}\sigma^{p}_{i}(X)\right)^{1/p}
\end{displaymath}
where $\sigma_{i}(X)$ denotes the $i$-th largest singular value of $X$.
\end{definition}
\vspace{3mm}

In the following, we will list some special cases of the Schatten-$p$ norm ($0\!<\!p\!<\!\infty$).
\begin{itemize}
  \item When $0\!<\!p\!<\!1$, the Schatten-$p$ norm is a quasi-norm, and it is non-convex and violates the triangle inequality.
  \item When $p\!=\!1$, the Schatten-1 norm (also known as the nuclear norm or trace norm) of $X$ is defined as
        \begin{displaymath}
        \|X\|_{\ast}=\sum^{n}_{i=1}\sigma_{i}(X).
        \end{displaymath}
  \item When $p\!=\!2$, the Schatten-2 norm is more commonly called the Frobenius norm\footnote{Note that the Frobenius norm is the induced norm of the $\ell_{2}$-norm on matrices.} defined as
      \begin{displaymath}
      \|X\|_{F}=\sqrt{\sum^{n}_{i=1}\sigma^{2}_{i}(X)}=\sqrt{\sum_{i,j}X^{2}_{ij}}.
      \end{displaymath}
\end{itemize}

\section*{APPENDIX A: Proof of Lemma 2}
To prove Lemma~2, we first define the doubly stochastic matrix, and give the following lemma.

\begin{definition}
A square matrix is doubly stochastic if its elements are non-negative real numbers, and the sum of elements of each row or column is equal to $1$.
\end{definition}

\begin{lemma}\label{lem7}
Let $P\!\in\! \mathbb{R}^{n\times n}$ be a doubly stochastic matrix, and if
\begin{equation}\label{equ151}
\!\!0\leq x_{1}\leq x_{2}\leq\ldots \leq x_{n},\;y_{1}\geq y_{2}\geq\ldots \geq y_{n}\geq 0,
\end{equation}
then
\begin{equation*}
\sum^{n}_{i,j=1}p_{ij}x_{i}y_{j}\geq \sum^{n}_{k=1}x_{k}y_{k}.
\end{equation*}
\end{lemma}

The proof of Lemma~\ref{lem7} is essentially similar to that of the lemma in~\cite{mirsky:ti}, thus we give the following proof sketch for this lemma.

\begin{IEEEproof}
Using~\eqref{equ151}, there exist non-negative numbers $\alpha_{i}$ and $\beta_{j}$ for all $1\leq i,j\leq n$ such that
\begin{equation*}
x_{k}=\sum_{1\leq i\leq k}\alpha_{i}, \;\; y_{k}=\sum_{k\leq j\leq n} \beta_{j} \,\;\textup{for all}\;\, k=1,\ldots,n.
\end{equation*}

Let $\delta_{ij}$ denote the Kronecker delta (i.e., $\delta_{ij}\!=\!1$ if $i\!=\!j$, and $\delta_{ij}\!=\!0$ otherwise), we have
\begin{equation*}
\begin{split}
&\sum^{n}_{k=1}x_{k}y_{k}-\sum^{n}_{i,j=1}p_{ij}x_{i}y_{j}=\sum^{n}_{i,j=1}(\delta_{ij}-p_{ij})x_{i}y_{j}\\
=&\sum_{1\leq i,j\leq n}(\delta_{ij}-p_{ij})\sum_{1\leq r\leq i}\!\alpha_{r}\sum_{j\leq s\leq n}\!\beta_{s}\\
=&\sum_{1\leq r,s\leq n}\alpha_{r}\beta_{s}\sum_{r\leq i\leq n,1\leq j\leq s}(\delta_{ij}-p_{ij}).
\end{split}
\end{equation*}
If $r\leq s$, by the lemma in~\cite{mirsky:ti} we know that
\begin{equation*}
\begin{split}
\sum_{1\leq i< r,1\leq j\leq s}(\delta_{ij}-p_{ij})\geq 0,\;\;\textup{and}
\end{split}
\end{equation*}
\begin{equation*}
\sum_{r\leq i\leq n,1\leq j\leq s}(\delta_{ij}-p_{ij})+\sum_{1\leq i< r,1\leq j\leq s}(\delta_{ij}-p_{ij})=0.
\end{equation*}
Therefore, we have
\begin{equation*}
\sum_{r\leq i\leq n,1\leq j\leq s}(\delta_{ij}-p_{ij})\leq 0.
\end{equation*}
The same result can be obtained in a similar way for $r\!\geq\! s$.
\end{IEEEproof}

\textbf{Proof of Lemma 2:}
\begin{IEEEproof}
Using the properties of the trace, we know that
\begin{equation*}
\textrm{Tr}(X^{T}Y)=\sum^{n}_{i,j=1}(U_{X})^{2}_{ij}\lambda_{i}\tau_{j}.
\end{equation*}

Note that $U_{X}$ is a unitary matrix, i.e., $U^{T}_{X}U_{X}\!=\!U_{X}U^{T}_{X}\!=\!I_{n}$, which implies that $((U_{X})^{2}_{ij})$ is a doubly stochastic matrix. By Lemma~\ref{lem7}, we have $\textrm{Tr}(X^{T}Y)\geq \sum^{n}_{i=1}\lambda_{i}\tau_{i}$.
\end{IEEEproof}

\section*{APPENDIX B: Proofs of Theorems 1 and 2}
\textbf{Proof of Theorem~1:}
\begin{IEEEproof}
Using Lemma 3, for any factor matrices $U\!\in\! \mathbb{R}^{m\times d}$ and $V\!\in\! \mathbb{R}^{n\times d}$ with the constraint $X\!=\!UV^{T}$, we have
\begin{equation*}
\left(\frac{\|U\|_{\ast}+\|V\|_{\ast}}{2}\right)^{2}\geq\|X\|_{S_{1/2}}.
\end{equation*}

On the other hand, let $U^{\star}\!=\!L_{X}\Sigma^{1/2}_{X}$ and $V^{\star}\!=\!R_{X}\Sigma^{1/2}_{X}$, where $X\!=\!L_{X}\Sigma_{X}R^{T}_{X}$ is the SVD of $X$ as in Lemma 3, then we have
\begin{displaymath}
X=U^{\star}(V^{\star})^{T} \;\;\textup{and}\;\; \|X\|_{S_{1/2}}=(\|U^{\star}\|_{\ast}+\|V^{\star}\|_{\ast})^{2}/4.
\end{displaymath}

Therefore, we have
\begin{displaymath}
\min_{U,V:X=UV^{T}}\!\left(\frac{\|U\|_{\ast}+\|V\|_{\ast}}{2}\right)^{2}=\|X\|_{S_{1/2}}.
\end{displaymath}
This completes the proof.
\end{IEEEproof}

\vspace{5mm}

\textbf{Proof of Theorem~2:}
\begin{IEEEproof}
Using Lemma~4, for any $U\!\in\!\mathbb{R}^{m\times d}$ and $V\!\in\!\mathbb{R}^{n\times d}$ with the constraint $X\!=\!UV^{T}$, we have
\begin{equation*}
\left(\frac{\|U\|^{2}_{F}+2\|V\|_{\ast}}{3}\right)^{3/2}\geq\|X\|_{S_{2/3}}.
\end{equation*}

On the other hand, let $U^{\star}\!=\!L_{X}\Sigma^{1/3}_{X}$ and $V^{\star}\!=\!R_{X}\Sigma^{2/3}_{X}$, where $X\!=\!L_{X}\Sigma_{X}R^{T}_{X}$ is the SVD of $X$ as in Lemma~3, then we have
\begin{displaymath}
X=U^{\star}(V^{\star})^{T}\;\;\textup{and}\;\; \|X\|_{S_{2/3}}=[(\|U^{\star}\|^{2}_{F}+2\|V^{\star}\|_{\ast})/3]^{3/2}.
\end{displaymath}

Thus, we have
\begin{displaymath}
\min_{U,V:X=UV^{T}}\!\left(\frac{\|U\|^{2}_{F}+2\|V\|_{\ast}}{3}\right)^{3/2}=\|X\|_{S_{2/3}}.
\end{displaymath}
This completes the proof.
\end{IEEEproof}

\section*{APPENDIX C: Proof of Property 4}

\begin{IEEEproof}
The proof of Property 4 involves some properties of the $\ell_{p}$-norm, which we recall as follows. For any vector $x$ in $\mathbb{R}^{n}$ and $0<p_{2}\leq p_{1}\leq1$, the following inequalities hold:
\begin{displaymath}
\|x\|_{\ell_{1}}\leq\|x\|_{\ell_{p_{1}}},\quad \|x\|_{\ell_{p_{1}}}\leq \|x\|_{\ell_{p_{2}}}\leq n^{\frac{1}{p_{2}}-\frac{1}{p_{1}}}\|x\|_{\ell_{p_{1}}}.
\end{displaymath}

Let $X$ be an ${m\times n}$ matrix of rank $r$, and denote its compact SVD by $X=U_{m\times r}\Sigma_{r\times r}V^{T}_{n\times r}$. By Theorems 1 and 2, and the properties of the $\ell_{p}$-norm mentioned above, we have
\begin{displaymath}
\begin{split}
\|X\|_{\ast}&=\|\textrm{diag}(\Sigma_{r\times r})\|_{\ell_{1}}\leq \|\textrm{diag}(\Sigma_{r\times r})\|_{\ell_{\frac{1}{2}}}=\|X\|_{\textup{D-N}}\leq \textup{rank}(X)\|X\|_{\ast},\\
\|X\|_{\ast}&=\|\textrm{diag}(\Sigma_{r\times r})\|_{\ell_{1}}\leq \|\textrm{diag}(\Sigma_{r\times r})\|_{\ell_{\frac{2}{3}}}=\|X\|_{\textup{F-N}}\leq \sqrt{\textup{rank}(X)}\|X\|_{\ast}.
\end{split}
\end{displaymath}
In addition,
\begin{displaymath}
\|X\|_{\textup{F-N}}=\|\textrm{diag}(\Sigma_{r\times r})\|_{\ell_{\frac{2}{3}}}\leq\|\textrm{diag}(\Sigma_{r\times r})\|_{\ell_{\frac{1}{2}}}=\|X\|_{\textup{D-N}}.
\end{displaymath}

Therefore, we have
\begin{displaymath}
\|X\|_{\ast}\leq\|X\|_{\textup{F-N}}\leq\|X\|_{\textup{D-N}}\leq \textup{rank}(X)\|X\|_{\ast}.
\end{displaymath}
This completes the proof.
\end{IEEEproof}

\begin{algorithm}[t]
\caption{ADMM for solving $(\textup{S+L})_{2/3}$ problem (18)}
\label{alg2}
\renewcommand{\algorithmicrequire}{\textbf{Input:}}
\renewcommand{\algorithmicensure}{\textbf{Initialize:}}
\renewcommand{\algorithmicoutput}{\textbf{Output:}}
\begin{algorithmic}[1]
\REQUIRE $D\!\in\!\mathbb{R}^{m\times n}$, the given rank $d$ and $\lambda$.
\ENSURE $\mu_{0}$, $\rho\!>\!1$, $k\!=\!0$, and $\epsilon$.\\
\WHILE {not converged}
\WHILE {not converged}
\STATE {Update $U_{k+1}$ and $V_{k+1}$ by \eqref{equ19} and \eqref{equ20}, respectively.}
\STATE {Update $\widehat{V}_{k+1}$ by $\widehat{V}_{k+1}=\mathcal{D}_{2\lambda/3\mu_{k}}\!\left(V_{k+1}\!-\!(Y_{2})_{k}/\mu_{k}\right)$.}
\STATE {Update $L_{k+1}$ and $S_{k+1}$ by \eqref{equ23} and (33) in this paper, respectively.}
\ENDWHILE \;\,$\verb|//|$ Inner loop
\STATE {Update the multipliers $Y^{k+\!1}_{1}$, $Y^{k+\!1}_{2}$ and $Y^{k+\!1}_{3}$ by \\$\quad Y^{k+\!1}_{1}\!=\!Y^{k}_{1}\!+\!\mu_{k}(\widehat{V}_{k+\!1}\!-\!V_{k+\!1})$, $Y^{k+\!1}_{2}\!=\!Y^{k}_{2}\!+\!\mu_{k}(U_{k+\!1}V^{T}_{k+\!1}\!-\!L_{k+\!1})$, and $Y^{k+\!1}_{3}\!=\!Y^{k}_{3}\!+\!\mu_{k}(L_{k+\!1}\!+\!S_{k+\!1}\!-\!D)$.}
\STATE {Update $\mu_{k+1}$ by $\mu_{k+1}=\rho\mu_{k}$.} 
\STATE $k\leftarrow k+1$.
\ENDWHILE \;\,$\verb|//|$ Outer loop
\OUTPUT $U_{k+1}$ and $V_{k+1}$.
\end{algorithmic}
\end{algorithm}

\section*{APPENDIX D: Solving (18) via ADMM}
Similar to Algorithm 1, we also propose an efficient algorithm based on the alternating direction method of multipliers (ADMM) to solve (18), whose augmented Lagrangian function is given by
\begin{equation*}
\begin{split}
\mathcal{L}_{\mu}(U,V,L,S,\widehat{V},Y_{1},Y_{2},Y_{3})=&\,\frac{\lambda}{3}\!\left(\|U\|^{2}_{F}\!+\!2\|\widehat{V}\|_{*}\right)+\|\mathcal{P}_{\Omega}(S)\|^{2/3}_{\ell_{2/3}}\!+\!\left\langle Y_{1},\widehat{V}\!-\!V\right\rangle\!+\!\left\langle Y_{2},UV^{T}\!-\!L\right\rangle\!+\!\left\langle Y^{}_{3},L\!+\!S\!-\!D\right\rangle\\
&+\frac{\mu}{2}\left(\|UV^{T}-L\|^{2}_{F}+\|\widehat{V}-V\|^{2}_{F}+\|L+S-D\|^{2}_{F}\right)
\end{split}
\end{equation*}
where $Y_{1}\!\in\!\mathbb{R}^{n\times d}$, $Y_{2}\!\in\!\mathbb{R}^{m\times n}$ and $Y_{3}\!\in\!\mathbb{R}^{m\times n}$ are the matrices of Lagrange multipliers.

\subsection*{Update of $U_{k+1}$ and $V_{k+1}$:}

For updating ${U}_{k+1}$ and ${V}_{k+1}$, we consider the following optimization problems:
\begin{equation}\label{equ17}
\min_{U}\frac{\lambda}{3}\|U\|^{2}_{F}+\frac{\mu_{k}}{2}\|UV^{T}_{k}-L_{k}+\mu^{-\!1}_{k}Y^{k}_{2}\|^{2}_{F},
\end{equation}
\begin{equation}\label{equ18}
\min_{V}\|\widehat{V}_{k}-V+\mu^{-\!1}_{\!k}Y^{k}_{1}\|^{2}_{F}+\|U_{\!k+\!1}V^{T}\!-L_{k}+\mu^{-\!1}_{\!k}Y^{k}_{2}\|^{2}_{F},
\end{equation}
and their optimal solutions can be given by
\begin{equation}\label{equ19}
U_{k+1}=\mu_{k}P_{k}V_{k}\left(\frac{2\lambda}{3}I_{d}+\mu_{k}V^{T}_{k}V_{k}\right)^{-\!1},
\end{equation}
\begin{equation}\label{equ20}
V_{k+1}=\left[\widehat{V}_{k}+\mu^{-\!1}_{k}Y^{k}_{1}+P^{T}_{k}U_{k+\!1}\right]\left(I_{d}+U^{T}_{k+\!1}U_{k+\!1}\right)^{-\!1},
\end{equation}
where $P_{k}\!=\!L_{k}-\mu^{\!-\!1}_{k}Y^{k}_{2}$.

\subsection*{Update of $\widehat{V}_{k+1}$:}

To update $\widehat{V}_{k+1}$, we fix the other variables and solve the following optimization problem
\begin{equation}\label{equ21}
\min_{\widehat{V}}\frac{2\lambda}{3}\|\widehat{V}\|_{*}+\frac{\mu_{k}}{2}\|\widehat{V}-V_{k+1}+Y^{k}_{1}/\mu_{k}\|^{2}_{F}.
\end{equation}
Similar to (23) and (24), the closed-form solution of \eqref{equ21} can also be obtained by the SVT operator~\cite{cai:svt} defined as follows.

\begin{definition}\label{def3}
Let $Y$ be a matrix of size $m\!\times\!n$ ($m\geq n$), and $U_{Y}\Sigma_{Y}V^{T}_{Y}$ be its SVD. Then the singular value thresholding (SVT) operator
$\mathcal{D}_{\tau}$ is defined as~\cite{cai:svt}:
\begin{equation*}
\mathcal{D}_{\tau}(Y)=U_{Y}\mathcal{S}_{\tau}(\Sigma_{Y})V^{T}_{Y},
\end{equation*}
where $\mathcal{S}_{\tau}(x)=\max(|x|-\tau,0)\cdot\textup{sgn}(x)$ is the soft shrinkage operator~\cite{daubechies:it,donoho:st,hale:fp}.
\end{definition}

\subsection*{Update of $L_{k+1}$:}
For updating $L_{k+\!1}$, we consider the following optimization problem:
\begin{equation}\label{equ22}
\min_{L} \|U_{k+\!1}\!V^{T}_{k+\!1}\!-\!L\!+\!\mu^{\!-\!1}_{k}Y^{k}_{2}\|^{2}_{F}+\|L\!+\!S_{k}\!-\!D\!+\!\mu^{-\!1}_{k}Y^{k}_{3}\|^{2}_{F}.
\end{equation}
Since \eqref{equ22} is a least squares problem, and thus its closed-form solution is given by
\begin{equation}\label{equ23}
L_{k+\!1}=\frac{1}{2}\!\left(U_{k+\!1}\!V^{T}_{k+\!1}+\mu^{-\!1}_{k}Y^{k}_{2}-S_{k}+D-\mu^{-\!1}_{k}Y^{k}_{3}\right).
\end{equation}

Together with the update scheme of $S_{k+1}$, as stated in (33) in this paper, we develop an efficient ADMM algorithm to solve the Frobenius/nuclear hybrid norm penalized RPCA problem (18), as outlined in \textbf{Algorithm}~\ref{alg2}.~\\

\section*{APPENDIX E: Proof of Theorem 3}
In this part, we first prove the boundedness of multipliers and some variables of Algorithm 1, and then we analyze the convergence of Algorithm 1. To prove the boundedness, we first give the following lemma.

\begin{lemma}[\cite{lin:almm}]\label{lem5}
Let $\mathcal{H}$ be a real Hilbert space endowed with an inner product $\langle\cdot,\cdot\rangle$ and a corresponding norm $\|\!\cdot\!\|$, and $y\!\in\! \partial\|x\|$, where $\partial f(x)$ denotes the subgradient of $f(x)$. Then $\|y\|^{*}\!=\!1$ if $x\neq  0$, and $\|y\|^{*}\!\leq\! 1$ if $x\!=\!0$, where $\|\!\cdot\!\|^{*}$ is the dual norm of $\|\!\cdot\!\|$. For instance, the dual norm of the nuclear norm is the spectral norm, $\|\!\cdot\!\|_{2}$, i.e., the largest singular value.
\end{lemma}

\begin{lemma}[Boundedness]\label{lem6}
Let $Y^{k+\!1}_{1}\!=\!Y^{k}_{1}\!+\!\mu_{k}(\widehat{U}_{k+\!1}\!-\!U_{k+\!1})$, $Y^{k+\!1}_{2}\!=\!Y^{k}_{2}\!+\!\mu_{k}(\widehat{V}_{k+\!1}\!-\!V_{k+\!1})$, $Y^{k+\!1}_{3}\!=\!Y^{k}_{3}\!+\!\mu_{k}(U_{k+\!1}\!V^{T}_{k+\!1}\!-\!L_{k+\!1})$ and $Y^{k+\!1}_{4}\!=\!Y^{k}_{4}\!+\!\mu_{k}(L_{k+\!1}\!+\!S_{k+\!1}\!-\!D)$. Suppose that $\mu_{k}$ is non-decreasing and $\sum^{\infty}_{k=0}\!\frac{\mu_{k+\!1}}{\mu^{4/3}_{k}}\!<\!\infty$, then the sequences $\{(U_{k},V_{k})\}$, $\{(\widehat{U}_{k},\widehat{V}_{k})\}$, $\{(Y^{k}_{1},Y^{k}_{2}, Y^{k}_{4}/\!\sqrt[3]{\mu_{k-\!1}})\}$, $\{L_{k}\}$ and $\{S_{k}\}$ produced by Algorithm 1 are all bounded.
\end{lemma}

\begin{IEEEproof}
Let $\mathscr{U}_{k}\!:=\!(U_{k},V_{k})$, $\mathscr{V}_{k}\!:=\!(\widehat{U}_{k},\widehat{V}_{k})$ and $\mathscr{Y}_{k}\!:=\!(Y^{k}_{1},Y^{k}_{2},Y^{k}_{3},Y^{k}_{4})$. By the first-order optimality conditions of the augmented Lagrangian function of (17) with respect to $\widehat{U}$ and $\widehat{V}$ (i.e., Problems (23) and (24)), we have
\begin{displaymath}
0\in \partial_{\widehat{U}}\mathcal{L}_{\mu_{k}}(\mathscr{U}_{k+1},\mathscr{V}_{k+1},L_{k+1},S_{k+1},\mathscr{Y}_{k})\;\;\textup{and}\;\; 0\in \partial_{\widehat{V}}\mathcal{L}_{\mu_{k}}(\mathscr{U}_{k+1},\mathscr{V}_{k+1},L_{k+1},S_{k+1},\mathscr{Y}_{k}),
\end{displaymath}
i.e., $-Y^{k+1}_{1}\!\in\!\frac{\lambda}{2}\partial\|\widehat{U}_{k+1}\|_{\ast}$ and $-Y^{k+1}_{2}\!\in\!\frac{\lambda}{2}\partial\|\widehat{V}_{k+1}\|_{\ast}$, where $Y^{k+1}_{1}\!=\!\mu_{k}(\widehat{U}_{k+\!1}\!-\!U_{k+\!1}\!+\!Y^{k}_{1}/\mu_{k})$ and $Y^{k+1}_{2}\!=\!\mu_{k}(\widehat{V}_{k+\!1}\!-\!V_{k+\!1}\!+\!Y^{k}_{2}/\mu_{k})$.
By Lemma~\ref{lem5}, we obtain
\begin{displaymath}
\|Y^{k+1}_{1}\|_{2}\leq \frac{\lambda}{2}\;\;\textup{and}\;\;\|Y^{k+1}_{2}\|_{2}\leq \frac{\lambda}{2},
\end{displaymath}
which implies that the sequences $\{Y^{k}_{1}\}$ and $\{Y^{k}_{2}\}$ are bounded.

Next we prove the boundedness of $\{Y^{k}_{4}\!/\!\sqrt[3]{\mu_{k-1}}\}$. Using (27), it is easy to show that $\mathcal{P}^{\perp}_{\Omega}(Y^{k+1}_{4})\!=\!0$, which implies that the sequence $\{\mathcal{P}^{\perp}_{\Omega}(Y^{k+1}_{4})\}$ is bounded. Let $A\!=\!D\!-\!L_{k+1}\!-\!\mu^{-1}_{k}Y^{k}_{4}$, and $\Phi(S)\!:=\!\|\mathcal{P}_{\Omega}(S)\|^{1/2}_{\ell_{1/2}}$. To prove the boundedness of $\{Y^{k}_{4}\!/\!\sqrt[3]{\mu_{k-1}}\}$, we consider the following two cases.

\textbf{Case 1}: $|a_{ij}|>\frac{3}{2\sqrt[3]{\mu^{2}_{k}}},\,(i,j)\!\in\!\Omega.$

If $|a_{ij}|\!>\!\frac{3}{2\sqrt[3]{\mu^{2}_{k}}}$, and using Proposition 1, then $|[S_{k+1}]_{ij}|>0$. The first-order optimality condition of (27) implies that
\begin{displaymath}
[\partial\Phi(S_{k+1})]_{ij}+[Y^{k+1}_{4}]_{ij}=0,
\end{displaymath}
where $[\partial\Phi(S_{k+1})]_{ij}$ denotes the gradient of the penalty $\Phi(S)$ at $[S_{k+1}]_{ij}$. Since $\partial\Phi(s_{ij})=\textup{sign}(s_{ij})/(2\sqrt{|s_{ij}|})$, then
\begin{displaymath}
\left|[Y^{k+1}_{4}]_{ij}\right|=\left|\frac{1}{2\sqrt{|[S_{k+1}]_{ij}|}}\right|.
\end{displaymath}
Using Proposition 1, we have
\begin{displaymath}
\left|\frac{[Y^{k+1}_{4}]_{ij}}{\sqrt[3]{\mu_{k}}}\right|=\left|\frac{1}{2\sqrt[3]{\mu_{k}}\sqrt{|[S_{k+1}]_{ij}|}}\right|=\frac{\sqrt{3}}{2\sqrt{2|a_{ij}(1+\cos(\frac{2\pi-2\phi_{\gamma}(a_{ij})}{3}))|}\sqrt[3]{\mu_{k}}}\leq \frac{1}{\sqrt{2}},
\end{displaymath}
where $\gamma\!=\!2/\mu_{k}$.

\textbf{Case 2}: $|a_{ij}|\leq\frac{3}{2\sqrt[3]{\mu^{2}_{k}}},\,(i,j)\!\in\!\Omega.$

If $|a_{ij}|\leq\frac{3}{2\sqrt[3]{\mu^{2}_{k}}}$, and using Proposition 1, we have $[S_{k+1}]_{ij}=0$. Since
\begin{displaymath}
|[Y^{k+1}_{4}/\mu_{k}]_{ij}|=|[L_{k+1}+\mu^{-1}_{k}Y^{k}_{4}-D+S_{k+1}]_{ij}|=|[L_{k+1}+\mu^{-1}_{k}Y^{k}_{4}-D]_{ij}|=|a_{ij}|\leq\frac{3}{2\sqrt[3]{\mu^{2}_{k}}},
\end{displaymath}
then
\begin{displaymath}
\left|\frac{[Y^{k+1}_{4}]_{ij}}{\sqrt[3]{\mu_{k}}}\right|\leq\frac{3}{2},\;\,\textup{and}\,\;\|Y^{k+1}_{4}\|_{F}/\sqrt[3]{\mu_{k}}=\|\mathcal{P}_{\Omega}(Y^{k+1}_{4})\|_{F}/\sqrt[3]{\mu_{k}}\leq \frac{3|\Omega|}{2}.
\end{displaymath}
Therefore, $\{Y^{k}_{4}\!/\!\sqrt[3]{\mu_{k-\!1}}\}$ is bounded.

By the iterative scheme of Algorithm 1, we have
\begin{equation*}
\begin{split}
\mathcal{L}_{\mu_{k}}(\mathscr{U}_{k+\!1},\mathscr{V}_{k+\;1},L_{k+\!1},S_{k+\!1},\mathscr{Y}_{k})\leq&\mathcal{L}_{\mu_{k}}(\mathscr{U}_{k+1},\mathscr{V}_{k+1},L_{k},S_{k},\mathscr{Y}_{k})\\
\leq&\mathcal{L}_{\mu_{k}}(\mathscr{U}_{k},\mathscr{V}_{k},L_{k},S_{k},\mathscr{Y}_{k})\\
=&\mathcal{L}_{\mu_{k-\!1}}(\mathscr{U}_{k},\mathscr{V}_{k},L_{k},S_{k},\mathscr{Y}_{k-\!1})+\alpha_{k}\sum^{3}_{j=1}\|Y^{k}_{j}\!-\!Y^{k-\!1}_{j}\|^2_{F}+\beta_{k}\frac{\|Y^{k}_{4}\!-\!Y^{k-\!1}_{4}\|^2_{F}}{\sqrt[3]{\mu^{2}_{k-\!1}}},
\end{split}
\end{equation*}
where $\alpha_{k}\!=\!\frac{\mu_{k-\!1}+\mu_{k}}{2(\mu_{k-\!1})^{2}}$, $\beta_{k}\!=\!\frac{\mu_{k-\!1}+\mu_{k}}{2(\mu_{k-1})^{4/3}}$, and the above equality holds due to the definition of the augmented Lagrangian function $\mathcal{L}_{\mu}(U,V,L,S,\widehat{U},\widehat{V},\{Y_{i}\})$. Since $\mu_{k}$ is non-decreasing, and $\sum^{\infty}_{k=1}(\mu_{k}/\mu^{4/3}_{k-\!1})<\infty$, then
\begin{displaymath}
\sum^\infty_{k=1}\beta_{k}=\sum^\infty_{k=1}\frac{\mu_{k-1}+\mu_{k}}{2\mu^{4/3}_{k-1}}\leq\sum^\infty_{k=1}\frac{\mu_{k}}{\mu^{4/3}_{k-1}}<\infty,
\end{displaymath}
\begin{displaymath}
\sum^\infty_{k=1}\alpha_{k}=\sum^\infty_{k=1}\frac{\mu_{k-1}+\mu_{k}}{2\mu^{2}_{k-1}}\leq\sum^\infty_{k=1}\frac{\mu_{k}}{\mu^{2}_{k-1}}<\sum^\infty_{k=1}\frac{\mu_{k}}{\mu^{4/3}_{k-1}}<\infty.
\end{displaymath}

Since $\{\|Y^{k}_{4}\|_{F}/\!\sqrt[3]{\mu_{k-\!1}}\}$ is bounded, and $\mu_{k}\!=\!\rho\mu_{k-\!1}$ and $\rho\!>\!1$, then $\{\|Y^{k-\!1}_{4}\|_{F}/\!\sqrt[3]{\mu_{k-\!1}}\}$ is also bounded, which implies that $\{{\|Y^{k}_{4}\!-\!Y^{k-1}_{4}\|^2_{F}}/\!{\sqrt[3]{\mu^{2}_{k-\!1}}}\}$ is bounded.
Then $\{\mathcal{L}_{\mu_{k}}(\mathscr{U}_{k+\!1},\mathscr{V}_{k+\!1},L_{k+\!1},S_{k+\!1},\mathscr{Y}_{k})\}$ is upper-bounded due to the boundedness of the sequences of all Lagrange multipliers, i.e., $\{Y^{k}_{1}\}$, $\{Y^{k}_{2}\}$, $\{Y^{k}_{3}\}$ and $\{{\|Y^{k}_{4}\!-\!Y^{k-1}_{4}\|^2_{F}}/\!{\sqrt[3]{\mu^{2}_{k}}}\}$.
\begin{displaymath}
\frac{\lambda}{2}(\|\widehat{U}_{k}\|_{\ast}\!+\!\|\widehat{V}_{k}\|_{\ast})+\|\mathcal{P}_{\Omega}(S_{k})\|^{1/2}_{\ell_{1/2}}=\mathcal{L}_{\mu_{k-\!1}}(\mathscr{U}_{k},\mathscr{V}_{k},L_{k},S_{k},\mathscr{Y}_{k-1})-\frac{1}{2}\sum^{4}_{i=1}\frac{\|Y^{k}_{i}\|^2_{F}\!-\!\|Y^{k-1}_{i}\|^2_{F}}{\mu_{k-1}}
\end{displaymath}
is upper-bounded (note that the above equality holds due to the definition of $\mathcal{L}_{\mu}(U,V,L,S,\widehat{U},\widehat{V},\{Y_{i}\})$), thus $\{S_{k}\}$, $\{\widehat{U}_{k}\}$ and $\{\widehat{V}_{k}\}$ are all bounded.

Similarly, by $U_{k}=\widehat{U}_{k}\!-\![Y^{k}_{1}\!-\!Y^{k-\!1}_{1}]/{\mu_{k-\!1}}$, $V_{k}=\widehat{V}_{k}\!-\![Y^{k}_{2}\!-\!Y^{k-\!1}_{2}]/{\mu_{k-\!1}}$, $L_{k}=U_{k}V^{T}_{k}\!-\![Y^{k}_{3}\!-\!Y^{k-\!1}_{3}]/{\mu_{k-\!1}}$ and the boundedness of $\{\widehat{U}_{k}\}$, $\{\widehat{V}_{k}\}$, $\{Y^{k}_{i}\}$ $(i\!=\!1,2,3)$, and $\{Y^{k}_{4}\!/\!\sqrt[3]{\mu_{k-\!1}}\}$, thus $\{U_{k}\}$, $\{V_{k}\}$ and $\{L_{k}\}$ are also bounded. This means that each bounded sequence must have a convergent subsequence due to the Bolzano-Weierstrass theorem.
\end{IEEEproof}

\vspace{5mm}
\textbf{Proof of Theorem 3:}
\begin{IEEEproof}
$\textbf{(I)}$ $\widehat{U}_{k+\!1}-U_{k+\!1}\!=\![Y^{k+\!1}_{1}\!-\!Y^{k}_{1}]/\mu_{k}$, $\widehat{V}_{k+\!1}-V_{k+\!1}\!=\![Y^{k+\!1}_{2}\!-\!Y^{k}_{2}]/\mu_{k}$, $U_{k+\!1}\!V^{T}_{k+\!1}-L_{k+\!1}\!=[Y^{k+\!1}_{3}\!-\!Y^{k}_{3}]/\mu_{k}$ and $L_{k+\!1}\!+\!S_{k+\!1}\!-\!D\!=\![Y^{k+\!1}_{4}\!-\!Y^{k}_{4}]/\mu_{k}$. Due to the boundedness of $\{Y^{k}_{1}\}$, $\{Y^{k}_{2}\}$, $\{Y^{k}_{3}\}$ and $\{Y^{k}_{4}/\!\sqrt[3]{\mu_{k-1}}\}$, the non-decreasing property of $\{\mu_{k}\}$, and $\sum^{\infty}_{k=0}(\mu_{k+\!1}/\mu^{4/3}_{k})\!<\!\infty$, we have
\begin{equation*}
\begin{split}
\sum^\infty_{k=0}\|\widehat{U}_{k+\!1}\!-\!U_{k+\!1}\|_{F}\leq\sum^\infty_{k=0}\frac{\mu_{k+\!1}}{\mu^{2}_{k}}\|Y^{k+\!1}_{1}\!-\!Y^{k}_{1}\|_{F}<\infty,\,&\;\sum^\infty_{k=0}\|\widehat{V}_{k+\!1}\!-\!V_{k+\!1}\|_{F}\leq\sum^\infty_{k=0}\frac{\mu_{k+\!1}}{\mu^{2}_{k}}\|Y^{k+\!1}_{2}\!-\!Y^{k}_{2}\|_{F}<\infty,\\
\sum^\infty_{k=0}\|L_{k+\!1}\!-\!U_{k+\!1}V^{T}_{k+\!1}\|_{F}\leq\sum^\infty_{k=0}\frac{\mu_{k+\!1}}{\mu^{2}_{k}}\|Y^{k+\!1}_{3}\!-\!Y^{k}_{3}\|_{F}<\infty,\,&\;\sum^\infty_{k=0}\|L_{k+\!1}\!+\!S_{k+\!1}\!-\!D\|_{F}\leq\sum^\infty_{k=0}\frac{\mu_{k+\!1}}{\mu^{5/3}_{k}}\frac{\|Y^{k+\!1}_{4}\!-\!Y^{k}_{4}\|_{F}}{\mu^{1/3}_{k}}<\infty,
\end{split}
\end{equation*}
which implies that
\begin{equation*}
\begin{split}
\lim_{k\rightarrow\infty}\|\widehat{U}_{k+\!1}-U_{k+\!1}\|_{F}=&0,\,\;\lim_{k\rightarrow\infty}\|\widehat{V}_{k+\!1}-V_{k+\!1}\|_{F}=0,\,\;\lim_{k\rightarrow\infty}\|L_{k+\!1}-U_{k+\!1}V^{T}_{k+\!1}\|_{F}=0,\\
&\textup{and}\;\, \lim_{k\rightarrow\infty}\|L_{k+\!1}+S_{k+\!1}-D\|_{F}=0.
\end{split}
\end{equation*}
Hence, $\{(U_{k},V_{k},\widehat{U}_{k},\widehat{V}_{k},L_{k},S_{k})\}$ approaches to a feasible solution. In the following, we will prove that the sequences $\{U_{k}\}$ and $\{V_{k}\}$ are Cauchy sequences.

Using $Y^{k}_{1}\!=\!Y^{k-\!1}_{1}\!+\!\mu_{k-\!1}(\widehat{U}_{k}\!-\!U_{k})$, $Y^{k}_{2}\!=\!Y^{k-\!1}_{2}\!+\!\mu_{k-\!1}(\widehat{V}_{k}\!-\!V_{k})$ and $Y^{k}_{3}\!=\!Y^{k-\!1}_{3}\!+\!\mu_{k-\!1}(U_{k}V^{T}_{k}\!-\!L_{k})$, then the first-order optimality conditions of (19) and (20) with respect to $U$ and $V$ are written as follows:
\begin{equation}\label{equ38}
\begin{split}
&\left(U_{k+1}V^{T}_{k}\!-\!L_{k}\!+\!\frac{Y^{k}_{3}}{\mu_{k}}\right)\!V_{k}\!+\!\left(U_{k+1}\!-\!\widehat{U}_{k}\!-\!\frac{Y^{k}_{1}}{\mu_{k}}\right)\\
=&\left(U_{k+\!1}\!V^{T}_{k}\!-\!U_{k}\!V^{T}_{k}\!-\!\frac{Y^{k-\!1}_{3}}{\mu_{k-\!1}}\!+\!\frac{Y^{k}_{3}}{\mu_{k-\!1}}\!+\!\frac{Y^{k}_{3}}{\mu_{k}}\right)\!V_{k}+\!U_{k+\!1}\!-\!U_{k}\!+\!U_{k}\!-\!\widehat{U}_{k}\!-\!\frac{Y^{k}_{1}}{\mu_{k}}\\
=&\,(U_{k+\!1}\!-\!U_{k})(V^{T}_{k}V_{k}\!+\!I_{d})+\!\left(\frac{Y^{k}_{3}\!-\!Y^{k-\!1}_{3}}{\mu_{k-\!1}}\!+\!\frac{Y^{k}_{3}}{\mu_{k}}\right)\!V_{k}+\frac{ Y^{k-\!1}_{1}\!-\!Y^{k}_{1}}{\mu_{k-\!1}}-\frac{Y^{k}_{1}}{\mu_{k}}\\
=&\,0,
\end{split}
\end{equation}
\begin{equation}\label{equ39}
\begin{split}
&\left(V_{k+\!1}U^{T}_{k+\!1}\!-\!L^{T}_{k}\!+\!\frac{(Y^{k}_{3})^{T}}{\mu_{k}}\right)\!U_{k+1}\!+\!\left(V_{k+1}\!-\!\widehat{V}_{k}\!-\!\frac{Y^{k}_{2}}{\mu_{k}}\right)\\
=&\left(V_{k+\!1}U^{T}_{k+\!1}\!-\!V_{k}U^{T}_{k}\!\!-\!\frac{(Y^{k-\!1}_{3})^{T}}{\mu_{k\!-\!1}}\!+\!\frac{(Y^{k}_{3})^{T}}{\mu_{k\!-\!1}}\!+\!\frac{(Y^{k}_{3})^{T}}{\mu_{k}}\right)\!U_{k\!+\!1}+V_{k+\!1}\!-\!V_{k}\!+\!V_{k}\!-\!\widehat{V}_{k}\!-\!\frac{Y^{k}_{2}}{\mu_{k}}\\
=&\,(V_{k+\!1}\!-\!V_{k})\!(U^{T}_{k+\!1}\!U_{k+\!1}\!+\!I_{d})\!+\!V_{k}(U^{T}_{k+\!1}\!-\!U^{T}_{k})U_{k+\!1}\!+\!\left(\frac{(Y^{k}_{3})^{T}\!-\!(Y^{k-\!1}_{3})^{T}}{\mu_{k-\!1}}\!+\!\frac{(Y^{k}_{3})^{T}}{\mu_{k}}\right)\!U_{k+\!1}\!+\!\frac{ Y^{k-\!1}_{2}\!-\!Y^{k}_{2}}{\mu_{k-\!1}}-\frac{Y^{k}_{2}}{\mu_{k}}\\
=&\,0.
\end{split}
\end{equation}
By (\ref{equ38}) and (\ref{equ39}), we obtain
\begin{equation*}
\begin{split}
&U_{k+1}-U_{k}\\
=&\,\left[\left(\frac{Y^{k-\!1}_{3}\!-\!Y^{k}_{3}}{\mu_{k-\!1}}-\frac{Y^{k}_{3}}{\mu_{k}}\right)\!V_{k}+\frac{ Y^{k}_{1}\!-\!Y^{k-\!1}_{1}}{\mu_{k-\!1}}+\frac{Y^{k}_{1}}{\mu_{k}}\right]\!\left(V^{T}_{k}\!V_{k}+I_{d}\right)^{-1},
\end{split}
\end{equation*}
\begin{equation*}
\begin{split}
&V_{k+1}-V_{k}\\
=&\left[V_{k}(U^{T}_{k}\!-U^{T}_{k+\!1})U_{k+\!1}\!+\!\left(\frac{(Y^{k-\!1}_{3})^{T}\!-\!(Y^{k}_{3})^{T}}{\mu_{k-\!1}}\!-\!\frac{(Y^{k}_{3})^{T}}{\mu_{k}}\right)\!U_{k+\!1}\!+\!\frac{ Y^{k}_{2}\!-\!Y^{k-\!1}_{2}}{\mu_{k-\!1}}\!+\!\frac{Y^{k}_{2}}{\mu_{k}}\right]\!\left(U^{T}_{k+\!1}U_{k+\!1}\!+\!I_{d}\right)^{-1}.
\end{split}
\end{equation*}

Recall that
\begin{displaymath}
\sum^{\infty}_{k=0}\frac{\mu_{k+1}}{\mu^{4/3}_{k}}<\infty,
\end{displaymath}
we have
\begin{displaymath}
\sum^{\infty}_{k=0}\|U_{k+1}-U_{k}\|_{F}\leq\sum^{\infty}_{k=0}\mu_{k}^{-1}\vartheta_{1}\leq\sum^{\infty}_{k=0}\frac{\mu_{k+1}}{\mu^{2}_{k}}\vartheta_{1}\leq\sum^{\infty}_{k=0}\frac{\mu_{k+1}}{\mu^{4/3}_{k}}\vartheta_{1}<\infty,
\end{displaymath}
where the constant $\vartheta_{1}$ is defined as
\begin{displaymath}
\vartheta_{1}=\max\left\{\left[\left(\rho\|Y^{k-\!1}_{3}\!-\!Y^{k}_{3}\|_{F}\!+\!\|Y^{k}_{3}\|_{F}\right)\|V_{k}\|_{F}\!+\!\rho\|Y^{k}_{1}\!-\!Y^{k-\!1}_{1}\|_{F}\!+\!\|Y^{k}_{1}\|_{F}\right] \!\|(V^{T}_{k}\!V_{k}\!+\!I_{d})^{-1}\|_{F},\,k=1,2,\cdots\right\}.
\end{displaymath}
In addition,
\begin{displaymath}
\begin{split}
&\sum^{\infty}_{k=0}\|V_{k+1}-V_{k}\|_{F}\\
\leq&\sum^{\infty}_{k=0}\frac{1}{\mu_{k}}\!\left[\left(\rho\|Y^{k-\!1}_{3}\!-\!Y^{k}_{3}\|_{F}\!+\!\|Y^{k}_{3}\|_{F}\right)\!\|U_{k+\!1}\|_{F}\!+\!\rho\|Y^{k}_{2}\!-\!Y^{k-\!1}_{2}\|_{F}\!+\!\|Y^{k}_{2}\|_{F}\right]\!\|(U^{T}_{k+\!1}U_{k+\!1}\!+\!I_{d})^{-\!1}\|_{F}\\
&+\sum^{\infty}_{k=0}\left(\|V_{k}\|_{F}\|U_{k+\!1}\|_{F}\|(U^{T}_{k+\!1}U_{k+\!1}\!+\!I_{d})^{-\!1}\|_{F}\right)\!\|U_{k+\!1}\!-\!U_{k}\|_{F}\\
\leq&\sum^{\infty}_{k=0}\vartheta_{2}\|U_{k+\!1}\!-\!U_{k}\|_{F}+\sum^{\infty}_{k=0}\frac{1}{\mu_{k}}\vartheta_{3}\leq\sum^{\infty}_{k=0}\vartheta_{2}\|U_{k+\!1}\!-\!U_{k}\|_{F}+\sum^{\infty}_{k=0}\frac{\mu_{k+1}}{\mu^{2}_{k}}\vartheta_{3}<\infty,
\end{split}
\end{displaymath}
where the constants $\vartheta_{2}$ and $\vartheta_{3}$ are defined as
\begin{displaymath}
\vartheta_{2}=\max\left\{\|V_{k}\|_{F}\|U_{k+\!1}\|_{F}\|(U^{T}_{k+\!1}U_{k+\!1}\!+\!I_{d})^{-\!1}\|_{F},\,k=1,2,\cdots\right\},
\end{displaymath}
\begin{displaymath}
\vartheta_{3}=\max\left\{\left[\left(\rho\|Y^{k-\!1}_{3}\!-\!Y^{k}_{3}\|_{F}\!+\!\|Y^{k}_{3}\|_{F}\right)\!\|U_{k+\!1}\|_{F}\!+\!\rho\|Y^{k}_{2}\!-\!Y^{k-\!1}_{2}\|_{F}\!+\!\|Y^{k}_{2}\|_{F}\right] \!\|(U^{T}_{k+\!1}\!U_{k+\!1}\!+\!I_{d})^{-1}\|_{F},\,k=1,2,\cdots\right\}.
\end{displaymath}

Consequently, both $\{U_{k}\}$ and $\{V_{k}\}$ are convergent sequences. Moreover, it is not difficult to verify that $\{U_{k}\}$ and $\{V_{k}\}$ are both Cauchy sequences.

Similarly, $\{\widehat{U}_{k}\}$, $\{\widehat{V}_{k}\}$, $\{S_{k}\}$ and $\{L_{k}\}$ are also Cauchy sequences. Practically, the stopping criterion of Algorithm 1 is satisfied within a finite number of iterations.

$\textbf{(II)}$ Let $(U_{\star},V_{\star},\widehat{U}_{\star},\widehat{V}_{\star},L_{\star}, S_{\star})$ be a critical point of (17), and $\Phi(S)=\|\mathcal{P}_{\Omega}(S)\|^{1/2}_{\ell_{1/2}}$. Applying the Fermat's rule as in~\cite{wang:admm} to the subproblem (27), we then obtain
\begin{equation*}
\begin{split}
&0\in \frac{\lambda}{2}\partial \|\widehat{U}_{\star}\|_{*}+(Y^{\star}_{1}) \;\, \textrm{and} \;\, 0\in \frac{\lambda}{2}\partial\|\widehat{V}_{\star}\|_{*}+(Y^{\star}_{2}),\\
&0\in \partial\Phi(S_{\star})+\mathcal{P}_{\Omega}(Y^{\star}_{4})\;\, \textrm{and} \;\, \mathcal{P}_{\Omega^{c}}(Y^{\star}_{4})=\textbf{0},\\
&L_{\star}=U_{\star}V^{T}_{\star},\,\widehat{U}_{\star}=U_{\star},\,\widehat{V}_{\star}=V_{\star},\,L_{\star}+S_{\star}=D.
\end{split}
\end{equation*}

Applying the Fermat's rule to (27), we have
\begin{equation}\label{equ41}
0\in \partial\Phi(S_{k+1})+\mathcal{P}_{\Omega}(Y^{k+1}_{4}),\;\;\mathcal{P}_{\Omega^{c}}(Y^{k+1}_{4})=0.
\end{equation}

In addition, the first-order optimal conditions for (23) and (24) are given by
\begin{equation}\label{equ42}
0\in\lambda\partial\|\widehat{U}_{k+1}\|_{*}+2Y^{k+1}_{1},\;\;0\in\lambda\partial\|\widehat{V}_{k+1}\|_{*}+2Y^{k+1}_{2}.
\end{equation}

Since $\{U_{k}\}$, $\{V_{k}\}$, $\{\widehat{U}_{k}\}$, $\{\widehat{V}_{k}\}$, $\{L_{k}\}$ and $\{S_{k}\}$ are Cauchy sequences, then
\begin{equation*}
\begin{split}
\lim_{k\rightarrow\infty}\|U_{k+\!1}\!-\!U_{k}\|=0,\,\;\lim_{k\rightarrow\infty}\|V_{k+\!1}\!-\!V_{k}\|= 0,\,\; \lim_{k\rightarrow\infty}\|\widehat{U}_{k+\!1}\!-\!\widehat{U}_{k}\|=0,\\
\lim_{k\rightarrow\infty}\|\widehat{V}_{k+\!1}\!-\!\widehat{V}_{k}\|= 0,\,\; \lim_{k\rightarrow\infty}\|L_{k+\!1}\!-\!L_{k}\|= 0,\,\;\lim_{k\rightarrow\infty}\|S_{k+\!1}\!-\!S_{k}\|=0.
\end{split}
\end{equation*}
Let $U_{\infty}$, $V_{\infty}$, $\widehat{U}_{\infty}$, $\widehat{V}_{\infty}$, $S_{\infty}$ and $L_{\infty}$ be their limit points, respectively. Together with the results in $\textbf{(I)}$, then we have that $U_{\infty}\!=\!\widehat{U}_{\infty}$, $V_{\infty}\!=\!\widehat{V}_{\infty}$, $L_{\infty}\!=\!U_{\infty}V^{T}_{\infty}$ and $L_{\infty}\!+\!S_{\infty}\!=\!D$. Using (\ref{equ41}) and (\ref{equ42}), the following holds
\begin{equation*}
\begin{split}
&0\in \frac{\lambda}{2}\partial\|\widehat{U}_{\infty}\|_{*}+(Y_{1})_{\infty}\;\,\textrm{and} \;\, 0\in \frac{\lambda}{2}\partial\|\widehat{V}_{\infty}\|_{*}+(Y_{2})_{\infty},\\
&0\in \partial\Phi(S_{\infty})+\mathcal{P}_{\Omega}((Y_{4})_{\infty})\;\,\textrm{and}\;\, \mathcal{P}_{\Omega^{c}}((Y_{4})_{\infty})=\textbf{0},\\
&L_{\infty}=U_{\infty}V^{T}_{\infty},\,\widehat{U}_{\infty}=U_{\infty},\,\widehat{V}_{\infty}=V_{\infty},\,L_{\infty}+S_{\infty}=D.
\end{split}
\end{equation*}

Therefore, any accumulation point $\{(U_{\infty},V_{\infty},\widehat{U}_{\infty}, \widehat{V}_{\infty}, L_{\infty}, S_{\infty})\}$ of the sequence $\{(U_{k},V_{k},\widehat{U}_{k}, \widehat{U}_{k}, L_{k}, S_{k})\}$ generated by Algorithm 1 satisfies the KKT conditions for the problem (17). That is, the sequence asymptotically satisfies the KKT conditions of (17). In particular, whenever the sequence $\{(U_{k}, V_{k}, L_{k}, S_{k})\}$ converges, it converges to a critical point of (15). This completes the proof.
\end{IEEEproof}

\section*{APPENDIX F: Stopping Criterion}
For the problem (15), the KKT conditions are
\begin{equation}\label{equ51}
\begin{split}
&0\in \frac{\lambda}{2}\partial \|U^{\star}\|_{*}+Y^{\star}V^{\star} \;\, \textrm{and} \;\, 0\in \frac{\lambda}{2}\partial\|V^{\star}\|_{*}+(Y^{\star})^{T}U^{\star},\\
&0\in \partial\Phi(S^{\star})+\mathcal{P}_{\Omega}(\widehat{Y}^{\star})\;\, \textrm{and} \;\, \mathcal{P}_{\Omega^{c}}(\widehat{Y}^{\star})=\textbf{0},\\
&L^{\star}=U^{\star}(V^{\star})^{T},\,\mathcal{P}_{\Omega}(L^{\star})+\mathcal{P}_{\Omega}(S^{\star})=\mathcal{P}_{\Omega}(D).
\end{split}
\end{equation}
Using \eqref{equ51}, we have
\begin{equation}\label{equ52}
-Y^{\star}\in \frac{\lambda}{2}\partial \|U^{\star}\|_{*}(V^{\star})^{\dag} \;\, \textrm{and} \;\, -Y^{\star}\in [(U^{\star})^{T}]^{\dag}(\frac{\lambda}{2}\partial\|V^{\star}\|_{*})^{T}
\end{equation}
where $(V^{\star})^{\dag}$ is the pseudo-inverse of $V^{\star}$. The two conditions hold if and only if
\begin{equation*}
\partial \|U^{\star}\|_{*}(V^{\star})^{\dag}\cap[(U^{\star})^{T}]^{\dag}(\partial\|V^{\star}\|_{*})^{T}\neq\emptyset.
\end{equation*}
Recalling the equivalence relationship between (15) and (17), the KKT conditions for (17) are given by
\begin{equation}\label{equ53}
\begin{split}
&0\in \frac{\lambda}{2}\partial \|\widehat{U}^{\star}\|_{*}+Y^{\star}_{1}\;\, \textrm{and} \;\, 0\in \frac{\lambda}{2}\partial\|\widehat{V}^{\star}\|_{*}+Y^{\star}_{2},\\
&0\in \partial\Phi(S^{\star})+\mathcal{P}_{\Omega}(\widehat{Y}^{\star})\;\, \textrm{and} \;\, \mathcal{P}_{\Omega^{c}}(\widehat{Y}^{\star})=\textbf{0},\\
&L^{\star}\!=\!U^{\star}(V^{\star})^{T},\widehat{U}^{\star}\!=\!U^{\star},\widehat{V}^{\star}\!=\!V^{\star},L^{\star}\!+\!S^{\star}\!=\!D.
\end{split}
\end{equation}
Hence, we use the following conditions as the stopping criteria for Algorihtm 1:
\begin{equation*}
\max\left\{{\epsilon_{1}}/{\|D\|_{F}},\,\epsilon_{2}\right\}<\epsilon
\end{equation*}
where $\epsilon_{1}\!=\!\max\{\|U_{k}\!V^{T}_{k}\!-\!L_{k}\|_{F},\|L_{k}\!+\!S_{k}\!-\!D\|_{F},\|Y^{k}_{1}(\widehat{V}_{k})^{\dagger}\!-\!(\widehat{U}^{T}_{k})^{\dagger}(Y^{k}_{2})^{T}\|_{F}\}$ and $\epsilon_{2}\!=\!\max\{\|\widehat{U}_{k}\!-\!U_{k}\|_{F}/\|U_{k}\|_{F},\|\widehat{V}_{k}\!-\!V_{k}\|_{F}/\|V_{k}\|_{F}\}$.

\begin{algorithm}[t]
\caption{Solving image recovery problem \eqref{equ62} via ADMM}
\label{alg3}
\renewcommand{\algorithmicrequire}{\textbf{Input:}}
\renewcommand{\algorithmicensure}{\textbf{Initialize:}}
\renewcommand{\algorithmicoutput}{\textbf{Output:}}
\begin{algorithmic}[1]
\REQUIRE $\mathcal{P}_{\Omega}(D)$, the given rank $d$, and $\lambda$.
\ENSURE $\mu_{0}\!=\!10^{-4}$, $\mu_{\max}\!=\!10^{20}$, $\rho\!>\!1$, $k\!=\!0$, and $\epsilon$.
\WHILE {not converged}
\STATE {$U_{k+1}=\left[(L_{k}+{Y^{k}_{3}}/{\mu_{k}})V_{k}+\!\widehat{U}_{k}-{Y^{k}_{1}}/{\mu_{k}}\right]\left(V^{T}_{k}V_{k}+I\right)^{-1}$.}
\STATE {$V_{k+1}=\left[(L_{k}+{Y^{k}_{3}}/{\mu_{k}})^{T}U_{k+1}+\widehat{V}_{k}-{Y^{k}_{2}}/{\mu_{k}}\right]\left(U^{T}_{k+\!1}U_{k+\!1}+I\right)^{-1}$.}
\STATE {$\widehat{U}_{k+1}=\mathcal{D}_{\lambda/(2\mu_{k})}\!\left(U_{k+1}+Y^{k}_{1}/\mu_{k}\right)$, and $\widehat{V}_{k+1}=\mathcal{D}_{\lambda/(2\mu_{k})}\!\left(V_{k+1}+{Y^{k}_{2}}/{\mu_{k}}\right)$. }
\STATE {$L_{k+1}=\mathcal{P}_{\Omega}\!\left(\frac{D+\mu_{k}U_{k+1}V^{T}_{k+1}-Y^{k}_{3}}{1+\mu_{k}}\right)+\mathcal{P}^{\perp}_{\Omega}\!\left(U_{k+1}V^{T}_{k+1}-Y^{k}_{3}/\mu_{k}\right)$.}
\STATE {$Y^{k+\!1}_{1}=Y^{k}_{1}+\mu_{k}\!\left(U_{k+\!1}-\widehat{U}_{k+\!1}\right)$, $Y^{k+\!1}_{2}=Y^{k}_{2}+\mu_{k}\!\left(V_{k+\!1}-\widehat{V}_{k+\!1}\right)$, and $Y^{k+\!1}_{3}=Y^{k}_{3}+\mu_{k}\!\left(L_{k+\!1}-U_{k+\!1}V^{T}_{k+\!1}\right)$.}
\STATE {$\mu_{k+1}=\min\left(\rho\mu_{k},\,\mu_{\max}\right)$.}
\STATE $k\leftarrow k+1$.
\ENDWHILE
\OUTPUT $U_{k+1}$ and $V_{k+1}$.
\end{algorithmic}
\end{algorithm}

\section*{APPENDIX G: Algorithms for Image Recovery}
In this part, we propose two efficient ADMM algorithms (as outlined in \textbf{Algorithms}~\ref{alg3} and \ref{alg4}) to solve the following D-N/F-N penalty regularized least squares problems for matrix completion:
\begin{equation}\label{equ62}
\min_{U,V,L} \frac{\lambda}{2}\!\left(\|U\|_{*}\!+\!\|V\|_{*}\right)+\frac{1}{2}\|\mathcal{P}_{\Omega}(L)-\mathcal{P}_{\Omega}(D)\|^{2}_{F},\;\textup{s.t.}, L=UV^{T},
\end{equation}

\begin{equation}\label{equ63}
\min_{U,V,L} \frac{\lambda}{3}\!\left(\|U\|^{2}_{F}\!+\!2\|V\|_{*}\right)+\frac{1}{2}\|\mathcal{P}_{\Omega}(L)-\mathcal{P}_{\Omega}(D)\|^{2}_{F},\;\textup{s.t.}, L=UV^{T}.
\end{equation}

Similar to (17) and (18), we also introduce the matrices $\widehat{U}$ and $\widehat{V}$ as auxiliary variables to \eqref{equ62} (i.e., (35) in this paper), and obtain the following equivalent formulation,
\begin{equation}\label{equ11}
\begin{split}
&\min_{U,V,\widehat{U},\widehat{V},L} \frac{\lambda}{2}(\|\widehat{U}\|_{\ast}\!+\!\|\widehat{V}\|_{\ast})+\frac{1}{2}\|\mathcal{P}_{\Omega}(L)\!-\!\mathcal{P}_{\Omega}(D)\|^{2}_{F},\\
&\;\;\;\;\textup{s.t.},\,L =UV^{T},\,U=\widehat{U},\,V=\widehat{V}.
\end{split}
\end{equation}

The augmented Lagrangian function of \eqref{equ11} is
\begin{equation*}
\begin{split}
\mathcal{L}_{\mu}=&\frac{\lambda}{2}(\|\widehat{U}\|_{\ast}\!+\!\|\widehat{V}\|_{\ast})\!+\!\frac{1}{2}\|\mathcal{P}_{\Omega}(L)\!-\!\mathcal{P}_{\Omega}(D)\|^{2}_{F}+\!\langle Y_{1},U\!-\!\widehat{U}\rangle\!+\!\langle Y_{2},V\!-\!\widehat{V}\rangle\!+\!\langle Y_{3},L\!-\!UV^{T}\rangle\\
&\quad\quad\qquad+\!\frac{\mu}{2}(\|U\!-\!\widehat{U}\|^{2}_{F}\!+\!\|V\!-\!\widehat{V}\|^{2}_{F}\!+\!\|L\!-\!UV^{T}\|^{2}_{F})
\end{split}
\end{equation*}
where $Y_{i}\,(i\!=\!1,2,3)$ are the matrices of Lagrange multipliers.

\subsection*{Updating $U_{k+1}$ and $V_{k+1}$:}
By fixing the other variables at their latest values, and removing the terms that do not depend on $U$ and $V$ and adding some proper terms that do not depend on $U$ and $V$, the optimization problems with respect to $U$ and $V$ are formulated as follows:
\begin{equation}\label{equ12}
\|U-\widehat{U}_{k}+Y^{k}_{1}/\mu_{k}\|^{2}_{F}+\|L_{k}-UV^{T}_{k}+Y^{k}_{3}/\mu_{k}\|^{2}_{F},
\end{equation}
\begin{equation}\label{equ13}
\|V - \widehat{V}_{k} + Y^{k}_{2}/\mu_{k}\|^{2}_{F}+\|L_{k}-U_{k+1}V^{T}+Y^{k}_{3}/\mu_{k}\|^{2}_{F}.
\end{equation}
Since both \eqref{equ12} and \eqref{equ13} are smooth convex optimization problems, their closed-form solutions are given by
\begin{equation}\label{equ14}
U_{k+\!1}=\left[(L_{k}+{Y^{k}_{3}}/{\mu_{k}})V_{k}+\widehat{U}_{k}-{Y^{k}_{1}}/{\mu_{k}}\right]\left(V^{T}_{k}V_{k}+I\right)^{-1},
\end{equation}
\begin{equation}\label{equ15}
V_{k+\!1}=\left[(L_{k}+{Y^{k}_{3}}/{\mu_{k}})^{T}U_{k+\!1}+\widehat{V}_{k}-{Y^{k}_{2}}/{\mu_{k}}\!\right]\left(U^{T}_{k+\!1}U_{k+\!1}+I\right)^{-1}.
\end{equation}

\subsection*{Updating $\widehat{U}_{k+1}$ and $\widehat{V}_{k+1}$:}
By keeping all other variables fixed, $\widehat{U}_{k+1}$ is updated by solving the following problem:
\begin{equation}\label{equ16}
\frac{\lambda}{2}\|\widehat{U}\|_{\ast}+\frac{\mu_{k}}{2}\|U_{k+1}-\widehat{U}+Y^{k}_{1}/\mu_{k}\|^{2}_{F}.
\end{equation}
To solve \eqref{equ16}, the SVT operator~\cite{cai:svt} is considered as follows:
\begin{equation}\label{equ17}
\widehat{U}_{k+1}\!=\!\mathcal{D}_{\lambda/(2\mu_{k})}\!\left(U_{k+1}\!+\!Y^{k}_{1}/\mu_{k}\right).
\end{equation}

Similarly, $\widehat{V}_{k+1}$ is given by
\begin{equation}\label{equ18}
\widehat{V}_{k+1}=\mathcal{D}_{\lambda/(2\mu_{k})}\!\left(V_{k+1}+{Y^{k}_{2}}/{\mu_{k}}\right).
\end{equation}

\subsection*{Updating $L_{k+1}$:}
By fixing all other variables, the optimal $L_{k+1}$ is the solution to the following problem:
\begin{equation}\label{equ19}
\frac{1}{2}\|\mathcal{P}_{\Omega}(L)-\mathcal{P}_{\Omega}(D)\|^{2}_{F}+\frac{\mu_{k}}{2}\|L-U_{k+1}V^{T}_{k+1}+\frac{Y^{k}_{3}}{\mu_{k}}\|^{2}_{F}.
\end{equation}
Since \eqref{equ19} is a smooth convex optimization problem, it is easy to show that the optimal solution to \eqref{equ19} is
\begin{equation}\label{equ20}
\begin{split}
L_{k+1}=\mathcal{P}_{\Omega}\!\left(\frac{D+\mu_{k} U_{k+1}V^{T}_{k+1}-Y^{k}_{3}}{1+\mu_{k}}\right)
+\mathcal{P}^{\perp}_{\Omega}\!\left(U_{k+1}V^{T}_{k+1}-Y^{k}_{3}/\mu_{k}\right)
\end{split}
\end{equation}
where $\mathcal{P}^{\perp}_{\Omega}$ is the complementary operator of $\mathcal{P}_{\Omega}$, i.e., $\mathcal{P}^{\perp}_{\Omega}(D)_{ij}\!=\!0$ if $(i,j)\!\in\!\Omega$, and $\mathcal{P}^{\perp}_{\Omega}(D)_{ij}\!=\!D_{ij}$ otherwise.

Based on the description above, we develop an efficient ADMM algorithm for solving the double nuclear norm minimization problem \eqref{equ62}, as outlined in \textbf{Algorithm~\ref{alg3}}. Similarly, we also present an efficient ADMM algorithm to solve \eqref{equ63}, as outlined in \textbf{Algorithm~\ref{alg4}}.

\begin{algorithm}[t]
\caption{Solving image recovery problem \eqref{equ63} via ADMM}
\label{alg4}
\renewcommand{\algorithmicrequire}{\textbf{Input:}}
\renewcommand{\algorithmicensure}{\textbf{Initialize:}}
\renewcommand{\algorithmicoutput}{\textbf{Output:}}
\begin{algorithmic}[1]
\REQUIRE $\mathcal{P}_{\Omega}(D)$, the given rank $d$, and $\lambda$.
\ENSURE $\mu_{0}\!=\!10^{-4}$, $\mu_{\max}\!=\!10^{20}$, $\rho\!>\!1$, $k\!=\!0$, and $\epsilon$.
\WHILE {not converged}
\STATE {$U_{k+1}=\left[(\mu_{k}L_{k}+Y^{k}_{2})V_{k}\right]\left(\mu_{k}V^{T}_{k}V_{k}+(2\lambda/3)I\right)^{-1}$.}
\STATE {$V_{k+1}=\left[(L_{k}+{Y^{k}_{2}}/{\mu_{k}})^{T}U_{k+1}+\widehat{V}_{k}-{Y^{k}_{1}}/{\mu_{k}}\right]\left(U^{T}_{k+\!1}U_{k+\!1}+I\right)^{-1}$.}
\STATE {$\widehat{V}_{k+1}=\mathcal{D}_{2\lambda/(3\mu_{k})}\!\left(V_{k+1}+{Y^{k}_{1}}/{\mu_{k}}\right)$.}
\STATE {
$L_{k+1}=\mathcal{P}_{\Omega}\!\left(\frac{D+\mu_{k} U_{k+1}V^{T}_{k+1}-Y^{k}_{2}}{1+\mu_{k}}\right)+\mathcal{P}^{\perp}_{\Omega}\!\left(U_{k+1}V^{T}_{k+1}-Y^{k}_{2}/\mu_{k}\right)$.}
\STATE {$Y^{k+1}_{1}\!=\!Y^{k}_{1}\!+\!\mu_{k}\!\left(V_{k+1}\!-\!\widehat{V}_{k+1}\right)$ and $Y^{k+1}_{2}\!=\!Y^{k}_{2}\!+\!\mu_{k}\!\left(L_{k+1}\!-\!U_{k+1}V^{T}_{k+1}\right)$.}
\STATE {$\mu_{k+1}=\min\left(\rho\mu_{k},\,\mu_{\max}\right)$.}
\STATE $k\leftarrow k+1$.
\ENDWHILE
\OUTPUT $U_{k+1}$ and $V_{k+1}$.
\end{algorithmic}
\end{algorithm}

\section*{APPENDIX H: More Experimental Results}
In this paper, we compared both our methods with the state-of-the-art methods, such as LMaFit\footnote{\url{http://lmafit.blogs.rice.edu/}}~\cite{shen:alad}, RegL1\footnote{\url{https://sites.google.com/site/yinqiangzheng/}}~\cite{zheng:lrma}, Unifying~\cite{cabral:nnbf}, factEN\footnote{\url{http://cpslab.snu.ac.kr/people/eunwoo-kim}}~\cite{kim:factEN}, RPCA\footnote{\url{http://www.cis.pku.edu.cn/faculty/vision/zlin/zlin.htm}}~\cite{lin:almm}, PSVT\footnote{\url{http://thohkaistackr.wix.com/page}}~\cite{oh:psvt}, WNNM\footnote{\url{http://www4.comp.polyu.edu.hk/~cslzhang/}}~\cite{gu:wnnmj}, and LpSq\footnote{\url{https://sites.google.com/site/feipingnie/}}~\cite{nie:rmc}. The Matlab code of the proposed methods can be downloaded from the link\footnote{\url{https://www.dropbox.com/s/npyc2t5zkjlb7tt/Code_BFMNM.zip?dl=0}}.

\subsection*{Convergence Behavior}
Fig.\ \ref{fig3} illustrates the evolution of the relative squared error (RSE), i.e.\ $\|L\!-\!\overline{L}\|_{F}/\|\overline{L}\|_{F}$, and stop criterion over the iterations on corrupted matrices of size $1,000\!\times\!1,000$ with outlier ratio $5\%$, respectively. From the results, it is clear that both the stopping tolerance and RSE values of our two methods decrease fast, and they converge within only a small number of iterations, usually within 50 iterations.

\begin{figure}[t]
\centering
\subfigure[Stop criterion: $(\textup{S+L})_{1/2}$ (left) and $(\textup{S+L})_{2/3}$ (right)]{\includegraphics[width=0.241\linewidth]{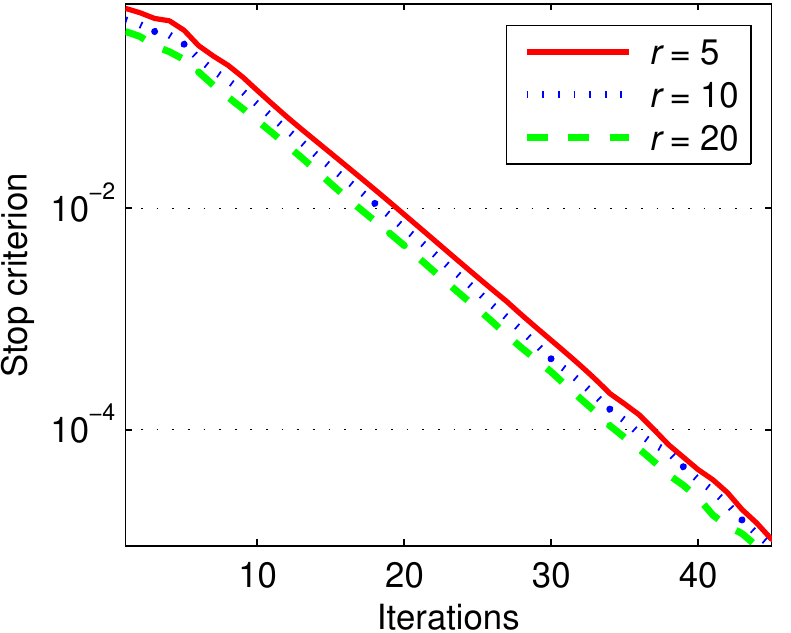}
\includegraphics[width=0.241\linewidth]{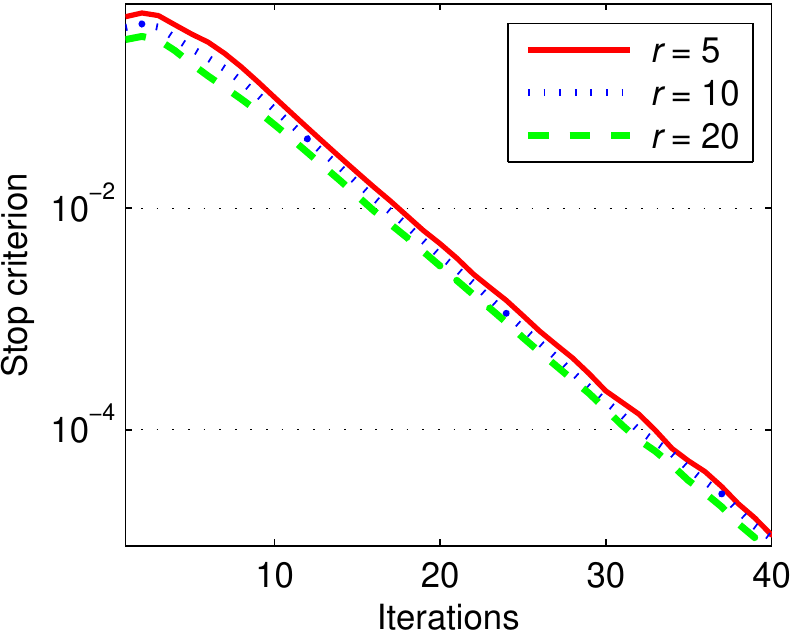}}\;\;\;
\subfigure[RSE: $(\textup{S+L})_{1/2}$ (left) and $(\textup{S+L})_{2/3}$ (right)]{\includegraphics[width=0.241\linewidth]{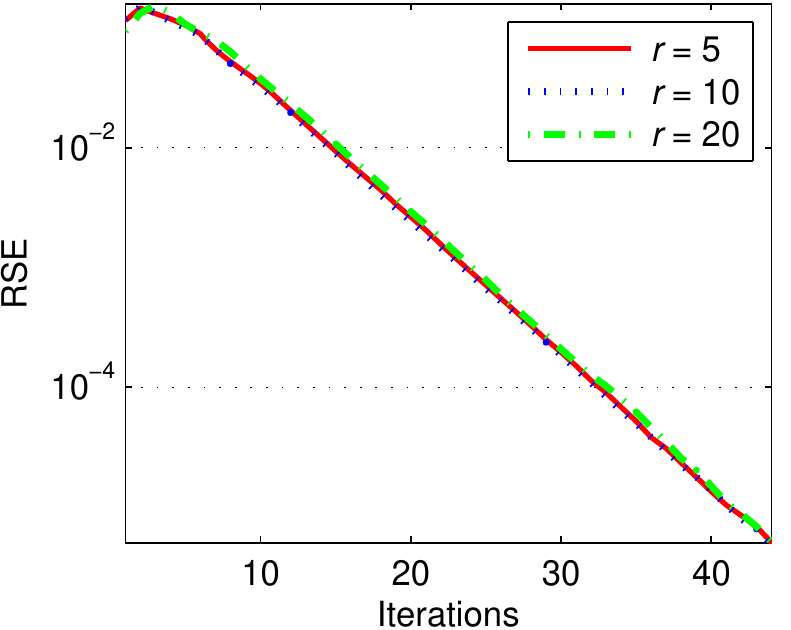}
\includegraphics[width=0.241\linewidth]{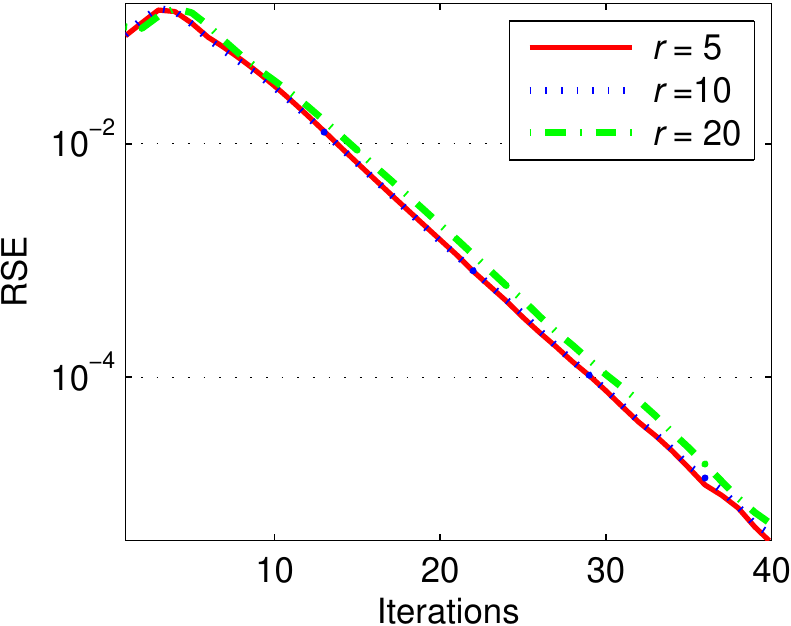}}
\caption{Convergence behavior of our $(\textup{S+L})_{1/2}$ and $(\textup{S+L})_{2/3}$ methods for the cases of the matrix rank 5, 10 and 20.}
\label{fig3}
\end{figure}

\subsection*{Robustness}
\begin{figure}[t]
\centering
\subfigure[RSE vs.\ $d$: $200\!\times\!200$ (left) and $500\!\times\!500$ (right)]{\includegraphics[width=0.241\linewidth]{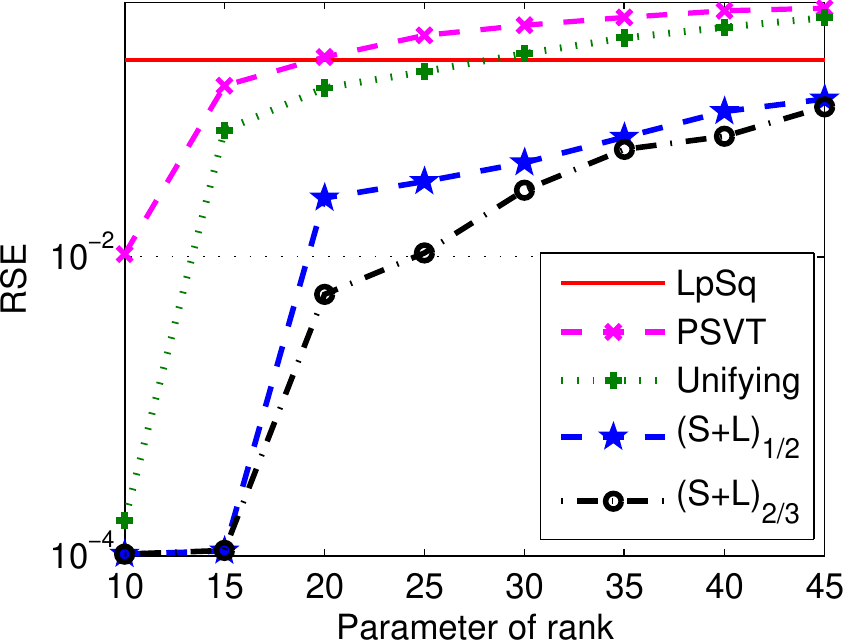}
\includegraphics[width=0.241\linewidth]{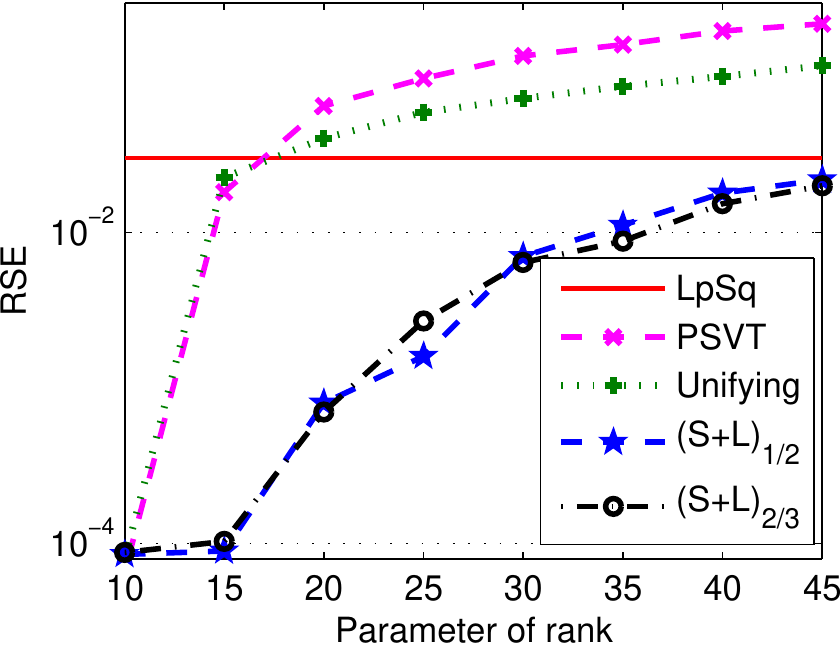}\label{fig41}}\;\;\;
\subfigure[RSE vs.\ $\lambda$: $200\!\times\!200$ (left) and $500\!\times\!500$ (right)]{\includegraphics[width=0.241\linewidth]{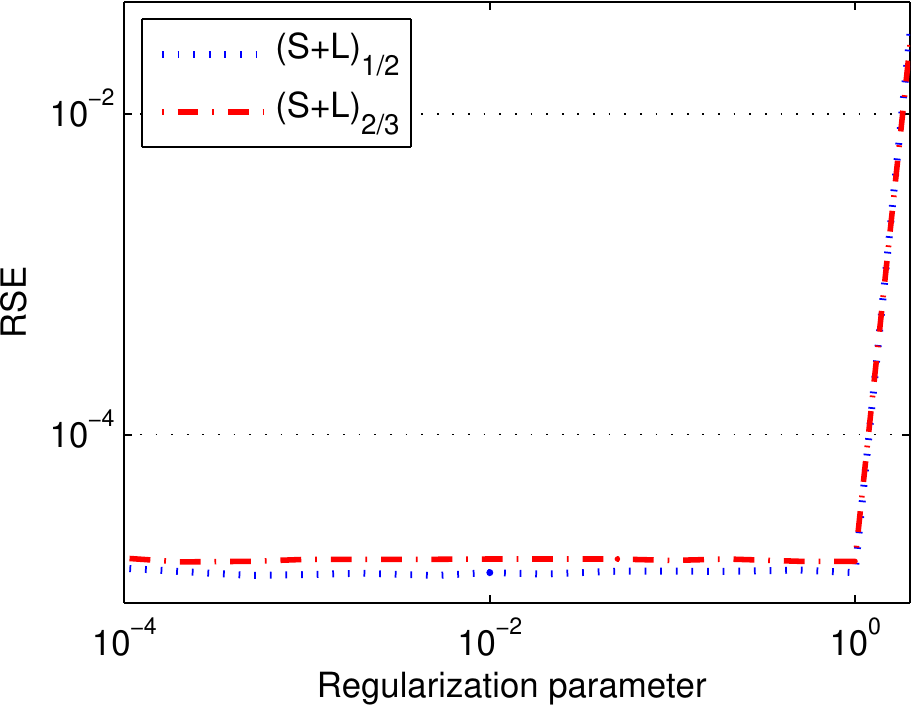}
\includegraphics[width=0.241\linewidth]{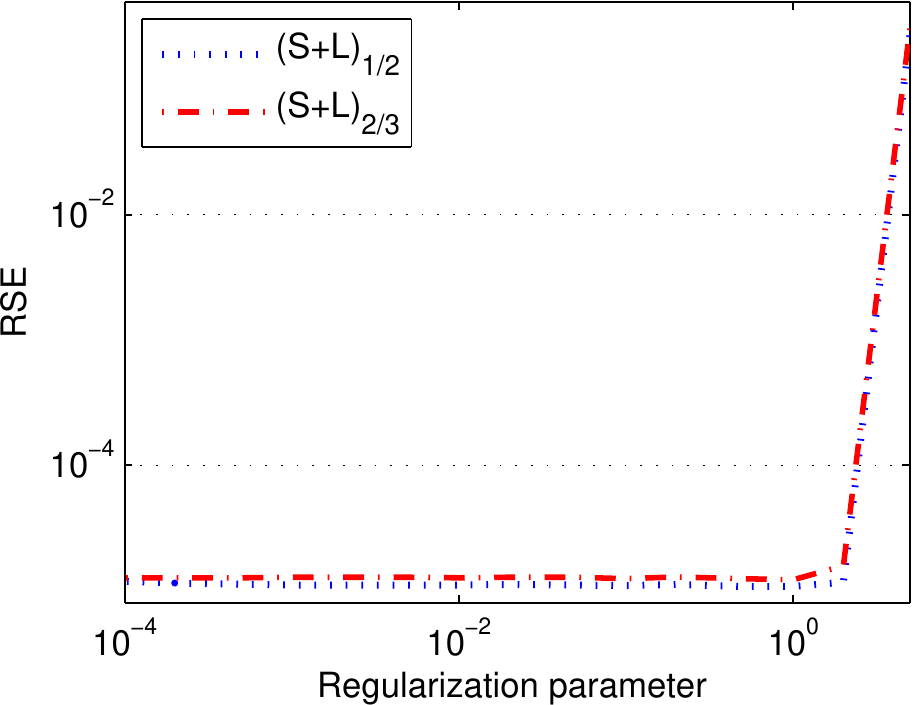}\label{fig42}}
\caption{Comparison of RSE results on corrupted matrices with different rank parameters (a) and regularization parameters (b).}
\label{fig4}
\end{figure}

Like the other non-convex methods such as PSVT and Unifying, the most important parameter of our methods is the rank parameter $d$. To verify the robustness of our methods with respect to $d$, we report the RSE results of PSVT, Unifying and our methods on corrupted matrices with outlier ratio 10\% in Fig.\ \ref{fig41}, in which we also present the results of the baseline method, LpSq~\cite{nie:rmc}. It is clear that both our methods perform much more robust than PSVT and Unifying, and consistently yield much better solutions than the other methods in all settings.

To verify the robustness of both our methods with respect to another important parameter (i.e.\ the regularization parameter $\lambda$), we also report the RSE results of our methods on corrupted matrices with outlier ratio 10\% in Fig.\ \ref{fig42}. Note that the rank parameter of both our methods is computed by our rank estimation procedure. From the resutls, one can see that both our methods demonstrate very robust performance over a wide range of the regularization parameter, e.g.\ from $10^{-4}$ to $10^{0}$.

\begin{table}[t]
\centering
\renewcommand{\arraystretch}{1.26}
\small
\caption{Information of the surveillance video sequences used in our experiments.}
\label{tab_sim1}
\begin{tabular}{r|c|c|c|c|c}
\hline
{Datasets} & {Bootstrap} & {Hall} &{Lobby}  &{Mall} & {WaterSurface}\\
\hline
{Size$\times\sharp$frames}  &$[120,160]\times1000$	&$[144,176]\times1000$ &$[128,160]\times1000$	&$[256,320]\times600$	&$[128,160]\times500$\\
{Description}     &Crowded scene  &Crowded scene   &Dynamic foreground	&Crowded scene	&Dynamic background\\
\hline
\end{tabular}
\end{table}

\begin{figure}[t]
\centering
\includegraphics[width=0.369\linewidth]{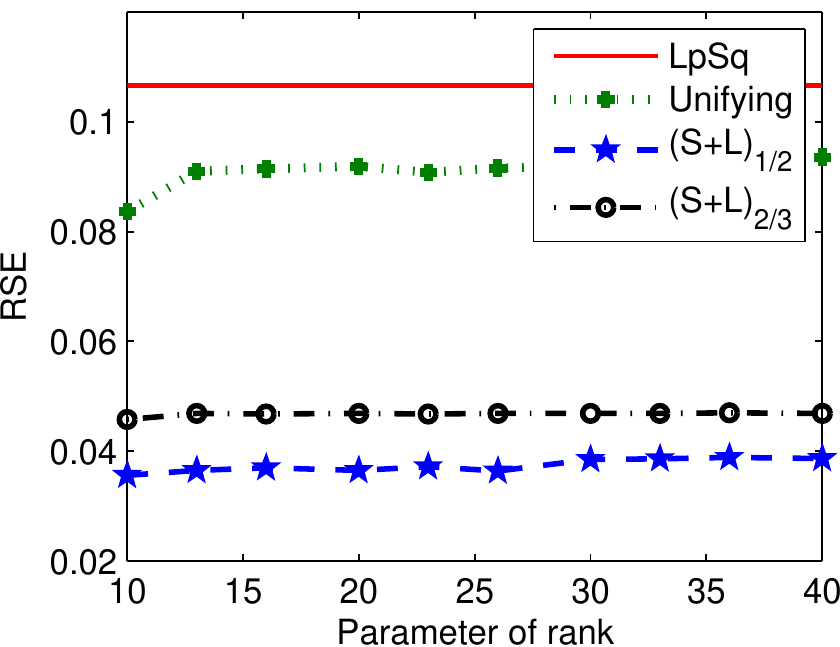}\quad\qquad\quad
\includegraphics[width=0.369\linewidth]{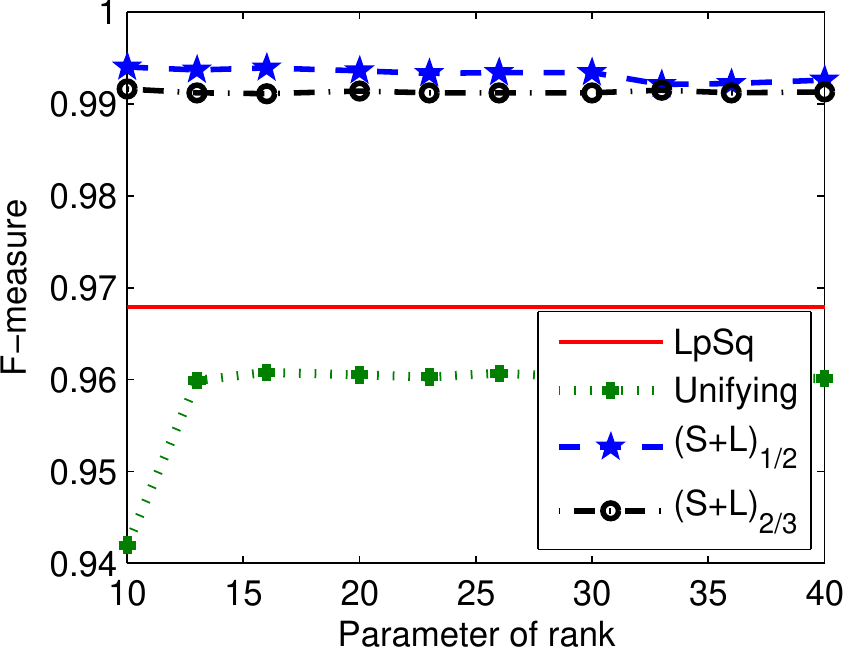}
\caption{The text removal results (including RSE (left) and F-measure (right)) of LpSq~\cite{nie:rmc}, Unifying~\cite{cabral:nnbf} and our methods with different rank parameters.}
\label{fig45}
\end{figure}

\begin{figure*}[th]
\centering
\includegraphics[width=0.157\linewidth]{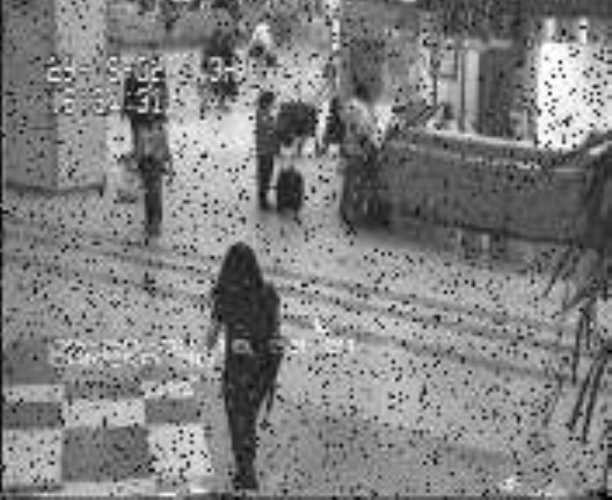}
\includegraphics[width=0.157\linewidth]{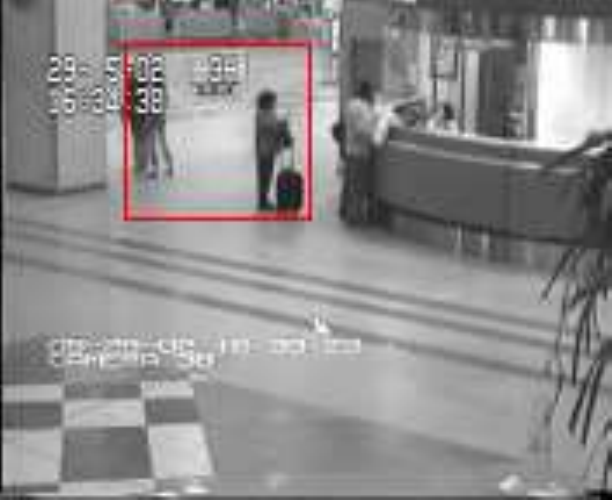}
\includegraphics[width=0.157\linewidth]{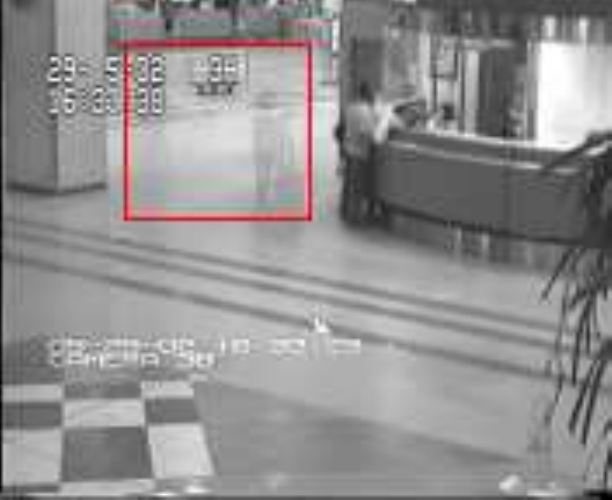}
\includegraphics[width=0.157\linewidth]{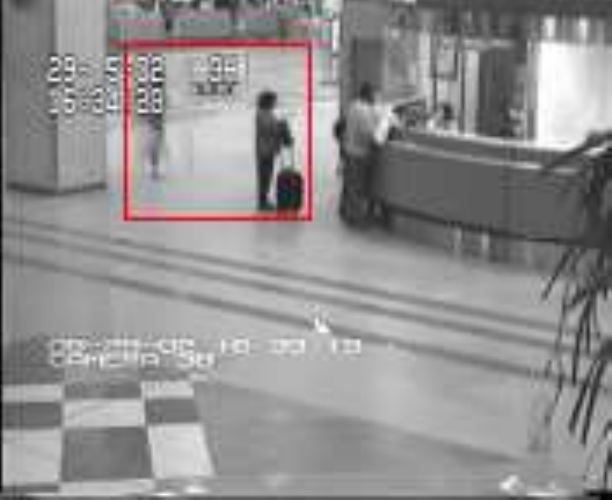}
\includegraphics[width=0.157\linewidth]{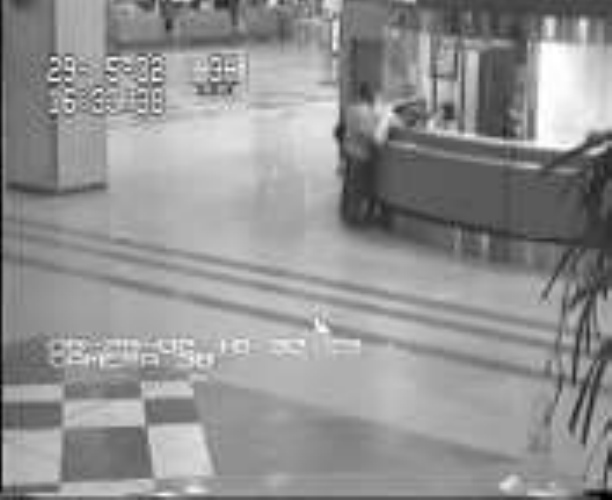}
\includegraphics[width=0.157\linewidth]{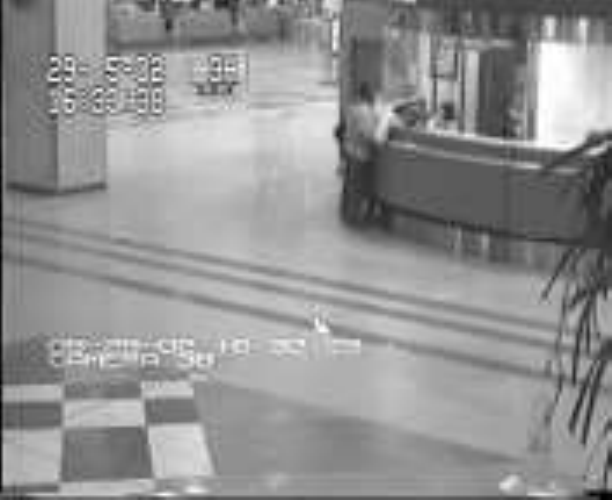}
\includegraphics[width=0.157\linewidth]{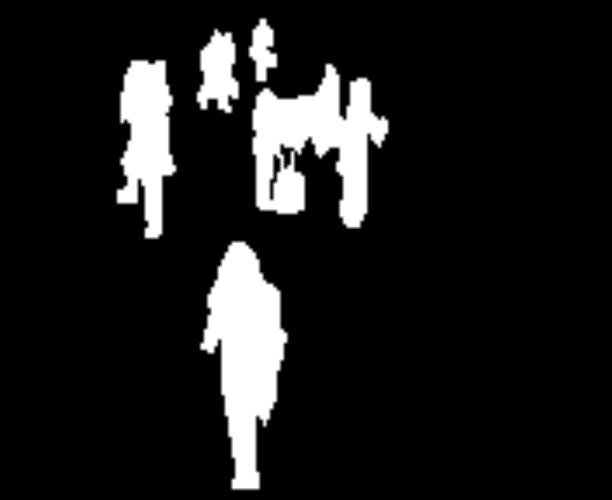}
\includegraphics[width=0.157\linewidth]{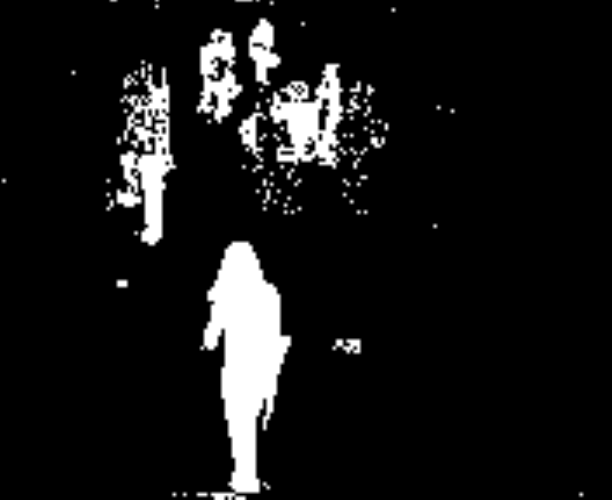}
\includegraphics[width=0.157\linewidth]{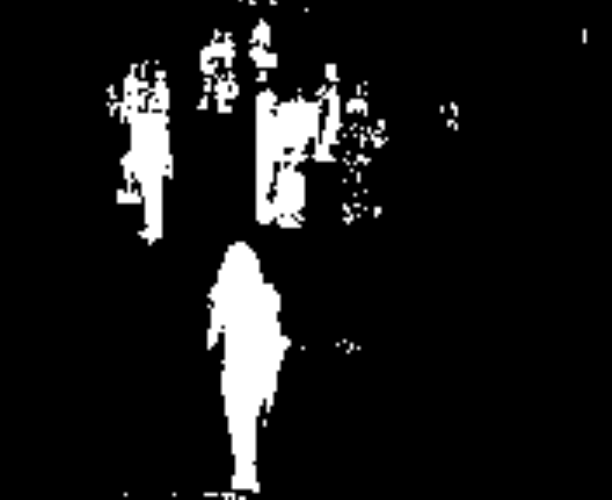}
\includegraphics[width=0.157\linewidth]{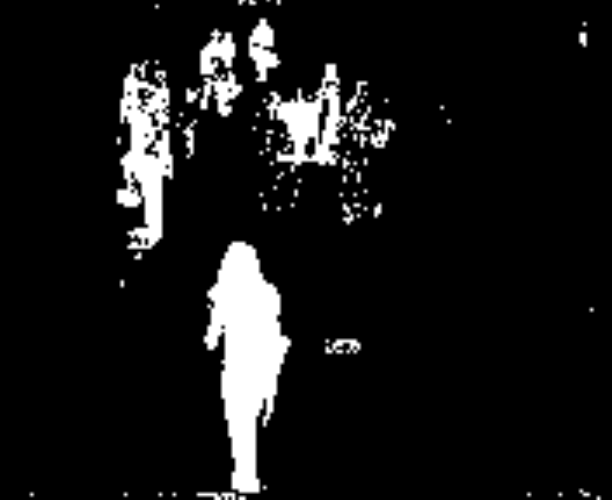}
\includegraphics[width=0.157\linewidth]{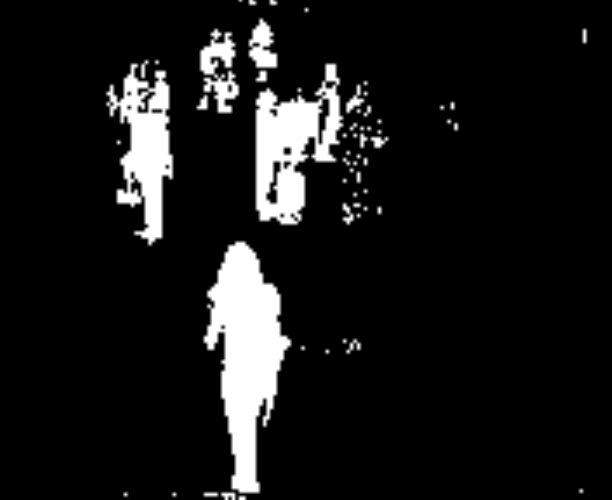}
\includegraphics[width=0.157\linewidth]{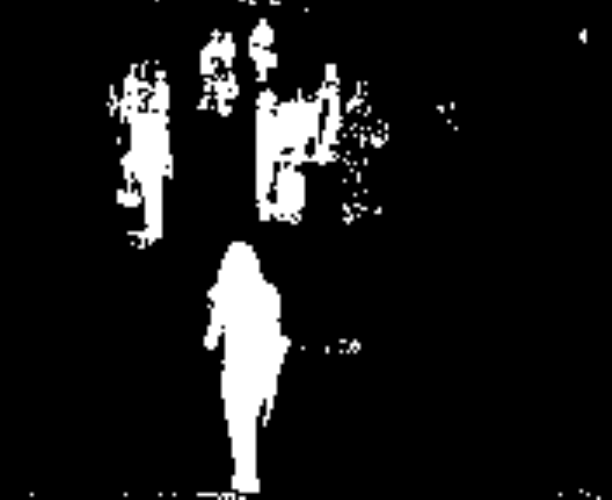}\\~\\
\vspace{-1.8mm}

\includegraphics[width=0.157\linewidth]{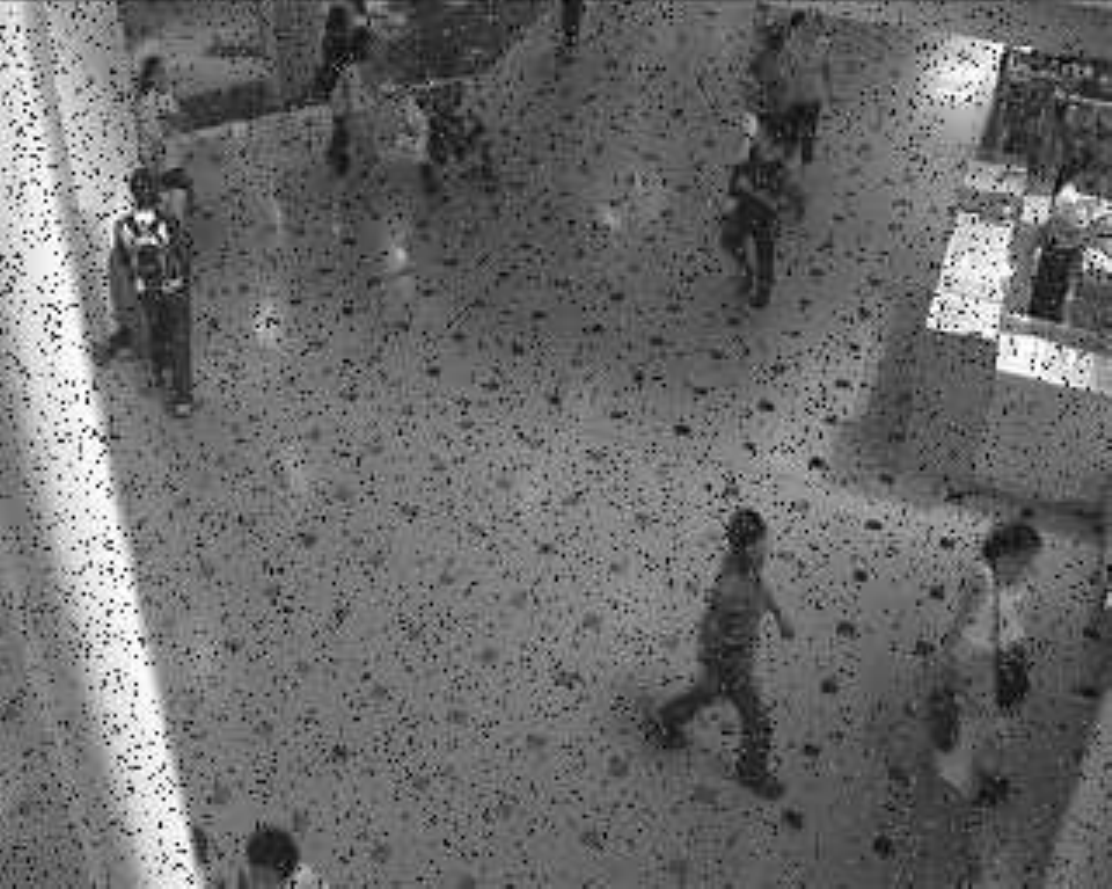}
\includegraphics[width=0.157\linewidth]{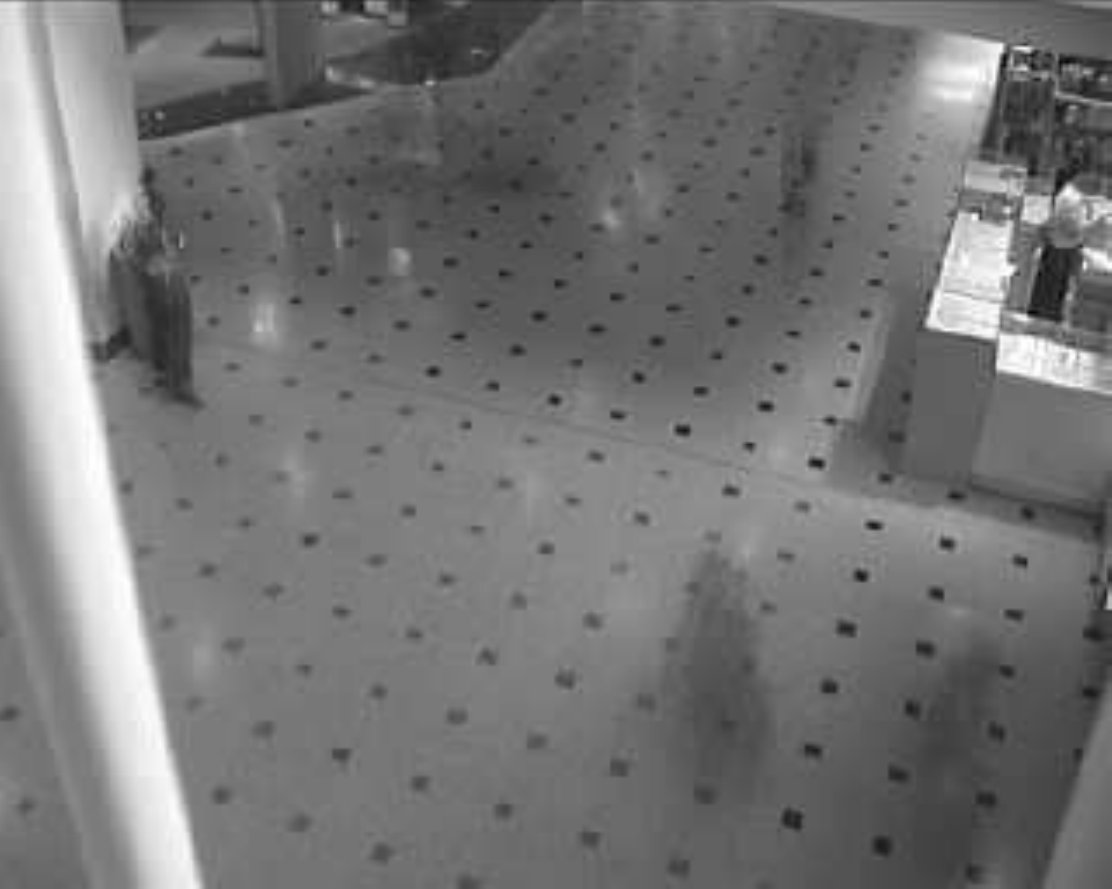}
\includegraphics[width=0.157\linewidth]{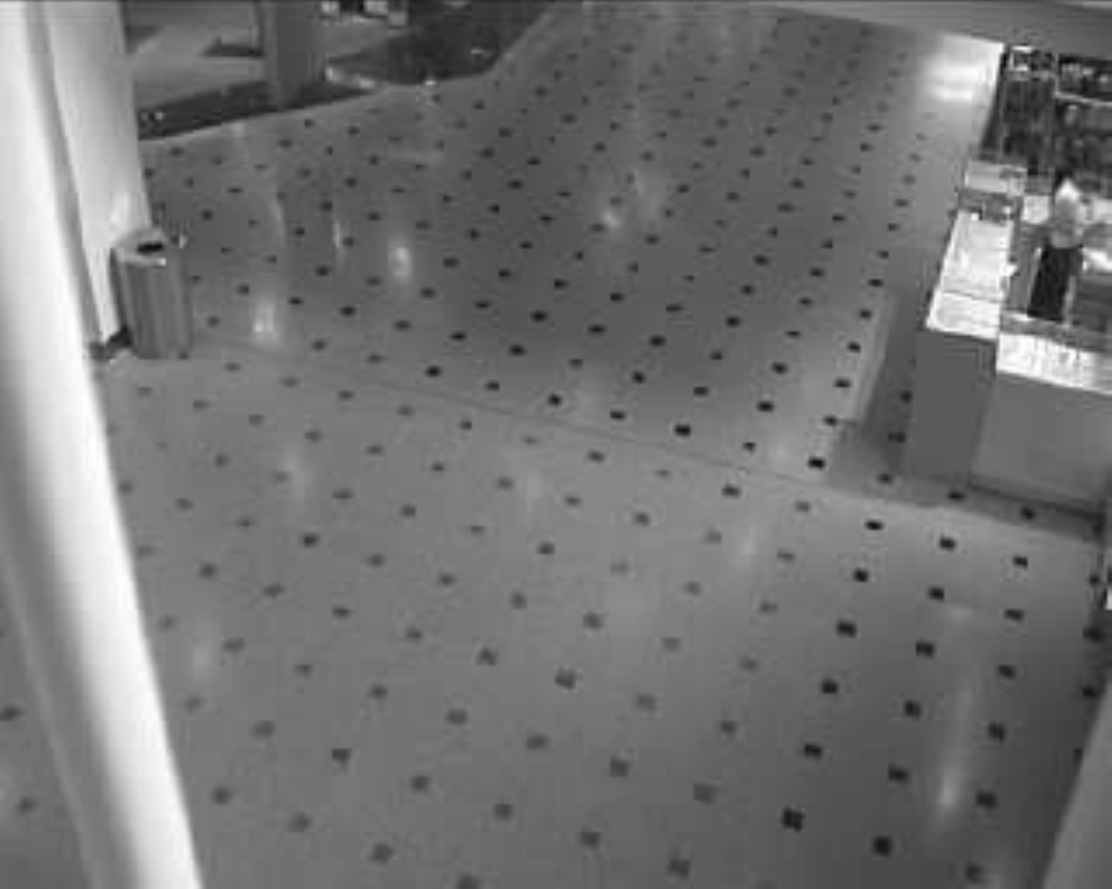}
\includegraphics[width=0.157\linewidth]{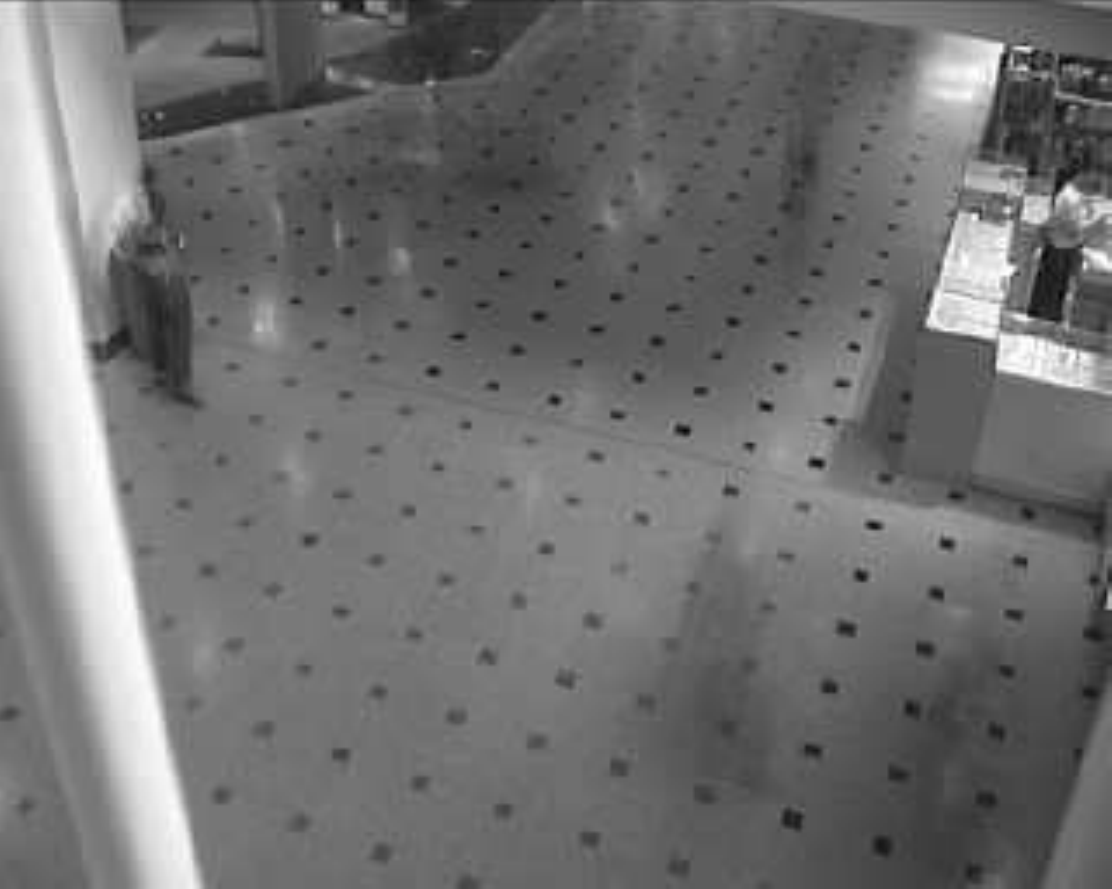}
\includegraphics[width=0.157\linewidth]{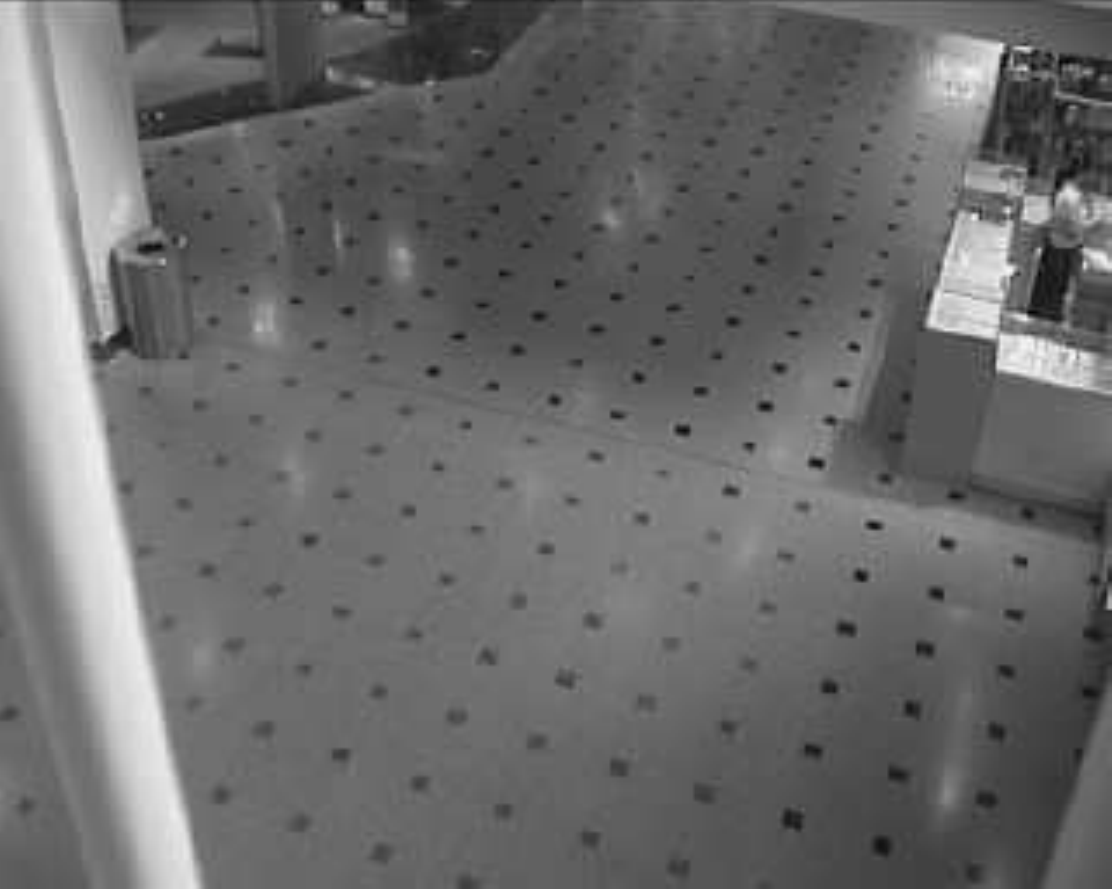}
\includegraphics[width=0.157\linewidth]{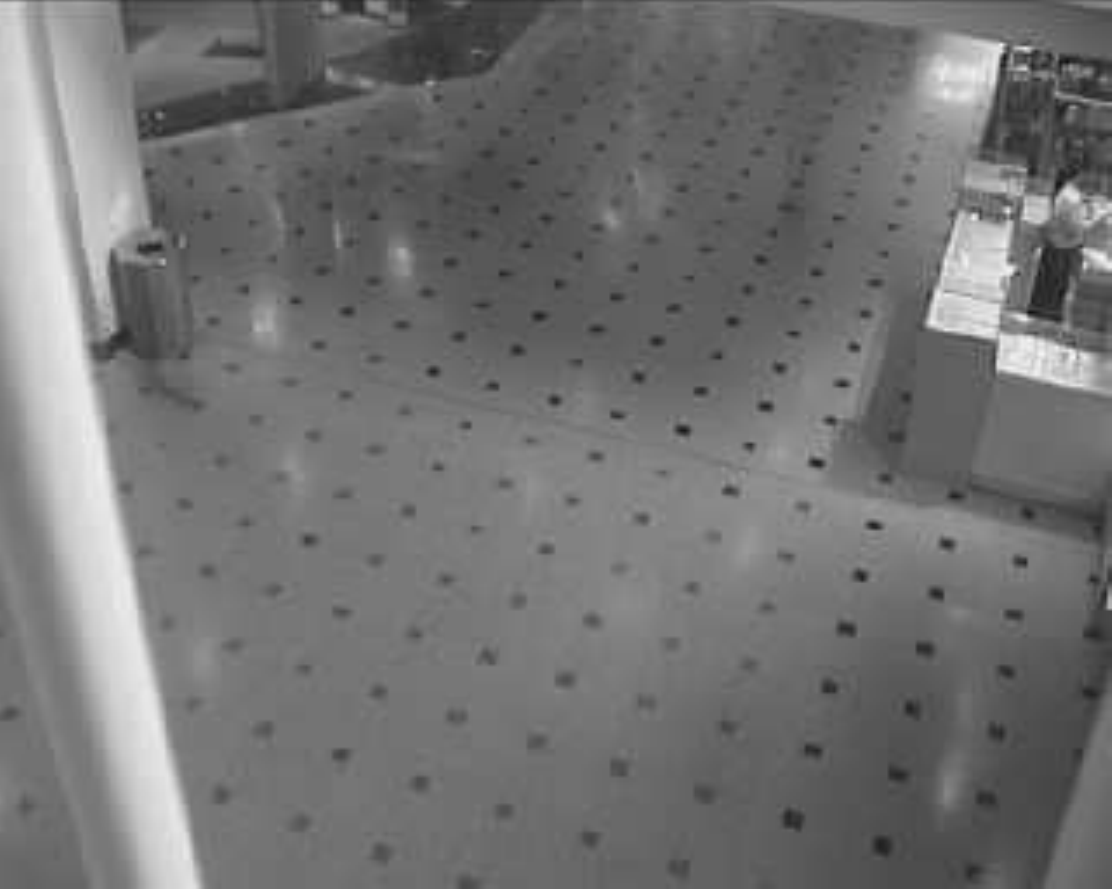}
\includegraphics[width=0.157\linewidth]{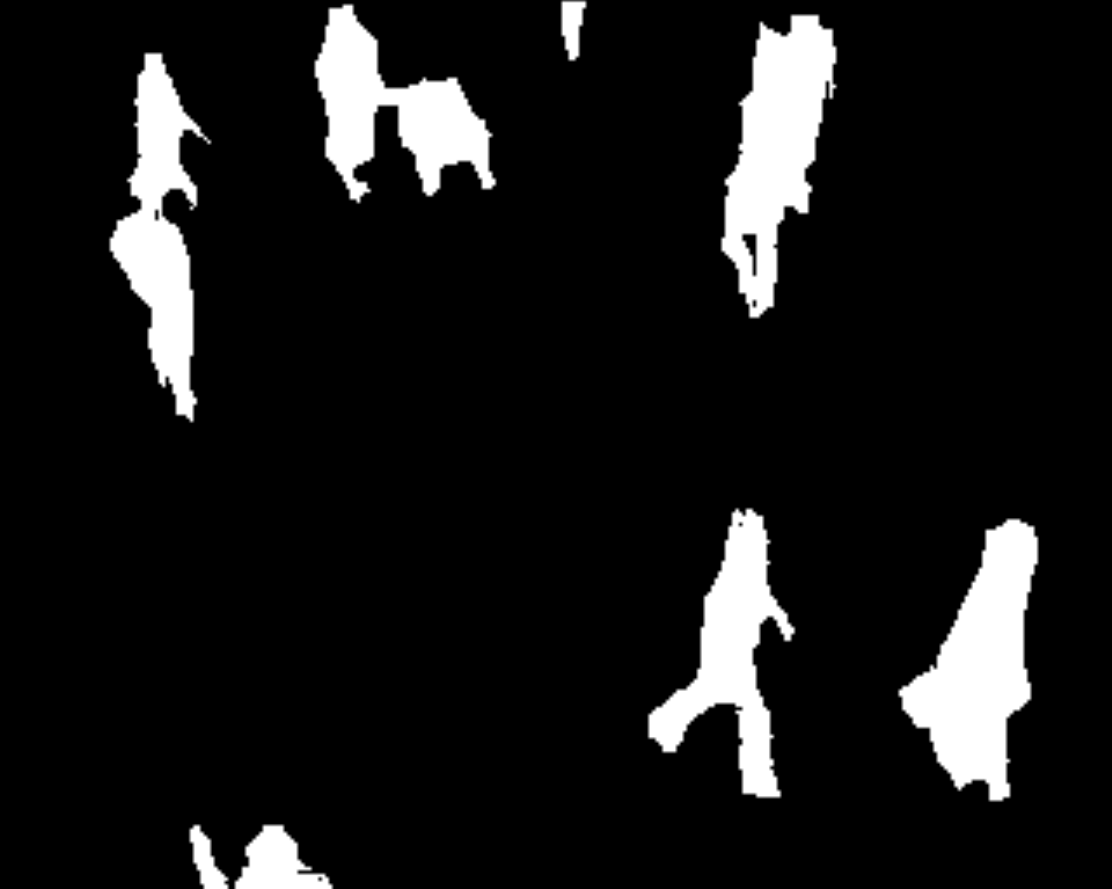}
\includegraphics[width=0.1572\linewidth]{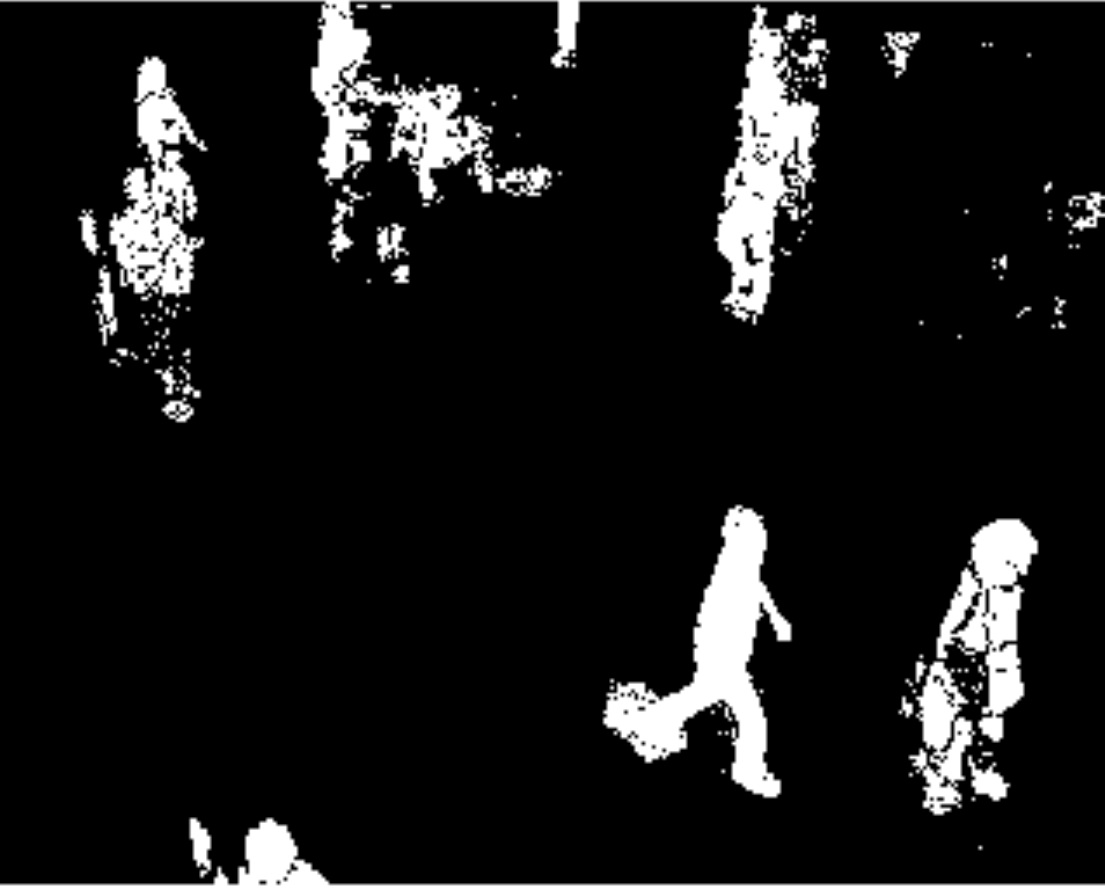}
\includegraphics[width=0.157\linewidth]{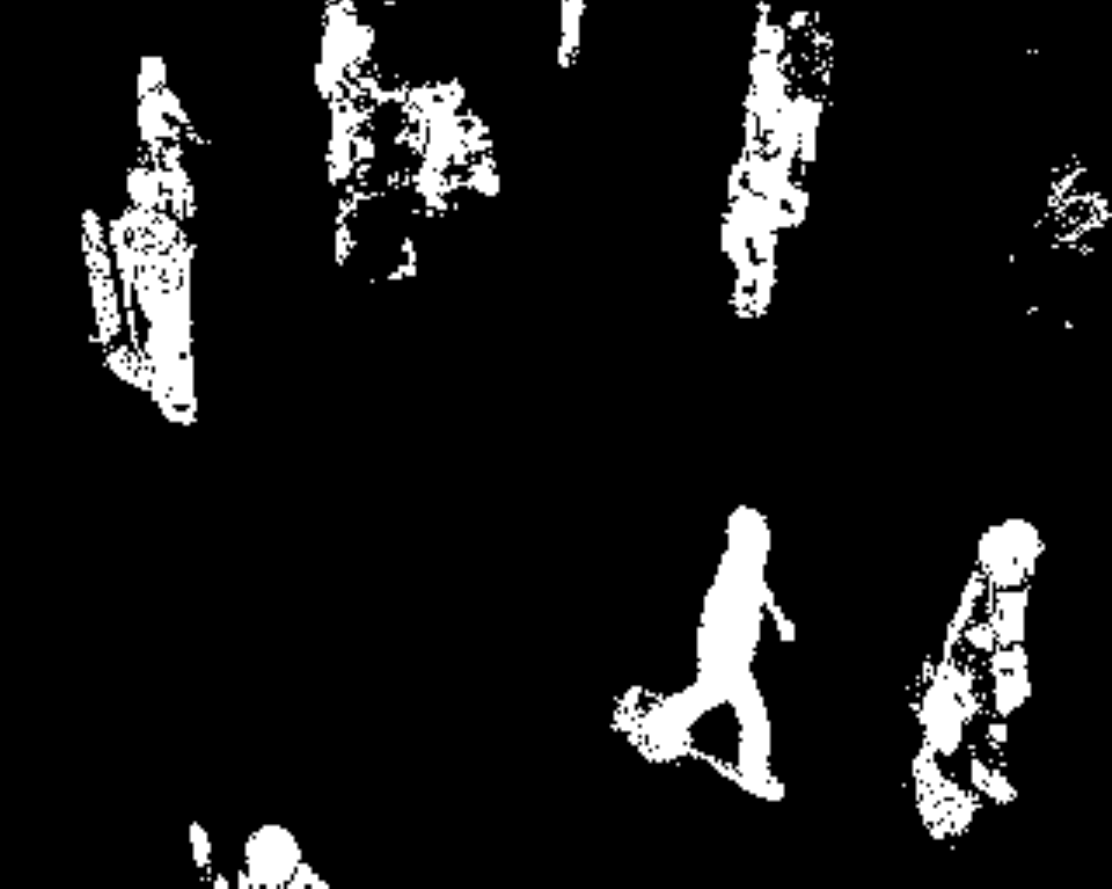}
\includegraphics[width=0.157\linewidth]{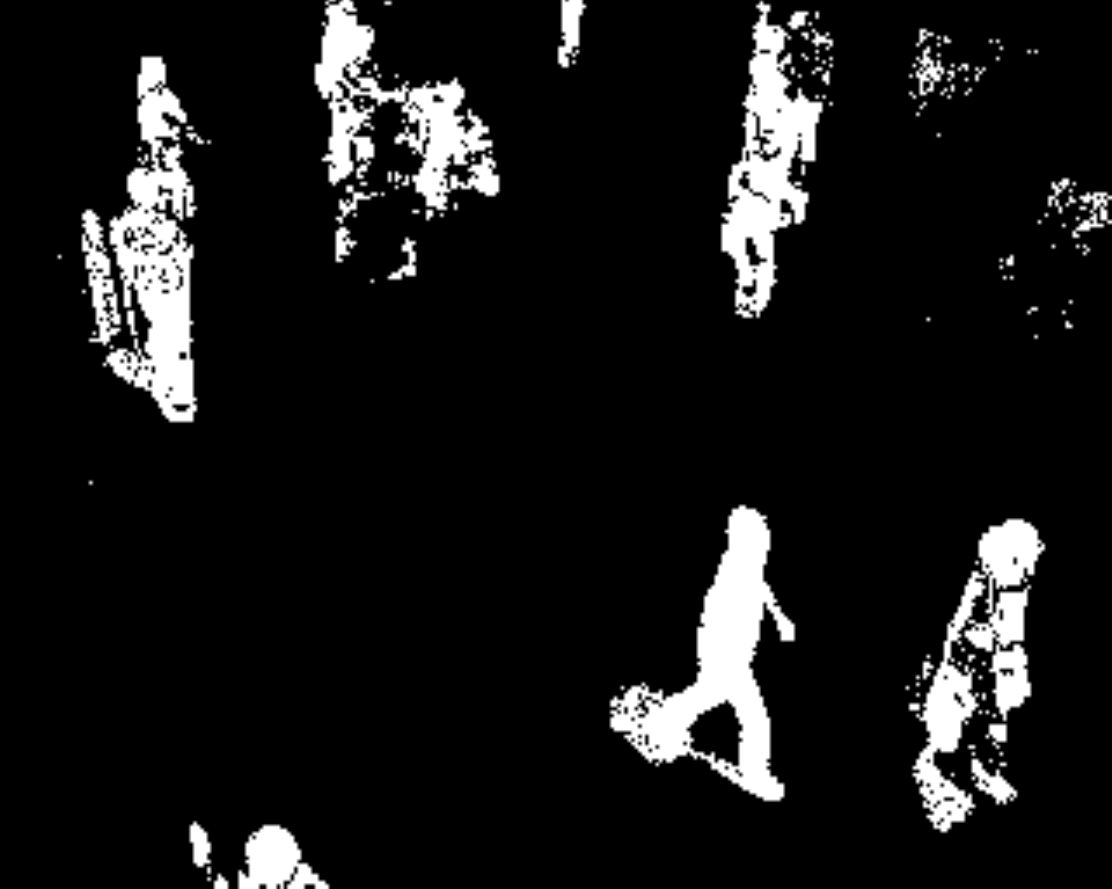}
\includegraphics[width=0.157\linewidth]{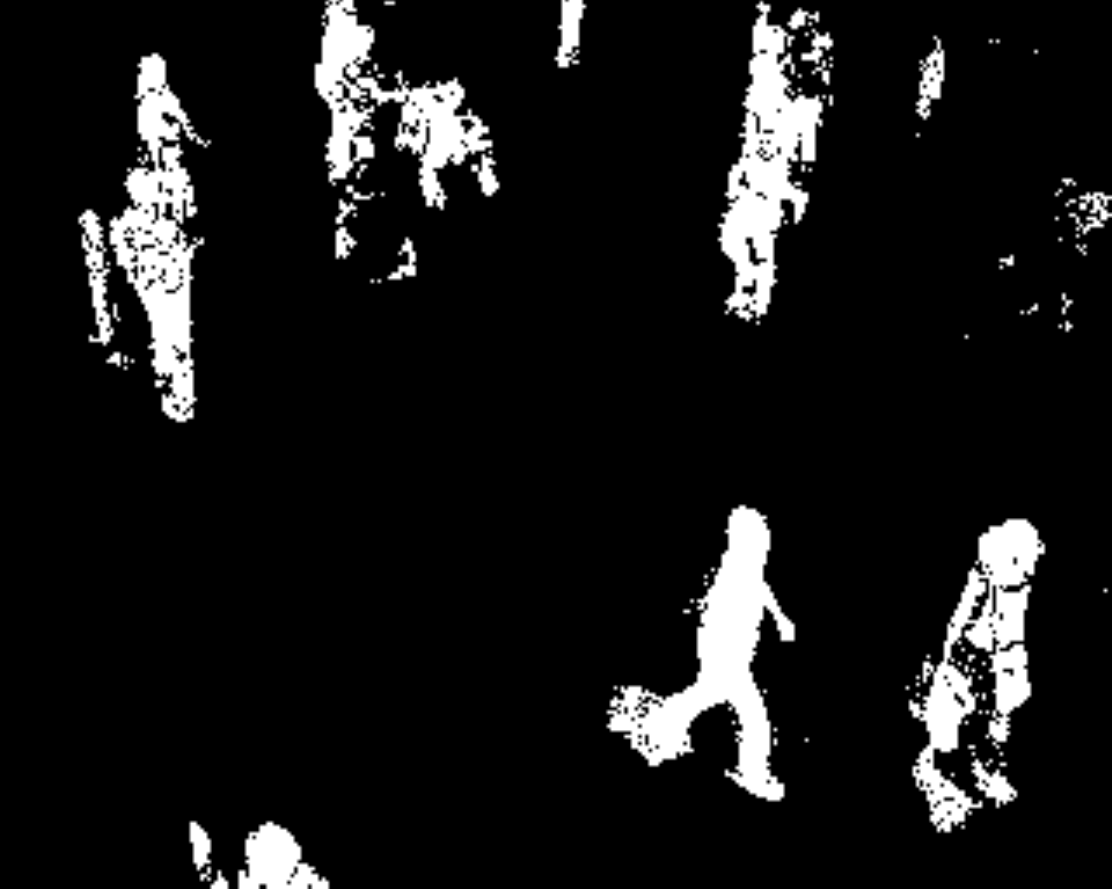}
\includegraphics[width=0.157\linewidth]{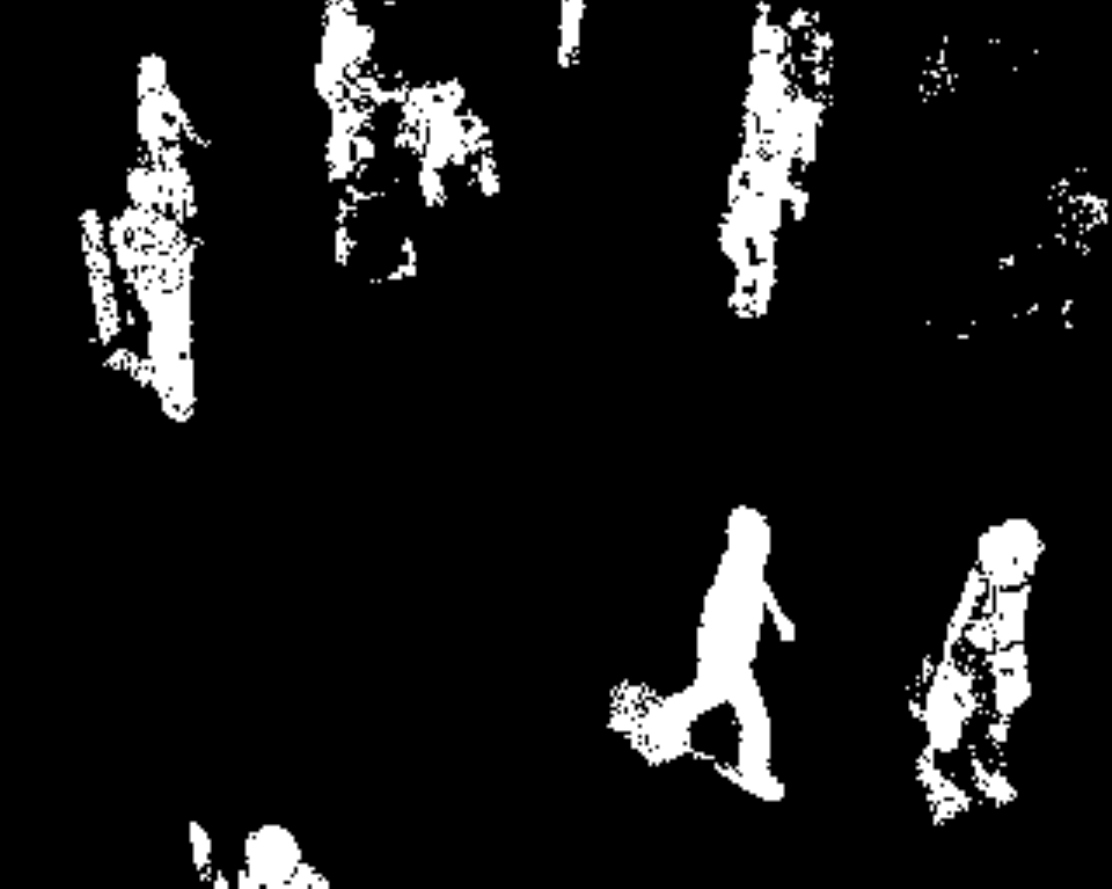}\\~\\
\vspace{-1.8mm}

\includegraphics[width=0.157\linewidth]{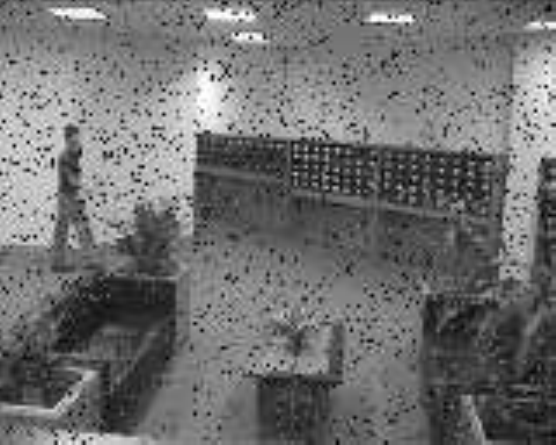}
\includegraphics[width=0.157\linewidth]{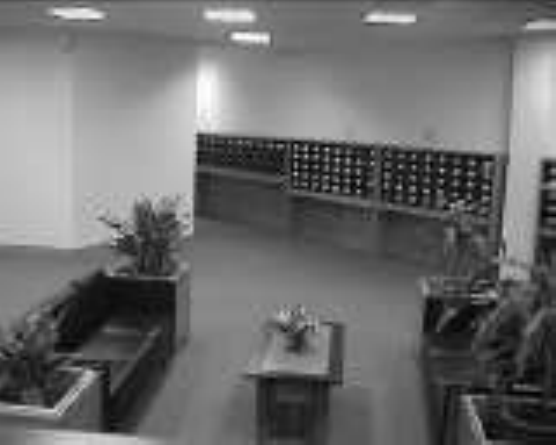}
\includegraphics[width=0.157\linewidth]{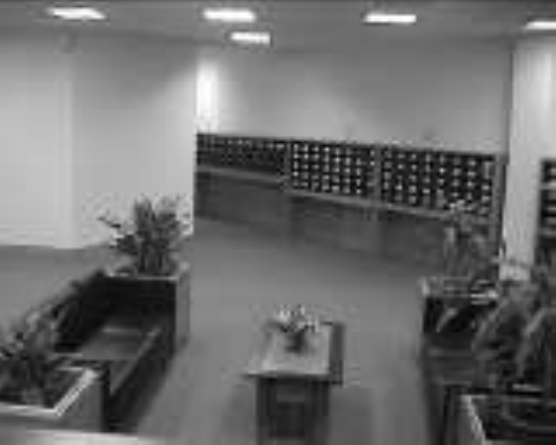}
\includegraphics[width=0.157\linewidth]{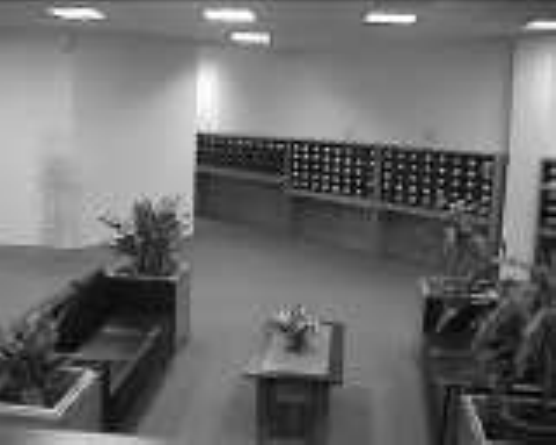}
\includegraphics[width=0.157\linewidth]{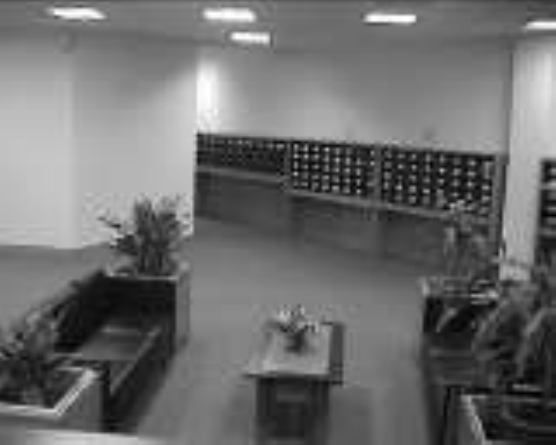}
\includegraphics[width=0.157\linewidth]{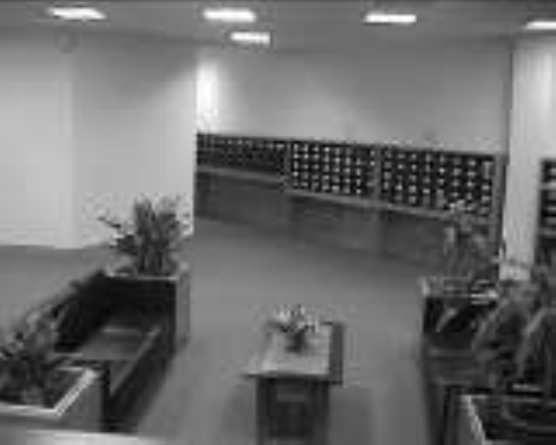}
\includegraphics[width=0.157\linewidth]{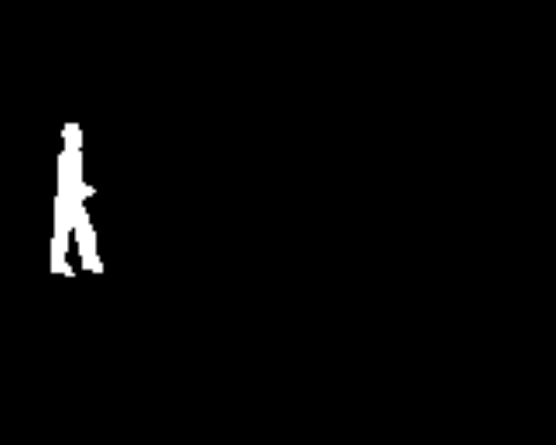}
\includegraphics[width=0.157\linewidth]{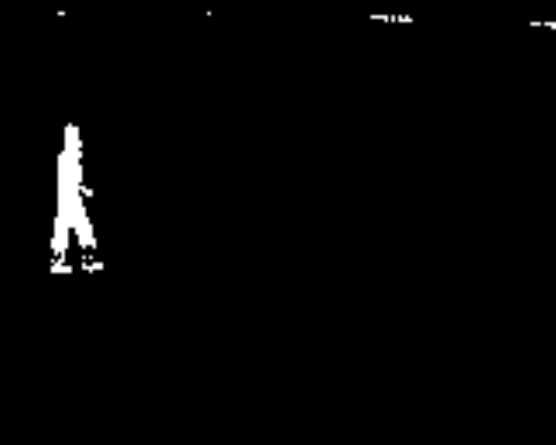}
\includegraphics[width=0.157\linewidth]{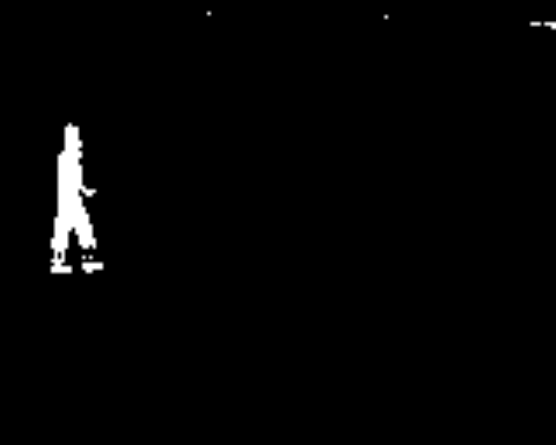}
\includegraphics[width=0.157\linewidth]{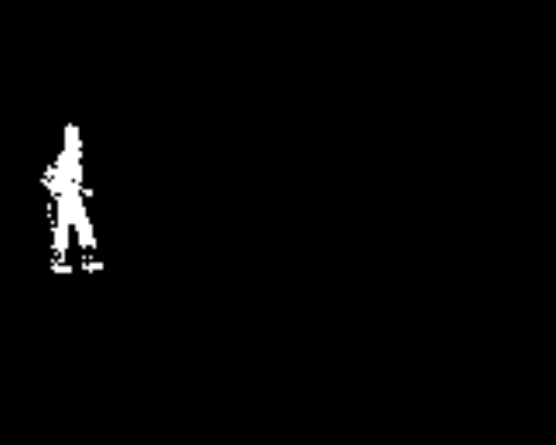}
\includegraphics[width=0.157\linewidth]{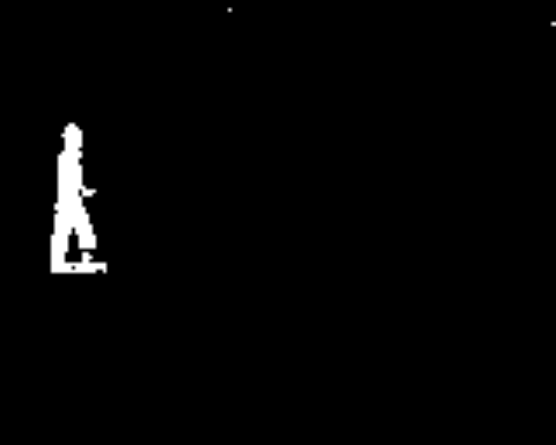}
\includegraphics[width=0.157\linewidth]{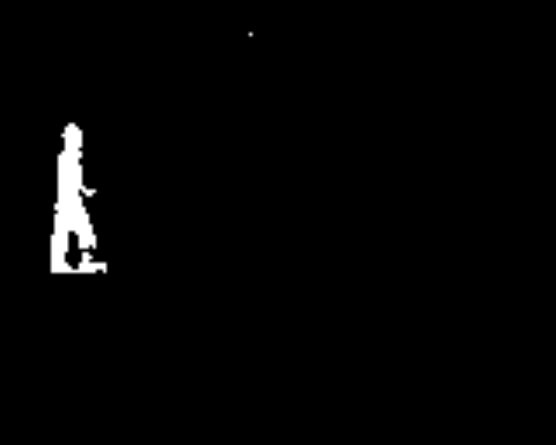}\\~\\
\vspace{-1.8mm}

\includegraphics[width=0.157\linewidth]{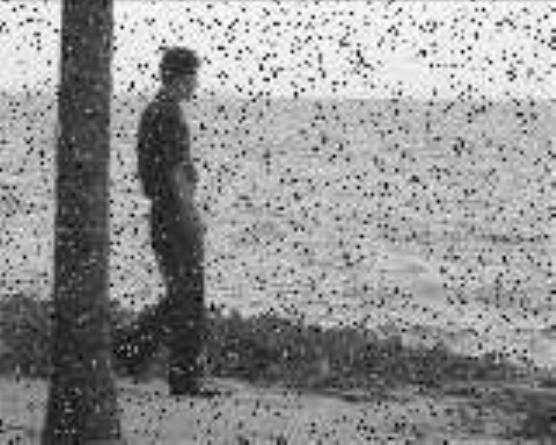}
\includegraphics[width=0.157\linewidth]{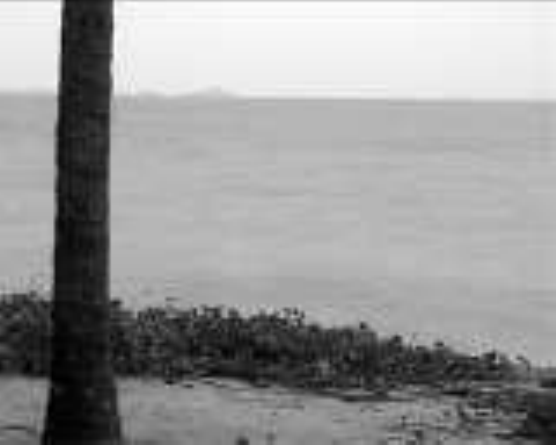}
\includegraphics[width=0.157\linewidth]{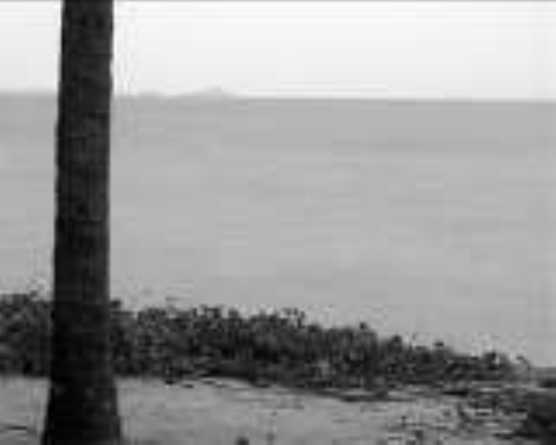}
\includegraphics[width=0.157\linewidth]{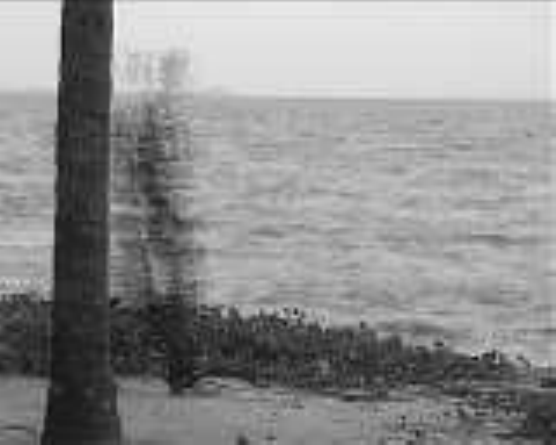}
\includegraphics[width=0.157\linewidth]{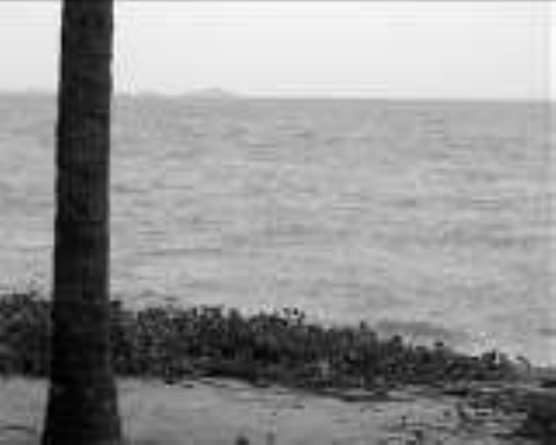}
\includegraphics[width=0.157\linewidth]{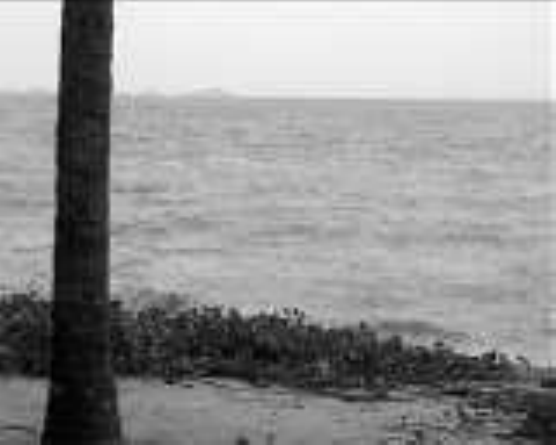}
\subfigure[\!Input, \!segmentation]{\includegraphics[width=0.157\linewidth]{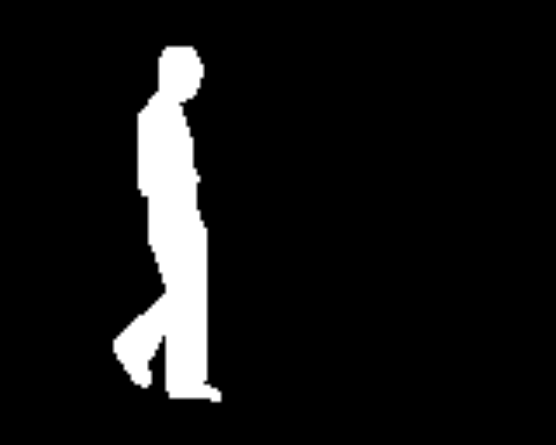}\label{fig5_first_case}}
\subfigure[RegL1~\cite{zheng:lrma}]{\includegraphics[width=0.157\linewidth]{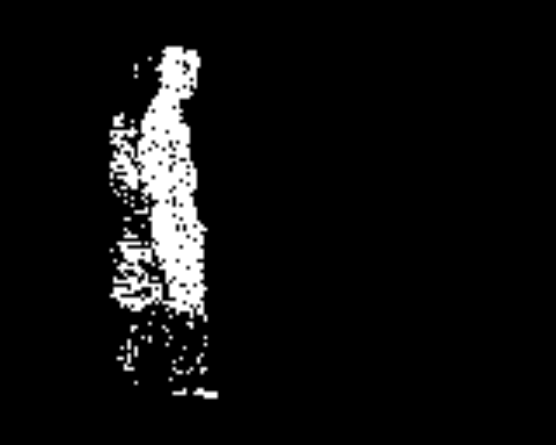}}
\subfigure[Unifying~\cite{cabral:nnbf}]{\includegraphics[width=0.157\linewidth]{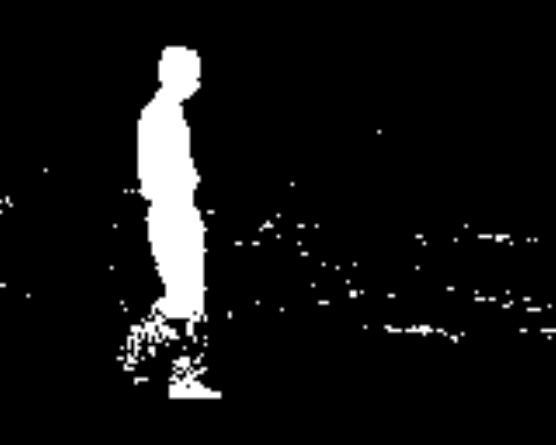}}
\subfigure[factEN~\cite{kim:factEN}]{\includegraphics[width=0.1568\linewidth]{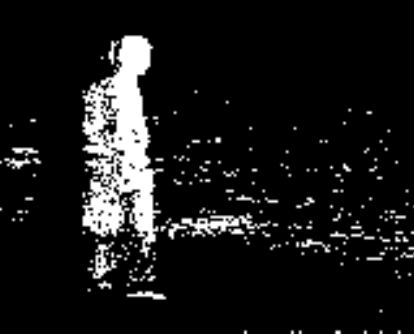}}
\subfigure[$(\textup{S+L})_{1/2}$]{\includegraphics[width=0.157\linewidth]{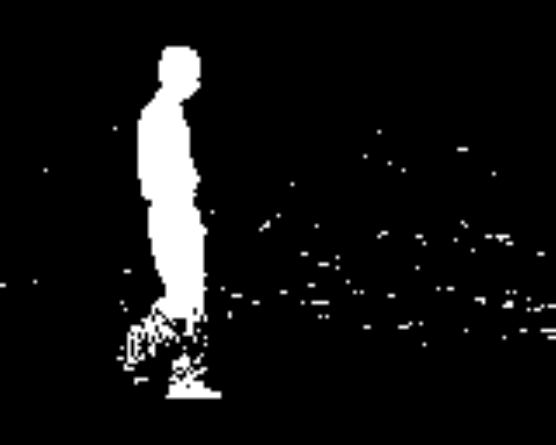}}
\subfigure[$(\textup{S+L})_{2/3}$]{\includegraphics[width=0.157\linewidth]{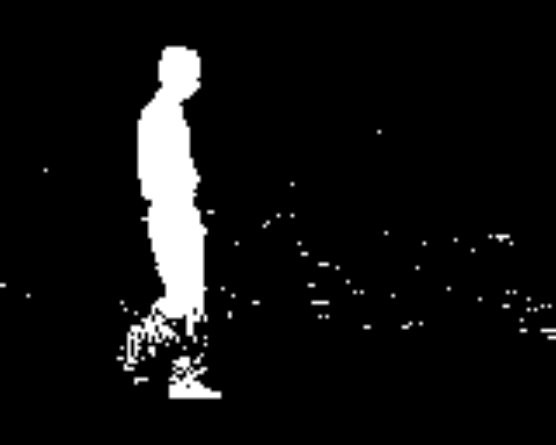}}
\caption{Background and foreground separation results of different algorithms on the Hall, Mall, Lobby and WaterSurface data sets. The one frame with missing data of each sequence (top) and its manual segmentation (bottom) are shown in (a). The results by different algorithms are presented from (b) to (f), respectively. The top panel is the recovered background and the bottom panel is the segmentation.}
\label{fig5}
\end{figure*}

\subsection*{Text Removal}
We report the text removal results of our methods with varying rank parameters (from 10 to 40), as shown in Fig.\ \ref{fig45}, where the rank of the original image is 10. We also present the results of the baseline method, LpSq~\cite{nie:rmc}. The results show that our methods significantly outperform the other methods in terms of RSE and F-measure, and they perform much more robust than Unifying with respect to the rank parameter.

\subsection*{Moving Object Detection}
We present the detailed descriptions for five surveillance video sequences: Bootstrap, Hall, Lobby, Mall and WaterSurface data sets, as shown in Table~\ref{tab_sim1}. Moreover, Fig.\ \ref{fig5} illustrates the foreground and background separation results on the Hall, Mall, Lobby and WaterSurface data sets.

\subsection*{Image Inpainting}
In this part, we first reported the average PSNR results of two proposed methods (i.e., D-N and F-N) with different ratios of random missing pixels from 95\% to 80\%, as shown in Fig.\ \ref{fig7}. Since the methods in~\cite{lu:lrm,hu:tnnr} are very slow, we only present the average inpainting results of APGL\footnote{\url{http://www.math.nus.edu.sg/~mattohkc/NNLS.html}}~\cite{toh:apg} and WNNM\footnote{\url{http://www4.comp.polyu.edu.hk/~cslzhang/}}~\cite{gu:wnnmj}, both of which use the fast SVD strategy and need to compute only partical SVD instead of the full one. Thus, APGL and WNNM are usually much faster than the methods~\cite{lu:lrm,hu:tnnr}. Considering that only a small fraction of pixels are randomly selected, thus we conducted 50 independent runs and report the average PSNR and standard deviation (std). The results show that both our methods consistently outperform APGL~\cite{toh:apg} and WNNM~\cite{gu:wnnmj} in all the settings. This experiment actually shows that both our methods have even greater advantage over existing methods in the cases when the numer of observed pixels is very limited, e.g., 5\% observed pixels.

\begin{figure}[th]
\centering
\subfigure[Image 1]{\includegraphics[width=0.241\linewidth]{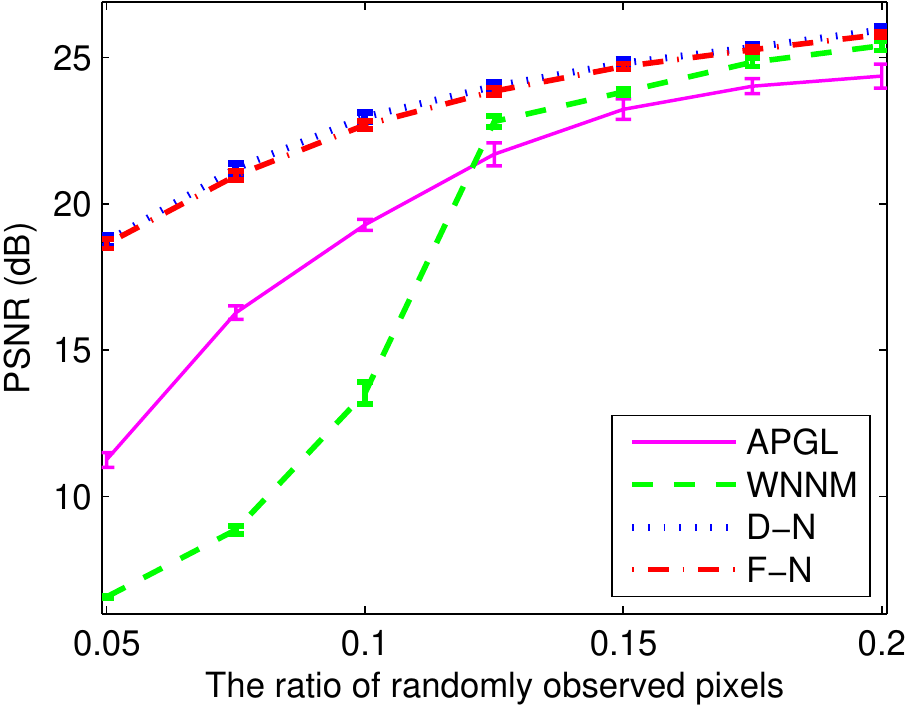}}\;
\subfigure[Image 2]{\includegraphics[width=0.241\linewidth]{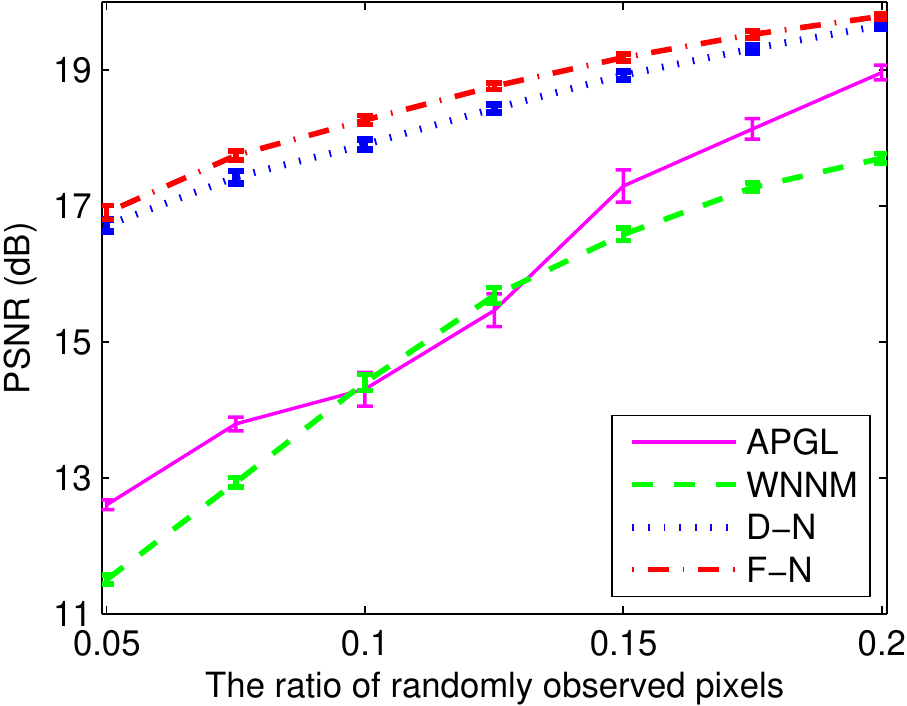}}\;
\subfigure[Image 3]{\includegraphics[width=0.241\linewidth]{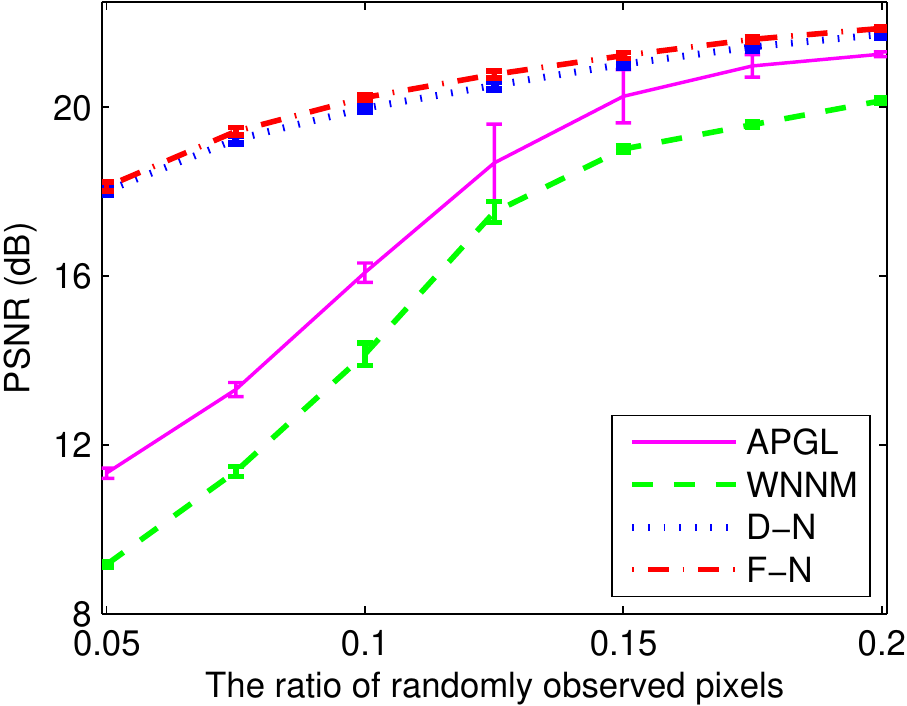}}\;
\subfigure[Image 4]{\includegraphics[width=0.241\linewidth]{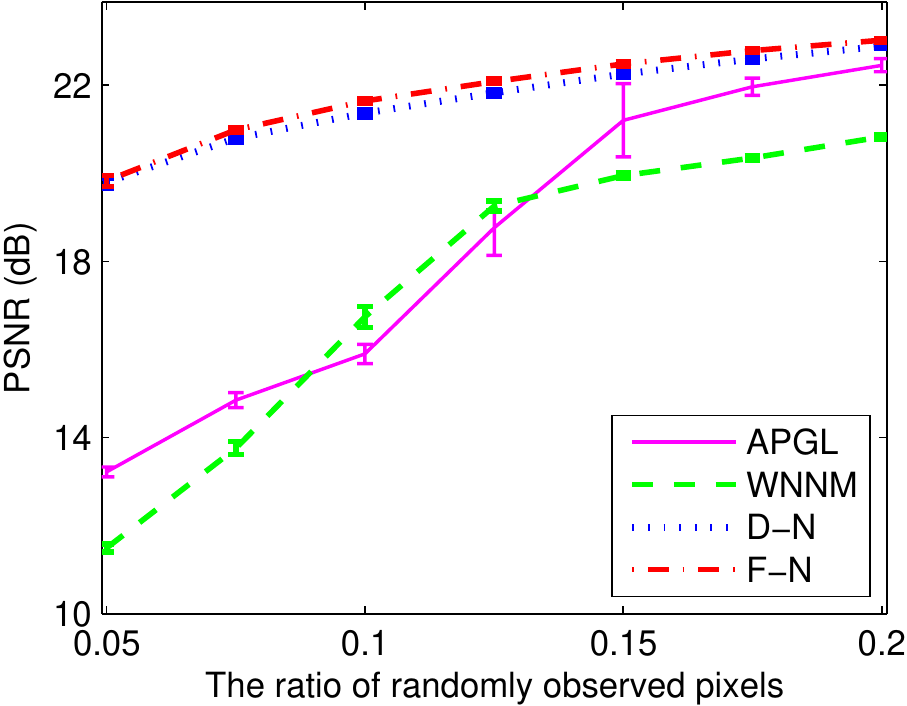}}
\subfigure[Image 5]{\includegraphics[width=0.241\linewidth]{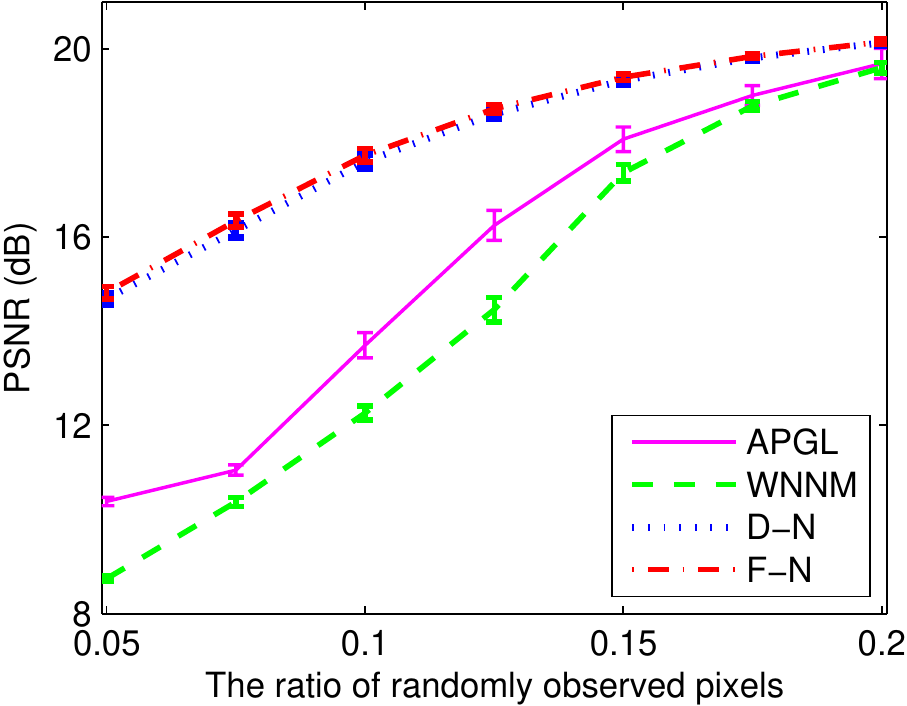}}\;
\subfigure[Image 6]{\includegraphics[width=0.241\linewidth]{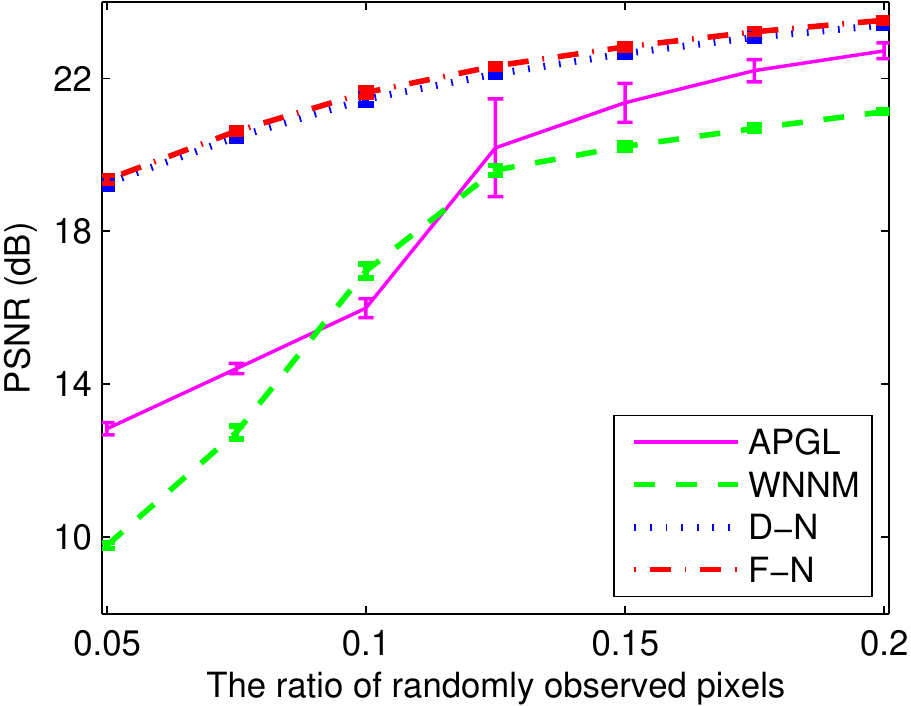}}\;
\subfigure[Image 7]{\includegraphics[width=0.241\linewidth]{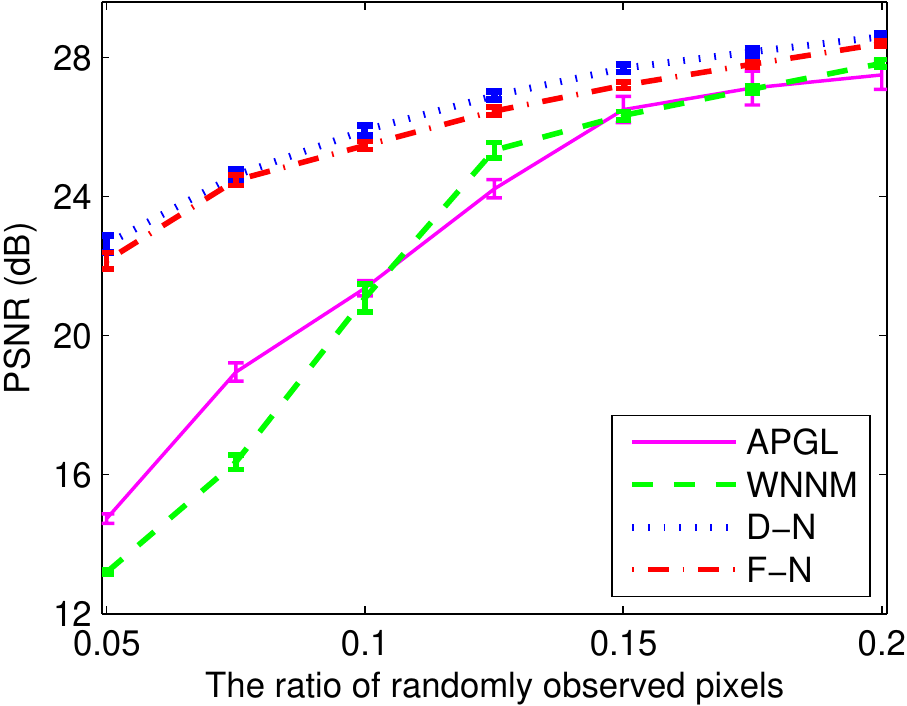}}\;
\subfigure[Image 8]{\includegraphics[width=0.241\linewidth]{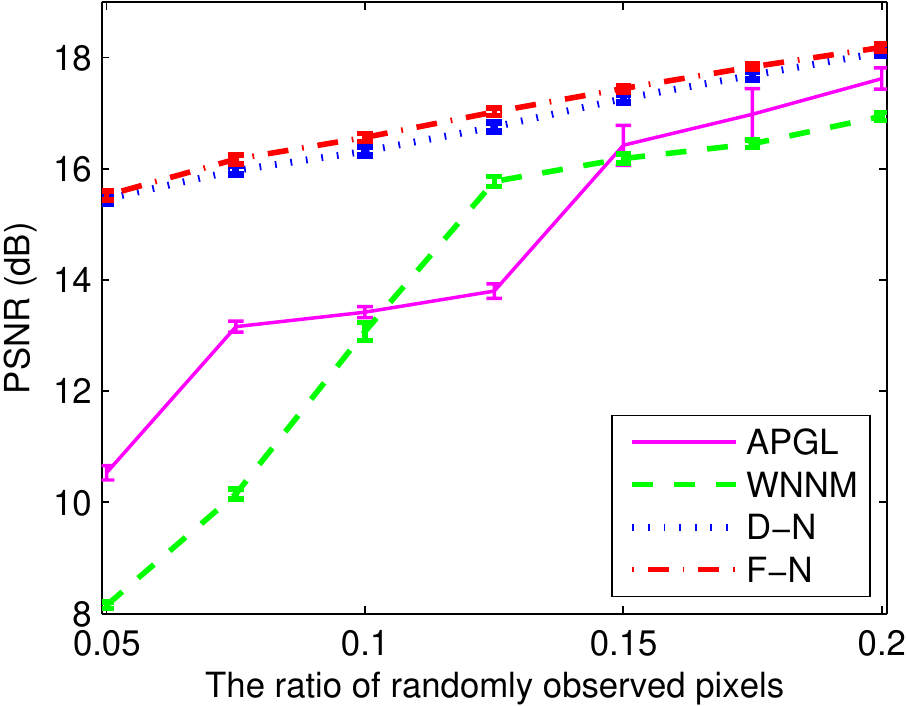}}
\caption{The average PSNR and standard deviation of APGL~\cite{toh:apg}, WNNM~\cite{gu:wnnmj} and both our methods for image inpainting vs.\ fraction of observed pixels (best viewed in colors).}
\label{fig7}
\end{figure}

As suggested in~\cite{hu:tnnr}, we set $d\!=\!9$ for our two methods and TNNR\footnote{\url{https://sites.google.com/site/zjuyaohu/}}~\cite{hu:tnnr}. To evaluate the robustness of our two methods with respect to their rank parameter, we report the average PSNR and standard deviation of two proposed methods with varying rank parameter $d$ from 7 to 15, as shown in Fig.\ \ref{fig8}. Moreover, we also present the average inpainting results of TNNR~\cite{hu:tnnr} over 50 independent runs. It is clear that two proposed methods perform much more robust than TNNR with repsect to the rank parameter.

\begin{figure}[th]
\centering
\subfigure[Image 1]{\includegraphics[width=0.241\linewidth]{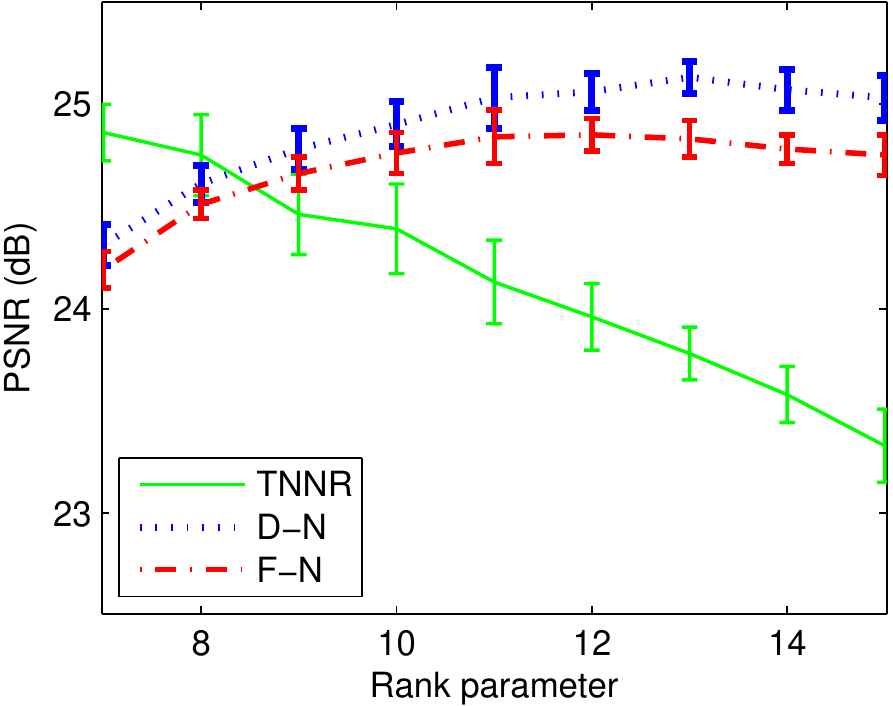}}\;
\subfigure[Image 2]{\includegraphics[width=0.241\linewidth]{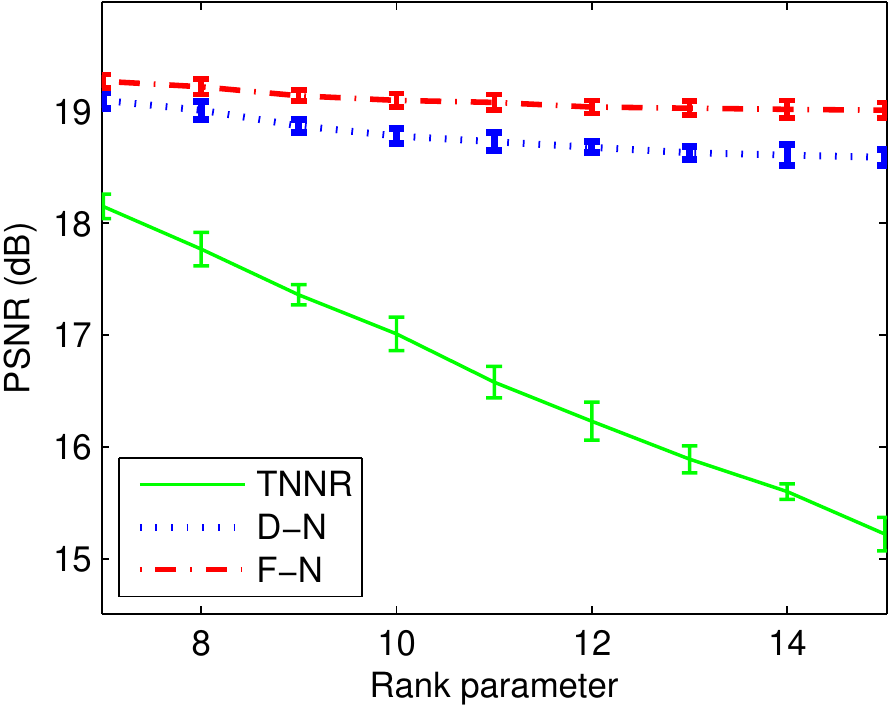}}\;
\subfigure[Image 3]{\includegraphics[width=0.241\linewidth]{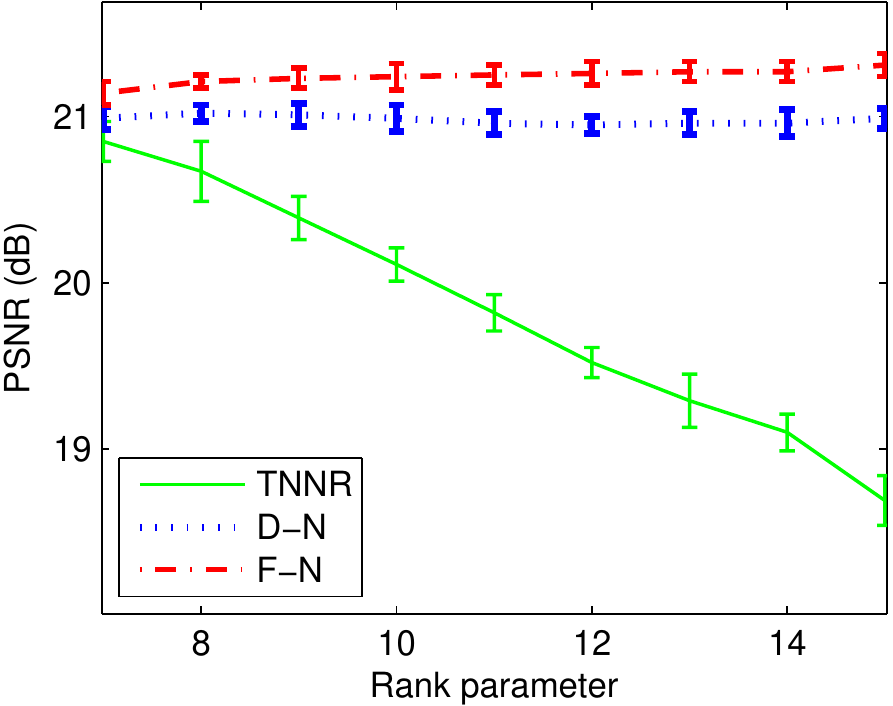}}\;
\subfigure[Image 4]{\includegraphics[width=0.241\linewidth]{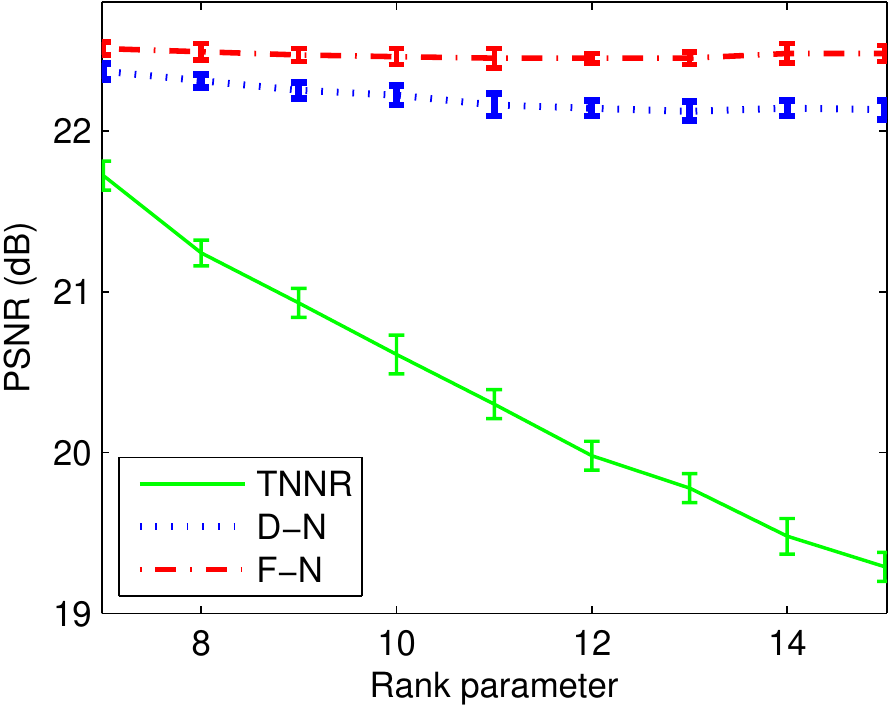}}
\subfigure[Image 5]{\includegraphics[width=0.241\linewidth]{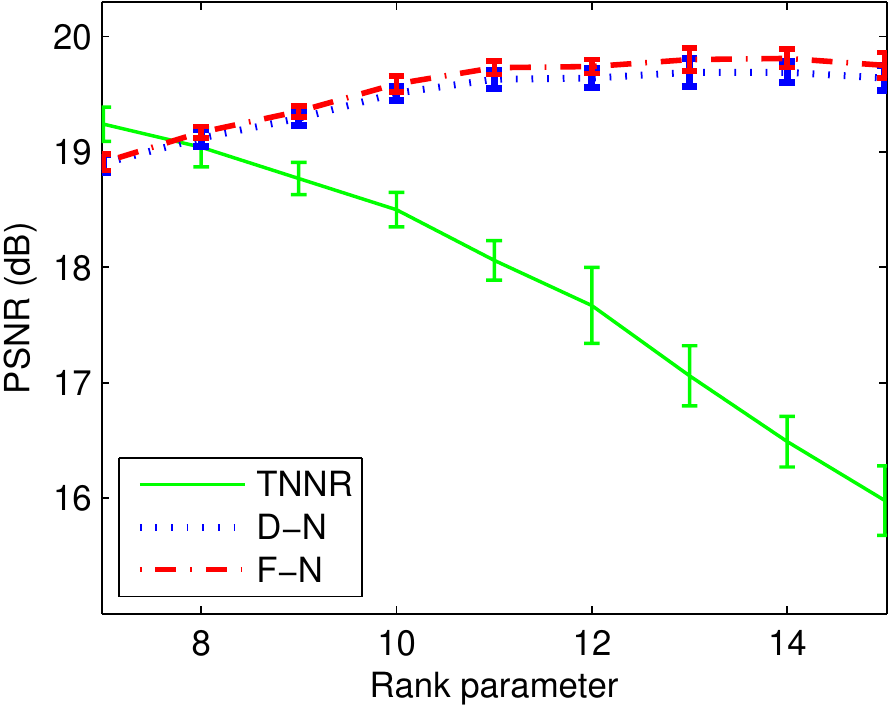}}\;
\subfigure[Image 6]{\includegraphics[width=0.241\linewidth]{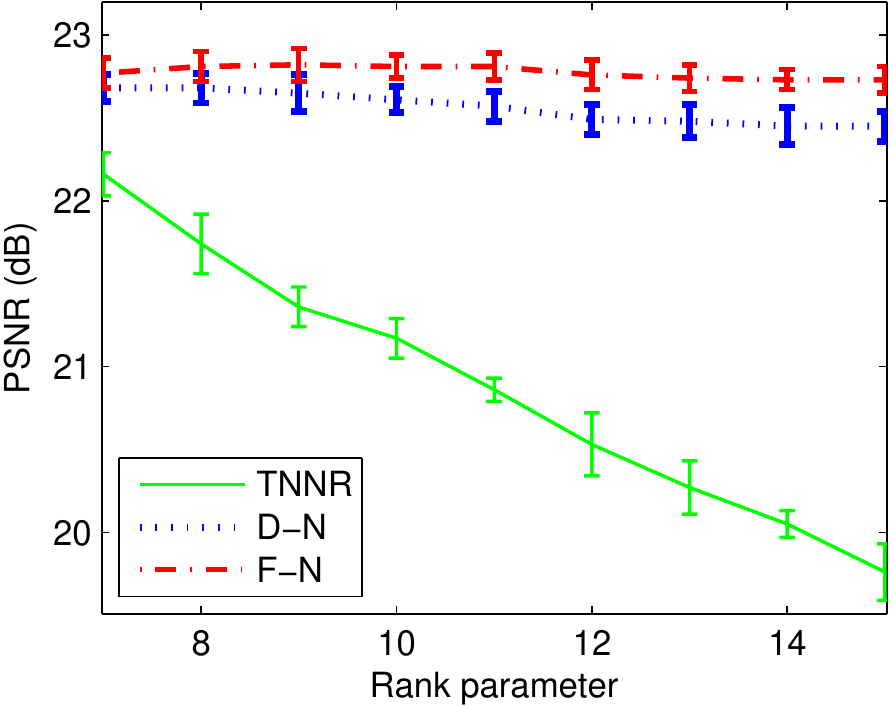}}\;
\subfigure[Image 7]{\includegraphics[width=0.241\linewidth]{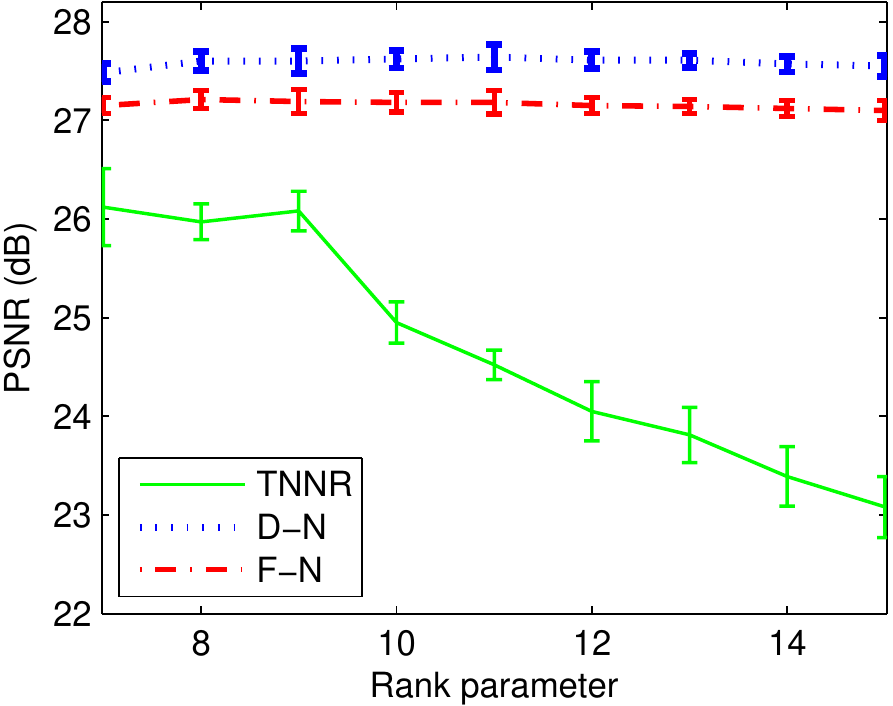}}\;
\subfigure[Image 8]{\includegraphics[width=0.241\linewidth]{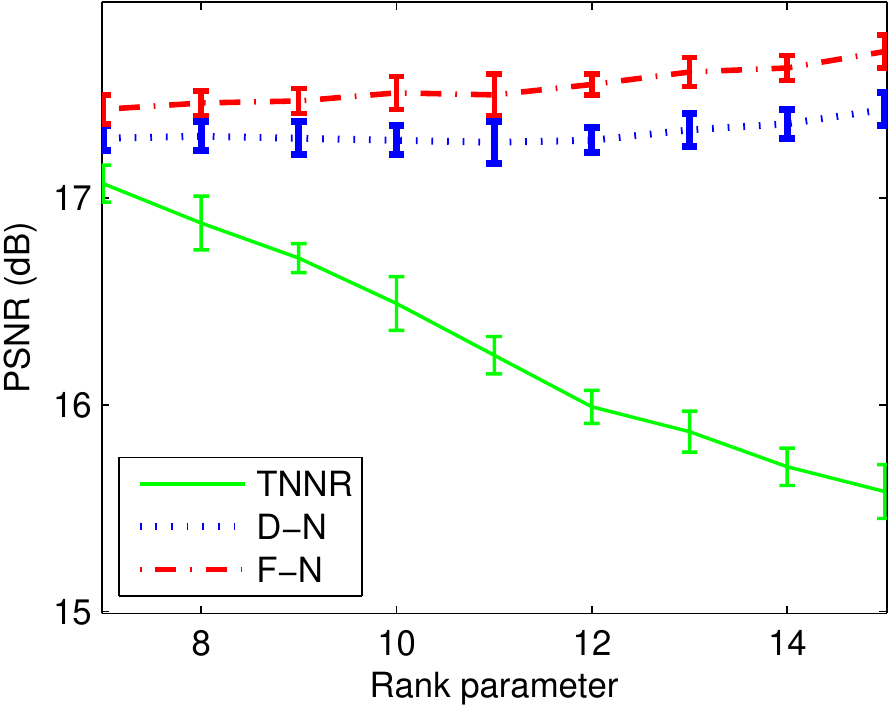}}
\caption{The average PSNR and standard deviation of TNNR~\cite{hu:tnnr} and both our methods for image inpainting vs.\ the rank parameter (best viewed in colors).}
\label{fig8}
\end{figure}


\end{document}